\pdfoutput=1
\documentclass[a4paper, 11pt, twoside]{Thesis}  
\graphicspath{Figures/}  
\usepackage{verbatim}  
\usepackage{vector}  
\usepackage{enumerate}
\hypersetup{urlcolor=blue, colorlinks=false}  

\usepackage{booktabs}
\usepackage{multirow}
\usepackage{graphicx}
\usepackage{lscape}
\usepackage{caption}
\usepackage{subfigure}  
\usepackage{array}
\usepackage{longtable}
\usepackage{rotating}
\usepackage{notoccite}
\usepackage{subfiles} 
\usepackage[export]{adjustbox}
\newenvironment{conditions}
  {\par\vspace{\abovedisplayskip}\noindent\begin{tabular}{>{$}l<{$} @{${}-\:{}$} l}}
  {\end{tabular}\par\vspace{\belowdisplayskip}}
  
\begin{document}
\frontmatter      

\UNIVERSITY{{Gdańsk University of Technology}}    
%
\school{{Faculty of Electrical and Control Engineering}}
\gradtime{{2022}}

%
\title  {Data augmentation and explainability for bias discovery and mitigation in deep learning}

\authors  {\texorpdfstring
            {\href{agnieszka.mikolajczyk@pg.edu.pl}{Agnieszka Mikołajczyk}}
            {Agnieszka Mikołajczyk}
            }
\addresses  {\groupname\\\deptname\\\univname}  
\date       {\today}
\subject    {Automation, electronic and electrical
engineering}

\setstretch{1.3}  

\fancyhead{}  
\rhead{\thepage}  
\lhead{}  

\pagestyle{fancy}  

\copyrightnotice{

\addtocontents{toc}{\vspace{1em}}  

\textsuperscript{\textcopyright  \authornames (2022).}

Aware of criminal liability for violations of the Act of 4th February 1994 on Copyright and Related Rights (Journal of Laws 2006, No. 90, item 631) and disciplinary actions set out in the Law on Higher Education (Journal of Laws 2012, item 572 with later amendments) \footnote{Act of 27th July 2005, Law on Higher Education: Chapter 7, Criminal responsibility of PhD students, Article 226.}, as well as civil liability, I declare, that the submitted PhD dissertation is my own work. 
I declare, that the submitted PhD dissertation is my own work performed under and in cooperation with the supervision of Michał Grochowski.

This submitted PhD dissertation has never before been the basis of an official procedure associated with the awarding of a PhD degree.

All the information contained in the above thesis which is derived from written and electronic sources is documented in a list of relevant literature in accordance with art. 34 of the Copyright and Related Rights Act. 
I confirm that this PhD dissertation is identical to the attached electronic version.  
\vfil\vfil\null
Signature:\\
\rule[1em]{25em}{0.5pt}  

Print Name: Agnieszka Mikołajczyk\\
\rule[1em]{25em}{0.5pt}  

Date:\\
\rule[1em]{25em}{0.5pt}  

}
\cleardoublepage  


\cleardoublepage
\abstract{
\addtocontents{toc}{}  
In recent years, deep neural networks have achieved remarkable performance across a range of tasks. The amount, nature, and quality of data used to train neural networks undeniably have a considerable impact on model performance. However, even the most widely accepted benchmarks often contain noisy, inconsistent or incorrectly labeled samples, as well as unintentional bias which can be introduced through data collection, preparation, or analysis.  
This dissertation explores the impact of bias in deep neural networks and presents methods for reducing its influence on model performance. The first part begins by categorizing and describing potential sources of bias and errors in data and models, with particular focus on bias in machine learning pipelines. Next chapter outlines a taxonomy and methods of Explainable AI as a way to justify predictions and control and improve the model. 
Then, as an example of a laborious manual data inspection and bias discovery process, a skin lesion dataset is manually examined. A Global Explanations for Bias Identification method is proposed as an alternative semi-automatic approach to manual data exploration for discovering potential biases in data. Relevant numerical methods and metrics are discussed for assessing the effects of the identified biases on the model. Whereas identifying errors and bias is critical, improving the model and reducing the number of flaws in the future is an absolute priority. 
Hence, the second part of the thesis focuses on mitigating the influence of bias on machine learning models. Three approaches are proposed and discussed: Style Transfer Data Augmentation, Targeted Data Augmentations, and Attribution Feedback. Style Transfer Data Augmentation aims to address shape and texture bias by merging a style of a malignant lesion with a conflicting shape of a benign one. Targeted Data Augmentations randomly insert possible biases into all images in the dataset during the training, as a way to make the process random and, thus, destroy spurious correlations. Lastly, Attribution Feedback is used to fine-tune the model to improve its accuracy by eliminating obvious mistakes and teaching it to ignore insignificant input parts via an attribution loss. The goal of these approaches is to reduce the influence of bias on machine learning models, rather than eliminate it entirely.
}

\cleardoublepage  




\setstretch{1.3}  
\ack{
\addtocontents{toc}{}  





This thesis would not have been possible without the support of many people. Many thanks to my supervisor, Michał Grochowski, who read my numerous revisions, always offering his time, advice, and encouragement. My sincere thanks to all members of my doctorate committee for patience and understanding during the five years of effort that went into preparing and writing this thesis. Many thanks to team members Arkadiusz Kwasigroch, Maria Ferlin, and Zuzanna Klawikowska, who offered guidance and support.

Thanks to the \textit{Gdańsk University of Technology} for awarding me numerous scholarships and providing me with the financial means to complete this research. Thanks to \textit{Polish National Science Center} for supporting me with the research grant \textit{Detecting and overcoming bias in data with explainable artificial intelligence} (Preludium no. \textit{UMO-2019/35/N/ST6/04052}). I am fortunate to have been a part of the \textit{Women in AI}: thank you for the fun ML projects, the lively AI conversations, and most importantly, the great companionship and support. My research wouldn't look the same without the community support: many thanks to the \textit{ML Gdańsk} team. Big thank you to \textit{Voicelab.ai} and Jacek Kawalec, who gave me access to the computational infrastructure, supported my volunteer ML projects, and gave me networking opportunities. Thanks to my \textit{NLP team}, Piotr Pęzik, for an endless supply of interesting ideas, Adam Wawrzyński, Wojciech Janowski, Filip Żarnecki, for lively discussions, and the great sense of humor.

Thanks to numerous friends who endured this long process with me, always offering support and love. Thank you, Paulina and Marysia, for being there for me, and believing in me. Many thanks to all Sky girls for fun and refreshing classes, where I could clear my cloudy mind and find my piece. As always, many thanks to my husband Krzysztof, who with support and love endured my weakest moments, always ready to comfort me with a good laugh and homemade waffles. Many thanks to my parents and brother, who always supported me in pursuing my dreams. 

}
\clearpage  

\pagestyle{fancy}  

\lhead{\emph{Contents}}  
\tableofcontents  

\lhead{\emph{List of Figures}}  
\listoffigures  

\lhead{\emph{List of Tables}}  
\listoftables  

\setstretch{1.5}  
\clearpage  
\lhead{\emph{Abbreviations}}  
\listofsymbols{ll}  
{
\textbf{ACC}	&	\textbf{Acc}uracy	\\				
\textbf{AI}	&	\textbf{A}rtificial	\textbf{I}ntelligence	\\			
\textbf{AM}	&	\textbf{A}ctivation	\textbf{M}aximization	\\			
\textbf{AR}	&	\textbf{A}ugmented	\textbf{R}eality	\\			
\textbf{AUC}	&	\textbf{A}rea	\textbf{U}nder	the	\textbf{C}urve	\\	
\textbf{CAM}	&	\textbf{C}lass	\textbf{A}ctivation	\textbf{M}ap	\\		
\textbf{CAV}	&	\textbf{C}oncept	\textbf{A}ctivation	\textbf{V}ectors	\\		
\textbf{CBA}	&	\textbf{C}ounterfactual	\textbf{B}ias	\textbf{I}nsertion	\\		
\textbf{CE}	&	\textbf{C}ross-\textbf{E}ntropy	\\				
\textbf{CF}	&	\textbf{C}ommon	\textbf{F}eatures	\\			
\textbf{CNN}	&	\textbf{C}onvolutional	\textbf{N}eural	\textbf{N}etwork	\\		
\textbf{DA}	&	\textbf{D}ata	\textbf{A}ugmentation	\\			
\textbf{DeepLIFT}	&	\textbf{D}eep	\textbf{L}earning	\textbf{I}mportant \textbf{F}ea\textbf{t}ures	\\			
\textbf{DNN}	&	\textbf{D}eep	\textbf{N}eural	\textbf{N}etwork	\\		
\textbf{DTD}	&	\textbf{D}eep	\textbf{T}aylor	\textbf{D}ecomposition	\\		
\textbf{GALE}	&	\textbf{G}lobal	\textbf{A}ggregations	of	\textbf{L}ocal	\textbf{E}xplanations	\\
\textbf{GAM}	&	\textbf{G}lobal	\textbf{A}ttribution	\textbf{M}apping	\\		
\textbf{GAN}	&	\textbf{G}enerative	\textbf{A}dversarial	\textbf{N}etwork	\\		
\textbf{GEBI}	&	\textbf{G}lobal	\textbf{E}xplanation	for	\textbf{B}ias	\textbf{I}dentification	\\
\textbf{GradCAM}	&	\textbf{Grad}ient-weighted	\textbf{C}lass	\textbf{A}ctivation	\textbf{M}apping	\\	
\textbf{IDR}	&	\textbf{I}nput	\textbf{D}ependence	\textbf{R}ate	\\		
\textbf{IIR}	&	\textbf{I}nput	\textbf{I}ndependence	\textbf{R}ate	\\		
\textbf{INFD}	&	Explanation	\textbf{Inf}i\textbf{d}elity	\\			
\textbf{ISIC}	&		\textbf{I}nternational	\textbf{S}kin	\textbf{I}maging	\textbf{C}ollaboration	\\
\textbf{LDA}	&	\textbf{L}atent	\textbf{D}irichlet	\textbf{A}llocation	\\		
\textbf{LIME}	&	\textbf{L}ocal	\textbf{I}nterpretable	\textbf{M}odel-agnostic	\textbf{E}xplanations	\\	
\textbf{LRP}	&	\textbf{L}ayer-wise	\textbf{R}elevance	\textbf{P}ropagation	\\		
\textbf{MCS}	&	\textbf{M}odel	\textbf{C}ontrast	\textbf{S}core	\\		
\textbf{ML}	&	\textbf{M}achine	\textbf{L}earning	\\			
\textbf{MSE}	&	\textbf{M}ean	\textbf{S}quared	\textbf{E}rror	\\		
\textbf{NLP}	&	\textbf{N}atural	\textbf{L}anguage	\textbf{P}rocessing	\\		
\textbf{NST}	&	\textbf{N}eural	\textbf{S}tyle	\textbf{T}ransfer	\\		
\textbf{PCA}	&	\textbf{P}rincipal	\textbf{C}omponent	\textbf{A}nalysis	\\		
\textbf{PM}	&	\textbf{P}roject	\textbf{M}anager	\\			
\textbf{RGB}	&	\textbf{R}ed	\textbf{G}reen	\textbf{B}lue	(color	model)	\\
\textbf{ROC}	&	\textbf{R}eceiver	\textbf{O}perating	\textbf{C}haracteristic	\\		
\textbf{ROI}	&	\textbf{R}egion	\textbf{O}f	\textbf{I}nterest	\\		
\textbf{SHAP}	&	\textbf{Sh}apley	\textbf{A}dditive	ex\textbf{p}lanation	\\		
\textbf{SpRay}	&	\textbf{S}pectral	\textbf{R}elevance	\textbf{A}nalysis	\\		
\textbf{ST}	&	\textbf{S}tyle	\textbf{T}ransfer	\\			
\textbf{STDA}	&	\textbf{S}tyle	\textbf{T}ransfer	\textbf{D}ata	\textbf{A}ugmentation	\\	
\textbf{t-SNE}	&	\textbf{t}-distributed	\textbf{S}tochastic	\textbf{N}eighbor	\textbf{E}mbedding	\\	
\textbf{TDA}	&	\textbf{T}argeted	\textbf{D}ata	\textbf{A}ugmentation	\\		
\textbf{VGG}	&	\textbf{V}isual	\textbf{G}eometry	\textbf{G}roup	(convolutional	model)	\\
\textbf{VR}	&	\textbf{V}irtual	\textbf{R}eality	\\			
\textbf{XAI}	&	E\textbf{x}plainable	\textbf{A}rtificial	\textbf{I}ntelligence	\\

}




\mainmatter	  
\pagestyle{fancy}  
\clearpage  

\lhead{\emph{Chapter 1: Introduction}}  
\chapter{Introduction}
In recent years, deep neural networks have achieved state-of-the-art performance in almost every suitable task. The amount, nature, and quality of datasets used to train neural networks have, without a doubt, a considerable impact on the model's performance. However, even the widely accepted benchmarks are sometimes noisy and contain inconsistent or incorrectly labeled samples. In addition, it is not uncommon to unintentionally introduce bias to data during the data collection, preparation, or analysis.

Those fragile black-box models often successfully hide their drawbacks behind high-performance metrics. High, but not necessarily valid and trustworthy. Sometimes it is easy to spot the problem, but biases are usually hidden well. Research proves that most models often treat bias as an essential feature and amplify it. 
Those observations raise an important question: should we blindly trust the machine learning systems based only on performance metrics? Those metrics are often calculated based on the test set coming from the same distribution. 
To thoroughly review the developed system, we should, at first, look back at the data used to train the models. 
Is the data adequately annotated, prepossessed and unbiased? If we refer to the machine learning model as the "engine" of the ML systems, the data would be the fuel that powers it. It is necessary to have good-quality data to run models: how can we expect the engine to work well with contaminated fuel? Similarly, we cannot expect 90\% accuracy when 15\% of it is incorrectly labeled. Or if the data we use is inconsistent or biased. 

Those errors and biases might lead to improper or unexpected reasoning. For instance, seemingly insignificant elements correlated with the class labels will probably affect the training process. The spurious correlation does not reveal the real cause of the problem and might exist only in the data itself, e.g., how it was collected or preprocessed. Such a case was observed during my experiments on classifying skin lesions into benign and malignant. 
The dermatologists support their diagnosis by carefully analyzing skin lesions with broad dermoscopic methods, analyzing their shape, colors, structure, and symmetry. The deep neural models get relevant features from the imperfect training data.

In the thesis, I discovered that image and camera artifacts strongly correlate with the skin lesion type in the dermoscopic datasets. Most artifacts like gel bubbles or ruler marks were easy to detect, which led the model to confuse the correlation with causation, causing a high accuracy effect. Although I still noted high results, those were not valid, as the dermatologists would never consider an artifact an essential premise for lesion diagnosis. This example is one of the many discovered in the past few years. 

Hence, there is a strong need to develop methods for further data examination and bias discovery. One of the ML gurus (Andrew Ng) said that ML Engineers should have a more data-centric approach to developing systems, including careful data inspection and examination. Focusing on architectures, models, and different regularization methods is necessary, as much as cautious and responsible data preparation.
The demand for clear reasoning and correct decisions is very high, primarily when deep black box systems are used in autonomous transportation, healthcare, legal systems, finances, and the military. Even the General Data Protection Regulation, a regulation in EU law on data protection and privacy for all individuals within the European Union, states that every person has the right to an explanation. Hence, the models should provide explanations of their predictions. Deep neural networks (DNNs) usually have tens of layers with millions of parameters, and very complex latent space, which make their decisions very hard to verify. Hence, how to interpret if the model uses correct features for prediction? For instance, if we train the model to differentiate between dogs and wolfs, how do we know if the model focuses on the animal's appearance rather than the background or other variables (like a collar on a dog's neck)? Or if we train the model to detect malignant tumors in lungs, but it learns to detect pen marking made by a radiologist instead? One of the approaches that allow interpreting deep model's decisions is using, i.e., local explainability methods, which aim to explain a single prediction.

In computer vision, local explanations are usually done with techniques highlighting input parts that significantly impact the prediction. Analyzing predictions only with local explanations is still a time-consuming and tiring task. Moreover, even globally-aware local explanations might overlook important and hard-to-notice large-scale patterns.
Hence, the thesis focus on the less developed category of global explanations. Global explanations are vital for discovering abnormalities in the model and comparing different models and even different datasets. 

Discovering the root of this problem is only the first of many steps to designing more robust and trustful systems. However, even if we know that the bias exists, we should ask ourselves another question: what exactly is the bias source, and how to eliminate or at least mitigate it? 
The thesis attempts to answer those questions twofold: first, by developing and providing methods to help find the root of the bias in data and models. Second, by improving the training process, the bias becomes irrelevant and, as a result -- ignored. The dissertation aims to propose techniques that make DNN models more explainable and trustworthy to address those challenges. 

Hence, I defined the hypothesis is as follows:

\begin{itemize}
    \item \textit{Identification and further mitigation of biases in data with explainable AI and data augmentation increase the model's accuracy and robustness against biases.}
    
\end{itemize}
One can distinguish two aspects of this hypothesis: 1: bias discovery (identification) and 2: bias mitigation.
The thesis structure reflects it.

In the first part of the dissertation, I aimed to categorize and describe potential biases and errors in data and models. Bias is relatively common in all research areas, including machine learning, as both data and ML pipelines are designed and prepared by biased humans. 
Next, I presented the taxonomy and methods of Explainable AI (XAI). I showed that XAI could be successfully used to justify predictions and to control and improve the model. 
I demonstrated a long and tiring process of manual data inspection and bias discovery in the example of the skin lesion dataset. Finally, I showed an alternative approach to manual data exploration: \textit{Global Explanations for Bias Identification} method, designed to semi-automatically discover potential biases in data. Alongside, I presented relevant numerical methods and metrics to asses whether found bias affected the model.

However, bias discovery is just the tip of the iceberg. 
Whereas identifying errors and bias is critical, improving the model and reducing the number of flaws in the future is an absolute priority. 
The next huge problem is how to mitigate the influence of discovered biases to make the models as expected.

Hence, in Part II, I aimed to reduce the bias's influence on the model as much as possible.  
Some approaches discussed data cleaning, for instance, by selectively not including biasing factors in training data, like removing information about \textit{gender} from metadata. Or by advanced preprocessing of images (i.e., erasing artifacts from the images), by using segmentation masks in the Region of Interest (ROI), or by generating new samples in the dataset with GANs and Style Transfer \cite{mikolajczyk2018data}. But even if we omit a conflicting variable, other variables might still be infected by it. Many experiments showed that it is tough to ignore bias completely. In the thesis, I introduced and tested a method of bias mitigation. First, I proposed \textit{Style Transfer Data Augmentation} to deal with the shape and texture bias by merging a  style of a malignant lesion with a conflicting shape of a benign one.

However, the real goal should not be to delete all biases in the data (or erase them in the case of images) but to ensure that the predictor can learn to ignore them. Researchers started to develop methods that tackle the problem differently in the last few years. They began to incorporate human expert knowledge into the training process to avoid artifactual data's negative influence. 
Here, the problem is approached twofold. First, I proposed to use \textit{Targeted Data Augmentations} -- randomly inserting possible biases into all images in the dataset. This way, bias in data might be reduced without affecting the outcome. 
Second, I proposed to use \textit{Attribution Feedback} to fine-tune the model to improve it by eliminating obvious mistakes and teaching it to ignore insignificant input parts. The attribution feedback is closely related to relevance guidance, which by highlighting relevant regions and suppressing unimportant ones, is enabling a better classification.


The first part of the thesis is organized as follows: in the \textit{Chapter \ref{chapter.bias_in_ML}: Bias in machine learning pipeline} I presented a comprehensive review and analysis of common data, model, and reasoning biases found in machine learning projects. Next, in \textit{Chapter \ref{chapter.xai}: Explainable Artificial Intelligence}, I described, analyzed, and presented the taxonomy behind explainable methods and models. In \textit{Chapter \ref{chapter.skin_lesion_bias}: Identifying bias with manual data inspection} I designed a methodology, described and followed it, and analyzed the results of manual data exploration towards bias identification. In the last chapter of Part I, \textit{Identifying bias with global explanations} (\ref{chapter.gebi}), I designed a methodology, proposed and conducted extensive experiments, and tested the results of a semi-automatic method for bias discovery using explainable AI. 

The second part of the thesis starts with the \textit{Chapter \ref{chapter.neural-style}: Mitigating shape and texture bias with data augmentation}, where I proposed, extensively tested, and analyzed the results of an effective data augmentation method for reducing shape and texture bias and increasing overall robustness. Next, in \textit{Chapter \ref{chapter.targeted}: Bias-targeted data augmentation}, I designed, developed a pipeline, and conducted advanced experiments on reducing the influence of several types of biases in data on models. Finally, I finished with the presentation of a novel method, supported by numerous experiments and analytical commentary in \textit{Chapter \ref{chapter.attribution}: Debiasing effect of training with attribution feedback} that is an alternative approach for bias mitigation and increasing robustness.

The following publications support the results:
\begin{enumerate}
\item Mikołajczyk, A., Majchrowska, S., and Carrasco-Limeros, S. \textit{The (de)biasing effect of GAN-based augmentation methods on skin lesion images}. International Conference on Medical Image Computing and Computer-Assisted Intervention (MICCAI). Springer, 2022

\item Mikołajczyk, A., Grochowski, M., and Kwasigroch, A. \textit{Towards explainable classifiers using the counterfactual approach-global explanations for discovering bias in data}. Journal of Artificial Intelligence and Soft Computing Research, 11(1), p. 51--67, 2021

\item Klawikowska, Z., Mikołajczyk, A., and Grochowski, M. \textit{Explainable AI for Inspecting Adversarial Attacks on Deep Neural Networks}. In International Conference on Artificial Intelligence and Soft Computing, p. 134--146, 2020

\item Mikołajczyk, A., and Grochowski, M. \textit{Style transfer-based image synthesis as an efficient regularization technique in deep learning}. In 24th International Conference on Methods and Models in Automation and Robotics, arXiv:1905.10974, 2019

\item Grochowski, M., Mikołajczyk, A., and Kwasigroch, A. \textit{Diagnosis of malignant melanoma by neural network ensemble-based system utilising hand-crafted skin lesion features}. Metrology and Measurement Systems, 26(1), p. 65--80, 2019

\item Mikołajczyk, A., and Grochowski, M. \textit{Data augmentation for improving deep learning in image classification problem}. In 2018 International Interdisciplinary PhD Workshop (IIPhDW), p. 117--122, 2018

\item Kwasigroch, A., Mikołajczyk, A., and Grochowski, M. \textit{Deep convolutional neural networks as a decision support tool in medical problems–malignant melanoma case study}. In Polish Control Conference, p. 848--856, 2017

\item Mikołajczyk, A., Kwasigroch, A., and Grochowski, M. \textit{Intelligent system supporting diagnosis of malignant melanoma}. In Polish Control Conference, p. 828--837, 2017
 \end{enumerate} 

\clearpage  
\part{XAI for discovering bias in data and models}

\lhead{\emph{Chapter 2: Bias in machine learning pipeline}}  

\chapter{Bias in machine learning pipeline} \label{chapter.bias_in_ML}

Bias can be broadly defined as \textit{a systematic deviation of results or inferences from the truth or processes leading to such deviation} \cite{k2014bias}. In machine learning, bias is often referred to as \textit{a systematic error from erroneous assumptions in the learning algorithm} \cite{mehrabi2021survey}. Unfortunately, we can find it in all areas of the research. Bias can be introduced to the research project at every possible stage, including the beginning, like literature review or data collection, the middle with monitoring and reacting to the progress, and the end: evaluation and closure of a research project \cite{k2014bias}. Avoiding bias demands full awareness and constant vigilance of all project members. And even then, there is still a place for making errors. To detect and mitigate bias, first, we need to resolve definitional ambiguities and outline the scope of the topic.

Hence, in this chapter, I presented an integrated, synthesized overview of the current state of knowledge regarding bias in the machine learning pipeline, including the review of biases at different stages of research and contemporary approaches to bias detection and mitigation.

\section{Bias in different stages of research}
\begin{figure}[!htb] 
\centering 
  \includegraphics[width=0.8\textwidth]{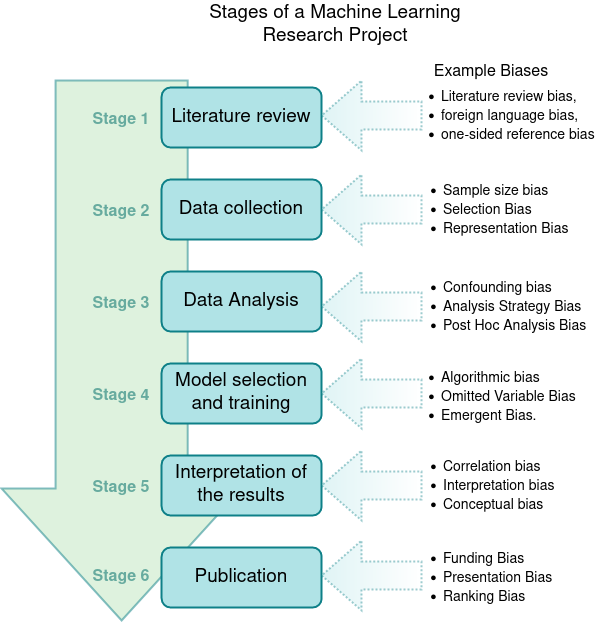}%
  \caption{Stages of machine learning project and possible biases that could be injected} \label{Fig.stagesofbias}
\end{figure}

In this chapter, I briefly introduced the reader to over forty bias types in various stages in the typical machine learning pipeline. The stages are presented visually in Figure~\ref{Fig.stagesofbias}. 

I distinguished six stages in a machine learning research project:
\begin{itemize}
    \item \textit{Literature review. }A literature review is a thorough inspection and analysis of the previously published works on a selected topic. In ML, this includes reviewing scholarly papers, open-source repositories, existing projects, and programming libraries.
    \item \textit{Data collection }is defined as gathering or measuring targeted variables in an established system, enabling one to answer relevant questions and evaluate outcomes.
    \item \textit{Data analysis. }Data analysis and exploration are essential steps of any research project. This stage is defined as inspecting, cleaning, preparing, and exploring data to discover meaningful information about the investigated process or relation. This stage often includes an exploratory data analysis to summarize the main characteristics of data collections, often using statistical graphics and other data visualization methods. Data analysis shapes the future directions of the next stage: model selection and training.
    \item \textit{Model selection and training.} This step includes choosing the appropriate model for the existing data and problem, designing the training pipeline, selecting the input features and hyper-parameters, finding the model's parameters by training, and validating it to achieve planned efficiency. The process heavily depends on the knowledge gathered from the previous data analysis stage.
    \item \textit{Interpretation of the results.} This stage covers analyzing the results of the experiments, including comparing with state-of-the-art or ground-truth, interpreting the results, and forming conclusions about experiments. 
    \item \textit{Publication.} A publication is the last stage that essentially focuses on publishing the experiments' results, sharing the code, presenting the methodology, and summarizing the project. 
\end{itemize}

\subsection{Literature Review}
In some cases, the methodology of reviewing the state-of-the-art can also induce bias. Although this stage is not as important as others in the process of designing the model, it can sometimes significantly impact the final study results. \textit{Literature review bias}, also called \textit{reading-up bias}, is defined as errors in reading-up on the field \cite{aleu2020assessing,k2014bias}.  

A commonly-known example is a \textit{literature search bias}, defined as an incomplete search due to poor keywords or search strategies or failure to include unpublished reports and hard-to-reach journals \cite{k2014bias}. Including this bias might lead to repeating failed experiments or addressing the problems that have been well-defined and researched in the past.
Sometimes, \textit{foreign language exclusion bias} \cite{k2014bias} happens when literature search is restricted to a single language and publications in foreign languages are ignored. This exclusion might result in a significant bias in selection \cite{dubben2005systematic}.
Some publications can be lost on the web because of both \textit{halo and horn effects}. For instance, poorly written or badly structured papers leave an impression of low-quality research and decrease the trust towards achieving results (horn effect). Even with outstanding achievements, such reports might remain unnoticed, whereas some average articles might be over-glorified due to the great impression of the journal or authorship (halo effect). 

In the machine learning community, a \textit{literature review bias} might be noticed during the review of the state-of-the-art models: those might be biased, for instance, by using different dataset split or training data shuffle. Some may report false results by testing models' performance on the validation set instead of the separate test set or testing too little data. 

Another type of literature review bias is \textit{one-sided reference bias} which happens when researchers restrict their references to only those studies that support their position \cite{gotzsche1987reference}. Some researchers might unintentionally induct a \textit{rhetoric bias} when they try to convince the reader without any scientific fact, or reason \cite{k2014bias}. Those two are closely related to what is often discussed in psychology: \textit{confirmation bias}. Confirmation bias is defined as \textit{the tendency to interpret new evidence as confirmation of one's existing beliefs or theories}\footnote{definition from Google's English dictionary provided by Oxford Languages}. This bias means that literature review or any other information searched for, interpreted, or analyzed, is systematically favored by the biased researcher towards their position or hypothesis \cite{oswald2012confirmation}. It is not strictly related to machine learning and can be found at any research stage.

\subsection{Data Collection}

Dataset has a significant impact on the final results in any data-driven model. In 1957 in the article \textit{Work with new electronic 'brains' opens field for army math experts} \cite{mellin1957work}, the phrase \textit{garbage in, garbage out} was used for the first time, referring to the software development process. Until today, it made a famous saying inside the ML community that explains that even if the model is well and correctly prepared, we will not get satisfying results if the data used is low-quality, incorrect, biased, or noisy.

Usually, researchers developing ML methods tend to use widely accepted benchmark datasets. However, surprisingly, even widely used benchmarks have been proved to be biased and noisy, so most of the algorithms are trained and tested on the same limited datasets. It might seem as not a big problem at first. Still, state-of-the-art models are often copy-pasted into real-world applications, where bias in data is crucial for successfully implementing the project into real life. 

In real-world projects, data preparation is usually very long and expensive. During the procedure, one might accidentally insert unwanted bias into the dataset. This chapter highlights the three substages in the dataset preparation stage: design of data acquisition, execution, and collection. 

Dataset may contain a bias that will impact the models' performance. First, the general design of the data acquisition process might influence data quality. A standardized protocol for acquiring and gathering data is the first step in minimizing potential biases influence. On the one hand, there is no standardized guideline for collecting data in machine learning. On the other hand, the enormous increase in available datasets defined some standards required for top ML conferences and journals. However, many commonly used datasets are still biased due to how the data was collected. In many cases, the bias was introduced in the first step of data collection: the design of the data collection process.  

One of the most common problems is a sample size referred to later as dataset size. It is well known that deep learning algorithms require vast amounts of data to be effectively trained. A small dataset makes training more challenging and makes the proper data randomization process harder. This problem is called a \textit{sample size bias}.  

The problematic design is not a rare case in ML datasets. It can often be found in crowdsourced datasets \cite{zheng2017truth} or even medical datasets \cite{gregory2012research}. Data might be accidentally collected in such a way that it introduces a \textit{selection bias}. Selection bias is defined as a deviation of data from the truth resulting from how samples were collected. It can arise when a) the sampling frame is incomplete or inaccurate, b) the sampling process was nonrandom, or c) some targets were excluded from data collection. 
For instance, the Open Litter Map \cite{lynch2018openlittermap}, currently the most extensive dataset of images with litter, introduces a strong \textit{sampling bias}, sometimes called \textit{sampling frame bias} and \textit{representation bias} \cite{mehrabi2019survey}.

\textit{Sampling Bias} (Representation Bias) is a bias in which data is acquired in such a way that not all samples have the same sampling probability, i.e., not all samples are equally likely to be selected in the study \cite{mehrabi2019survey}. In the case of an open litter map \cite{lynch2018openlittermap}, the main goal was gathering an immense amount of data possible instead of focusing on the quality. Waste images were collected and annotated by anonymous users that wanted to help the environment. However, because the website and application were first only available in English, most users were from the UK, and some were from the USA. The unnoticed problem of sampling bias in the dataset might result in the wrong assumption about the phenomenon of the study. For instance, one might assume "there is more litter in the UK than in India," whereas the real reason is that people in the UK upload more images than people in India. Such lacking of geographical diversity is a widespread problem in many datasets that are used all over the world \cite{mehrabi2019survey}.

Another problem of bias in data is a \textit{coverage error}. It can be observed when the sampling frame is faulty, i.e., a  correspondence between the target population and the one from which a sample is drawn is not one-to-one \cite{wolter1986some}. There are two main types of coverage error: under-coverage (non-coverage) and up-coverage. A \textit{non-coverage bias} can be found when some data samples are impossible or very hard to find \cite{k2014bias}. \textit{Up-coverage} happens when the same data sample is falsely recognized as two distinctive samples. For example, imagine we gather patients' data from a selected zone to examine whether they might or not be infected with a virus. If a single patient were registered twice by mistake, this would cause an up-coverage error. If some patients were accidentally missed in the registration, it would be an under-coverage error. Because of ML characteristics, coverage bias is quite vague and hard to define in machine learning. It seems to be more common in surveys or tabular data, i.e., response bias or illegal immigrant bias.

\textit{Nonrandom bias} exists when the selection process is affected by the human choice, e.g., when sampling is nonrandom \cite{mclachlan1984method,k2014bias}. It can occur, for instance, when gathering medical data but only in a few selected hospitals. Such a case happened in some skin lesion datasets \cite{mikolajczyk2021towards}. The pediatric hospital collected thousands of images with the same dermoscopic device. Most of them were benign – because children rarely develop malignant tumors \cite{hamm2011skin}. A great example of a nonrandom bias was mentioned in the book \textit{How to lie with statistics} by Huff Darrell \cite{huff1993lie}. It shows an example of sending a questionnaire about loving surveys and gathering the answers of those who responded. This survey collection method developed a nonrandom bias, as people who like responding to surveys are more likely to fill them than those who hate that. Instead, the researcher should gather several people that represent the population well and force them to complete the survey.

\begin{figure*}[!htb]
\centering
\begin{subfigure}[A dermoscopic image with inserted black frame]{
  \includegraphics[width=0.25\linewidth]{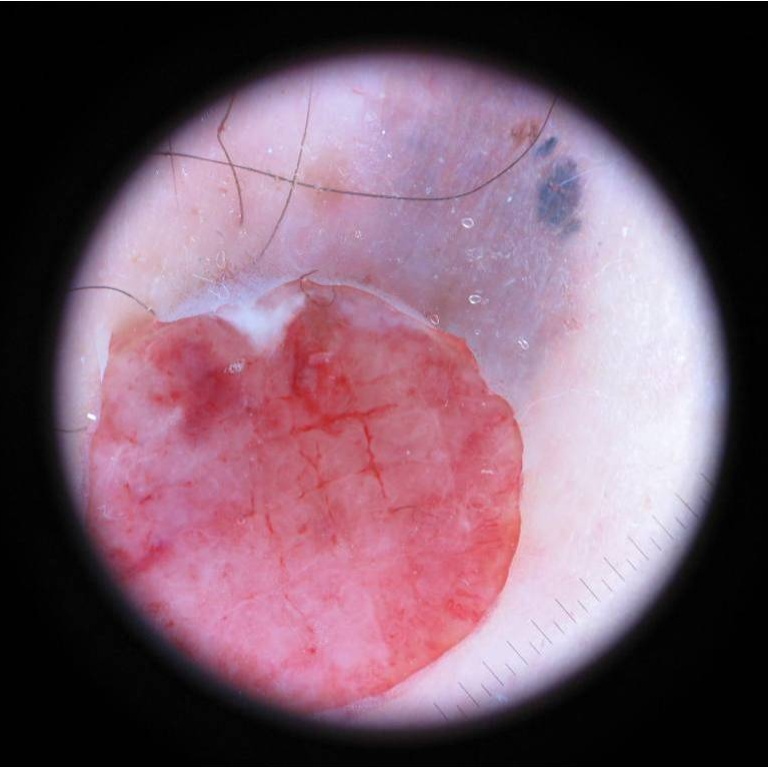}
  \includegraphics[width=0.25\linewidth]{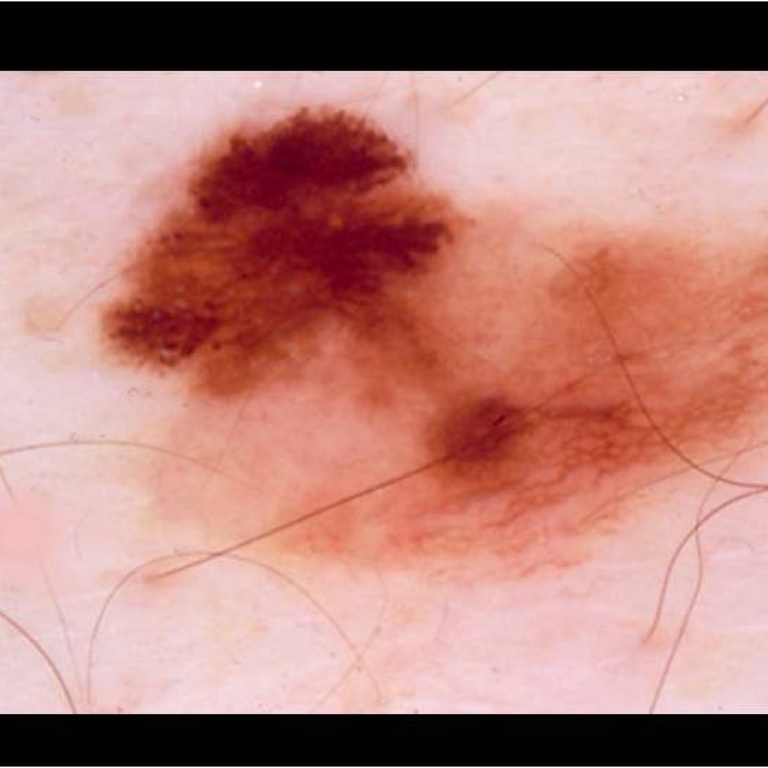}
  \includegraphics[width=0.25\linewidth]{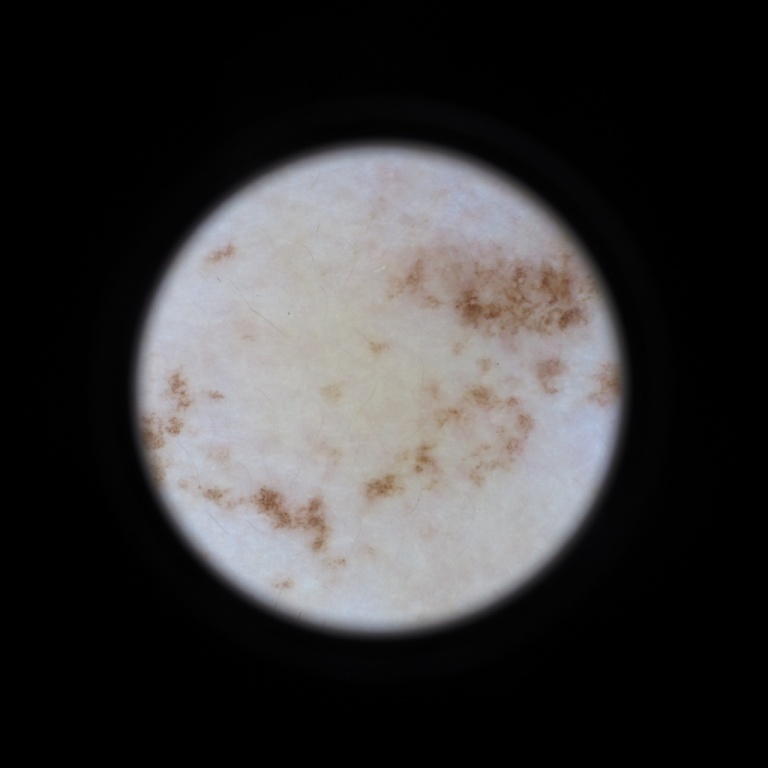}}
 \end{subfigure}
 \begin{subfigure}[Dermatoscopy of Acanthosis nigricans over the neck. Differences between common dermatoscopes (source: \cite{nirmal2017dermatoscopy})]{
  \includegraphics[width=0.75\linewidth]{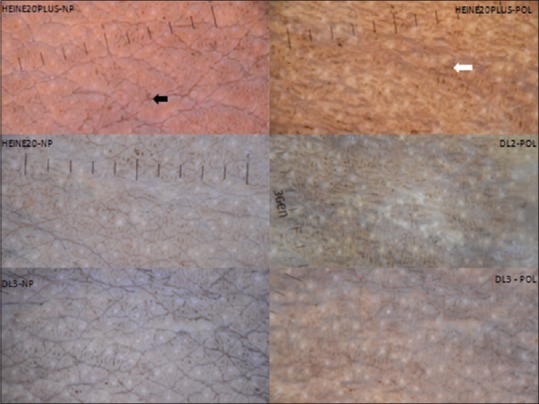}}
  \end{subfigure}
  \caption{Examples of instrument bias in dermoscopy presenting dermoscopic images.}
  \label{fig.bias.frames}
\end{figure*}

\textit{Instrument bias}. This type of bias results from imperfections in the instrument or method used to collect or manage the data \cite{he2012bias}. Devices used to collect the data can strongly affect the learning algorithm. Nirmal et al. analyzed the typical differences in dermatoscopes: medical aperture used to take images of skin lesions \cite{nirmal2017dermatoscopy}. They show, for instance, that some dermatoscopes \textit{show multiple cristae and sulci clearly, whereas polarized imaging allows better visualization of hyperpigmented dots and streaks} \cite{nirmal2017dermatoscopy}, as presented in Figure~\ref{fig.bias.frames}. Also, using a poor-quality dermatoscope that would miss specific visible structures of a skin lesion with a malignant characteristic, or alter the white balance, would result in the instrument bias. Such a poor device might hinder the classification of the lesion into malignant and benign by hiding valuable and significant information about the lesion. Moreover, a certain dermatoscope used to collect skin lesion images could modify the actual skin lesion by adding a round or rectangular black frame. The mentioned examples of the instrument bias are presented in Figure~\ref{fig.bias.frames}. 

In some cases, even the time of data collection is essential. The two examples are \textit{popularity bias} and \textit{temporal bias}. Temporal bias is defined as "systematic distortions across user populations or behaviors over time. "\cite{olteanu2019social}. The popularity bias comes from increased public interest in a subject of research \cite{ciampaglia2018algorithmic}. Recommendation systems \cite{abdollahpouri2019unfairness} have a common problem: popular
topics (items, movies, books) are recommended more often, whereas fewer popular are recommended less frequently or never \cite{abdollahpouri2019managing}. As a result, the popular items became even more popular, and niche items got lost in the sea of propositions.

Next, there is an \textit{observer bias}, sometimes called a \textit{research bias} or an \textit{experimenter bias}. It owes the name to its definition: it tends to observe what the observer wants to see \cite{mahtani2018catalogue}. A famous example of observer bias is research on the heritability of IQ conducted by the English educational psychologist Cyril Burt \cite{lamb1992biased,fletcher1991science, jensen1980bias}. During his studies, he truly believed that children with a higher socioeconomic status were more intelligent on average. His research led to creating a two-tier educational system in 1960s England, which sent middle- and upper-class children to elite schools. In contrast, working-class children were sent to less desirable schools \cite{burt1943ability}. Currently, he is well-known as a researcher who falsified his work \cite{lamb1992biased,fletcher1991science, jensen1980bias}. 

In data-driven systems, observer bias might appear when annotators use personal, subjective opinions to label data, resulting in incorrect annotations. Depending on the annotated data, it might be tough to differentiate emotional thoughts from objective observations. An example might be a sentiment analysis, where annotators must decide if the sentence (written or spoken) has a negative, neutral or positive meaning \cite{kiritchenko2018examining}. In some cases, the annotation process is even more advanced: e.g., in emotion recognition, annotators have to divide spoken conversation into seven different emotions: \textit{neutral, sad, angry, happy, surprised, fear, and disgust} \cite{koolagudi2012emotion}. Even in cases where annotators do not have to tag emotions, they might still let their habits into the procedure.
An example might be adding punctuation to the spoken text, which is particularly useful in punctuation restoration tasks \cite{yi2020focal, moro2017prosody}. The exact page of text is often tagged differently by annotators \cite{bohavc2017text} even when following the guidelines. Some people have preferences to stay with longer, complex sentences, whereas others prefer to keep them short \cite{bohavc2017text}. When an annotator uses his prejudice to label that, this sub-type of subjective bias is called an \textit{annotator bias} \cite{hellstrom2020bias}. This problem is usually handled with additional coefficients to measure the agreement between different annotators, i.e., Cohen's Kappa score \cite{artstein2017inter}.

However, the observer is not the only one who can add bias to the data. Another type of bias is \textit{observee bias}, widely known as \textit{subject bias}. It refers to the inaccurate data provided by the subjects. Choi et al. recognized multiple types of such bias, including subjects' preferences that give wrong information intentionally or unintentionally. It might alter the data whenever the obervee is the primary provider of the data (interviews, questionnaires, reports, etc.) \cite{k2014bias}. 

Collected data might also be biased by some inequities pre-existing in our society, like stereotypes or historically disadvantaged groups. Those are sometimes called \textit{cultural biases}, \textit{social biases} or \textit{stereotypical bias}, and are mostly found in various text corpora \cite{kaneko2022gender}. Nangia et al. define nine types of stereotypical bias: race, gender, socioeconomic status (occupation),  nationality, religion, age, sexual orientation, physical appearance, and disability bias \cite{nangia-etal-2020-crows}. According to Nangia et al., a sentence is stereotypical when an advantaged group (e.g., a high socioeconomic status) is associated with a pleasant attribute (e.g., \textit{People who live in a mansion are smart.}) or a disadvantaged group with an unpleasant adjective (e.g., \textit{People who live in trailer parks are careless.}).

Next, a bias connected to the data acquisition process includes \textit{data source bias} \cite{k2014bias}, including competing death bias, family history bias, and spatial bias \cite{k2014bias}. Those biases are primarily reported in medical papers. Since ML is often used to support work in hospitals and medical centers, it should also be considered. Another bias in decision support systems is \textit{automation bias}, which is defined as the tendency to over-trust them. This problem is often reported in medical decision support systems, where clinicians rely on the software too much and overlook contradictory information \cite{goddard2012automation}. Such data, if recorded and used to build new data collections, will affect new models.

\begin{figure}[!htb]
\centering
  \includegraphics[width=0.5\textwidth]{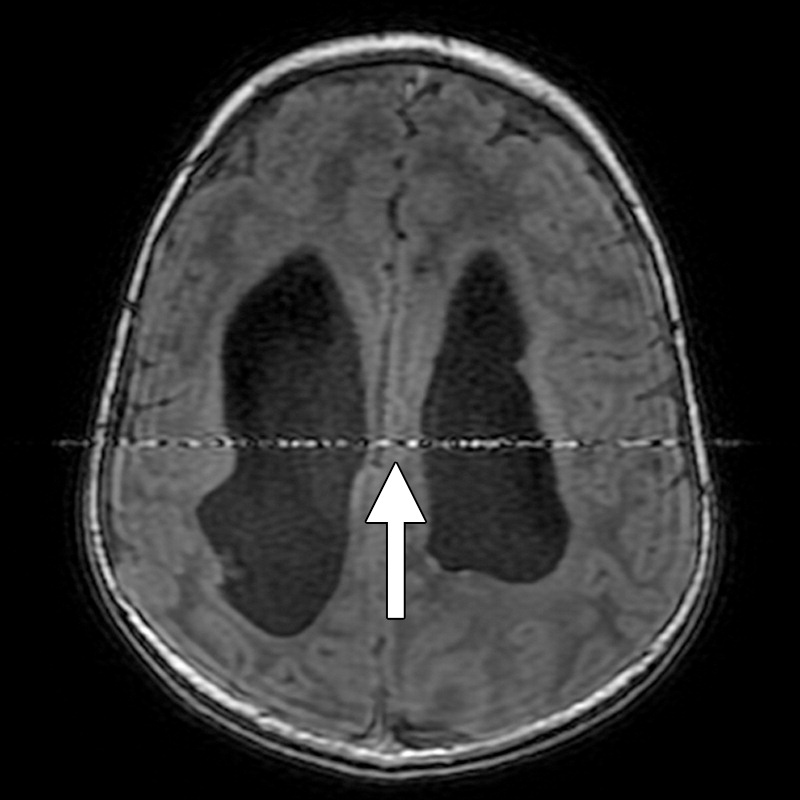}%
  \caption{Data handling bias is introduced by improper device calibration. MRI example -- zipper artifact (image source \cite{yanasak2014mr})} \label{fig.mri}
\end{figure}

Finally, we have the \textit{data handling bias}. This bias describes how data is handled, which sometimes might distort the output. For instance, scanning the medical images to move them from analog format to digital might add unwanted artifacts. Nathan E. Yanasak et al. presented several domain-specific artifacts that might appear when improperly calibrating parallel magnetic resonance (MR) imaging \cite{yanasak2014mr}. As illustrated in Figure~\ref{fig.mri}, they include chemical shifts, zippers, ghosting, and others. The ML algorithm might wrongly consider such artifacts as an essential feature. 

\subsection{Data Analysis} \label{section.data_analysis}

Even when the dataset acquisition process is well-thought and well-organized, data is collected carefully, following the guidelines, the bias might still be in Stage 3: Data analysis. Analysis bias is defined as the result of errors in data analysis.  

One of the most mentioned problems in the data analysis stage is a \textit{Confounding bias}. Confounding has been studied since the early '70s by epidemiologists, statisticians, doctors, and mathematicians \cite{mccroskey1966ethos,axelson1978aspects,greenland1980control}. Epidemiologists define a confounder as a pre-exposure variable associated with exposure and the outcome conditional on the exposure, possibly dependent on other covariates \cite{miettinen1974confounding}. 
In statistics, a \textit{confounder} (also known as a confounding variable, confounding factor, extraneous determinant, or lurking variable) is a variable that influences both the dependent variable (i.e., disease) and independent variable (the studied factor), causing a spurious association \cite{pearl2009causal,vanderweele2013definition}. 

To better understand a confounding factor, consider studying the relationship between drinking coffee daily and having heart problems \cite{tulchinsky2014measuring}. It might look like coffee causes heart problems because coffee drinkers statistically have more cardiovascular diseases. However,  coffee drinkers smoke more cigarettes than non-coffee drinkers. We might notice that smoking is a\textit{ confounding variable} in the study of the association between coffee drinking and heart disease. A higher probability of heart disease might be due to smoking rather than coffee drinking. More recent studies have shown coffee drinking to have substantial benefits in heart health and the prevention of dementia \cite{bidel2013emerging}.

\begin{figure}[!htb]
\centering
  \includegraphics[width=0.5\textwidth]{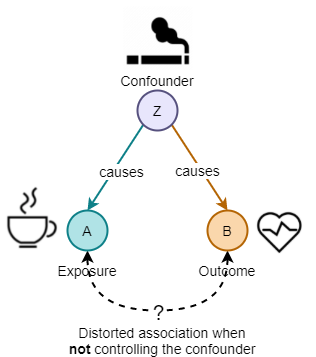}%
  \caption{Confounding bias. Smoking is an example of distorted association when studying the relationship between drinking coffee daily and having  heart problems when not controlling confounding factors.} \label{fig.confounder}
\end{figure}

According to \cite{mcnamee2003confounding} confounding bias, a systematic error can occur in epidemiological studies in measuring the association between exposure and the health outcome caused by mixing the exposure of primary interest with extraneous risk factors.

It was said that unlike selection or information bias \cite{althubaiti2016information}, confounding is one type of bias that can be adjusted after data gathering using statistical models \cite{pourhoseingholi2012control}. However, to decide whether a variable is working independently, a biological or social mechanism must cause exposure to the disease or health outcome \cite{alexander2015confounding}.

A confounding bias is a widely recognized problem in social sciences and causal modeling. In ML, it receives much less attention \cite{landeiro2017controlling}. However, the problem still exists and might be easily omitted due to the high dimensionality of the current issues solved by deep learning. 

\begin{figure}[!htb]
\centering
  \includegraphics[width=0.5\textwidth]{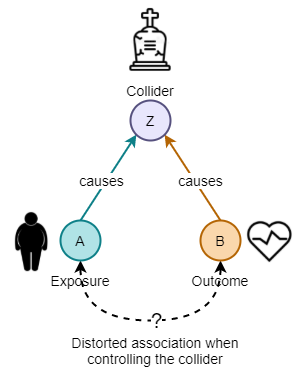}%
  \caption{The collider bias. Example of distorted association when studying the relationship between obesity and having  heart problems when controlling collider factor -- mortality rate.} \label{fig.collider}
\end{figure}

Next to the confounding bias stands \textit{collider bias}, also known as \textit{collider-stratification bias} \cite{10.2307/3703850} or \textit{reversal paradox} \cite{10.1093/aje/kwi002}). Collider bias is a causally influenced association between two or more exposures when a shared outcome (collider) is included in the model as a covariate \cite{day2016robust}. The main difference between confounding and collider bias is that confounders should be controlled when estimating causal associations, whereas colliders are not, as presented in Figure~\ref{fig.collider}). An exciting example of the collider is an obesity paradox \cite{viallon2016can}. An obesity paradox says that people with cardiovascular diseases and obesity have lower mortality rates than those without obesity. In a sample with only people with cardiovascular diseases, such observation creates a distorted association of the preventive effect of obesity on mortality. It is well known that obesity increases mortality rates \cite{viallon2016can}. In that case, the mortality rate is a collider - it is affected not only by obesity but also by other unmeasured factors.

Other frequently mentioned types of bias are \textit{analysis strategy} and \textit{post hoc analysis} biases.

\begin{figure}
\centering
  \includegraphics[width=0.8\textwidth]{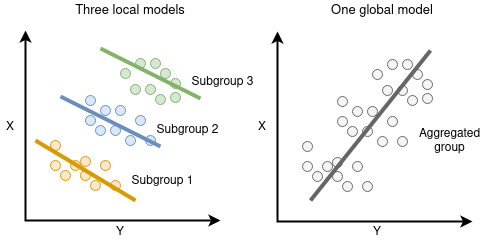}%
  \caption{The Simpson's Paradox - when the relationship between two variables differs within subgroups and its aggregation \cite{mehrabi2019survey}} \label{fig.simpsons}
\end{figure}

The \textit{Reversal paradox }(sometimes called amalgamation paradox) happens when the association between two (or more) variables can be reversed when another variable is statistically controlled for \cite{messick1981reversal}. The most known subtype of the reversal paradox is \textit{Simpson's Paradox} (Yule-Simpson effect). Simpson's paradox can be observed when the relationship between two variables differs within subgroups, and their aggregation \cite{tu2008simpson}. The Simpson's paradox is presented in Figure \ref{fig.simpsons}: the relationship between two variables  -- $X$ and $Y$ illustrated on axes --  is different for subgroups (higher $X$ means lower $Y$) and its aggregation (higher $X$ means higher $Y$).

\subsection{Model selection and training}

One of the most important steps in a research project is selecting a proper model for the corresponding problem. Even when the data is free from bias, the final predictions still might be biased. When the model is the source of bias, it is called an \textit{Algorithmic Bias} \cite{baeza2018bias}. Some sources also define an algorithmic bias as amplifying and adversely impacting existing inequities in an algorithm, e.g., socioeconomic status, race, ethnic background, religion, gender, disability, or sexual orientation \cite{panch2019artificial}. The general design based on feedback loops is also criticized widely, as researchers say that the self-reinforcing feedback might amplify the inequities \cite{nisan2001algorithmic,lum2016statistical, edelman2017racial}. The problem of bias amplification is often mentioned, e.g., in recommending engines \cite{lloyd2018bias}, word embeddings \cite{bolukbasi2016man}, or any other discriminate model \cite{mayson2018bias}. Let us assume we have a dataset for cat vs. dog classification. In the dataset, if an animal is sitting on the grass, it is a dog in 70\% of cases. However, after training the model, the predictions showed that 85\% of animals on the grass were classified as dogs, as the grass became an essential feature for the classifier, amplifying already existing bias. The problem was highlighted by Zhao et al. in the paper \textit{Men also like shopping: Reducing gender bias amplification using corpus-level constraints} \cite{zhao2017men}. A similar problem shows a significant \textit{gender bias} in commonly used benchmark datasets. The women were more often found in the kitchen, and as a result, introduced a strong gender bias in the algorithm \cite{zhao2017men}. Also, many NLP corpora have been proven to be influenced by gender bias \cite{kaneko2022gender}. Those and other biases might be amplified by algorithmic bias.

Some even call those biased models self-fulfilling prophecies \cite{cowgill2019economics}. Currently, the algorithmic bias is often discussed due to the increasing popularity of \textit{algorithmic fairness}, i.e., the concerns that algorithms may discriminate against certain groups \cite{edelman2017racial}. It is well known that algorithms (models) can inherit questionable values from data and acquire or amplify biases during the training \cite{wong2019democratizing}. However, some researchers believe that selecting an adequate model (or training procedure) will eliminate biased predictions \cite{cowgill2019economics}. The subject of "unfair algorithms" caught public attention: numerous people shared their stories on social media (e.g., Twitter) on how they were victims of algorithmic fairness. A real-life example of a discussion on the problematic algorithmic (un)fairness would be \cite{corbett2017algorithmic} a COMPAS -- Correctional Offender Management Profiling for Alternative Sanctions. COMPAS is a decision support tool used in the US to predict recidivism risk, i.e., a criminal defendant will re-offend. The reported problem with the device is that it gave significantly higher false-positive rates against black people \cite{juliaangwin_2016}. Such calculated risk strongly affected the judge's decision. It seemed that the algorithm more often classified black people as people at a high risk of committing a crime again - and made mistakes more often \cite{juliaangwin_2016}.

\begin{table}[!htb] \label{tab.propublica}
\caption{ProPublica's table (2016) reporting model errors at the study cut point (Low vs.  Not Low) for the General Recidivism Risk Scale \cite{juliaangwin_2016}}
\centering
\begin{tabular}{llll}
\hline
\textit{COMPAS Risk Prediction} & \textit{Reoffend} & \textit{White} & \textit{Black} \\ \hline
High Risk                      & No                & 23.5\%         & 44.9\%         \\
Low Risk                       & Yes               & 47.7\%         & 28.0\%         \\ \hline
\end{tabular}

\end{table}

That controversial article claimed that almost half of black people were mistakenly classified as high-risk re-offenders. In contrast, nearly half of white people were incorrectly classified as low risk, as presented in Table~\ref{tab.compas}. Later, another publication explained why those results were wrong and presented contradictory statistics showing properly calculated statistics \cite{dieterich2016compas}. This is a great example of \textit{Simpson's Paradox} -- to read more about this error, see Simpson's Paradox description in Subsection~\ref{section.data_analysis}: \textit{Data Analysis}.

\begin{table}[!htb]
\caption{Northpointe Inc. Research Department table (2016) reporting model errors at the study cut point (Low vs.  Not Low) for the General Recidivism Risk Scale \cite{dieterich2016compas}}\label{tab.compas}
\centering
\begin{tabular}{llll}
\hline
\textit{COMPAS Risk Prediction} & \textit{Reoffend} & \textit{White} & \textit{Black} \\ \hline
High Risk                      & No                & 41.0\%           & 37.0\%           \\
Low Risk                       & Yes               & 29.0\%           & 35.0\%           \\ \hline
\end{tabular}
\end{table}

Similar news and raising public awareness leads to higher demand for eliminating algorithmic bias and "fairer algorithms design." \cite{wong2019democratizing}
Hence, an \textit{algorithmic bias} might be defined as a bias that is amplified or introduced by the model.

A model can also inherit algorithmic bias. In ML, it is not uncommon to use more than one model to perform a task. A standard study in computer vision, detection is still often done as a two-stage process. First, the object of interest is localized on the image, and then it is classified by the further model \cite{majchrowska2022deep}. Similarly, many action recognition models work. First, a tool for pose estimation is used, and the coordinates are passed to another model \cite{li2021ta2n}.
Moreover, trained models are sometimes used to label new data quickly. Or to fine-tune models to the domain task instead of training from scratch. Sometimes pretrained models are used as feature extractors, and the following algorithm takes care of the target job. If the used model is biased, the next model in the sequence can inherit these tendencies. This bias is called \textit{inherited bias}. The term was introduced by Hellstrom et al.\cite{hellstrom2020bias}. Sun \cite{sun2019mitigating} (as noted by \cite{hellstrom2020bias}) identifies several NLP tasks that may cause an inherited bias: machine translation, caption generation, speech recognition, sentiment analysis, language modeling, and word embeddings \cite{sun2019mitigating}. An example is presented of how different tools for sentiment analysis predict different sentiments for the same utterances but other subject's gender. In one of my very recent studies \cite{mikolajczyk2022debiasing}, it was proved that generative models are vulnerable to catching and enhancing biases from data. GANs showed that they not only recreate bias in data but also significantly enhance it. For instance, one of the artifacts that was more prominent in one of the classes was never generated in the other (even though it naturally occurred in other class). Further models fed with data generated by GANs inherited those enhanced biases, resulting in even less robust models than those trained with no data augmentation at all.

Algorithmic bias can also come from the model itself. There is a noticeable \textit{learning bias} of deep networks toward low-frequency functions \cite{rahaman2019spectral}. It is well known that models prioritize learning simple patterns that generalize across data samples. Following that, Rahaman et al. \cite{rahaman2019spectral} investigated the shape of the data manifold by presenting both higher and lower frequencies to the models. They observed that lower frequencies are generally easier to learn (and learned first), and high frequencies get easier to train with increasing complexity. They called this tendency to favor smooth functions a \textit{spectral bias}. 

Recently, the image frequency in image classification was examined. It was discovered that some convolutional neural networks classify images by texture rather than by shape \cite{hermann2020origins}. The researchers examined the effect -- they experimented with mixing images with conflicting shapes and textures, e.g., a cat picture with an elephant texture. The CNNs tendency to classify images by texture despite particular objects' shapes was called a \textit{texture bias}. Authors showed that a random-crop data augmentation increases texture bias, and appearance-modifying data augmentation reduces it. On the other side, a \textit{shape bias} uses features primarily based on the item's shape in contrast to the texture.
 
Similarly, generative models tend to generate specific frequencies, making it easy to differentiate from the real ones \cite{schwarz2021frequency}. This tendency is called a \textit{frequency bias} \cite{schwarz2021frequency} or spatial frequency bias \cite{khayatkhoei2020spatial}. The paper shows that generating images leaves a trace of systematic artifacts that could be recognized as fake solely by spectra analysis.

Next, there is an \textit{omitted variable bias} \cite{cinelli2020making}, which covers the case when one or more important variables are, purposely or not, omitted. It might mean, for instance, training the model without providing a critical input feature. It is not easy to choose and select all crucial variables that might affect a final prediction. For example, when predicting stock market prices, one might try to analyze data in previous months and years and consider performing sentiment analysis on newspapers or web articles about companies valued on the market \cite{cakra2015stock}.

However, sometimes not only the design or feature selection affects the final result. At the time, the model might have been considered unbiased. Yet, a few months or years later, it could be burdened with an \textit{emergent bias} \cite{friedman1996bias}. Emergent bias arises in the context of use with real users \cite{friedman1996bias}. This bias typically emerges a while after training is finished due to changing societal knowledge, population, or even cultural values. Moreover, bias can occur when used by a population with different values than those assumed in the design. An example would be predicting the risk of obesity based on somebody's living area, which might change over time due to constantly changing eating habits, health awareness campaigns, or even changes in how obesity is defined. Emergent bias can be divided into more sub-types as described in \cite{friedman1996bias}.

Improper or poorly performed evaluation can also affect a model. This type of bias is usually referred to as \textit{evaluation bias} introduced during the model's evaluation \cite{gordon1995evaluation}. The definition includes poorly selected evaluation data (e.g., inappropriate benchmarks) or inadequate metrics that do not measure the model's performance \cite{suresh2019framework}. An example would be choosing the model based on an average accuracy in class imbalance: a model might quickly achieve 90\% accuracy by always predicting the same class when ninety percent of the whole dataset belongs to that class. Similarly, an \textit{illusion of control bias} happens when a designer achieves a high accuracy (or other metrics) and believes in controlling it \cite{masis2021interpretable}. However, high results on a test set do not always mean that model generalizes well. Models usually need to measure metrics on a test set and the behavior with outliers, examining it with explainable AI methods.

Finally, at the deployment stage of the model preparation, a \textit{deployment bias} can occur \cite{baker2021algorithmic}. A system is used or interpreted in inappropriate ways \cite{suresh2019framework}, e.g., a model is used for a different purpose than the initially designed purpose. 

\subsection{Results interpretation }

\begin{figure}[!htb]
\centering
  \includegraphics[width=0.85\textwidth]{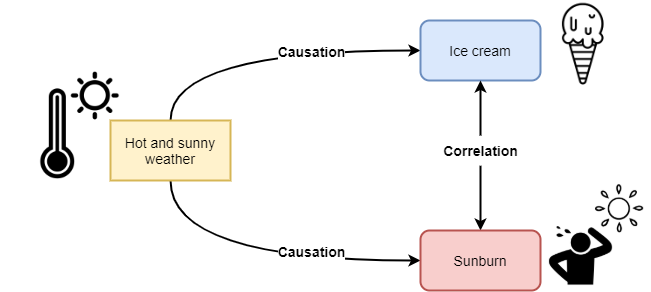}%
  \caption{Correlation bias - when the correlation between two or more variables is incorrectly mistaken with causation.} \label{fig.correlation}
\end{figure}

Even if we analyzed the data correctly, we still might erroneously interpret the results. This error is generally called an \textit{interpretation bias}. For instance, there is a \textit{correlation bias}, also known as \textit{cause-effect bias}, which, as the name suggests, happens when the correlation is mistaken with causation. Correlation means a relationship or pattern exists between the values of two variables \cite{altman2015association}. Causation means that one event causes another event to occur \cite{altman2015association}. Mistaking both leads to erroneous assumptions that bias the results of research. A commonly mentioned example of a cause-effect bias is a hot day example. When it is hot outside, people gladly go out and buy ice-creams. They are also more prone to sunburn, as they spend more time in intense sun. Hence, there is an apparent correlation between rising ice-cream sales and the number of sunburned people (Figure \ref{fig.correlation}). But does it mean that ice-creams are causing sunburns? 

The other type of interpretation bias is a \textit{conceptual bias}, also known as an \textit{assumption bias}, that arises from faulty logic, wrong premises, or mistaken beliefs of the investigator \cite{k2014bias}. It is very similar to what is often mentioned in psychology \textit{belief bias}, defined as a tendency to evaluate the argument's validity based on personal beliefs \cite{markovits1989belief}. The research on adults shows that subjects' performance is surprisingly often influenced by empirical factors instead of the premises or the conclusion. In an early work on the belief bias, the Markovits et al. \cite{markovits1989belief} presented to over 150 subjects two premises that were supposed to work as Ground Truth and a conclusion to mark as a $True$ or $False$. One of the examples was whether cats are animals or not, as presented in Figure~\ref{fig.cats}. It turns out that most people tend to answer due to their personal belief rather than the premises and select the answer as $False$, even though premises clearly show that cats can't be animals if they don't like water.

\begin{figure}[!htb]
\centering
    \begin{lstlisting}
    
    Task: For each problem, decide if the given conclusion follows logically
    from the premises. Circle YES if, and only if, you judge that the
    conclusion can he derived unequivocally from the given premises,
    otherwise circle NO.

    Example syllogisms:
    (1) Premise 1: All animals love water.
        Premise 2: Cats do not like water.
        Conclusion: Cats are not animals. [YES/NO]
        
    (2) Premise 1: All flowers have petals.
        Premise 2: Roses have petals.
        Conclusion: Roses are flowers.   
        
    \end{lstlisting}
    \caption{Belief bias examination. Questionnaire testing if subjects value more the premises or personal beliefs -- example from literature \cite{markovits1989belief}}
 \label{fig.cats}
\end{figure}

\subsection{Publication}

Finally, after all the previous stages, the researcher might try to publish the results. However, the research might be infused with bias even at this final stage.
 Interestingly, in machine learning, a new bias emerged called \textit{resubmission bias} \cite{stelmakh2021prior}. This bias creates a horn effect on the submitted manuscript previously rejected at a different venue. The impact of the resubmission bias on the overall score received by submissions appears to be small. Still, top machine learning conferences are highly competitive, and even tiny changes in review scores may significantly impact the outcome. For instance, the ICML 2012 conference data shows that papers with a mean reviewer score of 2.67/4.0 were six times more likely to be accepted than papers with a mean score of 2.33 \cite{stelmakh2021prior}. 

Another common phenomenon is a \textit{funding bias} that emerges when a party reporting results report them to satisfy the research study's funding agency or financial supporter \cite{mehrabi2019survey}. Valuable papers might be easily omitted, or their impact reduced due to the \textit{presentation bias}. Presentation bias is defined as a bias resulting from how the research topic (information) is presented \cite{mehrabi2019survey}. Finally, the \textit{ranking bias} shows that top-tier journals and conferences get much more attention than the local ones, even if the quality and scope of research are exact \cite{mehrabi2019survey}. As a result, it affects search engines and crowd-sourcing applications \cite{lerman2014leveraging}.

\section{Bias detection}

The previous sections introduced many possible biases that may accidentally contaminate data or models. Spreading the information about the consequences of poorly approached data or models preparation is one way to stop bias from spreading. But what if unwanted tendencies are already here, in gathered data or used models? How to make sure that our models are free from systematic errors? This is where bias detection is needed.

Research on bias analysis focuses mainly on detecting causal connections between input features and predictions of the trained models \cite{balakrishnan2021towards}. One of the bias detection methods could be a manual inspection of data and models, which relies mainly on observational studies. Statistical methods may help understand complicated statistics and reveal hidden spurious correlations that may influence the model. But is it possible to manually annotate or inspect every vast dataset that fuels deep learning algorithms? One of the most popular datasets, ImageNet, currently has over 14 mln images, Amazon Reviews over 82 mln of text reviews, and Common Voice over 1000 hours of speech. The manual inspection of those might be Sisyphean labor. 

Local explainability methods such as attribution maps visualized as heatmaps or visualizations based on prediction perturbation can boost manual review. Schaaf et al. \cite{schaaf2021towards} compared different attribution maps and their ability to detect bias. They introduce and use several metrics like the Relevance Mass Accuracy or Area over the perturbation curve (AOPC), which help evaluate how relevant attribution maps are. Large values of AOPC mean that perturbation significantly decreases prediction accuracy, indicating that the attribution method efficiently detects the relevant image regions \cite{schaaf2021towards}. As the authors conclude, they were able to find quantitative evidence that attribution maps can be used to detect some data biases. However, the analyses also show that attribution maps sometimes provide misleading explanations.

\begin{figure*}[!htb] 
\centering
\textit{Example visualization of occlusion-based attribution maps}

    \begin{subfigure}[Small skin lesion with smooth borders on the center of the image with strongly textured skin]{
      \includegraphics[width=0.25\linewidth]{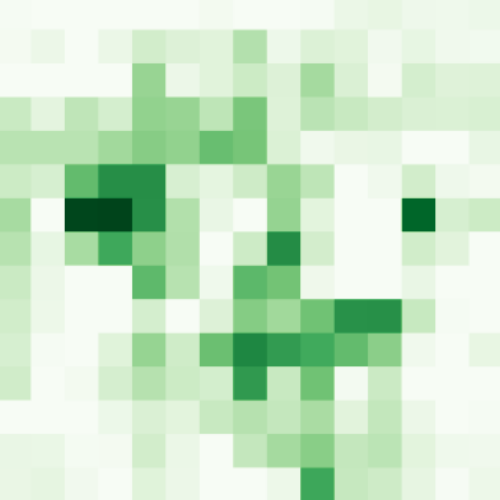}}
    \end{subfigure}
    \qquad
    \begin{subfigure}[Large protruding skin lesion with well-defined borders]{
        \includegraphics[width=0.25\linewidth]{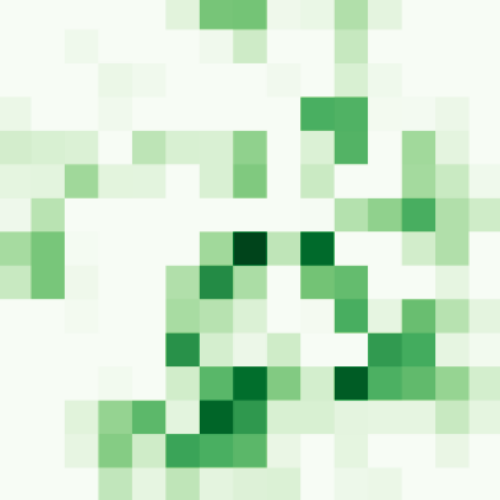}} 
    \end{subfigure}
    \qquad
    \begin{subfigure}[Medium round skin lesion with irregular border with streaks and atypical dots]{
        \includegraphics[width=0.25\linewidth]{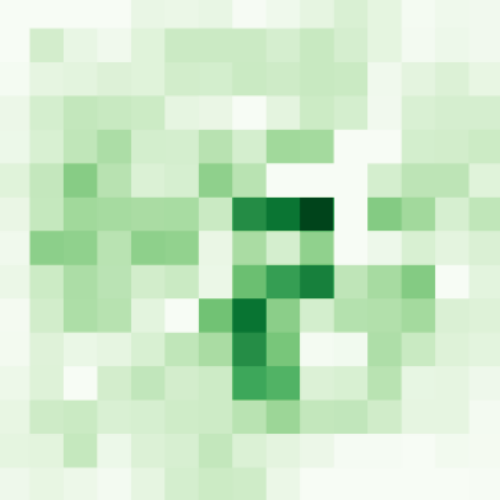}}
    \end{subfigure}
    \qquad
  \caption{Example visualization of occlusion-based explanations. In the heatmap, a darker green color means stronger attribution. Visualized with captum \cite{kokhlikyan2020captum}.}\label{fig.gebi.attribution}
\end{figure*}

One of the approaches to (semi) automated bias discovery is using global explainability methods. One of such methods is SpRAy, \cite{lapuschkin2019unmasking} described in Chapter~\ref{chapter.xai}. The idea of the method is to generate attribution maps of all data instances and then cluster those attribution maps to reveal some hidden patterns in the model's reasoning. Global explanation lets a user avoid a time-consuming manual analysis of individual attribution maps but requires a manual review of resulting clusters. However, there is one significant flaw with the method. The method uses only the attribution maps. Analyzing the attribution maps without an input image is a challenging task. Figure~\ref{fig.gebi.attribution} presents a few examples of attribution maps of skin lesions without source images. It is hard to tell what grabbed the attention of the prediction. 
Other global explainability methods are presented in Chapter~\ref{chapter.xai}.

Another approach, proposed by Stock et al. \cite{stock2018convnets}, used an adversarial version of model criticism initially proposed by Kim et al. \cite{kim2016examples}, and a  feature-based explanation to uncover potential biases. Model criticism summarizes the relations between input features learned by a model using a carefully selected subset of examples (prototypes). Such a tool helps in manual data and model investigations and could be automated shortly. 

Balakrishnan et al. \cite{balakrishnan2021towards} proposed using Generative Models for building causal benchmarks. Generative models manipulate input features, e.g., gender and skin tone, to reveal potential causal links between feature variation and prediction changes. However, as the author mentions, those Generative Models are hardly controllable, and hidden confounders can still be present in benchmarks. 

Another approach proposed by Serna et al. \cite{serna2021ifbid} is an Inference-Free Bias Detection that, in contrast to other approaches, tries to detect bias by investigating the models through their weights. In the paper \cite{serna2021ifbid}, bias is detected with an additional detector model, which tries to detect bias encoded in the parameters of the trained model. The detector model takes as input weights of the trained model: the architecture depends on the weights/filters of each layer and has a dense layer that concatenates all the outputs of each weight/filter module. The performance is proved to be quite good. Still, it can be used to caution the user during the inference rather than detect new bias in training data, as it requires a training dataset to find biases.

The literature about bias detection is relatively poor. There are no guidelines or widely used algorithms to help bias discovery in models and data. The only available option is manual annotation and manual data inspecting.
In the thesis, I proposed using global explainability methods as a tool for global evaluation of the model and, as a result, the data. The proposition is presented in Chapter~\ref{chapter.gebi}.

\section{Bias mitigation} \label{section.bias-mitigation}

Bias mitigation methods from classical literature usually operate on simple, often linear models \cite{wang2020towards}. However, such approaches are not even close to solving the problem in the deep learning era. Nobody wants to resign from high-efficiency models for simpler linear algorithms. And yet, completely ignoring possible biases in data and models is not a solution. Those problems have a tangible impact on our lives, as deep learning-based models are more often used in practice. It is well documented that models reflect the bias in data and often amplify it \cite{zhao2017men}.
This started a new area of research towards mitigating biases in data and models for safer, more robust, and fairer deep models without sacrificing their size or architecture. 

An often approach is fairness through blindness \cite{wang2020towards}. The idea is simple. If we think a variable might bias the model, we should not include it as a set of input features. For instance, we might not want to include information about the candidate's gender when evaluating the potential job candidate's resume. If a model does encode information about a protected variable, it cannot be biased -- that is the logic. In reality, some information about the gender might be encoded in the resume, e.g., feminine hobby connected to the gender or gender-specific adjectives. Removing all potential biases is often a very hard, if not impossible, challenge.

Hence, other approaches emerged. For instance, Zhao et al. \cite{zhao2017men} proposed an inference update scheme to match a target distribution to remove bias: their method introduces corpus-level constraints so that selected features co-occur no more often than in the original training distribution. Next, Dwork et al. \cite{dwork2018decoupled} proposed a scheme for decoupling classifiers that can be added to any black-box machine learning algorithm. They can be used to learn different classifiers for different groups. Another branch is adversarial bias mitigation, i.e., supervised learning
here, the task is to predict an output variable $Y$ given an input variable $X$, while remaining unbiased with respect to some variable $Z$ \cite{zhang2018mitigating}. The approach proposed by Zhang et al. \cite{zhang2018mitigating} used the output layer of the predictor as an input to another model called the adversary network, which attempts to predict $Z$. The idea was improved by Le Bras et al. \cite{le2020adversarial}, who proposed the idea of Adversarial Debasing Filters. The proposed algorithm used linear classifiers trained on different random data subsets at each filtering phase. Then, the linear classifier's predictions are collected, predictability score is calculated. High predictability scores are undesirable as their feature representation can be negatively exploited -- hence Le Bras et al. \cite{le2020adversarial} proposed simply removing the top $n$ instances with high scores. The process is then repeated several times to reduce the bias influence.

Finally, there is attention guidance. Early works about attention guidance in computer vision focused on improving the segmentation task \cite{huang2019brain}, making classification better with attention approaches used in Natural Language Processing \cite{barata2019deep}, or even using attention maps to zoom closer to the region of interest \cite{li2019zoom}. However, attention in terms of vision transformers is different from attribution maps. Researchers argue that they can be used interchangeably. In 2019, Jain et al. \cite{jain2019attention} explained why they think that \textit{Attention is not an explanation}, whereas \cite{wiegreffe2019attention} claimed that \textit{Attention is not not an explanation}. The conflict did not die with time, as many other papers regarding this matter appeared \cite{grimsley2020attention,tutek2020staying,woody2004more}.

One of the similar emerging approaches is attention guidance \cite{tang2018attention}. The guidance provided with, for example, attention maps highlights relevant regions and suppresses unimportant ones, enabling a better classification. A similar method is based on self-erasing networks that prohibit attention from spreading to unexpected background regions by erasing unwanted areas \cite{hou2018self}. Some researchers proposed different ways to solve this problem, such as rule extraction, built-in knowledge, or built-in-graphs \cite{chai2020human}. Yet, there is still a long way to go to achieve full transparency of the DNNs reasoning process and incorporate it into the training process.  
In the thesis, the bias mitigation problem is approached in two ways: with data augmentation, i.e., randomly inserting bias during the training to make it insignificant (Chapter \ref{chapter.targeted}, and adversarial attribution training, Chapter \ref{chapter.attribution}). 

\clearpage  
\lhead{\emph{Chapter 3: Explainable Artificial Intelligence}}  
\chapter{Explainable Artificial Intelligence }
\label{chapter.xai}
\section{Taxonomy} \label{section.taxonomy}

According to Cambridge Dictionary, in everyday life, the verb \textit{explain} means to \textit{make something clear or easy to understand by giving reasons for it or details about it} \footnote{\href{https://dictionary.cambridge.org/us/dictionary/learner-english/explain}{\textit{explanation} according to Cambridge Dictionary (US): Learner English}}. In machine learning, the term \textit{explain}, or rather \textit{explainability}, refers to the \textit{ability to explain} the inner model workings. In other words, explainability is used to make reasoning and predictions of the AI model clear and easy to understand. Explainability helps \textit{to understand the logic of the model in general} and \textit{to understand the reasoning behind the prediction}. 

Achieving explainability of the models is usually done by giving \textit{an explanation} to the user. An explanation can be generated by an external algorithm or the model itself. But what is an explanation? In the online Cambridge Dictionary, an explanation means \textit{the details or other information that someone} (e.g., an external algorithm or the model) \textit{gives to make something} (e.g., a model or a prediction) \textit{clear or easy to understand}. An explanation can also be defined as \textit{a reason or an excuse for doing something}. In Explainable Artificial Intelligence (XAI), an explanation can be defined as additional meta-information used to describe an input instance's feature importance or relevance towards a particular output classification \cite{das2020opportunities}. 
Once we find out the definitions of the explainability of AI and what an explanation is, we might start to think about what makes a good explanation. There are two different goals for explainability: achieving full \textit{interpretability} and \textit{completeness} \cite{gilpin2018explaining}.
Interpretability is a clear and understandable description for humans of the internals of a system.
The goal of completeness is to describe the operation of a system accurately.
I introduced an overall review of existing local and global explanations. Each algorithm will be categorized into global or local, intrinsic or post-hoc, model-specific, or model-agnostic explanations in the following description. 

\begin{figure*}[!htb]
\centering
\includegraphics[width=\linewidth]{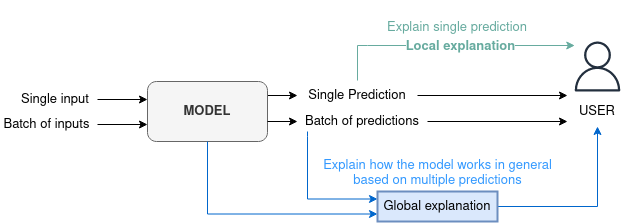}
  \caption{Global and local explanations.}
  \label{fig.xai.global-local}
\end{figure*}

Explainability methods can be divided depending on whether they apply to a single prediction or to how the model works in general: local and global explanations. \textit{Local explanations} aim to explain a single prediction, usually to justify or explain a reasoning process behind it \cite{molnar2020interpretable}. The \textit{global explanations} are generalizing over multiple predictions (sometimes even whole dataset) to explain the inner model workings \cite{molnar2020interpretable}. Global explanations are used for a better insight into the model, to understand the reason behind the group of predictions, or, as I presented in the Chapter \ref{chapter.gebi} to identify bias in data. Figure \ref{fig.xai.global-local} presents the global and local distinction.

\begin{figure*}[!htb]
\centering
\includegraphics[width=\linewidth]{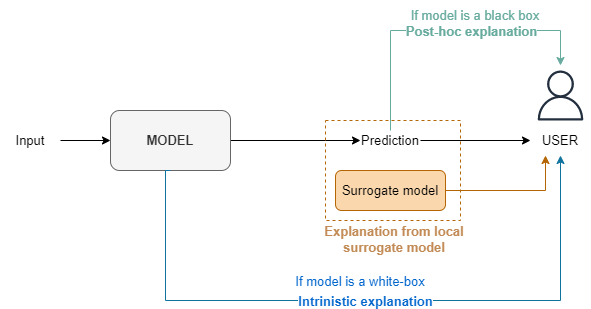}
  \caption{The system's transparency: post-hoc, intrinsic, and surrogate models.}
  \label{fig.xai.posthoc-intrinistic}
\end{figure*}

There is a division depending on the model's or system's transparency. The system can be either purposely designed to be \textit{intrinsically explainable}, or explained with \textit{post-hoc methods} afterward. Hence, intrinsic explainability refers to explainable models due to their relatively simple architecture \cite{pintelas2020grey}. Those models are sometimes even called \textit{white-box} or \textit{transparent}. For instance, short decision trees or linear models are sometimes explained by design \cite{arrieta2020explainable}. A post-hoc (from Latin post hoc, "after this") is an explanation type used to describe the reasoning of models with complicated structures as it refers to an answer given after making a prediction. It explains larger and more complex models like convolutional neural networks or transformer-based architectures. A common subtype of post-hoc models is \textit{surrogate models}. 
The surrogate model is an engineering term relating to using a more superficial approximation model of the outcome of the process instead of modeling the process itself \cite{queipo2005surrogate}.
Those kinds of metamodels are used, for instance, when the process is very complicated to recreate or takes too much computational effort. 
As a result, they also found a way to explain predictions of deep AI models.
Such simpler, approximate linear models of complicated and highly non-linear models are much easier to interpret and hence widely called \textit{interpretable models} \cite{messalas2019model}. The difference between surrogate models and post-hoc and intrinsic explanations is illustrated in Figure \ref{fig.xai.posthoc-intrinistic}.

\begin{figure*}[!htb]
\centering
\includegraphics[width=\linewidth]{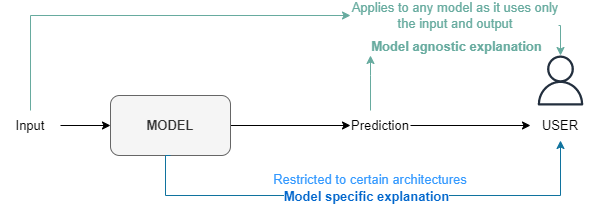}
  \caption{Model agnostic vs. model specific explanations.}
  \label{fig.xai.agnostic-specific}
\end{figure*}

Another type of XAI categorization is divided into model-specific and model-agnostic explanations \cite{molnar2020interpretable}. Model-specific methods are designed to work with one specific kind of model or group of them (i.e., to work with CNNs), whereas model-agnostic methods can work with any model, e.g., treat the model as a black box. The distinction between model-agnostic and model-specific explanations is presented in Figure \ref{fig.xai.agnostic-specific}.

Additionally, it is worth noticing that different XAI algorithms aim to answer other questions depending on model, user, and task \cite{mohseni2018multidisciplinary}. Mohseni et al. differentiate the six common explanations based on questions: \textit{how?}, \textit{what?}, and \textit{why?}. Explanation types, along with their definitions, are listed below.

Explanatory information types:
\begin{itemize}
 \item \textit{How?} explanations aim to explain how the model works in general. Hence those kinds of explanations usually fall in the subcategory of global explanations. Common approaches are, for instance, visualized decision boundaries \cite{van2008visualizing} and model graphs \cite{lakkaraju2016interpretable}.
 \item \textit{Why?} explanations are targeted to explain a particular prediction: e.g., why does such prediction occur? In contrast to \textit{how} explanations, they are usually local, i.e., they explain a single prediction. \textit{Why}  explanations try to raise light on what logic in the model led to such prediction or what features influenced it. The typical approach is to use feature attribution that highlights which parts have the most significant impact on the prediction.
 \item \textit{Why-not?} explanations are usually used to reveal why the output differs from excepted. Similarly, as in the case of \textit{why} explanations, this approach mainly uses local explanations and very often features importance.
 \item \textit{What-if?} and \textit{How-to?} are both counterfactual explanations, e.g., they are used to examine how prediction changes regarding the input's hypothetical changes. A theoretical change could be designed, for instance, to check if changes in a single feature could change the prediction. For example, a target user could formulate a question, \textit{What input changes would change the classification result?} or \textit{Would this prediction be the same if this feature is removed?}
 \item \textit{What-else?} explanations are used to present other similar inputs with similar predictions to the target user \cite{cai2019human}. The mechanism behind the what-else explanations is straightforward: the inner model's representation should pick up similar samples to the original input samples from the training dataset. This explanation is popular and easy to achieve but might sometimes be misleading.
\end{itemize}

\section{Methods of Local explainability}
As mentioned in the previous section, the local analysis aims to explain a single model prediction. Local explanations are usually used for prediction justification and verification. The are various methods of local explainability. In computer vision, one of the common approaches is to visualize the pixels that excite the output the most, e.g., in the form of a visual map. Those visualizations are heatmaps, saliency maps, attribution maps, and others. In general, a heatmap is defined as \textit{a data visualization technique that shows the magnitude of a phenomenon as color in two dimensions} \cite{wilkinson2009history}. A saliency map is defined as an image that highlights an important region (e.g., on which people's eyes focus first to reflect the degree of pixel importance \cite{guo2009novel}. 
In explainable AI \textit{heatmaps}, \textit{saliency maps} and \textit{ (pixel) attribution maps} refer to visual maps of \textit{the input stimuli that excite feature maps} \cite{selvaraju2017grad}, e.g., they show which regions of an input image had the most significant influence on the prediction. In most cases, the goal of the maps is the same: to visualize the input's importance through different means and methods.

Other commonly-mentioned names are sensitivity maps, relevance maps, feature contribution maps, and gradient-based attribution maps, in which generation methods differ, but the goal is similar. 

Here I presented selected local explainability methods, commented, and explained them step-by-step. To almost every method, I added a preliminary \textit{step zero} that is not necessarily a part of the algorithm but is needed to perform it accurately.

\subsection{GradCAM}
\begin{figure*}[!htb]
\centering
\textbf{Example visualization of gradient-based explanations}

    \begin{subfigure}[Original image]{
      \includegraphics[width=0.25\linewidth]{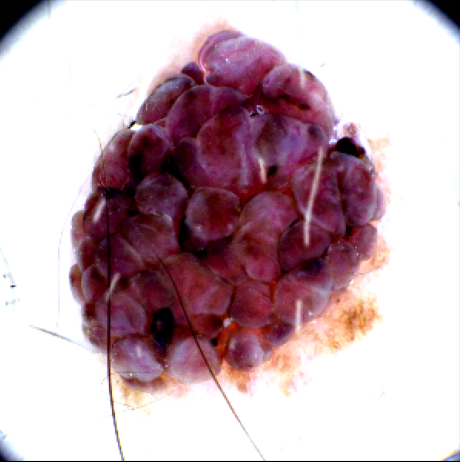}}
    \end{subfigure}
    \qquad
    \begin{subfigure}[Saliency-based visualization]{
        \includegraphics[width=0.25\linewidth]{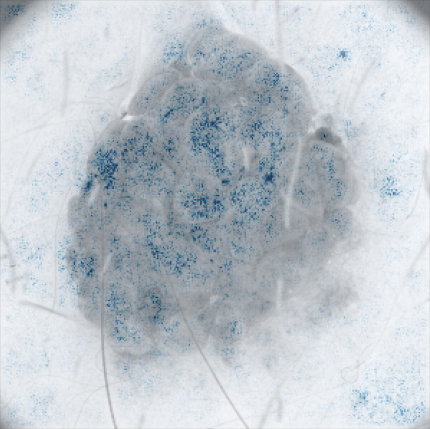}} 
    \end{subfigure}
    \qquad
    \begin{subfigure}[Integrated gradients]{
        \includegraphics[width=0.25\linewidth]{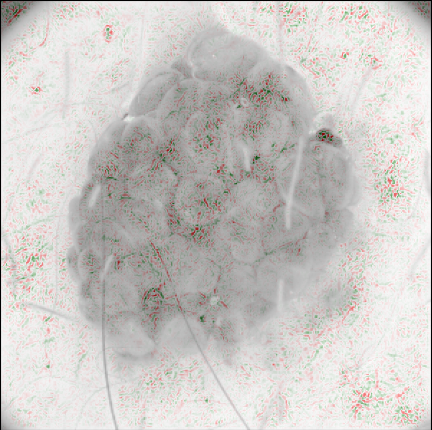}}
    \end{subfigure}
    \qquad
  \caption{Example visualization of gradient-based explanations. In a saliency-based heatmap, darker blue color means stronger attribution. In integrated gradients, green colors mean positive attribution, whereas red ones are negative. Visualized with Captum \cite{kokhlikyan2020captum}.}
  \label{fig.xai.grad}
\end{figure*}

A \textit{Gradient-weighted Class Activation Mapping} (Grad-CAM) is designed to visualize and better understand convolutional neural networks. The method combines class activation maps \cite{zeiler2014visualizing} with guided backpropagation \cite{springenberg2014striving} allowing for better and more detailed heatmaps. Those heatmaps can be visualized at any layer in the model, which allows for examination of the model layer-by-layer. Example gradient-based explanations are presented in Figure~\ref{fig.xai.grad}.

The steps of the Grad-CAM method are as follows:
\begin{enumerate}[\itshape   Step 1:]
\setlength{\itemsep}{1pt}
 \setcounter{enumi}{-1}
 \item Select an image and a class to explain.
 \item Forward propagate selected image through the model to generate the prediction. Save the loss value.
 \item Set gradients to zero for all not selected classes and one for the desired class.
 \item Compute the gradients through signal backpropagation from the second step through rectified convolutional feature maps of interest.
 \item Calculate alpha values (weights) by combining mean gradient values with convolutional outputs to compute the coarse grade-CAM localization (blue heatmap), which represents where the model has to make the particular decision. 
 \item Multiply the heatmap with alpha values with feature map activations to get Guided Grad-CAM visualizations.
\end{enumerate}

The greatest strength of Grad-CAM is its computational efficiency. Those gradient-based algorithms usually require a single backward pass through the network to generate a heatmap. Additionally, the simplicity of the approach opens the method to broader publicity. The weak point is its noisiness, lack of clarity, and understandability making the explanations hard to interpret. The architecture design can affect the visualizations as it does not work well with batch normalization or pooling layers, where the gradient is discontinuous. It is not compatible with certain activation functions, i.e., non-linear activations like ReLU are non-differentiable in specific ranges, as they provide discontinuous gradients \cite{laurent2018multilinear}. Some researchers even show that gradient-based explanations are more sensitive to the model's architecture than the parameters' values \cite{adebayo2018local}. They show that randomly initialized networks give very similar explanations to the trained ones. The suspicion behind that phenomenon is that gradient-based local explanations are dominated by lower-level features, which are connected with the considerable importance of the model's architecture and provide a solid prior affecting learned representations \cite{adebayo2018local}. Sadly, reported research shows that gradient-based methods might not be a reliable source of explanations, as they tend to be insensitive to the model's parameters and the data itself \cite{adebayo2018sanity}. 

\subsection{Activation Maximization}
Activation maximization (AM) is one of many deep visualization approaches designed for deep neural networks. Activation Maximization provides a post-hoc, model-specific local explanation. AM generates an input (e.g., image) that maximally activates all neurons or a group of neurons (features or group of features) \cite{zeiler2014visualizing, nguyen2016multifaceted}. The method uses the model's weights -- learned knowledge -- to determine the different parts of inputs (image regions) that activate a neuron, e.g., to visualize what triggers the set of neurons the most.

The steps for Multifaceted Feature Visualization, one of the improved activation maximization methods, are as follows:
\begin{enumerate}[\itshape   Step 1:]
\setlength{\itemsep}{1pt}
 \setcounter{enumi}{-1}
 \item Select a set of images and number of features to find.
 \item Pass all images through the network and compute outputs for the selected layer.
 \item Reduce the dimensionality of each output (e.g., with PCA).
 \item Visualize the entire set of outputs with t-SNE to produce a 2D embedding.
 \item Cluster each embedding with k-means clustering. The number of clusters is equal to the number of features to find.
 \item Compute a mean image by averaging the $n$ nearest images to the cluster centroid.
 \item Run activation maximization but initialized with a mean image instead of a random one.
\end{enumerate}

Activation maximization is an excellent source of knowledge about what the model has learned at different layers or filters. However, it is designed to help developers understand their models, not to address the problems of reliability or trust. It does not fully explain the inner model workings, and even when explanations are interesting, they are often ambiguous.

\subsection{LIME}
\textit{Local Interpretable Model-Agnostic Explanation }(LIME) explains predictions by approximating it locally with an interpretable, surrogate model \cite{lime}. 
The LIME method is a counterfactual approach to the explanations.
A user disturbs the input, e.g., covering a part of the analyzed image and checking how the prediction changed. The process of removing specific features and checking the prediction is repeated several times. 
Those steps create a local surrogate model that explains what part of input has the most decisive impact on the final result (or which has none). 

\begin{figure}[!htb]
\centering
 \includegraphics[width=0.8\textwidth]{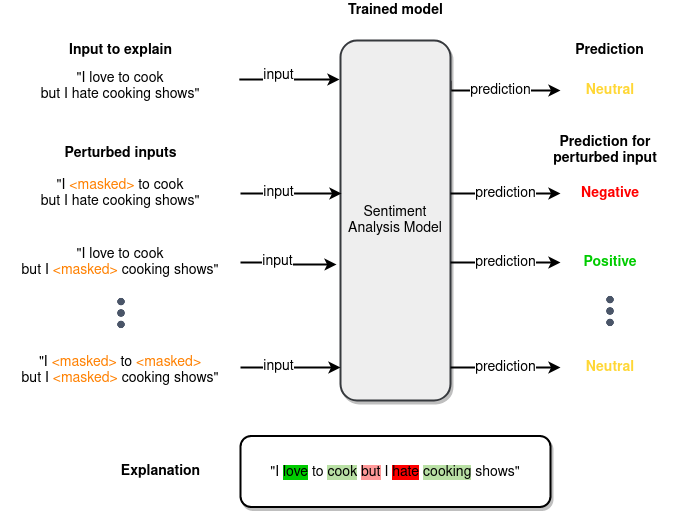}%
 \caption{LIME- example} \label{fig.lime}
\end{figure}

The idea behind the LIME can be easily illustrated by giving an example. Let us consider the case of explaining the model trained to predict the sentiment of a given text. The user wants to explain a sentence, \textit{I love to cook, but I hate cooking shows}, given the prediction of neutral sentiment. To explain it with LIME, we need to perturb the input first, for instance, by masking selected parts of a text, i.e., if we hide the word \textit{love} (\textit{I [MASK] to cook but I hate cooking shows}), the prediction changes from \textit{neutral} to \textit{negative}, but if we mask the word \textit{hate} (
\textit{I love to cook but I [MASK] cooking shows}) we will get a \textit{positive} output from the model. Masking other than those two words did not change a prediction. This example is illustrated in Figure~\ref{fig.lime}.

However, when applying LIME for images, masking each pixel one by one is a very ineffective approach. Hence, the LIME for image classification uses a clustering method first, which is used on the image, to divide it into clusters. Then, the image is perturbed by hiding those clusters. An example of a perturbed image and an example explanation is shown in Figure~\ref{fig.lime}.

The steps behind the LIME for image classification are as follows:
\begin{enumerate}[\itshape   Step 1:]
\setlength{\itemsep}{1pt}
 \setcounter{enumi}{-1}
 \item Select an image to explain and number of steps \textit{n}.
 \item Split  image $x$ into $n$ superpixels (segments) i.e. with quickshift algorithm \cite{vedaldi2008quick}.
 \item Create dataset $Z$ by computing replacements images $\bar{x}$ -- perturb the image by removing selected segments i.e., paint them black or the main color of superpixel. 
 \item Use dataset $Z$ of perturbed samples with the associated labels to optimize Eq. \ref{equation.lime1} to get an explanation $\xi(x)$. 
\end{enumerate}

The equation for optimizing explanation $\xi(x)$ is provided below:

\begin{equation} \label{equation.lime1}
\xi (x) = \underset{g \varepsilon  G}{argmin}  L(f, g, \pi_x) + \Omega(g) 
\end{equation}


LIME is a marvelous tool for generating straightforward explanations. The great strength is the simplicity of the approach: it is easy to grasp even for people without any technical background. However, it tends to work slowly on bigger models or larger pieces of data, as it requires multiple passes of input thorough the model. It raises another problem connected to hyperparameter selection. We might end with a different result depending on how we partition single input. Hence, changing parameters means different explanations. LIME seems to be also sensitive to the data and label shift. Rahnama et al. point out that the random perturbation of features cannot be considered a reliable data generation method for LIME \cite{rahnama2019study}. Last but not least, LIME is proved to be easily deceived by presenting perturbed with adversarial attacks examples \cite{slack2020fooling}. Nevertheless, it is still one of the most commonly known and used explanations.

\subsection{SHAP}
\begin{figure}[!htb]
\centering
 \includegraphics[width=0.9\textwidth]{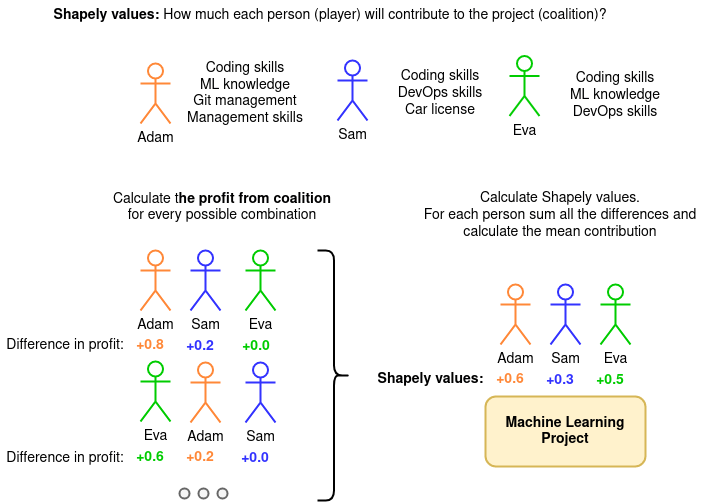}%
 \caption{Shapely values - illustrative explanation} \label{fig.shapely}
\end{figure}

\textit{Shapley Additive exPlanations} (SHAP) is another model-agnostic, post-hoc explanation method that uses Shapley values from the game theory to assess a feature's importance \cite{lundberg2017unified}. SHAP values measure how each input feature affects a model prediction, i.e., how much each part contributes to the output. Higher values mean higher importance of the feature - both positive and negative. It uses the Shapely values to calculate how each feature (player) contributed to the predicted output (coalition). To understand Shapely values, we can imagine the situation where the ML project manager (PM) wants to know which person contributes to his project the most. As most team members share similar skill sets, PM has to consider every possible combination, for instance, to check whether a new person joining adds unique value to the team. The summed difference in the profit from the coalition after adding a new person is called Shapely value. Shapely values with explained examples are presented in Figure~\ref{fig.shapely}.

\begin{figure}[!htb]
\centering
 \includegraphics[width=0.9\textwidth]{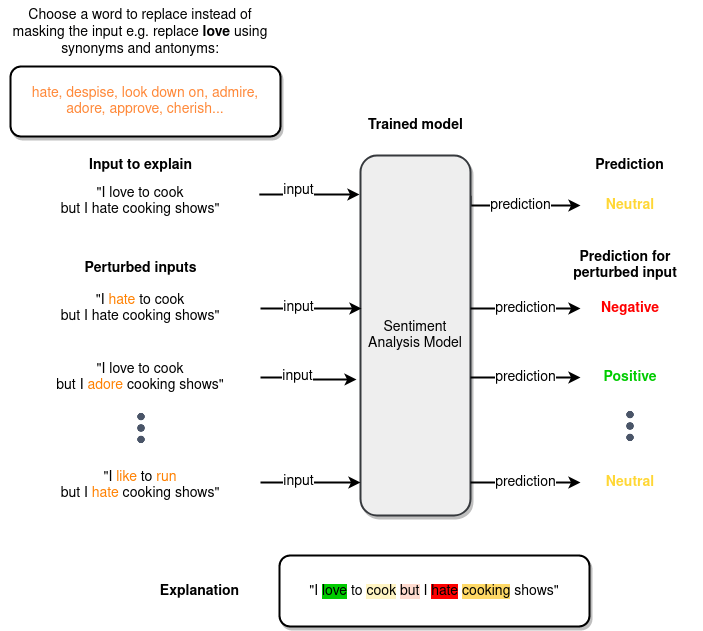}%
 \caption{SHAP- example} \label{fig.shap}
\end{figure}

The general idea behind SHAP is similar to LIME: the input is perturbed, fed to the network, and the feature importances are then calculated. 
The main difference is how the feature importance is calculated: in LIME, one creates a local, surrogate model around the prediction one wishes to explain. In SHAP, one tries to find out the exact contribution of each attribute. 
Let us consider the case of explaining the sentiment prediction model again. The user wants to explain a sentence, \textit{I love to cook, but I hate cooking shows}, given the prediction of neutral sentiment. To explain it with SHAP, we perturb the input by replacing selected parts of the text with a word from a different example, i.e., if we replace the word \textit{love} with the word \textit{hate} (\textit{I hate to cook but I hate cooking shows}), the prediction changes from \textit{neutral} to \textit{negative}, but if we replace the word \textit{hate} with \textit{adore} (
I\textit{ love to cook but I adore cooking shows}) we will get a \textit{positive} output from the model. We can try this by replacing more than a single word, for instance, \textit{love} to \textit{like}, and \textit{cook} to \textit{run}, which will change our prediction to neutral. This example is illustrated in Figure~\ref{fig.shap}.

The simplified steps behind the SHAP for image classification are as follows:
\begin{enumerate}[\itshape   Step 1:]
\setlength{\itemsep}{1pt}
 \setcounter{enumi}{-1}
 \item Select an image to explain and number of steps \textit{n}.
 \item Split image x into \textit{n} superpixels (segments).
 \item Perturb the image by removing selected segment and replacing them with pieces of images from other instances.
 \item Pass perturbed image through the network and get the prediction. 
 \item Repeat Steps three and four to detect which parts of an image (which segment) have the greatest impact on the final prediction.
 \item Show the most important segment of the image and hide irrelevant ones.
\end{enumerate}

As SHAP is basically a LIME with Shapely values, it also shares similar advantages and disadvantages. Contrary to LIME, it has a solid theoretical background in game theory, making it more trustworthy.  
On the one hand, SHAP explanations are more accurate than LIME, and thanks to using LIME methodology with Shapely values, it searches wider parameters space than the approach used in LIME. On the other hand, SHAP is even more computationally expensive, especially for large datasets, making it a slow but still reliable explanation method.

\subsection{Layer-wise Relevance Propagation}
The \textit{Layer-Wise Relevance Propagation} is a widely used method for explaining local predictions: it is a post-hoc and model-specific method. The general concept is to perform a pixel-wise decomposition that shows how pixels contribute to the positive and negative classification results. Hence, the goal is to attribute a contribution -- called the relevance, to each pixel of a corresponding prediction. Montavon et al. \cite{montavon2019layer} summarized the state-of-the-art to explain how to apply LRP to deep neural networks with a few rules such  as  Basic  Rule  (LRP-$\theta$),  Epsilon  Rule  (LRP-$\epsilon$), and  Gamma  Rule  (LRP-$\gamma$) as  one  rule:
\begin{equation}
R_{j} = \sum_{k}^{}\frac{a_{j}\cdot p (w_{jk})}{\varepsilon + \sum_{o,j}^{}a_{j}\cdot p (w_{jk})}R_{k}
\end{equation}
where $w_{jk}$ are weights between layer $j$ and $k$, $a_{j}$ are activations of neurons at $j$-th layer, and LRP$-\theta/\epsilon/\gamma$ are special cases. The role of $\epsilon$ is to provide numerical stability. The $\gamma$ controls how much positive contributions are favored. The LRP method, along with the rules, are described in detail in \cite{montavon2019layer}.

The steps behind the LRP for image classification are as follows:
\begin{enumerate}[\itshape   Step 1:]
\setlength{\itemsep}{1pt}
 \setcounter{enumi}{-1}
 \item Select an image, and class (output) to explain.
 \item Pass the image through the model and get the prediction.
 \item Starting from the last layer, assign the relevance score to the output neuron, equal to the predicted value.
 \item Move one layer forward and assign each neuron a relevance value. Each neuron redistributes relevance to the lower layer as much as it has received from the previous layer. Repeat this step until all the relevance scores are calculated. When redistributing relevance scores, follow the LRP rules depending on the type of each layer.
 \item Visualize the relevance on the selected layer.
\end{enumerate}

\begin{figure}[!htb]
\centering
    \begin{subfigure}[Original images]{
      \includegraphics[width=0.23\linewidth]{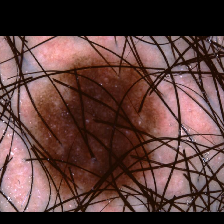}
      \includegraphics[width=0.23\linewidth]{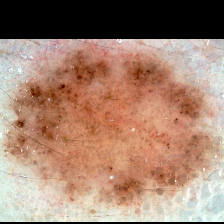}
      \includegraphics[width=0.23\linewidth]{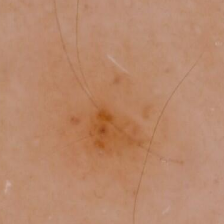}
      \includegraphics[width=0.23\linewidth]{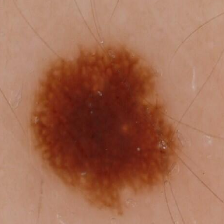}}
    \end{subfigure}
    \qquad
   \begin{subfigure}[Attribution maps]{
      \includegraphics[width=0.23\linewidth]{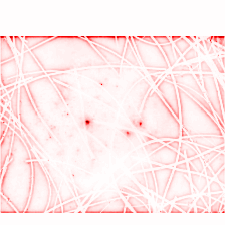}
      \includegraphics[width=0.23\linewidth]{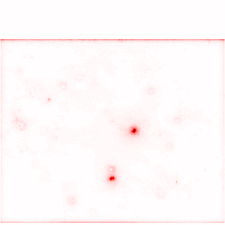}
      \includegraphics[width=0.23\linewidth]{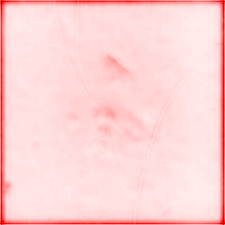}
      \includegraphics[width=0.23\linewidth]{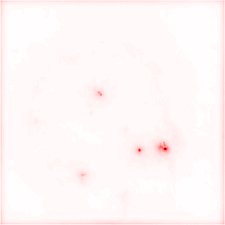}}
    \end{subfigure}
  \caption{Example visualization of relevance-based explanations - Deep Taylor Decomposition explanations.}
  \label{fig.xai.dtd}
\end{figure}

Another view on layer-wise relevance propagation was proposed by Montavon et al., as they proposed to see LRP as a Deep Taylor Decomposition (DTD) extended to work on deep learning models \cite{montavonDTD}. As the authors explain, the deep Taylor decomposition method proposed in \cite{montavonDTD}, was inspired by the divide-and-conquer paradigm. It is designed to exploit the property of deep neural networks' architecture, as they structurally decompose input into a set of simpler subfunctions that relate quantities in adjacent layers \cite{montavon2019layer}. Example DTD explanations are presented in Figure~\ref{fig.xai.dtd}.

The significant disadvantage of relevance-based approaches is that they are model-dependent. LRP introduces rules for different layers explaining how the relevance should be redistributed: e.g., Spatial Pooling Layers, Batch Normalization Layers, and Input Layers \cite{montavon2019layer}. Also, it leaves for the user hyperparameters search, which requires an expert's knowledge. The parameters $\gamma$ and $\epsilon$ should be different depending on which layer we want to create an explanation. The authors of the method suggest using different values for the upper, lower, and middle layers.  
However, most disadvantages disappear after the first stage: when the implementation and hyperparameter selection are made. It is computationally effective. It is faster than perturbation-based approaches. Alike, it is an attractive method for visualizing the reasoning behind predictions as it gives much less noisy output than, i.e., gradient-based approaches. Finally, LRP can be applied to every layer in the neural network and is extensible to many ML models and tasks.

Considering the mentioned above methods of local explainability, I decided to use attribution-based heatmaps mostly as local explainability methods. Both layer-wise relevance propagation and deep Taylor decomposition methods were used to explain predictions to verify the model's reasoning. They give relatively good-quality explanations with efficient computation times. LRP and DTD were further used in the \textit{Chapter \ref{chapter.skin_lesion_bias}: Identifying bias with manual data inspection as the main explainability tool for manual data inspection. Moreover, they were aggregated and used together as a global explainer described in Chapter \ref{chapter.gebi}: Identifying bias with global explanations}.
LRP was additionally used to analyze the shape and texture influence on the prediction in \textit{Chapter \ref{chapter.neural-style}: Mitigating shape and texture bias with data augmentation}. 

Although many disadvantages, the gradient-based heatmaps were used for experiments with attribution feedback described in \textit{Chapter \ref{chapter.attribution}: Debiasing effect of training with attribution feedback}. Their advantage is that they are differentiable, easier to implement on any architecture than LRP/DTD, and due to the best computation times necessary to efficiently perform multiple experiments, including thousands of training steps with numerous back and forward-propagation passes. 

\section{Methods of Global explainability}

Global explainability methods are less developed than the local ones. Most approaches consider aggregating multiple local explanations to create a more comprehensive global one. This includes generating multiple heatmaps and then clustering them or their representations to find some prediction strategies. Sometimes the clustering is also performed on the pair-wise rank distance matrices calculated from normalized attribution maps. Another type of global explanations is an idea based on concepts rather than features, i.e., showing what patterns or prediction strategies impact the model most. The section reviews the most common global approaches to explainability.

\subsection{Concept Activation Vectors} \label{subsection.CAV}
\textit{Concept Activation Vectors (CAV)} is a post-hoc method of global explanation that uses a set of pre-defined features (concepts) to explain the model's reasoning.

\begin{figure}[!htb]
\centering
 \includegraphics[width=1\textwidth]{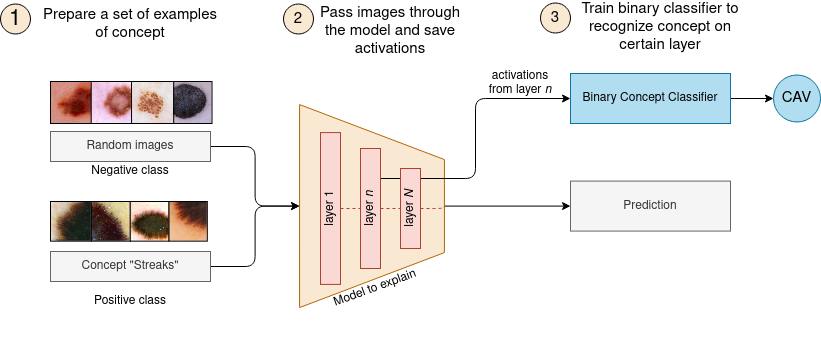}%
 \caption{Concept Activation Vectors - computation workflow.} \label{fig.cav}
\end{figure}

The first step of CAV is the selection/definition of so-called \textit{concepts}. Each concept is represented by a set of labeled examples. For example, we could define a \textit{striped pattern} concept visualized by images with vertical and horizontal stripes with different colors or thicknesses. Concepts are used to examine how much outputs on particular layers or outputs (predictions) will change when an input changes toward the concept. This allows observing how significant each concept was for the prediction. The computation workflow of CAVs is presented in Figure \ref{fig.cav}.

The steps behind the CAV for image classification are as follows:
\begin{enumerate}[\itshape   Step 1:]
\setlength{\itemsep}{1pt}
 \setcounter{enumi}{-1}
 \item Choose a set of examples with the concept of interest (positive class) and a set of random examples without it (negative class)
 \item Pass all example sets through the network and save selected layers activations
 \item Train a binary linear classifier to recognize the concept on particular network layers. The trained classifier is a learned Concept Activation Vector.
\end{enumerate}

A method of modifying input towards a concept creates a significant difference in explanations between saliency maps and CAVs. In the case of heatmaps, the model is examined for robustness, e.g., we look for changes in the output class in the magnitude of a single input (e.g., pixels or words). But in the case of CAVs, we examine changes in the direction of a concept.

\begin{figure}[!htb]
\centering
 \includegraphics[width=1\textwidth]{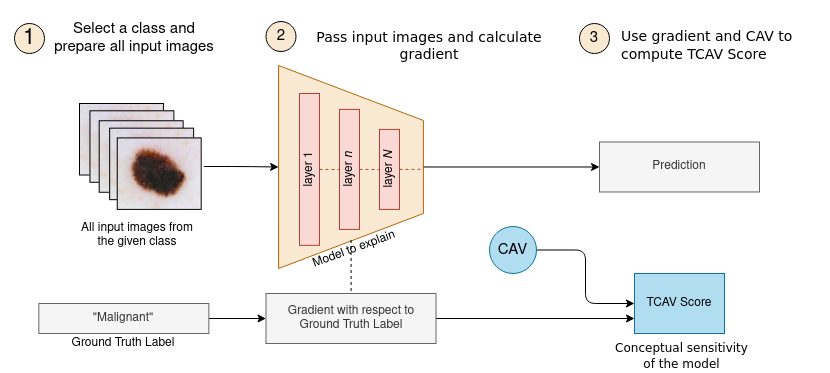}%
 \caption{Testing with Concept Activation Vectors - TCAV workflow.} \label{fig.tcav}
\end{figure}

Hence CAV is defined are per-concept metrics instead of per-feature metrics.
One of the benefits of such an approach is that it does not restrict the model's interpretations to features extracted from training data \cite{kim2018interpretability}.
However, CAVs need preliminary work before running it, like example selection (data annotation) or classifier training. Selecting good examples might also be tricky and challenging, especially when represented concepts are atypical or abstract. Concepts selection is time-consuming and makes it more customizable than other global explanation methods. Training also requires special knowledge, but CAV is relatively easy to use after finishing and allows a careful user to discover hidden errors in the predictor. The workflow for testing with CAV is presented in Figure \ref{fig.tcav}.

\subsection{Spectral Relevance Analysis}
\textit{Spectral Relevance Analysis} (SpRAy) is a semi-supervised method of global explainability that uses local explanations to generate the summarized explanation \cite{lapuschkin2019unmasking, mikolajczyk2021towards}. 
The SpRAy uses attribution maps as the local explanations and aggregates them later with spectral clustering. 
The idea is that clustering on attribution maps will reveal some hidden patterns forming on the attribution maps and allow the user to screen through a large dataset to find co-occurring patterns without manual, time-consuming analysis of individual explanations \cite{mikolajczyk2021towards}.
The final step of SpRAy is a visual inspection of clusters of interest by the user.

The steps of Spectral Relevance Analysis are as follows: 
\begin{enumerate}[\itshape   Step 1:]
\setlength{\itemsep}{1pt}
 \setcounter{enumi}{-1}
 \item Select images and a class to explain.
 \item Compute relevance scores with LRP and generate attribution maps.
 \item Normalize and preprocess the attribution maps.
 \item Perform spectral clustering on normalized attribution maps.
 \item Perform eigengap analysis to find interesting clusters.
 \item (optional). Visualize selected clusters with t-SNE.
\end{enumerate}

The method has great potential, but still, there is room for further improvements leading to better reflecting global dependencies. For instance, the analysis is based only on the relevance maps, which show the region of interest and its intensity. Still, it does not show why the classifier focused on that region. Hence, SpRAy might be biased against localization of attention and shape of the model's attention while completely ignoring the most crucial part of the explanation: why the attention focuses there. Additionally, it tends to concentrate on grouping together images with higher relevance in the same region. As a result, it does not matter what is "under" the attribution. The method analyzes only the heatmaps and does not consider what the actual input image represents, e.g., images with high attribution values in the upper-left corner of the image could be grouped even if the highlighted object is not similar. 
Therefore, in \textit{Chapter \ref{chapter.gebi}: Identifying bias with global explanations}, I proposed approaching the problem of global analysis in a similar yet different manner: to analyze input data and attention maps.

\subsection{Global Aggregations of Local Explanations}
\textit{Local Interpretable Model-Agnostic Explanations }can also be used for global explanations \cite{ribeiro2016why}. Global LIME importance is relatively straightforward: the user is presented with explanations of a few instances that might give a user some insight into the model. The vital step is a selection of cases to show to the user -- a pick step. Selected instances should cover the most significant features, but at the same time, the set of all cases should present various essential traits. As explained in the original paper, we should avoid picking samples with similar explanations \cite{ribeiro2016why}. 
However, as the authors of Global Aggregations of Local Explanations (GALE) explain, the way to aggregate and select local explanations is not straightforward \cite{van2019global}. Mainly because the attribution scores are not determined concerning other data instances. To answer this problem, they proposed a few functions to select a representative and informative subset of samples: Global LIME Importance, Global Average Importance, and Global homogeneity-weighted importance. The method was presented in the NLP example, but it can also be applied to image classification. 

\subsection{Global Attribution Mapping}
\textit{Global Attribution Mapping} (GAM) \cite{ibrahim2019global} uses local, post-hoc attribution techniques to build a global one. Depending on a used local attribution, it can be both model agnostic and model specific. The idea behind the method is to use each local attribution as a weighted conjoined ranking of features and then group similar attributions to find patterns. Each rank is calculated with the weighted Kendall's Tau or Spearman's Rho squared distances. The original method used the LIME \cite{lime}, Integrated Gradients, and DeepLIFT explainability methods as local attribution-mapping techniques. The original GAM method was validated on both real and artificial data and achieved satisfying results.

The steps of the GAM \cite{ibrahim2019global} method are as follows:
\begin{enumerate}[\itshape   Step 1:]
\setlength{\itemsep}{1pt}
 \setcounter{enumi}{-1}
 \item Select images and a class to explain.
 \item Compute attribution maps for samples of the selected class.
 \item Normalize and preprocess attribution maps in the same manner.
 \item Compute pair-wise rank distance matrix.
 \item Perform clustering with rank distances to measure similarity among attributions maps.
 \item Visualize top features.
\end{enumerate}

Global attribution mapping, similarly to  SpRAy method, is relatively easy to grasp and use. Any local explanation method can be used for aggregation, making it customizable and flexible. However, as in any cluster-based method, it is in the user's gesture to analyze and understand received clusters.

The above-mentioned global explainability methods found their application in several works. They help evaluate models and take a great part in the quick data collection analysis. Despite numerous advantages, they are still not as common as they should be, and they do not work well with every type of data. The SpRAy method shows great potential in the potential bias identification. Still, it's oversensitive to shapes on the attribution maps as it analyzes the heatmap alone. In the \textit{Chapter \ref{chapter.gebi}: Identifying bias with global explanations} I propose a few improvements which led to much better results on the investigated dataset of dermoscopic images.
\section{Evaluation of explainability methods}
Although numerous advantages, there are still some registered problems with explainability methods. One of them is often mentioned in subjective manual evaluation, \cite{lin2020you} also called Subjective Ratings \cite{hase2020evaluating} and human-in-the-loop evaluations \cite{das2020opportunities}. Human evaluation is a base method in commonly used algorithms such as Class Saliency Visualisation \cite{simonyan2013deep}, LIME \cite{lime} or GradCAM \cite{selvaraju2017grad}.
However, human-based subjective methods are not only time-consuming but sometimes inaccurate or even might introduce additional bias \cite{buccinca2020proxy}. On the other hand, most currently developed automatic evaluation methods are computational-costly and often inefficient such as methods that test accuracy degradation by perturbing the most relevant region \cite{fong2017interpretable,samek2016evaluating,yeh2019fidelity} . Moreover, some of these methods cause a distribution shift in the testing data, which violates an assumption of training data and the testing data coming from the same distribution \cite{lin2020you,hooker2018benchmark}. A rising number of explanation methods, difficulty in their comparison created, and imperfect early-stage approaches to evaluation created a massive need for better explanation evaluation metrics. Das et al. \cite{das2020opportunities} provide several requirements that an XAI algorithm should meet. Among them are:
\begin{itemize}
 \item \textit{Similarity}: Data instances with similar embeddings should generate similar explanations.
 \item \textit{Separability}: Data instances with dissimilar embeddings should generate different explanations.
 \item \textit{Consistency}: Same data instances with a single change must generate explanations that underline the difference.
 \item \textit{Bias Detection}: Bias in data instances should be detectable.
\end{itemize}

Additionally, Implementation Constraints should be met when designing an XAI method, i.e., computational requirements should be minimal \cite{das2020opportunities}. 

A brief overview of the requirements mentioned above for XAI is presented below, based on the \cite{yang2019benchmarking}.

\subsection{The sensitivity of explanations}
\textit{Sensitivity} evaluates how explanations change when inputs or models change \cite{yang2019benchmarking}.
Two principal metrics are used to measure the sensitivity of explanations: sensitivity and infidelity.
Both are designed to measure how the prediction will change in case of perturbation.
The sensitivity measures what happens to the explanation when a small perturbation to the input is introduced. In contrast, infidelity measures the outcome in case of significant perturbations.
The differentiation between sensitivity and infidelity measures was proposed in 2019 by Yeh et al. \cite{yeh2019fidelity}. In the paper, sensitivity is defined as the change in the explanation with a small (insignificant) input perturbation.

\begin{definition} \textbf{Explanation Sensitivity}
Given a model $f$, functional explanation $\phi$, and a given input neighborhood radius $r$, the max-sensitivity is defined as:
\begin{equation}
    SENS_{MAX }(\phi,f,x,r)= \underset{ ||y - x ||\leqslant r}{max} ||\phi(f,y) - \phi(f,x)||
\end{equation}
where $r$ is a lower and upper bound of the uniform distribution ${\mathcal {U}}\{-r,r\}$.
\end{definition}

In general, lower sensitivity means better explanations.
Explanations with high sensitivity might make users distrust explanations and could be significantly more vulnerable to adversarial attacks, as Ghorbani et al. \cite{ghorbani2019interpretation} show for gradient-based explanations.

The proposed term \textit{infidelity} quantifies the degree to which an explanation capture how the prediction changes when the input is significantly perturbed \cite{yeh2019fidelity}.
The work is motivated by the fact that a "good" explanation should specify how it will show changes in prediction.
The infidelity measure is the expected difference between the two terms: (a) the scalar product of the input perturbation to the explanation and (b) the output perturbation \cite{yeh2019fidelity}. 

\begin{definition} \textbf{Explanation Infidelity}
Given a model $f$, a sample input $x$ and output $y$, functional explanation $\phi$, and a random variable $I \in \mathbb{R}^{d}$ with probability measure $\mu_{I}$ representing meaningful perturbations, the Explanation Infidelity of $\theta$ is defined as:

\begin{equation}
    INFD(\phi,f,x)= \mathbb{E}_{{I} \sim \mu_I }[({I}^T \phi(f,x) - (f(x)-f(x-I))) )^2]
\end{equation}

In the infidelity measure, a random variable $I$ represents the significant perturbations applied to around input $x$. There are four different types of perturbations that are particularly important in terms of fidelity \cite{yeh2019fidelity}:

\begin{itemize}
    \item {\textit{Difference to a baseline}, which is defined as a difference between input $x$ and baseline $x_0$: $I = x - x_0$}
    \item {\textit{Subset of difference to baseline}, which is defined as a difference between selected subset $S_k$ and the input, corresponding to the perturbation $I$ in the correlation measure: $S_k \subseteq [d], I_{S_k} = x - x[x_{S_k} =(X_0)_{S_k}] $}
    \item {\textit{Difference to noisy baseline} that measures the difference between input and a baseline perturbed by the Gaussian Noise: $I = x - z_0$, where 
    $z_0 = x_0 + \epsilon $ and  $\epsilon\sim \mathcal (0,\sigma ^2)$}
    \item {\textit{Difference to multiple baselines} is defined as a difference between input $x$ and multiple baselines $x_{0_n}$: $I = x - x_{0_n}$}
\end{itemize}

\end{definition}

The infidelity measure is usually calculated with two main approaches: using a gaussian random vector as a perturbation (noise baseline) or removing subsets of pixels (square removal) \cite{yeh2019fidelity}.

When designing the explanation method, both metrics should be kept in mind. The process should not be too sensitive to the small, insignificant changes but should show significant perturbations.

\subsection{The correctness of explanations}
One of the most significant matters to evaluate when designing the XAI method is asking whether \textit{the explanation is correct?}
The correctness of explanations can be measured with pre-existing methodologies proposed in the past. One of them is measuring its \textit{completeness}.
 The term completeness was first described in 2015 in the paper that introduced Layer-wise Relevance Propagation \cite{bach2015pixel}, and formalized later on in 2017 by Sundararajan et al. in the paper \textit{"Axiomatic Attribution for Deep Networks"} \cite{pmlrv70sundararajan17a}.
 Completeness is satisfied when the sum of feature attributions equals the difference in the selected input and baseline predictions.
 The goal of completeness is to describe the model or system in an accurate -- but not necessarily easy to understand -- way \cite{gilpin2018explaining}.
 Hence, completeness might be understood as the measure of correctness of the explanations.
 The never-ending problem of ML engineers is the trade-off between accurate explanations and explanations that are easy to understand. Interpretability tries to explain the internals of a model in a way that is understandable to humans and completeness to make them accurate. Completeness is defined as the sum of attributions equals the difference between the models' output at the specified input and the selected baseline.

Later on, in 2018, the metric called \textit{Faithfulness} was introduced by Alvarez-Melis et al. \cite{alvarezmelis2018robust} to answer the question \textit{Are relevance scores indicative of "true" importance?}
The faithfulness can be measured with the simple removal of observing input features or concepts and the model's prediction.
When a feature considered necessary by the explanation is removed, the prediction should also drop and change. This metric allows understanding of the "relevant" features is genuinely relevant.

Another interesting metric that allows measuring if the generated explanation is correct is Monotonicity, introduced in 2019 by \cite{luss2021leveraging}. Monotonicity measures if adding more positive evidence to the input increases classification probability in the specified class \cite{arya2019one}. The inputs associated with positive attributes gradually increase, and the predictions' changes are noted. 

\subsection{Feature importance}
The evaluation with knowledge of feature importance is still a quite fresh field. Early approaches suggested manual attribution maps assessment with the TCAV method described in Section~\ref{subsection.CAV} \cite{kim2018interpretability}. However, it is again a subjective method that can be affected by the tester's knowledge. 
Another approach is a \textit{Relative Feature Importance} (RFI) \cite{yang2019benchmarking} that allows quantitatively measuring and evaluating feature attribution methods. 
The method uses an artificial dataset with well-known relative feature importance. It partially mitigates the problem of ground-truth labels for measuring feature importance. Additionally, relative feature importance can use a more straightforward test for more advanced issues without handcrafting or annotating new datasets. 
Some might reject the method concluding that one should not be certain that assigning correct attributions to an artificial dataset will do the same for authentic images. The authors rationalize the approach by saying that if attribution fails an easy test, it will likely fail the difficult one. 
Relative feature importance uses three other metrics to compute it:
\begin{itemize}
    \item \textit{Model contrast score} measures the difference in attributions between two models given the same input. The model contrast score (MCS) is equal to the difference of concept attributions between a model $f_1$ that considers a certain concept more important than another model $f_2$:
    \begin{equation}
        MCS = G_c(f_1,X_{corr})-G_c(f_2,X_{corr})
    \end{equation}
    where $G_c$ is a concept attribution, and $X_{corr}$ is a set of correctly classified inputs.
    \item \textit{Input dependence rate} measures the difference in attributions between two different inputs to the same model. Input dependence rate (IDR) is defined as the percentage of correctly classified inputs where Common Features (CF - a set of pixels with some semantic meaning) is attributed less than the original regions covered by CF. 
    \begin{equation}
        IDR = \frac{1}{|X_{cf}|}_{(x_{cf},x_{\sim cf}) \in (X_{cf},X_{\sim cf}))}\sum \mathds{1}(g_c(f,x_{cf}) < g_c(f,x_{\sim cf}))
    \end{equation}
    where $g_c$ is an average attribution of pixels in region $c$, $cf$ are common features, $f$ is a classifier, $x_{cf}$ and $x_{\sim cf}$ are inputs with and without $cf$.
    \item  \textit{Input independence rate}  measures the difference in attributions between two functionally similar inputs. Input Independence Rate (IIR) expects similar attributions between two inputs that lead to the same prediction.
    \begin{equation}
        IIR = arg \underset{\delta}{min} || f(x+\delta)||_2 - \eta_1 ||\delta||_2 + \textsl{R}
    \end{equation}
    where the formula $- \eta_1 ||\delta||_2$ force  $\delta \neq 0$, and $\textsl{R}$ is an additional regularization term that encourages perturbation $\delta$ to look more realistic looking. See orignal paper for more insight \cite{yang2019benchmarking}.
\end{itemize}

\subsection{Humans in the loop}
This group of evaluation approaches is fulfilled by evaluating whether users can correctly process a selected task with the insight from explainability methods or by assessing how often users can accurately predict model behavior after seeing the explanations \cite{yang2019benchmarking}. Such studies use, for instance, human judgment as to the evaluation criteria of explanation methods \cite{lakkaraju2016interpretable}. 

Lakkaraju et al.\cite{lakkaraju2016interpretable} proposed asking the users descriptive questions or multiple choice questions to evaluate the extent to which humans can understand the model's decisions when using XAI.
An example of a descriptive question would be: "Please write a short paragraph describing the characteristics of malignant skin lesion based on the attribution maps provided above.". 
The average time spent answering questions and the mean number of words used to write a descriptive answer were used to review the response.

Another approach proposed by Biessmann et al. \cite{biessmann2021quality} used XAI to support annotators during the labeling procedure. They measured the annotation accuracy and speed and then compared metrics along with different XAI approaches. Plain GradCAM explanations led to the lowest annotation accuracy, LRP achieved slightly higher results, and  Guided BackProp explanations resulted in the highest metrics. Biessmann shows that human-in-the-loop evaluation is necessary as other metrics did not precisely determine which explanation method was the best. 

Although many evaluation methods appeared in the past, there is still a significant lack of widely accepted benchmarks and baselines designed for explainability methods. New methods appear but are often rejected by the researchers who criticize the approach and propose new, supposedly better metrics. In most cases, the evaluation is not measured and is based on subjective human evaluation. There is also a need for methods designed purposely for global explanations like SpRAy or CAVs. Because there is no widely accepted standard for such evaluation, I decided to assess the proposed method in Chapter \ref{chapter.gebi} with a human evaluation of clustered embedding space evaluation. Additionally, discovered patterns were examined to determine whether and how much they impact the predictions. The details are presented in the \textit{Chapter \ref{chapter.gebi}: Identifying bias with global explanations.}





\clearpage  
\lhead{\emph{Chapter 4: Identifying bias with manual data inspection}}  
\chapter{Identifying bias with manual data inspection} \label{chapter.skin_lesion_bias}
\section{Introduction}

If we refer to the model as the main "drive" of the machine learning systems, the data would be the fuel that powers it. It is necessary to have good quality data to run models: how can we expect the engine to work well with contaminated fuel? Similarly, we cannot expect 90\% accuracy when a 15\% of data is incorrectly labeled. Or if it is inconsistent or biased. 

It might seem that widely accepted benchmarks are perfectly prepared, and we do not have to worry about labeling errors or annotation consistency. However, the research shows that even the most common data collections are mislabeled. Pleiss et al. show that, on average, the labeling error rate is equal to 3.3\% \cite{pleiss2020identifying}. In the case of weakly-labeled datasets, the error reaches over 17\% for \textit{WebVision50}, the data collection of 2 million images scraped from \textit{Flickr}, and \textit{Google Image Search} downloaded using search queries \cite{li2017webvision}. Their research shows that removing mislabeled data, even at the cost of losing some data points, always improves the model's error rate. They prove that mislabeled samples can be identified automatically by measuring a sample's contribution to generalization with margins. Additionally, they present a website showing each discovered mislabeled sample \footnote{Label errors in commonly used ML benchmarks: https://labelerrors.com/}.

\begin{figure*}[!htb]
\centering
\includegraphics[width=\linewidth]{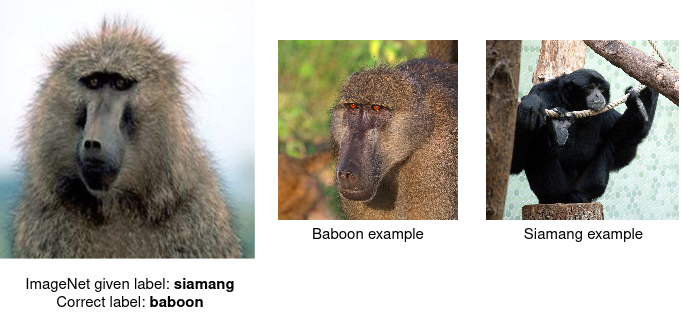}
  \caption{Mislabeled samples problem. ImageNet example discovered by \cite{pleiss2020identifying}.}
  \label{fig.data.mislabeling}
\end{figure*}

\begin{figure*}[!htb]
\centering
\includegraphics[width=\linewidth]{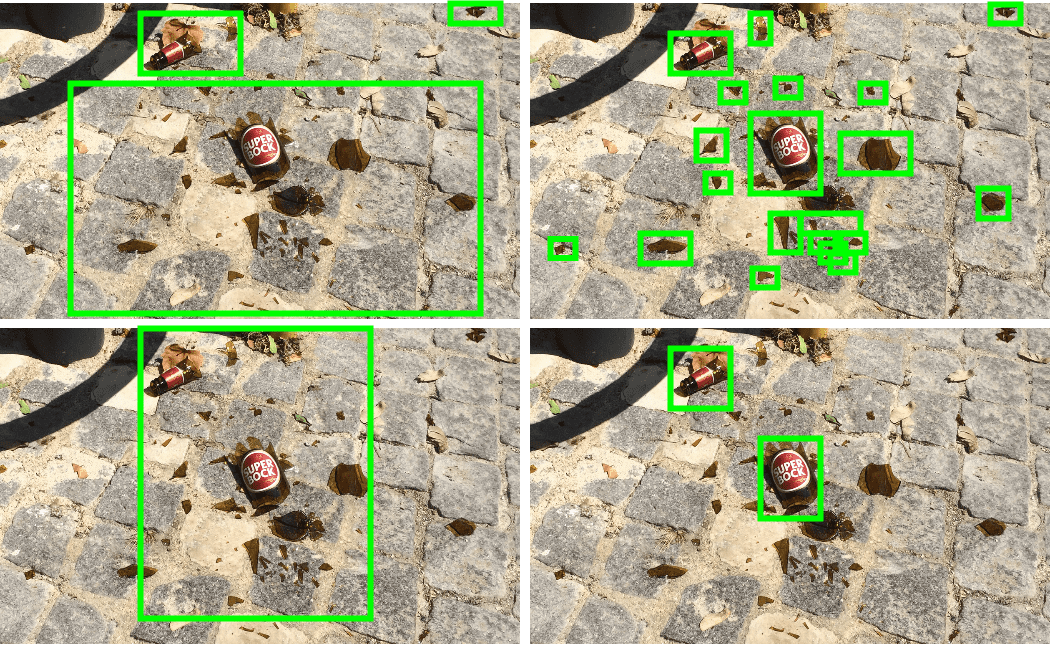}
  \caption{Labeling consistency problem. Waste detection example: how to label broken glass bottles? Annotated image come from the TACO dataset \cite{proencca2020taco}}
  \label{fig.data.labeling_consistency}
\end{figure*}

The problems in data do not end in mislabeled data. Data might be labeled correctly, and yet, we might not be able to train the model properly in the case of label inconsistency. In that case, it is worth checking inter-annotator agreement: e.g., select a slice of data for labeling by two or more annotators and compare results. The annotation guidelines should be investigated thoroughly in case of high differences between annotators. Each annotator should label the data in the same and consistent manner. In many projects, the annotation guideline preparation is not a single-step task but a continuous development process. The inconsistencies are often found during the annotation process or even after when the training results are unsatisfactory. An example of data labeling inconsistency occurred for me personally during a volunteer project for detecting waste in images and video \footnote{\textit{Detect waste in Pomerania}: Non-profit project for detecting waste in the environment. detectwaste.ml}. Annotation seemed straightforward at first: one bounding box per object (e.g., bottle) and corresponding category (e.g., plastic). Problematic cases included occlusion, very long objects, and partially visible or decomposed waste. Even stashed objects like a pile of trash were quite common (should each object be annotated separately or as a whole pile with a single bounding box?). Next, in the case of a scattered bottle, should each piece of glass is annotated separately or shattered glass as a single, destroyed bottle? We cannot expect good results if annotators label it differently (see the example in Figure~\ref{fig.data.labeling_consistency}). However, we cannot always reach out to annotators, especially when using open-source, public benchmarks. Some approaches were also proposed to tackle this problem, e.g., by Zeng et al. \cite{zeng2021validating} who proposed automated validation of label consistency in named entity recognition datasets.

The next problem is bias in data. Even high-quality data might be biased during the data collection, preparation, or analysis. Those biases are not uncommon and can be found in various datasets. Sometimes it is easy to spot the problem, but the problems are well hidden in most cases, as the model produces high-accuracy results. Stock et al. \cite{stock2018convnets} discovered a biased problem in ImageNet in recognizing basketball, ping pong, and volleyball players. The uneven distribution of players' skin color increased several times, resulting in the classifier that focuses more on the player's skin color rather than the ball, t-shirt, or background \cite{stock2018convnets}. 

Above mentioned problems showed that manual data inspection is often necessary. Data should be examined regarding labeling quality, consistency, and bias. I presented the time-consuming process of extensive data exploration toward the third problem: bias in data. 

In \textit{Chapter~\ref{chapter.bias_in_ML}: bias in machine learning pipeline}, I presented over 50 different biases connected to how the data was collected, annotated, and analyzed. Here, I frequently used the term' bias in data, by which I mainly referred to the four most common data biases in machine learning: 
\begin{itemize}
    \item \textit{observer bias} \cite{mahtani2018catalogue} which might appear when annotators use personal opinion to label data,
    \item \textit{sampling bias} when data is acquired in such a way that not all samples have the same sampling probability \cite{mehrabi2019survey},
    \item  \textit{data handling bias} when how data is handled distort the classifier's output,
    \item \textit{instrument bias} meaning imperfections in the instrument or method used to collect the data \cite{he2012bias}.
\end{itemize}

Additionally, I explored the subject of the \textit{shape vs. texture bias}, coming from the category of algorithmic biases, which can be encountered when using pretrained convolutional neural networks. Those biases will be explored in the case study of the skin lesion dataset.

I analyzed, described, and proposed a set of possibly biasing artifacts in the skin lesion dataset. Next, I manually annotated four thousand dermoscopic images (two thousand per class) regarding the presence of artifacts. Examined artifacts are comprehensively described along with annotation process details, and image examples.
Using annotations, I focused on analyzing the correlations between the artifacts presence and lesion types, artifacts distribution per class, and correlation between artifacts themselves. This study was designed to help discover whether the data is biased. Additionally, I examined how efficiently a model can be trained to classify skin lesions based only on the skin and artifacts (masked skin lesions) compared to standard training. This experiment helps to understand how well the model can predict the skin lesion type without seeing the actual lesion and as effect, how strongly data is biased.

\section{Skin lesion dataset}

\begin{figure}[!htb]
\centering
  \includegraphics[width=0.8\textwidth]{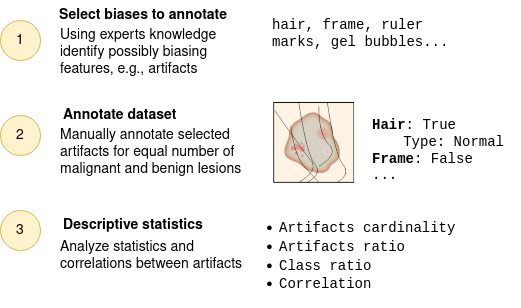}%
  \caption{The process of data annotation and examination.}
  \label{fig.manual-annotation}
\end{figure}

The artifacts in the skin lesion datasets are a broadly mentioned subject in a skin lesion classification. Yet, it is still unknown how significant is the problem. For a better understanding of a skin lesion dataset, or more precisely, the distribution of the artifacts, I have manually annotated 4000 images of skin lesions from ISIC 2019, and 2020 \cite{isic2019BCN20000,isic2019codella,isic2019ham10000,isic2020}. The data annotation and examination process is presented in Figure~\ref{fig.manual-annotation}.

In the manual annotation process, the selection of images for labeling was random. Each image was annotated in a binary multi-label fashion, i.e., if an artifact appeared, its value was set to $True$. The total count of present artifacts was not calculated (e.g., if a gel bubble and gel border appeared together, it was still annotated as $other = True$). Each image can have zero, one, or multiple artifacts annotated. Each image was annotated based on the artifacts I detected as potentially harmful: hair, frames, ruler marks, and other artifacts. The detailed statistics are presented in Section~\ref{section:statistcis}. 

The details behind the annotation are as follows.

\begin{figure*}[!htb]
\centering
\textbf{Images with hair}

    \begin{subfigure}[Dense hair]{
      \includegraphics[width=0.25\linewidth]{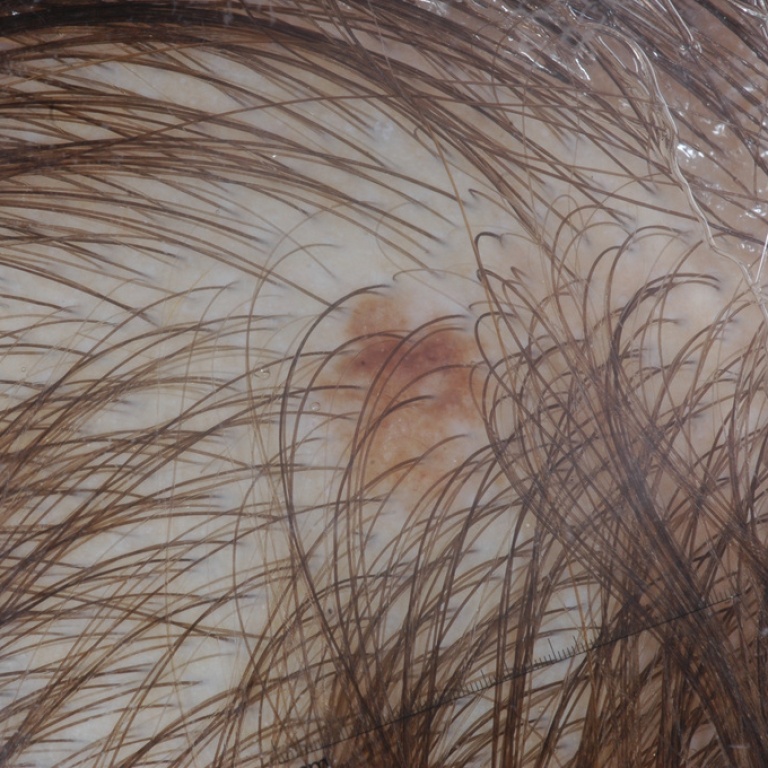}
      \includegraphics[width=0.25\linewidth]{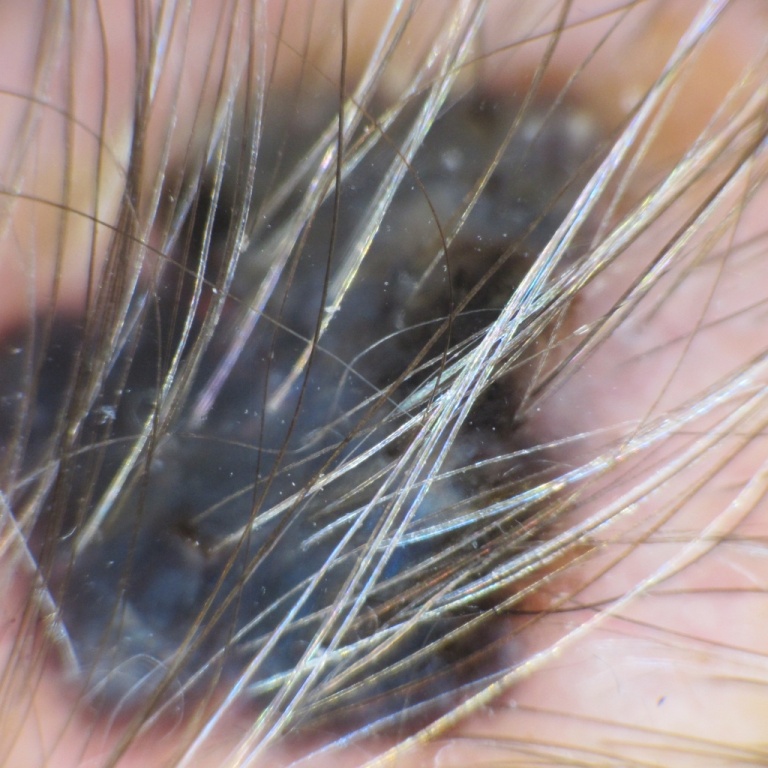}
      \includegraphics[width=0.25\linewidth]{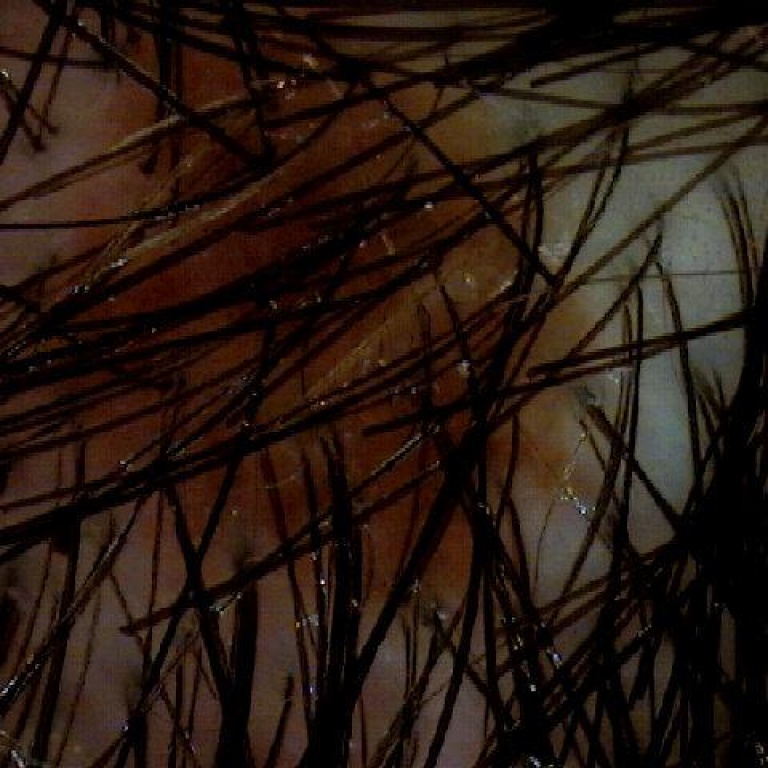}}
    \end{subfigure}
    \qquad
    \begin{subfigure}[Short hair]{
      \includegraphics[width=0.25\linewidth]{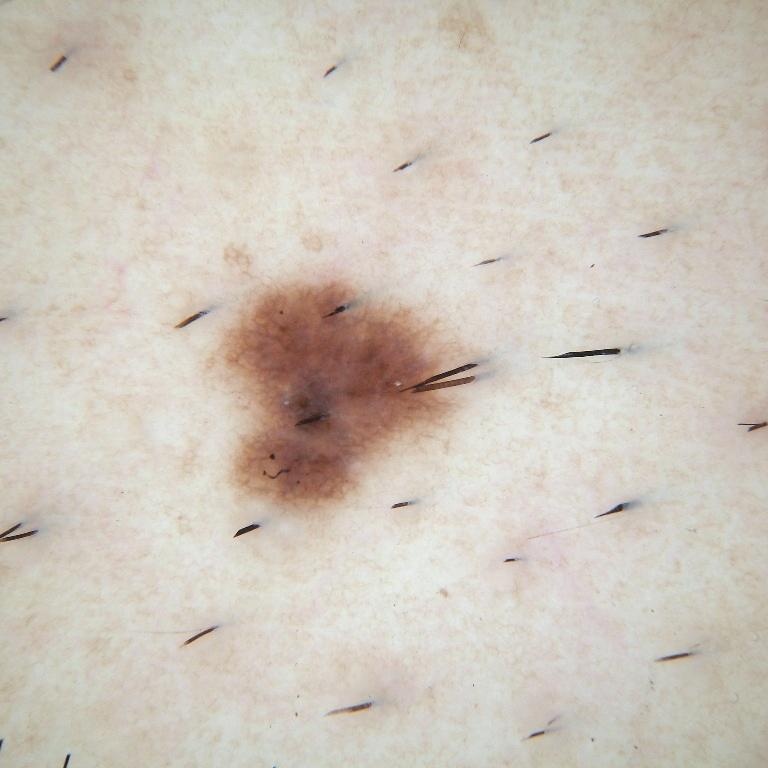}
      \includegraphics[width=0.25\linewidth]{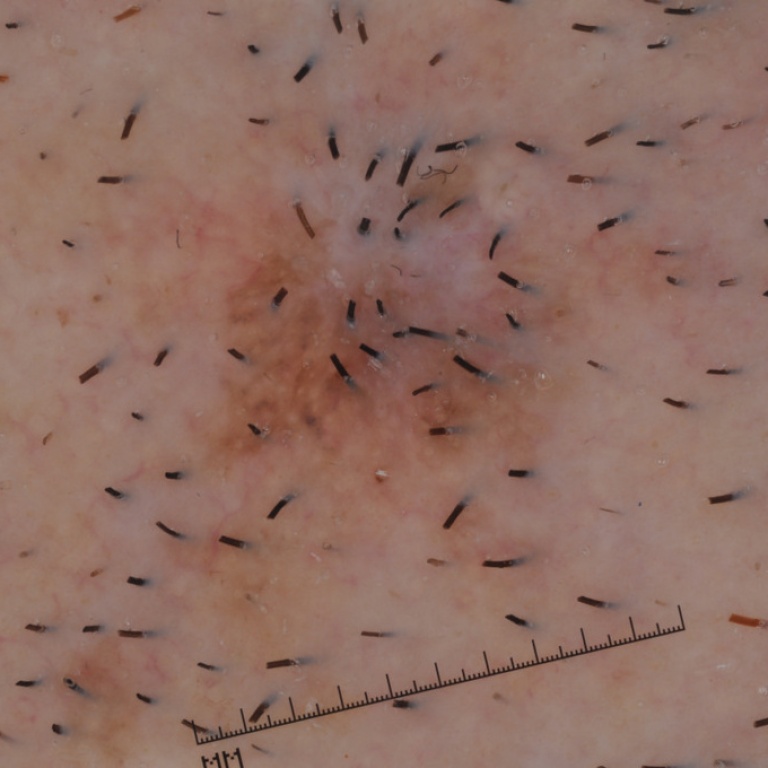}
      \includegraphics[width=0.25\linewidth]{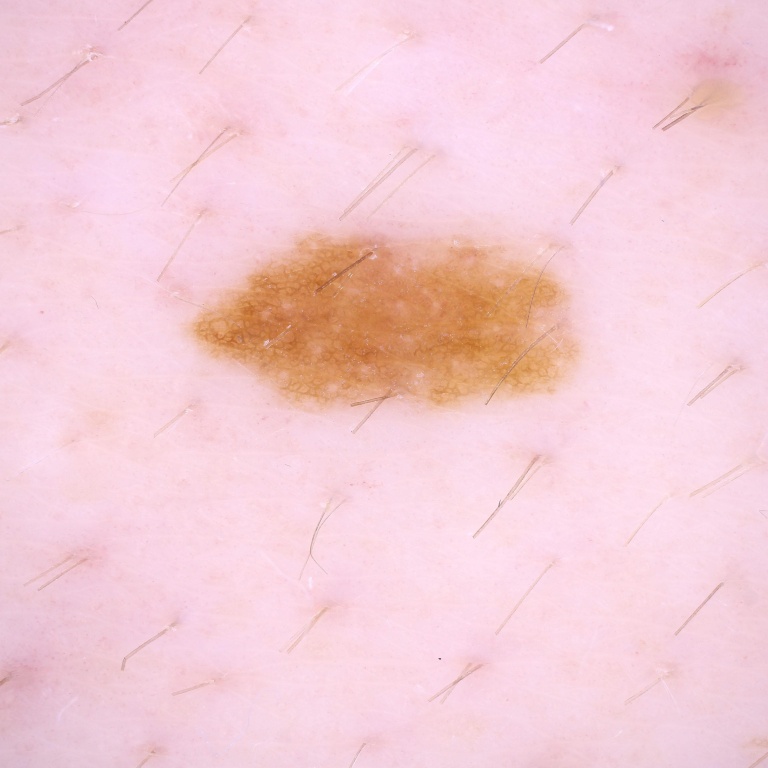}}
    \end{subfigure}
    \qquad
    \begin{subfigure}[Normal hair]{
      \includegraphics[width=0.25\linewidth]{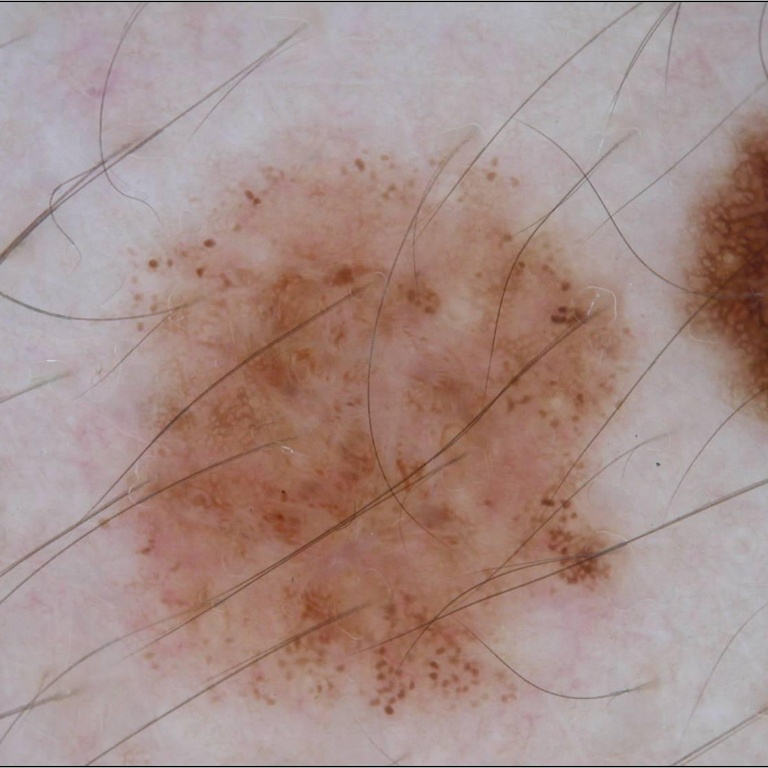}
      \includegraphics[width=0.25\linewidth]{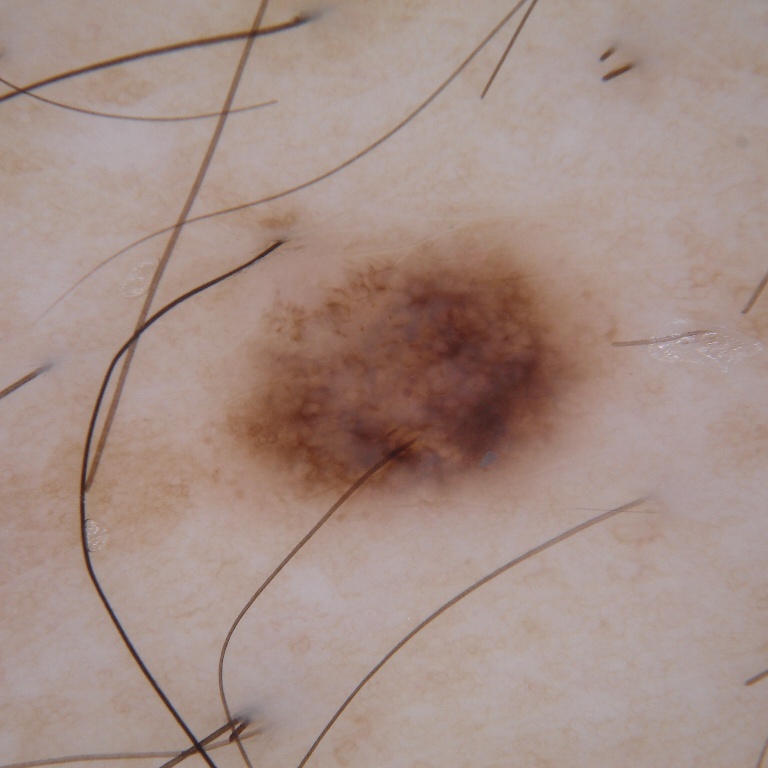}
      \includegraphics[width=0.25\linewidth]{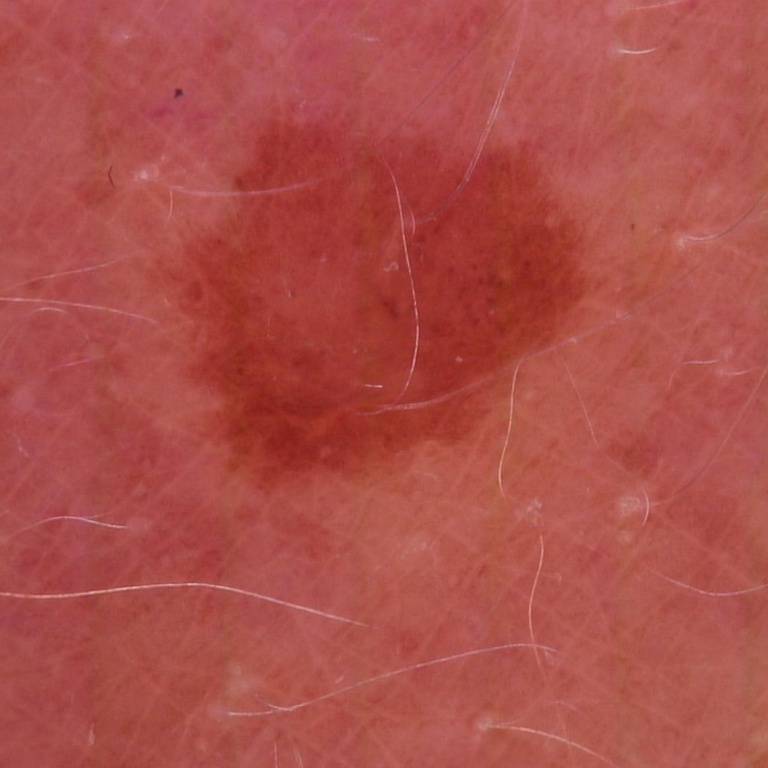}}
    \end{subfigure}
  \caption{Example of images with hair.}
  \label{fig.artifacts.other}
\end{figure*}

\begin{figure*}[!htb]
\centering
\textbf{Images with other artifacts}

    \begin{subfigure}[Gel bubbles]{
      \includegraphics[width=0.25\linewidth]{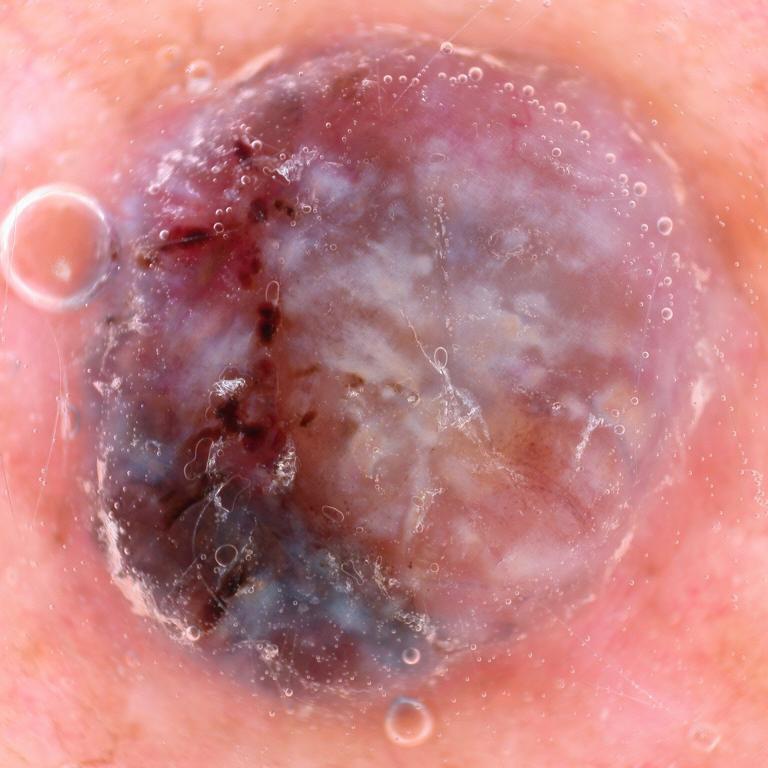}}
    \end{subfigure}
    \qquad
    \begin{subfigure}[Numbers or dates]{
        \includegraphics[width=0.25\linewidth]{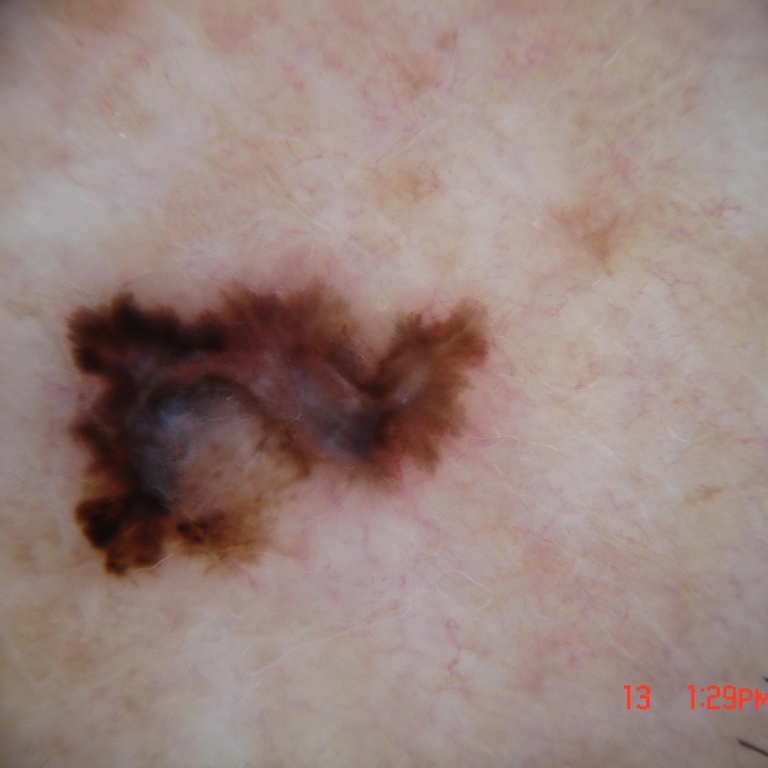}} 
    \end{subfigure}
    \qquad
    \begin{subfigure}[Patches]{
        \includegraphics[width=0.25\linewidth]{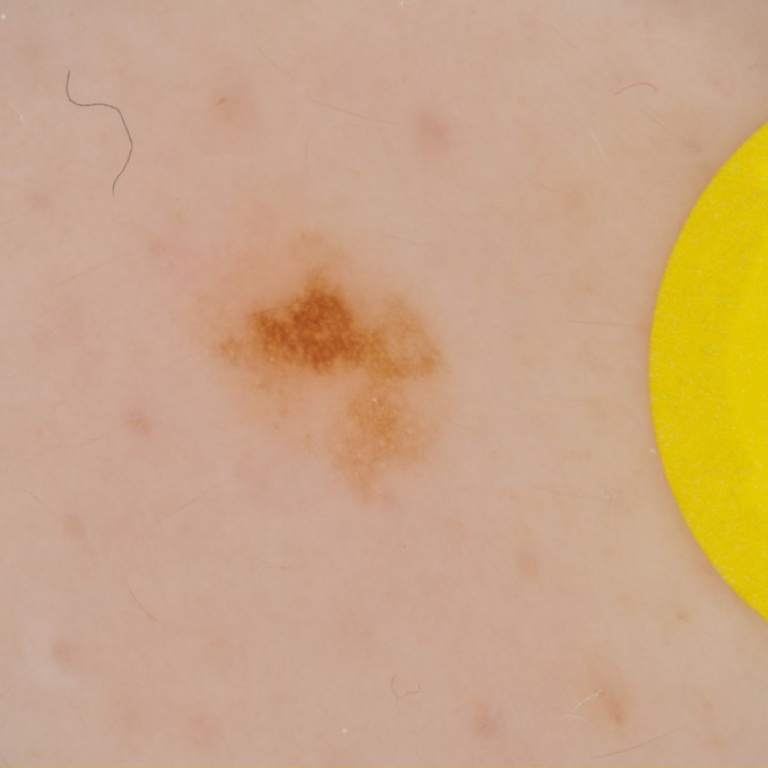}}
    \end{subfigure}
    \qquad
    \begin{subfigure}[Dust]{
      \includegraphics[width=0.25\linewidth]{Figures/artifacts/ruler/ISIC_0000516.jpg}}
     \end{subfigure}
     \qquad
    \begin{subfigure}[Ink circle]{
      \includegraphics[width=0.25\linewidth]{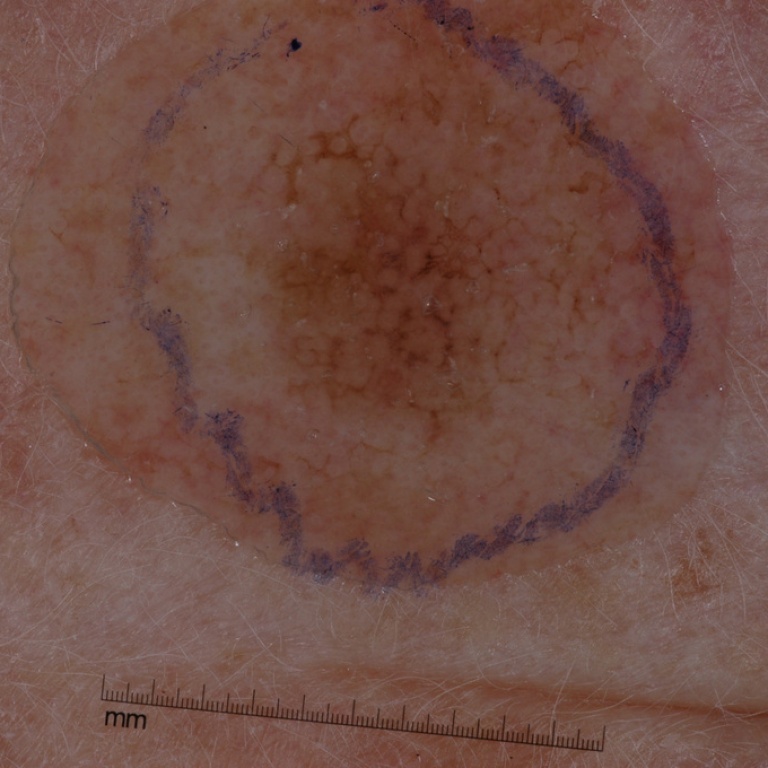}}
    \end{subfigure}
    \qquad
    \begin{subfigure}[Ink stains]{
      \includegraphics[width=0.25\linewidth]{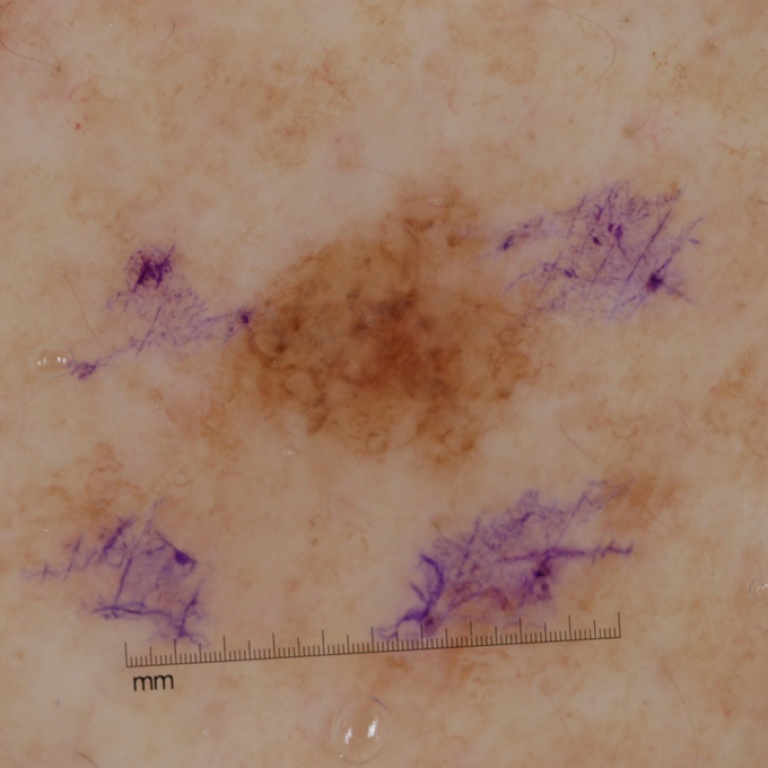}}
    \end{subfigure}
    \qquad
    \begin{subfigure}[Paper]{
      \includegraphics[width=0.25\linewidth]{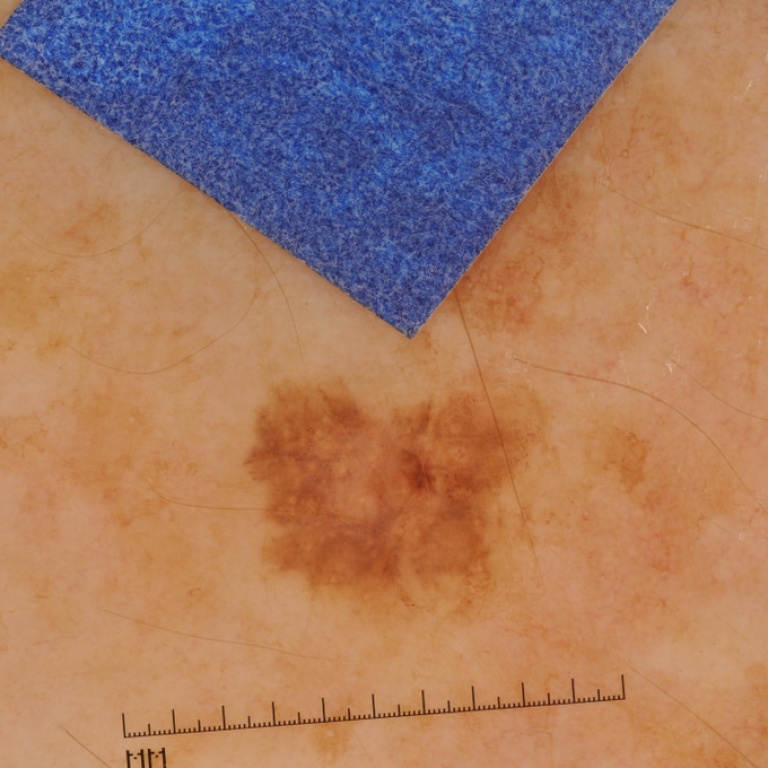}}
    \end{subfigure}
    \qquad
    \begin{subfigure}[Gel border]{
      \includegraphics[width=0.25\linewidth]{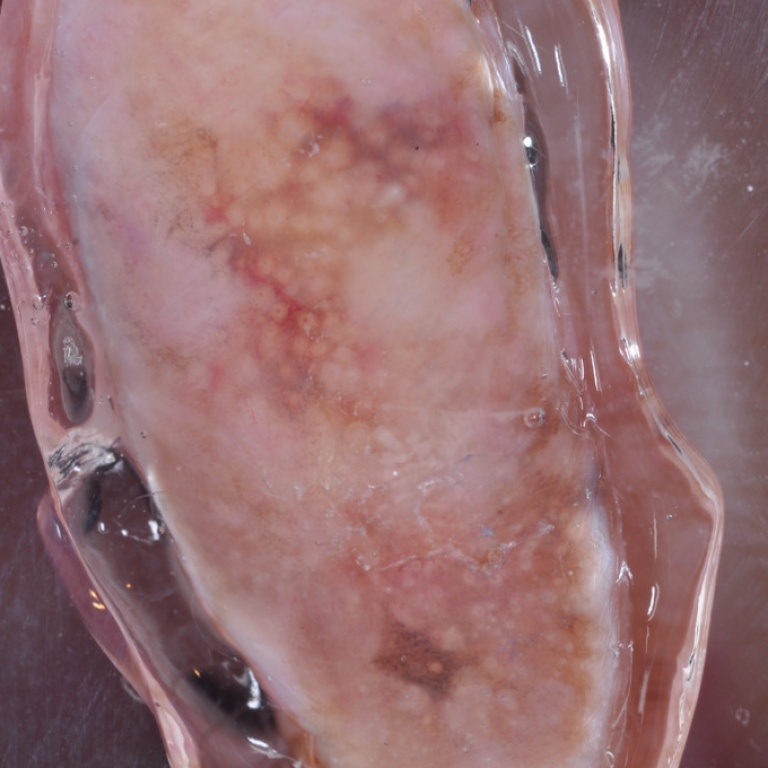}}
    \end{subfigure}
    \qquad
    \begin{subfigure}[Visible background]{
      \includegraphics[width=0.25\linewidth]{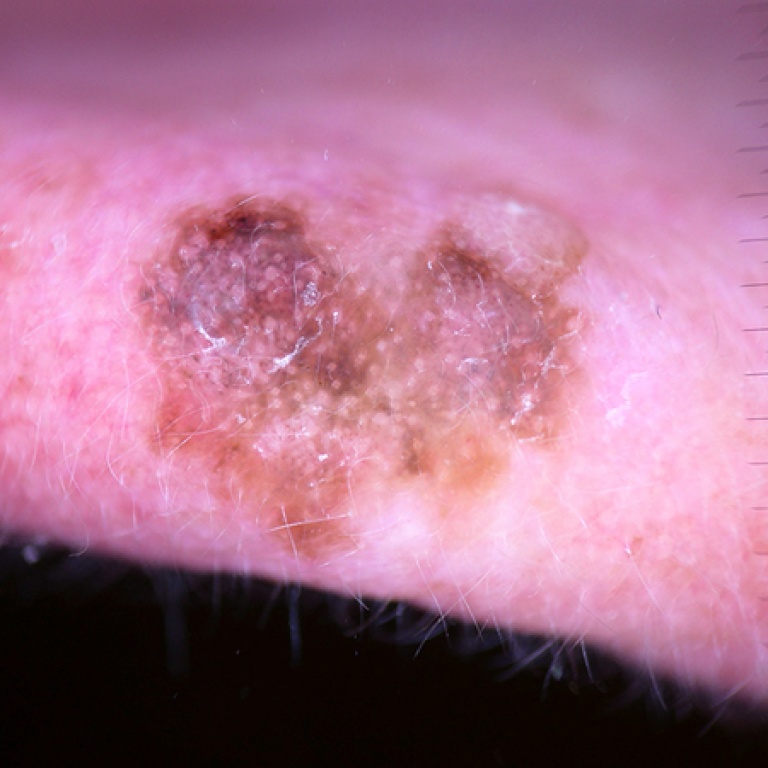}}
    \end{subfigure}
  \caption{Example of images with most common artifacts that were annotated as $other$.}
  \label{fig.artifacts.hair}
\end{figure*}

\textit{Hair} was annotated for both thick and thin hair of various colors from light blonde to black. Additionally, I added information on whether hair looks typical (normal hair) or not. Dense hair images where hair covers a significant part of an image and short hair that can be easily mistaken with some atypical, differential structures of skin lesions were annotated, as in Figure~\ref{fig.artifacts.hair}.
\begin{itemize}
    \item \textit{Normal hair} -- A typical hair of various colors and thicknesses.
    \item \textit{Short hair} -- short or shaved hair, usually dark, can easily be mistaken for some atypical, differential structures of skin lesions.
    \item \textit{Dense hair} -- dense or very thick hair covering a significant part of an image. Usually, the lesion with dense hair is located on the patient's head.
    \end{itemize}

\begin{figure*}[!htb]
\centering
\textbf{Images with frames}

    \begin{subfigure}[Round black frame]{
      \includegraphics[width=0.25\linewidth]{Figures/artifacts/frame/ISIC_0000004.jpg}}
    \end{subfigure}
    \qquad
    \begin{subfigure}[Very thin edge frame]{
        \includegraphics[width=0.25\linewidth]{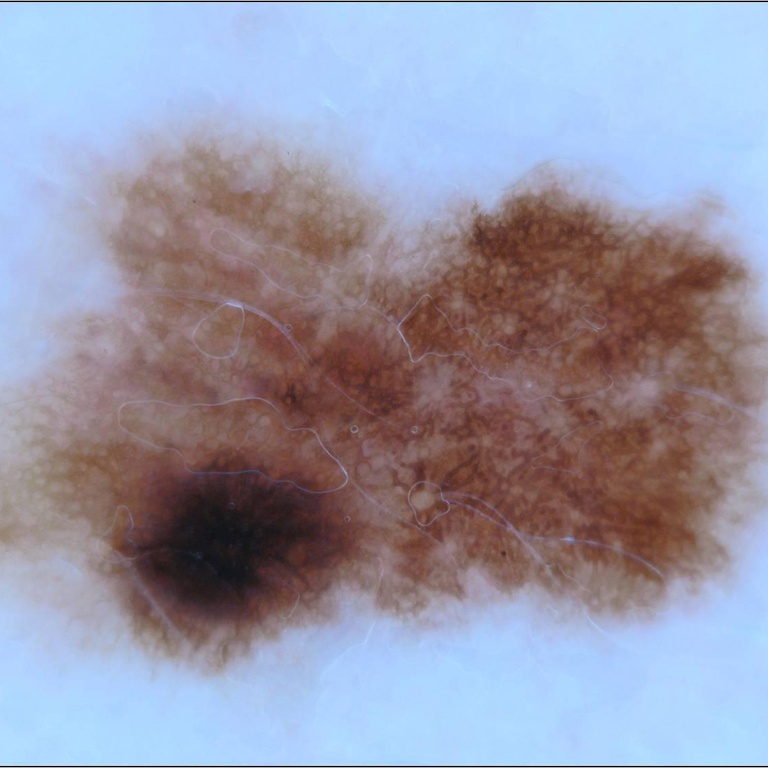}} 
    \end{subfigure}
    \qquad
    \begin{subfigure}[Vignette]{
        \includegraphics[width=0.25\linewidth]{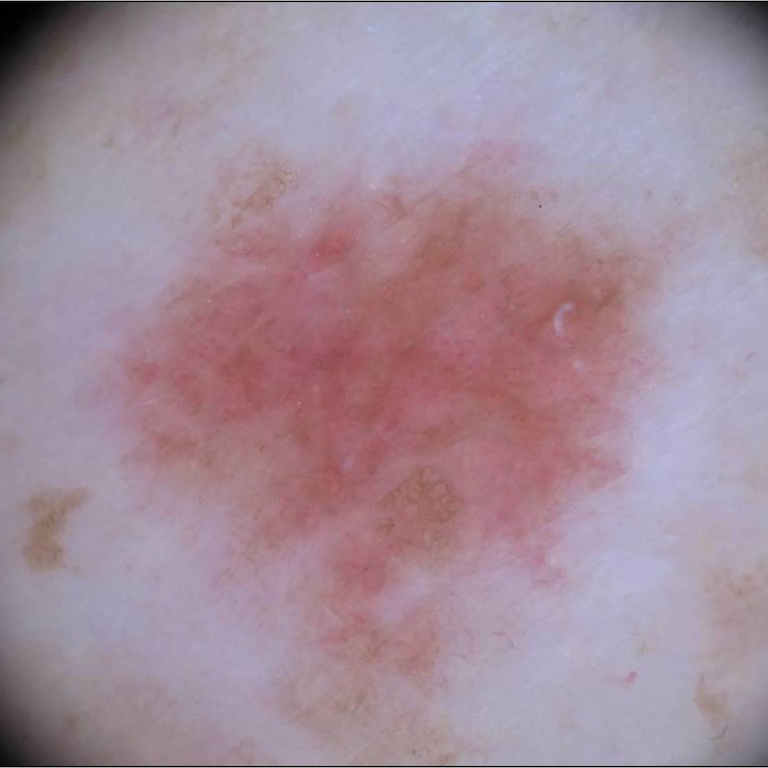}}
    \end{subfigure}
    \qquad
    \begin{subfigure}[Rectangular black frame]{
      \includegraphics[width=0.25\linewidth]{Figures/artifacts/frame/ISIC_0000170.jpg}}
     \end{subfigure}
     \qquad
    \begin{subfigure}[White round frame]{
      \includegraphics[width=0.25\linewidth]{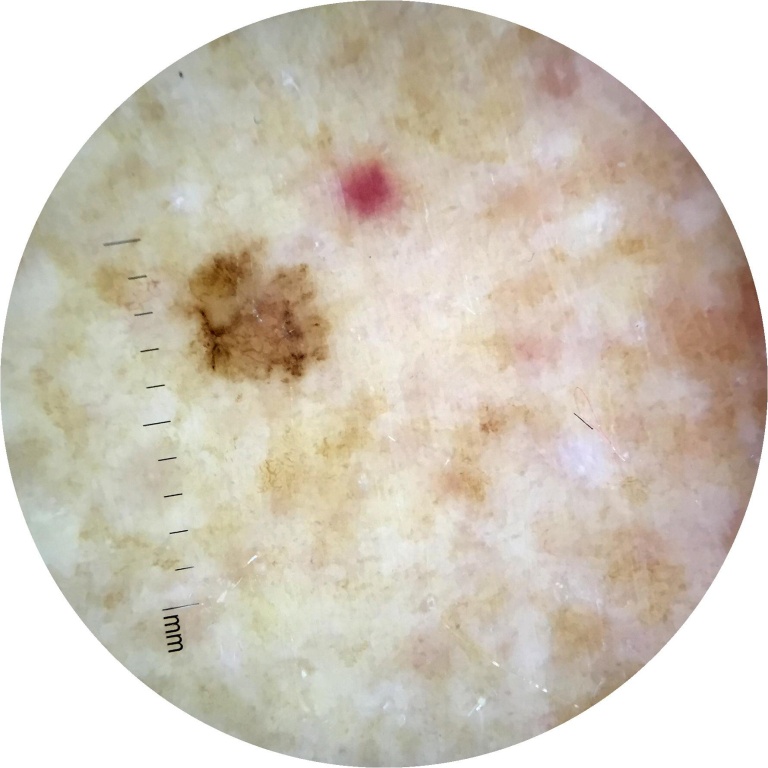}}
    \end{subfigure}
    \qquad
    \begin{subfigure}[Large round frame]{
      \includegraphics[width=0.25\linewidth]{Figures/artifacts/frame/ISIC_0059274.jpg}}
    \end{subfigure}
    
  \caption{Example of images with frames.}
  \label{fig.artifacts.frames}
\end{figure*}
\begin{figure*}[!htb]
\centering
\textbf{Images with ruler marks}

    \begin{subfigure}[Poorly visible ruler (right corner)]{
      \includegraphics[width=0.25\linewidth]{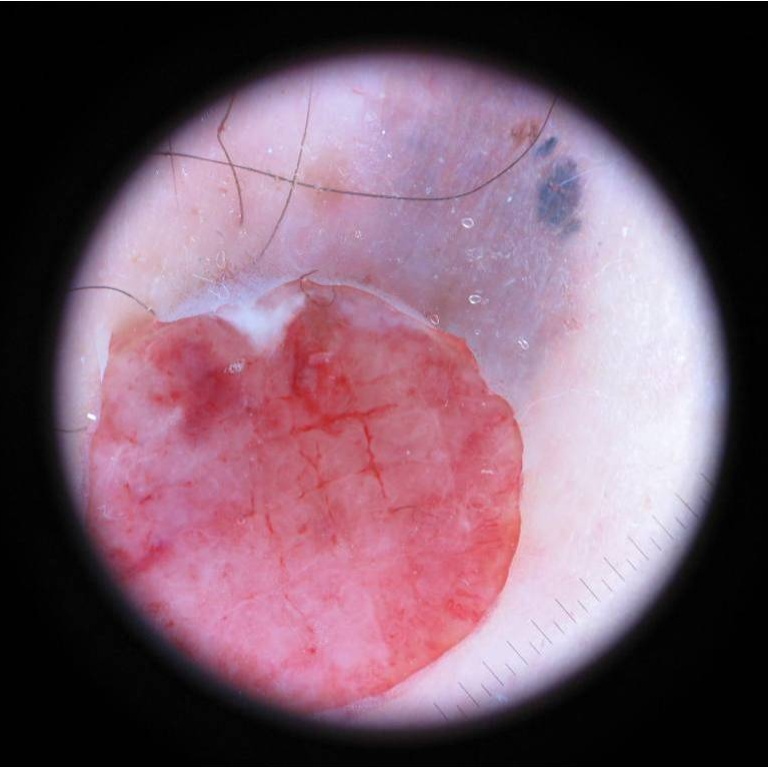}}
    \end{subfigure}
    \qquad
    \begin{subfigure}[Thick ruler with white background]{
        \includegraphics[width=0.25\linewidth]{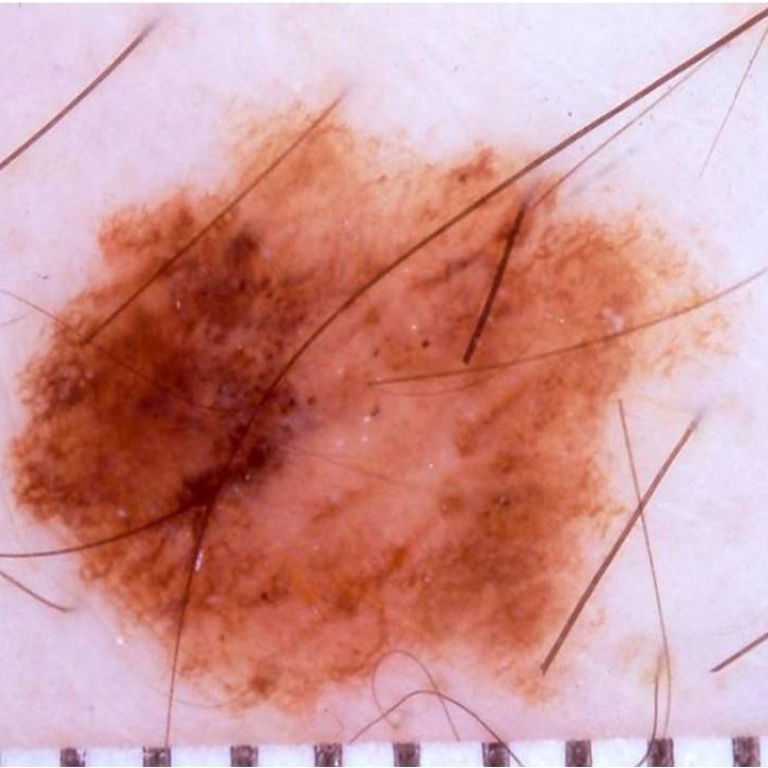}} 
    \end{subfigure}
    \qquad
    \begin{subfigure}[Standard ruler]{
        \includegraphics[width=0.25\linewidth]{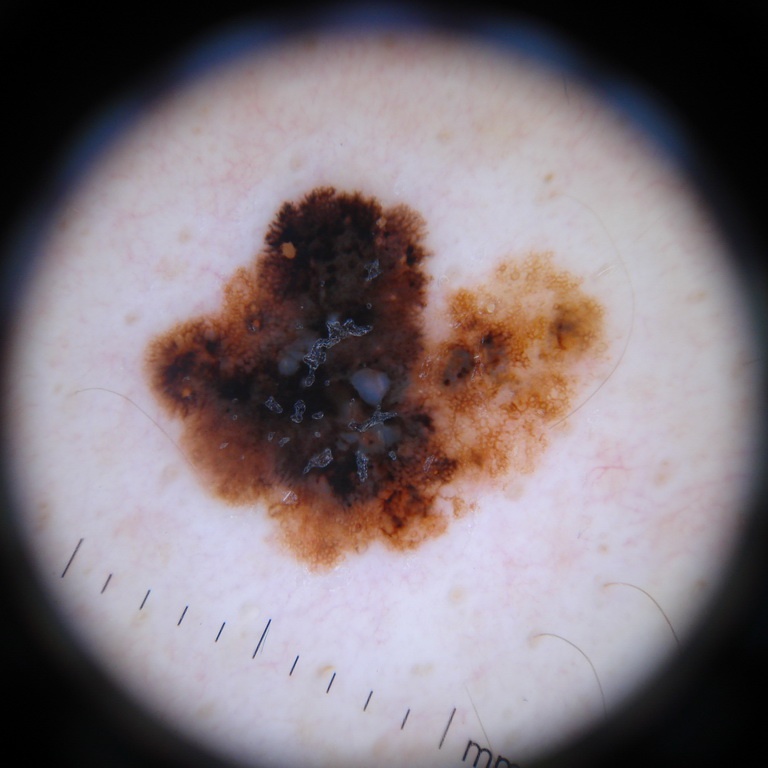}}
    \end{subfigure}
    \qquad
    \begin{subfigure}[Fully visible]{
      \includegraphics[width=0.25\linewidth]{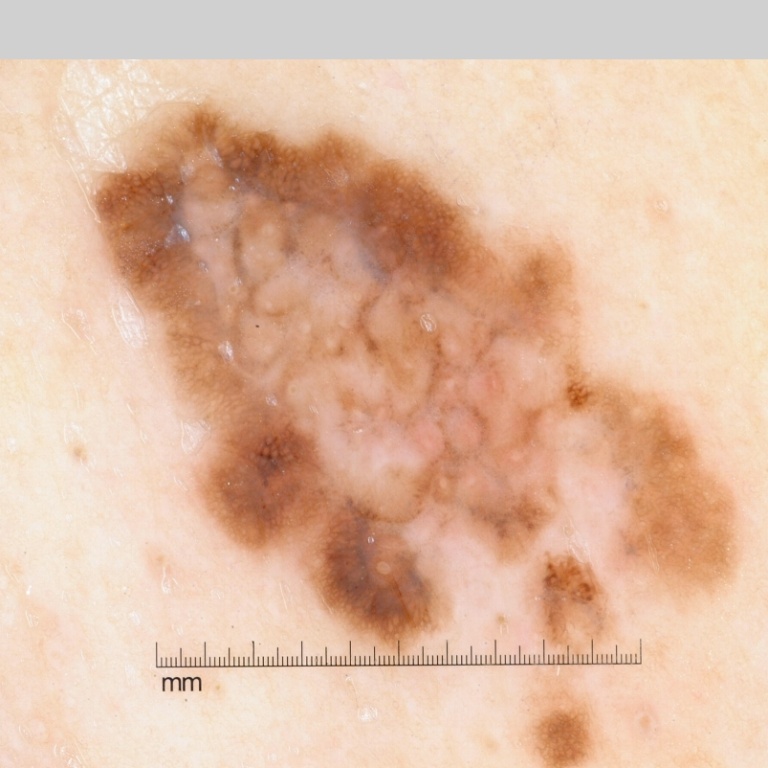}}
     \end{subfigure}
     \qquad
    \begin{subfigure}[Partially visible]{
      \includegraphics[width=0.25\linewidth]{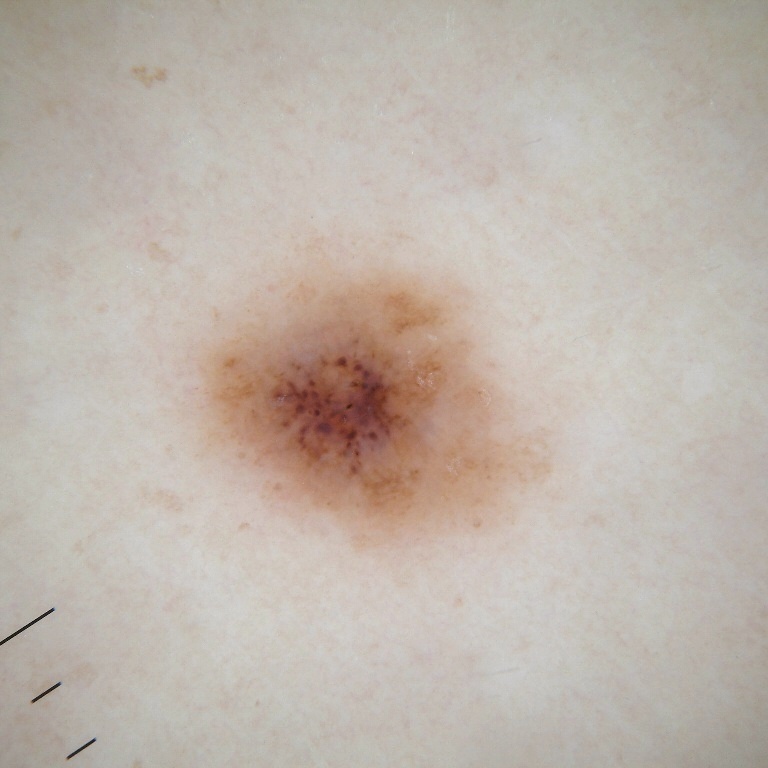}}
    \end{subfigure}
    \qquad
    \begin{subfigure}[Large with white backgroun]{
      \includegraphics[width=0.25\linewidth]{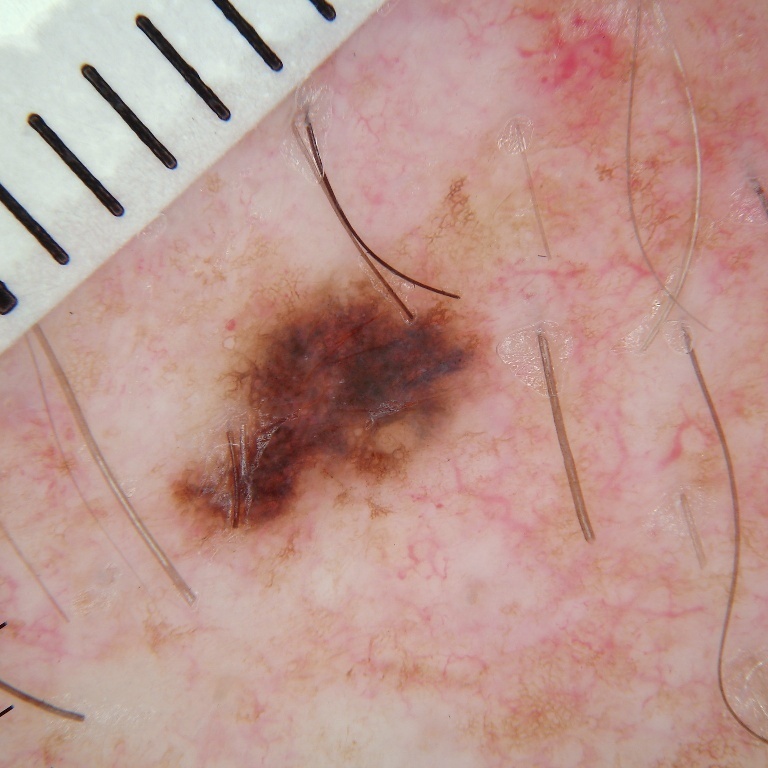}}
    \end{subfigure}
    
  \caption{Example of images with ruler marks.}
  \label{fig.artifacts.ruler}
\end{figure*}
    
\textit{Frames} are defined as both black and white round markings around the skin lesion, black rectangle edges, and vignette are defined as frames. Annotated examples are presented in Figure~\ref{fig.artifacts.frames}):
\begin{itemize}
    \item \textit{Round frames -- }frames with a round or oval shape. Often black, but sometimes can also be white. Come in different sizes, i.e., sometimes they cover only the corners, and sometimes the entire circle is visible.
    \item \textit{Rectangular frames -- } usually two gray or black rectangles at the upper and bottom part of the image. Like the round frames, they come in various sizes and can be thick and thin (only a few pixels).
    \item \textit{Vignette --} a photographic effect of darker border, blur, or a shadow at the periphery of the image. 
\end{itemize}

\textit{Ruler marks} are partially visible or evident ruler marks of different shapes and colors, as presented in Figure~\ref{fig.artifacts.ruler}). The annotation process included:
\begin{itemize}
    \item \textit{Physical ruler} -- a ruler that is physically placed near the skin lesion by the doctor. Usually thicker, with white background.
    \item\textit{ Built-in ruler} -- a physical ruler, built-in in the dermoscope. Usually less visible and transparent.
    \item \textit{Digital ruler} -- a ruler added with digital planimetry. Usually less visible, with no background.
\end{itemize}

\textit{Other artifacts} include any other artifacts that were not mentioned previously, including gel bubbles and borders, ink, patches, dates and numbers, parts of the background, light reflection, and dust. The annotated examples are presented in Figure~\ref{fig.artifacts.other}). In some cases\textit{other} artifacts are placed intentionally (surgical pen marking) by clinicians or accidentally by a person who photographed a lesion.
\begin{itemize}
    \item \textit{Dermoscopic gel or air bubbles} -- visible gel bubbles made out of isopropanol, liquid paraffin, ultrasound gel, or other substance, used during the dermoscopic procedure to provide better visibility of the skin lesion \cite{gewirtzman2003evaluation}.
    \item \textit{Ink or surgical pen markings} -- clinicians stain skin lesions with a violet surgical pen and wipe them off later. If the ring of ink remains, it might mean possible skin lesion disease (porokeratosis) \cite{navarrete2019ink}.
    \item \textit{Patches and papers} -- round or rectangular color patches are used for color calibration of dermoscopy images. They are usually only partially visible in the side or corner of the image. 
    \item \textit{Dates and numbers} -- automatically added dates by dermoscopic device or program. Usually in the corners.
    \item \textit{Parts of background }-- blurred arts of background visible when a skin lesion is located, e.g., on the foot or finger.
    \item \textit{Light reflection} -- white patches of reflected light visible when the gel is not well-distributed, and skin or gel reflects the light. 
    \item \textit{Dust} -- small pieces of tiny hair or dust visible on the skin lesion.
\end{itemize}

\begin{figure*}[!htb]
\centering
\textbf{Images without artifacts}

\includegraphics[width=0.24\linewidth]{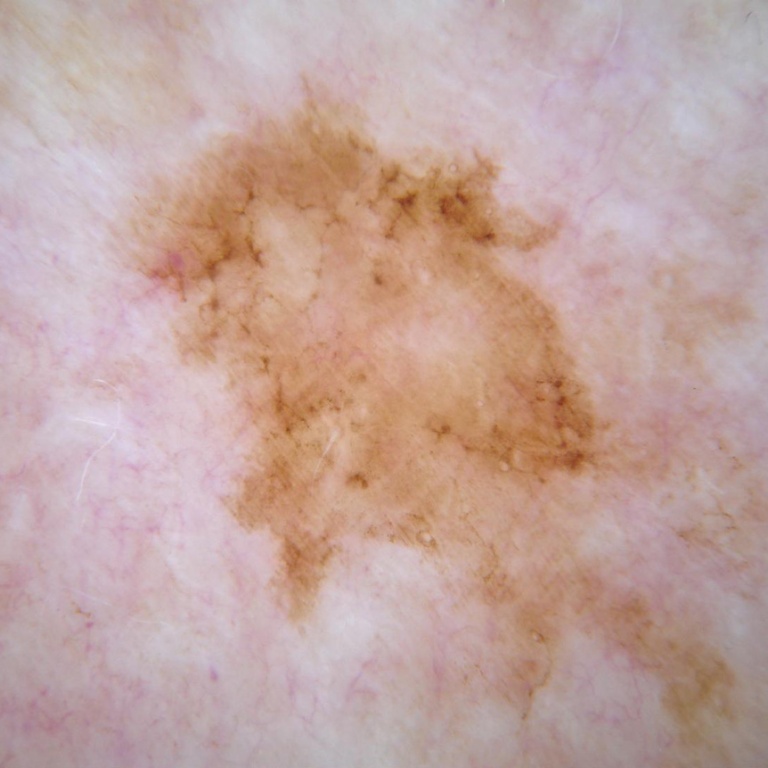} \includegraphics[width=0.24\linewidth]{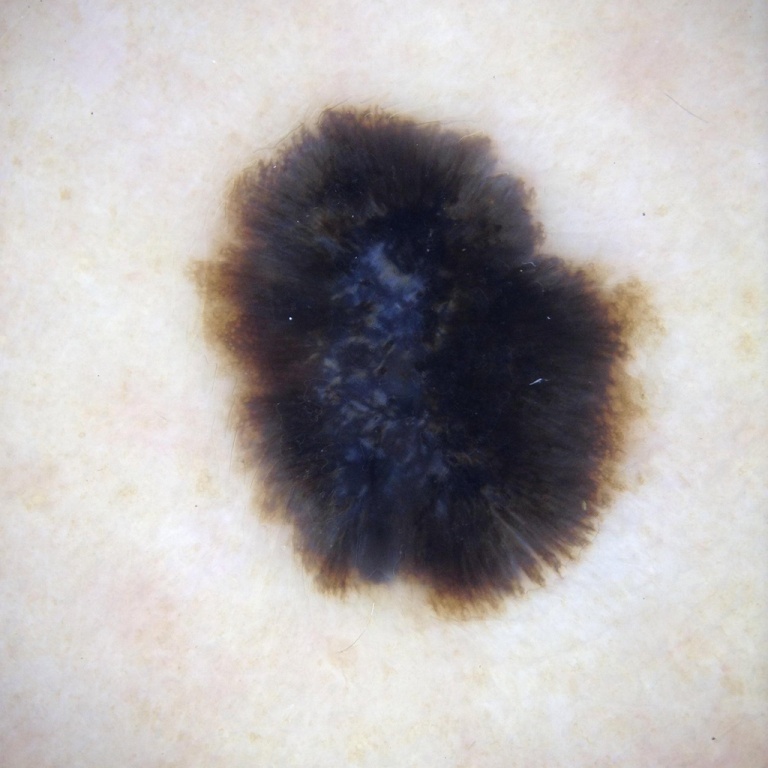} \includegraphics[width=0.24\linewidth]{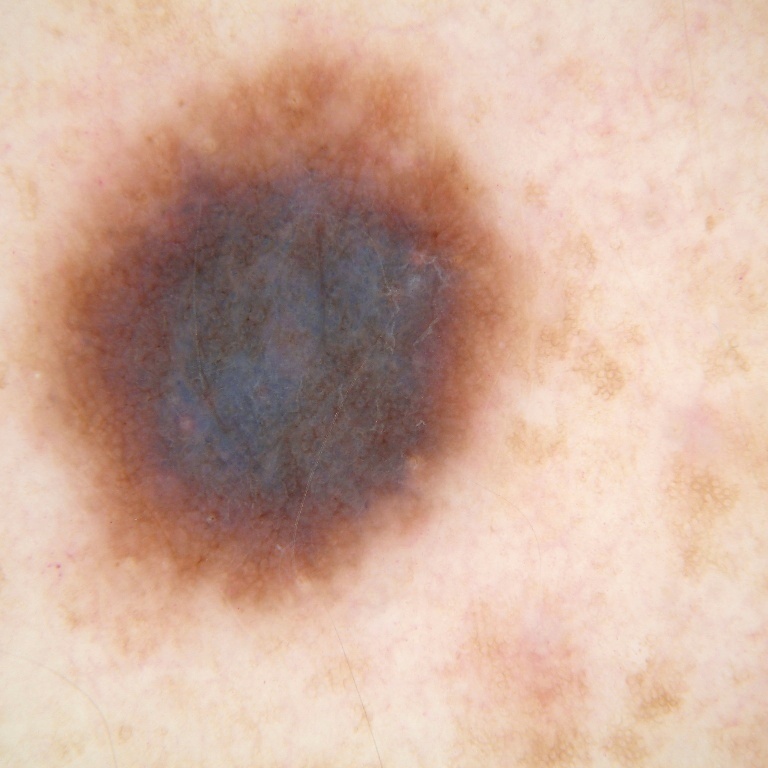} \includegraphics[width=0.24\linewidth]{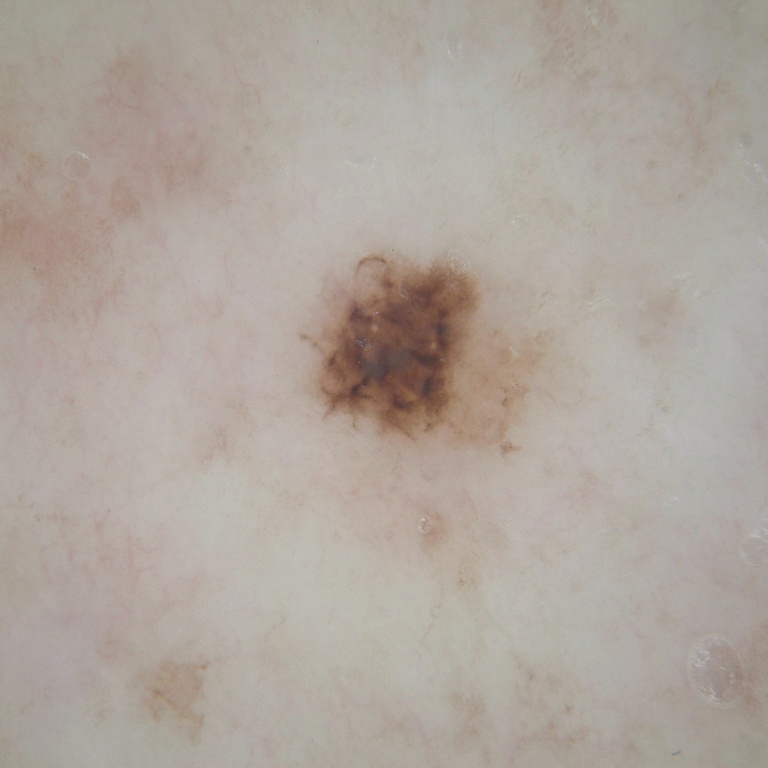}

  \caption{Example clean images without any annotated artifacts.}
  \label{fig.artifacts.clean}
\end{figure*}

\textit{Clean images} are images without any artifact visible. I additionally annotated clean images as a separate category to avoid accidental overlooking artifacts. I presented clean images free of artifacts in Figure~\ref{fig.artifacts.clean}. 

The data from ISIC challenge 2019 and 2020 was used for the experiments \cite{isic2019BCN20000,isic2019codella,isic2019ham10000,isic2020}. Additionally, I used data and publicly available annotations of artifacts provided by Bissoto \cite{bissoto19deconstructing}; it consists of multi-class and multi-label annotations of artifacts from ISIC 2018 and Atlas for skin Dermoscopy datasets.

Additionally, an Inter-Annotator Agreement score was measured using a small subsample of randomly selected data using a Cohen's kappa coefficient \cite{mchugh2012interrater}. A Cohen's kappa allows measuring interrater reliability when labeling data with categorical samples. In this study, the mean Cohen's kappa coefficient was over 70\%, with the highest values on \textit{ruler} annotations and lowest on the \textit{other}. The results are satisfactory, as the scores above 60\% are broadly accepted as a substantial agreement \cite{mchugh2012interrater}.

\section{Descriptive statistics} \label{section:statistcis}

Manual artifacts annotations are a helpful asset in dermoscopic images. In this section, brief descriptive statistics that summarize data collection are presented. 
I analyze artifacts and present the statistics behind manually annotated 4000 dermoscopic images. The calculated statistics are summarized in Table~\ref{tab:artifacts.annotation_stats}.

I calculated and presented the ratio of each artifact regarding all images examined. Those ratios are compared between types: benign and malignant. For the purpose of this study, I proposed to use three basic metrics: \textit{artifacts cardinality}, an \textit{artifact ratio} and \textit{class ratio.} 

\textit{Artifacts cardinality}. The cardinality of certain artifacts set within a class ($\lvert benign \rvert$, $\lvert malignant \rvert$) is the total number of elements (images) where a certain artifact is present. It cannot be larger than the number of images annotated per class.

\textit{Artifact ratio}. The artifact ratio $Q^{artifact}$ is the number of images with certain artifacts divided by the total number of images investigated. For instance, in case of a frame artifact: $Q^{frame} = \frac{ 	\lvert{frame} 	\rvert}{ 	\lvert images  	\rvert}$, where $Q^{frame}$ is frames artifact ratio, $\lvert{frame} \rvert$ the cardinality of frame artifacts (number of images with frames), and $\lvert images \rvert $ the total number of images annotated. The artifact ratio shows how many images have a certain artifact out of all. I calculated the artifacts ratio separately for each class.

\textit{Class ratio.} Class ratio $Q^{class}$ is equal to the fraction of artifacts ratios from two classes: $Q^{class, artifact} = \frac{Q^{artifact, mal}}{Q^{artifact,ben}}$. A class ratio close to 1 means that both classes have the same distribution of artifacts. If the class ratio is significantly lower or higher, the examined artifact is more common in one class than in the other. In case of using $\frac{mal}{ben}$ ratio, a higher number means that the artifact is more common in a malignant category, whereas the lower one appears more often for mild cases.

\begin{table}
\centering
\caption{Manually annotated artifacts in the skin lesion dataset ISIC 2019, \cite{isic2019BCN20000,isic2019codella,isic2019ham10000} and ISIC 2020 \cite{isic2020}.}
\label{tab:artifacts.annotation_stats}
\begin{tabular}{llllll} 
\toprule
type & $\lvert ben \rvert$  &    $Q^{artifact}$     & $\lvert mal \rvert$ &    $Q^{artifact}$     & $Q^{class}$  \\
\midrule
frame    & 104    & 5.20\%  & 521       & 26.05\% & 5.01             \\
hair     & 958    & 47.88\% & 868       & 43.40\% & 0.91             \\
dense    & 204    & 10.19\% & 99        & 4.95\%  & 0.49             \\
short    & 96     & 4.80\%  & 103       & 5.15\%  & 1.07             \\
ruler    & 422    & 21.09\% & 586       & 29.30\% & 1.39             \\
other    & 426    & 21.29\% & 818       & 40.90\% & 1.92             \\
none     & 538    & 26.89\% & 268       & 13.40\% & 0.50             \\ 
\midrule
total    & 2001   &         & 2000      &         &                  \\
\bottomrule
\end{tabular}
\end{table}

A little over a quarter of images in benign skin lesions are clean; they don't have any artifacts. In the case of malignant skin lesions, only half of that remains clean, about 13.4\%. This observation already shows a significant difference between the two classes.

An important feature is a \textit{frame}. Frames, dark and light, round and rectangle, appeared in around 5\% of benign images. In the case of malignant lesions, they appeared in 26\% of instances, which is five times higher. It is worth noticing that dark frames were discovered as a potential biasing factor using \textit{Global Explanation for Bias identification, a method proposed in Chapter~\ref{chapter.gebi}: Identifying bias with global explanations}. It is the highest result out of all artifacts. This might be the result of combining multiple sources of data into one. Considering that each institution uses a different dermatoscope (and some more often produce frames), and some provide more data about malignant lesions, it is safe to assume that it can lead to bias in data.

In the case of naturally occurring hair, the proportion of hair appearance in both classes is similar. Its ratio equals 0.91, making it slightly in favor towards benign lesions. Hair is also the most often observed artifact. Almost every second skin lesion is covered with thin or thick hair, both light or dark. However, when we look closely at densely-haired lesions, we notice that benign lesions are two times more likely to have it. It would be interesting to annotate even more data to evaluate if the $Q^{class}$ would reach one or stay at the same level. 

Ruler marks often occur on malignant images ($Q^{class}$ = 1.39). They appeared on 21\% of benign images and every third on the malignant lesion. This slight disproportion might result from a used dermatoscope (some have built-in rulers) or a doctor's approach (some manually place a ruler near the lesion).

Finally, there are \textit{other} artifacts. The other is the largest group of various artifacts as it covers multiple less common ones:  gel bubbles and borders, ink, patches, dates and numbers, parts of the background, light reflection, dust. Over 40\% of malignant lesions and over 21\% of benign skin lesions showed traces of those artifacts in the examined dataset. This makes another artifact group much more likely to appear with a malignant skin lesion in the image. Some ink markings, for instance, are added by a doctor as guiding surgical markings. 

Another interesting finding is measuring the correlation between the artifacts and skin lesion type (class). This again confirms the hypothesis that \textit{frames} are correlated with malignant skin lesions: there is a strong correlation between the frames and malignant skin lesions. It seems that malignant class is also correlated with \textit{other} artifacts. Another interesting observation is that \textit{rulers} are slightly correlated with \textit{frames} and malignant class, but almost zero correlation with \textit{other}. Hair is slightly correlated with the benign class, but it might also be a negligible value. Hair also seems to be negatively correlated with \textit{other} category: if there is hair in the image, it seems less likely to observe \textit{other} artifacts like surgical pen markings or gel bubbles. The results are presented in  Table~\ref{tab:correlation}.

\begin{table}[!htb]
\caption{Pearson correlation coefficients $r$ and p-values $P$ ($r(P)$) for testing non-correlation between the artifacts and skin lesion types in the manually annotated skin lesion dataset ISIC 2019, \cite{isic2019codella,isic2019ham10000} and ISIC 2020 \cite{isic2020}. }
\label{tab:correlation}
\resizebox{\textwidth}{!}{%
\begin{tabular}{llllll}
\hline
 & hair & frame & ruler & other & lesion type* \\ \hline
\multicolumn{1}{l|}{hair} &  & -0.0446 (0.0048) & -0.0879 (0.0) & -0.1211  (0.0) & -0.0449 (0.0045) \\
\multicolumn{1}{l|}{frame} & -0.0446  (0.0048) &  & 0.0865  (0.0) & -0.0421 (0.0077) & 0.2872  (0.0) \\
\multicolumn{1}{l|}{ruler} & -0.0879  (0.0) & 0.0865  (0.0) &  & 0.0094 (0.5505) & 0.0946  (0.0) \\
\multicolumn{1}{l|}{other} & -0.1211  (0.0) & -0.0421 (0.0077) & 0.0094 (0.5505) &  & 0.2118  (0.0) \\
\multicolumn{1}{l|}{lesion type*} & -0.0449 (0.0045) & 0.2872  (0.0) & 0.0946  (0.0) & 0.2118  (0.0) &  \\ \hline
\end{tabular}
}
\textit{* $0$ for benign, $1$ for malignant. Positive $r$ values means the correlation with malignant class, and negative with the benign one.}
\end{table}

I analyzed publicly available annotations of artifacts in ISIC 2018 and Atlas for skin Dermoscopy (Table for additional insight \ref{tab:artifacts.bissotto_stats}). 
Interestingly, there is a significant difference in certain artifacts ratios for the benign and malignant classes. I hypothesize that those differences come from the uneven distribution of annotated classes in Bissoto (2695 benign vs. 771 malignant) and different annotation guidelines. In those statistics, most $Q^{class}$ are close to one. However, interestingly, patches were observed in only 2 cases for a malignant and 188 for benign (Table~\ref{tab:artifacts.bissotto_stats}). This huge disproportion in training data might significantly impact the model. Ink is also more often visible on benign lesions ($Q^{class}$=0.58) than malignant ones. Similar to my annotations, ruler marks and frames are more likely to appear on malignant lesions. All \textit{other} artifacts are also more likely to occur with malignant lesions.

\begin{table}[]
\centering
\caption{The artifacts in the skin lesion dataset ISIC 2018 Task 1 and 2, and Atlas of Dermoscopy, according to annotations provided by Bissoto et al. \cite{bissoto19deconstructing}. The category $other$ was created by summing gel border, gel bubble, ink, and patches.} \label{tab:artifacts.bissotto_stats}

\begin{tabular}{@{}llllll@{}}
\toprule
type & $\lvert ben \rvert$  &    $Q^{artifact}$     & $\lvert mal \rvert$ &    $Q^{artifact}$     & $Q^{class}$  \\
 \midrule
frame (dark\_corner) & 728 & 27.01\% & 244 & 31.65\% & 1.17 \\
hair & 1679 & 62.30\% & 391 & 50.71\% & 0.81 \\
gel\_border & 590 & 21.89\% & 89 & 11.54\% & 1.00 \\
gel\_bubble & 1441 & 53.47\% & 444 & 57.59\% & 1.08 \\
ruler & 1049 & 38.92\% & 328 & 42.54\% & 1.09 \\
ink & 384 & 14.25\% & 64 & 8.30\% & 0.58 \\
patches & 188 & 6.98\% & 2 & 0.26\% & 0.04 \\
other & 439 & 16.29\% & 188 & 24.38\% & 1.50 \\ \midrule
total & 2695 &  & 771 &  &  \\ \bottomrule
\end{tabular}%

\end{table}

\section{The importance of texture and shape} \label{section.shape_texture}
In 2020  \cite{hermann2020origins}, it was discovered that Convolutional neural networks pretrained on ImageNet tend to classify images based on their texture rather than their shape -- texture quickly became a widely mentioned form of algorithmic bias. Authors showed that mixing the shape and texture of conflicting images, e.g., a cat's shape with an elephant skin pattern, predicts based on the texture. Hermann et al. \cite{hermann2020origins} proved that a texture bias hinders the robustness and makes the model more prone to mistakes in production, making texture bias a highly unpleasant side-effect of convolutional networks. 

As it might seem obvious that a cat picture with added elephant texture still represents a cat, it is not that easy to decide in case of skin lesions. \textit{Would a benign skin lesion become malignant if the malignant texture is added?}
The malignancy of skin lesions is decided by multiple factors, including the shape of the skin lesion and its texture. This section examines whether a texture or shape bias is better for skin lesions classification. The measurements are based on the respected dermoscopic methods of skin lesion categorization. In the past, several methods of skin lesion categorization were proposed, such as the ABCD rule \cite{argenziano2003dermoscopy}, the Menzies method\cite{braun2021mmd,argenziano2003dermoscopy}, or a 7-point checklist \cite{argenziano2021spcd,argenziano2003dermoscopy}. Each of them tried to break down the similarities between malignant lesions into a set of rules, which in the end, would make young clinicians quickly learn the basic rules of categorization.

Considering the importance of shape/texture bias, I analyzed the rules of the most common dermoscopy methods of skin lesion classification and assign them $Yes$ and $No$ labels, depending on whether they alter the shape or the texture in the image. The table presenting rules (Variables), scores, and descriptions of each method is shown in Table~\ref{tab:shape_texture}.

\begin{table}
\centering
\caption{An overview of the three most common dermoscopic methods for melanoma diagnosis in the view of possible shape or texture influence. Description, scores, and variables are literature-based \cite{argenziano2003dermoscopy,braun2021mmd,argenziano2021spcd}.}
\label{tab:shape_texture}
\resizebox{\linewidth}{!}{%
\begin{tabular}{p{4cm}p{1.5cm}p{1.5cm}p{2cm}p{9cm}} 
\toprule
\textbf{Variable} & \textbf{Shape} & \textbf{Texture} & \textbf{Score}   & \textbf{Description} \\ 
\midrule
\textbf{ABCD rule}             &                                    &                                      &        &  \\
Asymmetry                             & Yes                                & Yes                                  & 0–2.6            & Measure symmetry in perpendicular axes: contour, colours and structures                                                                                    \\
Border                                & Yes                                & No                                   & 0–0.8            & Divide to eight segments and count ones with abrupt ending of pigment pattern                                                                              \\
Color                                 & No                                 & No                                   & 0.5–3.0          & Count the number of following six color in the lesion: white, red, light brown, dark brown, blue-gray, and black.                                          \\
Dermoscopic structures                & No                                 & Yes                                  & 0.5–2.5          & Find and evaluate strcutureless areas, pigment network, branched streaks, dots, and globules                                                               \\
\textbf{Diagnosis:}                   & \multicolumn{4}{l}{\textbf{\textit{At least 4.76 points for suspicious, and at least 5.45 for malignant}}}                                                                                                                                                                  \\ 
\midrule
\textbf{Menzies method}             &                                    &                                      & \textbf{}        &                                                                                                                                                            \\
Single color                         & No                                 & No                                   & Negative & The colors scored are black, gray, blue, red, dark brown, and tan. White is not scored. A single color excludes the diagnosis of melanoma.                 \\
Symmetry of pattern                   & Yes                                & Yes                                  & Negative & Symmetrical pattern (colours, structure)                                                                                                                   \\
Positive features                     & No                                 & Yes                                  & Positive & Blue-white veil, multiple brown dots, pseudopods, radial streaming, scar-like depigmentation, multiple colors, multiple blue/grey dots, broadened network  \\
\textbf{Diagnosis:}                   & \multicolumn{4}{l}{\textbf{\textit{None of the negative features and at least one positive present for suspicious.}}}                                                             \\ 
\midrule
\textbf{7-point checklist}           &                                    &                                      & \textbf{}        &                                                                                                                                                            \\
Atypical pigment network              & No                                 & Yes                                  & 2                & Reticular lines, heterogeneous for color and thickness, asymmetrically distributed within the lesion                                                       \\
Blue-whitish veil                     & No                                 & No                                   & 2                & Structureless blue blotches with an overlying whitish haze                                                                                                 \\
Atypical vascular pattern             & No                                 & Yes                                  & 2                & Linear, dotted or globular vessels (polymorphic vessels), irregularly distributed                                                                          \\
Irregular streaks                     & Yes                                & Yes                                  & 1                & Radial streaks and pseudopods located at the lesion edge due to the melanoma radial growth phase                                                           \\
Irregular blotches                    & Yes                                & Yes                                  & 1                & Dots (less than 0.1 mm) and globules (larger than 0.1 mm) irregular in color, size, shape and distribution                                                 \\
Irregular dots/globules               & No                                 & Yes                                  & 1                & Structureless areas different in size and color (black, brown or gray) irregularly distributed                                                             \\
Regression structures                 & Yes                                & Yes                                  & 1                & White scar-like depigmentation or multiple scattered blue-gray granules within a hypopigmented background                                                  \\
\textbf{Diagnosis:}                   & \multicolumn{4}{l}{\textbf{\textit{Anything with score over one is suspicious, and over three requires deeper analysis.}}}                                                                                                                                                  \\
\bottomrule
\end{tabular}
}
\end{table}

In \textit{the ABCD} method, clinicians assign score $s$ to four variables that describe skin lesions: asymmetry ($s_A$), border ($s_B$), color ($s_C$), and dermoscopic structures ($s_D$). If the sum of points $s_{total}$ is larger than 4.76, then a skin lesion is considered suspicious, and when $s_{total}>5.45$, then it is diagnosed as malignant. Skin lesions will never be considered suspicious without malignant traces in the texture, even if all shape features are maximally scored according to the method. If the asymmetry is maximum ($s_A=2.6$ score), and similarly the border ($s_B=0.8$ score), there is still a large margin of over 1.3 points for the skin lesion to be considered suspicious. However, a texture itself will be enough to find it suspicious but not to consider it malignant. Hence, slight imperfections in shape are needed to classify it as malignant fully.

The \textit{Menzies method} uses three essential features to classify skin lesions as malignant: color, pattern symmetry, and structures. It does not assign any points; instead, it defines positive features that suggest skin lesion is malignant and negative ones that do the opposite. None of the negative attributes are present for suspicious skin lesions, and at least one positive feature is found. Hence, according to Menzies, both shape and texture features are needed to find the lesion suspicious. The shape feature is much less covered than the texture, and it plays a vital role in categorizing skin lesions as malignant.

Next, I analyzed the \textit{7-point checklist}. The method defines seven different characteristics of malignant skin lesions. If the feature is recognized, it is counted in the sum with an appropriate weight equal to one or two. Atypical pigment network, blue-whitish veil, and atypical vascular pattern are considered more significant, and their scores are doubled when encountered. The irregular streaks, blotches, dots/globules, and regression structures are summed. In the original method, anything with a score over three is considered highly suspicious, but current studies show that even a score equal to one should be examined. The method indicates that shape-based disproportions are enough for the lesion to be found suspicious, as the maximum possible score affected by shape is equal to three. However, with a maximum possible score of eight, the texture seems to have a more significant impact as multiple factors can modify it.

Presented in Table~\ref{tab:shape_texture} analysis of dermoscopy methods shows that a structure plays a significant role in contrast to classical image classification tasks. In all methods, the structure in the image was given higher importance scores than the shape of the lesion. However, it does not mean that the shape can be overlooked: in two out of three methods, the texture was not enough to consider a skin lesion malignant. Hence both shape and texture play a vital role in classification. Changes in shape do not necessarily mean a suspicious lesion, but shape asymmetry and texture irregularities almost certainly make the skin lesion suspicious. 

\section{Training with the covered object of interest}

Statistical analysis of the artifacts in skin lesion datasets shows that they might be a potential source of bias. Hence another question arises: can models use artifacts as deciding features? Bissoto et al. \cite{bissoto19deconstructing} suspected that a widely used dataset of skin lesions might be biased, so they conducted a series of experiments on that matter. They used segmentation masks for each lesion and modified the dataset by covering each lesion with a black segmentation mask. That modified dataset was used to train a convolutional neural network to differentiate benign and malignant skin lesions – but without any lesions in the dataset. Surprisingly, results showed that a model trained and tested on data without any lesions can classify it correctly with a performance (AUC) above 73\%, only ten percentage points less than the performance on the original data. Because the shape of the skin lesion is a critical feature for dermatologists, the researchers changed segmentation masks to black boxes and repeated the experiments. The results were even more terrifying because the performance was almost identical to the previous tests.

\begin{figure}[!htb]
\centering
  \includegraphics[width=\textwidth]{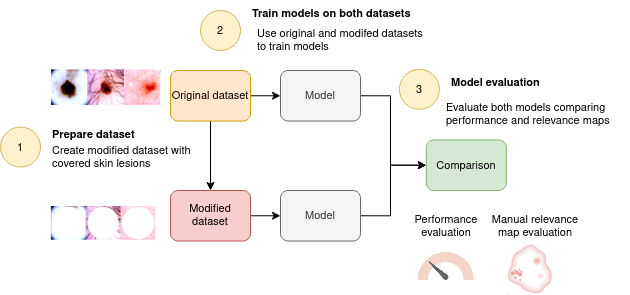}%
  \caption{The process of training with the covered object of interest.}
  \label{fig.covered-pipeline}
\end{figure}

To sum up, Bisotto et al. \cite{Bissoto_2019_CVPR_Workshops} showed that if we cover skin lesions in a dermoscopy dataset and train the model on that dataset, we might still get a decent classification accuracy. This surprising finding also suggests that models can use different features than we think they should when classifying. Those experiments are repeated in a new setup. Figure \ref{fig.covered-pipeline} presents the steps behind this experiment.

\begin{figure*}[!htb]
\centering
\textbf{Original images}

\includegraphics[width=\linewidth]{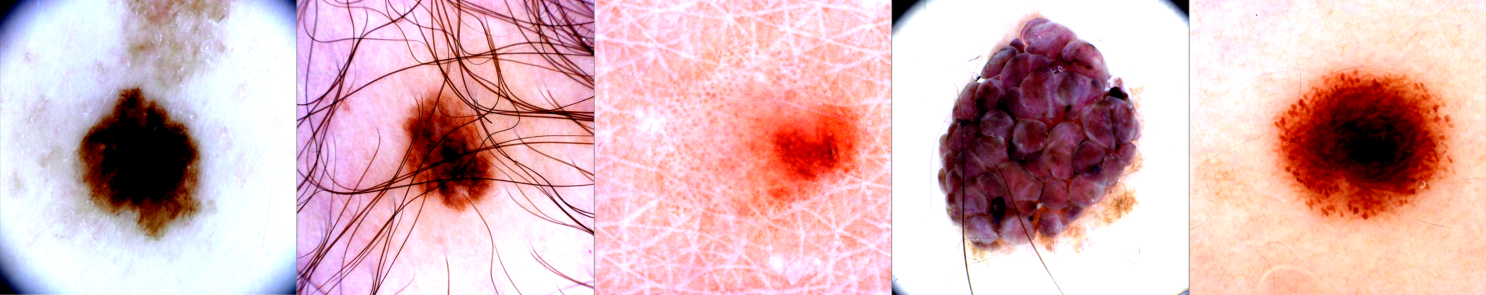}

\textbf{Covered images}
\includegraphics[width=\linewidth]{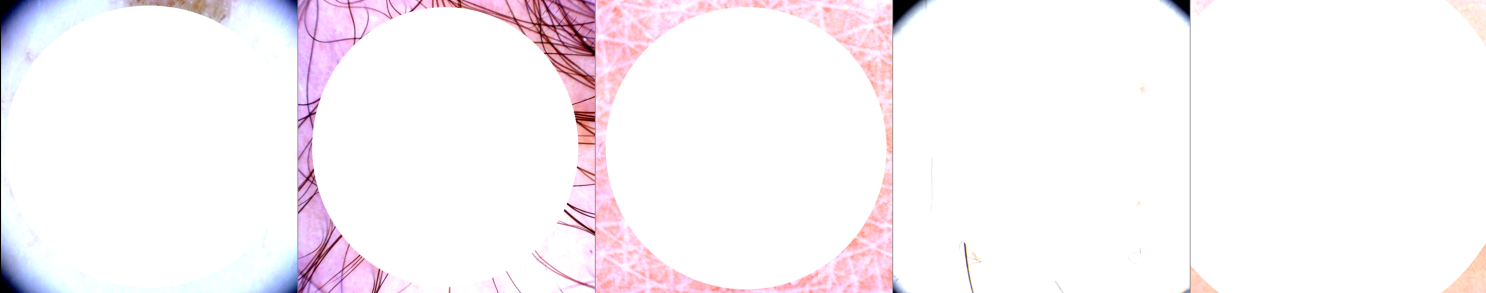} 

  \caption{An example set of modified samples. Each lesion is covered with round white patches of exact size. Modified samples are used to train CNN to differentiate between benign and malignant lesions.}
  \label{fig.artifacts.covered}
\end{figure*}

All skin lesions in the dataset are covered with white, round patches instead of a skin lesion segmentation mask, as presented in Figure~\ref{fig.artifacts.covered}). Modified samples are almost entirely covered; the only visible parts are fragments of skin and some leftover artifacts.

Then the model is trained on the modified dataset to recognize malignant and benign lesions. I compared those results with the same model trained on a standard dataset. The hyperparameters are the same for all experiments for the sake of comparison. Each model was initialized with pretrained weights on ImageNet, trained for ten epochs, and evaluated on the same test set.
The results are presented in Table~\ref{tab:artifact.debias}.

\begin{table}[]
\centering
\caption{The comparison of results for training different EfficientNet models with normal and covered skin lesions.}
\label{tab:artifact.debias}

\begin{tabular}{@{}lllll@{}}
\toprule
\textbf{model}  & \textbf{precision} & \textbf{recall} & \textbf{$F_1$} & \textbf{dataset} \\ \midrule
EfficientNet-B2 & 0.87               & 0.75            & 0.80        & standard         \\
                & 0.94               & 0.60            & 0.65        & covered          \\ \midrule
EfficientNet-B3 & 0.91               & 0.73            & 0.79        & standard         \\
                & 0.93               & 0.59            & 0.63        & covered          \\ \midrule
EfficientNet-B4 & 0.88               & 0.77            & 0.82        & standard         \\
                & 0.96               & 0.60            & 0.65        & covered          \\ \bottomrule
\end{tabular}%

\end{table}

The results show similar properties as in the case of Bissotto et al. \cite{Bissoto_2019_CVPR_Workshops}. The 15 pp loss in $F_1$ score was observed in my experiments when fully covering skin lesions. The model can learn to distinguish malignant skin lesions from benign based only on the scare information from the images and very often with no information about the lesion. Hence, it is vital to discover what features are the most important.
\begin{figure*}[!htb]
\centering
\textbf{Modified images}

\includegraphics[width=0.3\linewidth]{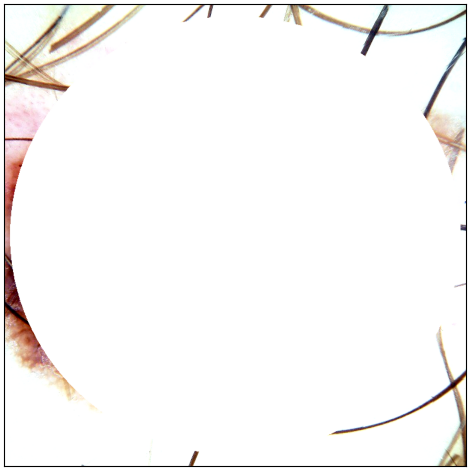}
\includegraphics[width=0.3\linewidth]{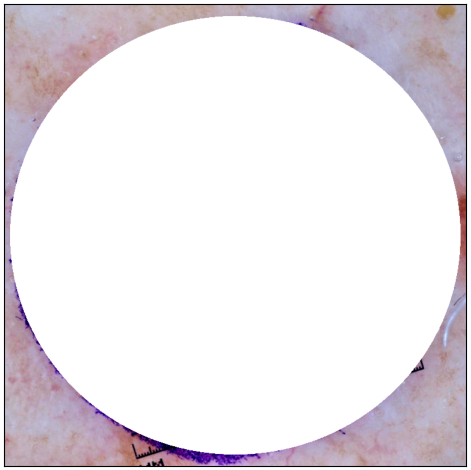}
\includegraphics[width=0.3\linewidth]{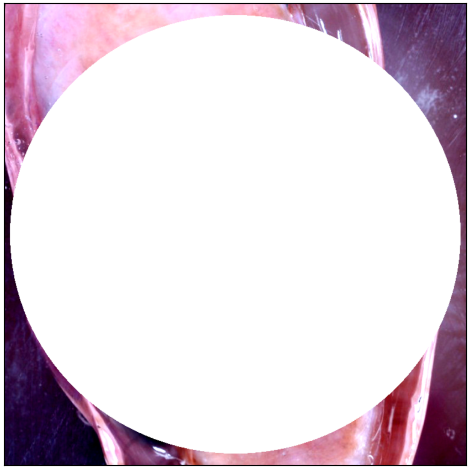}

\textbf{Heatmaps generated with a model trained on modified samples}

\includegraphics[width=0.3\linewidth]{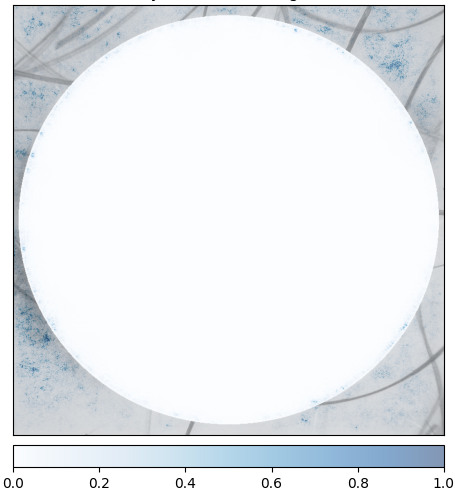}
\includegraphics[width=0.3\linewidth]{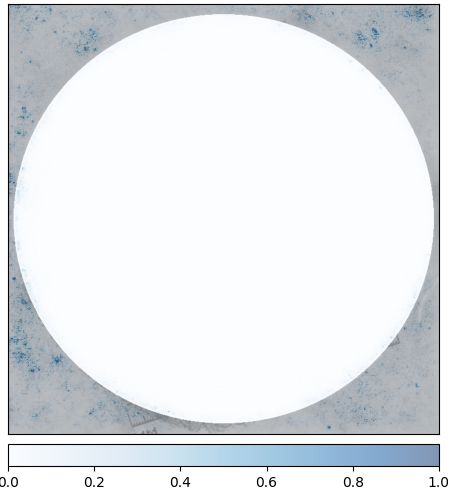}
\includegraphics[width=0.3\linewidth]{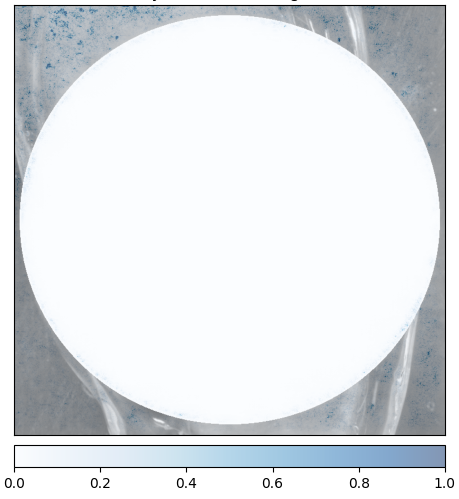}

\textbf{Heatmaps generated with a model trained on original samples}

\includegraphics[width=0.3\linewidth]{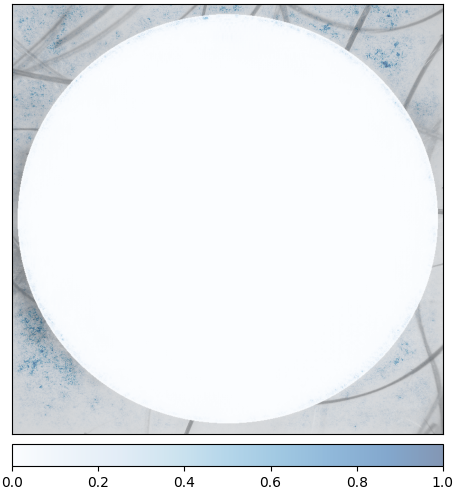}
\includegraphics[width=0.3\linewidth]{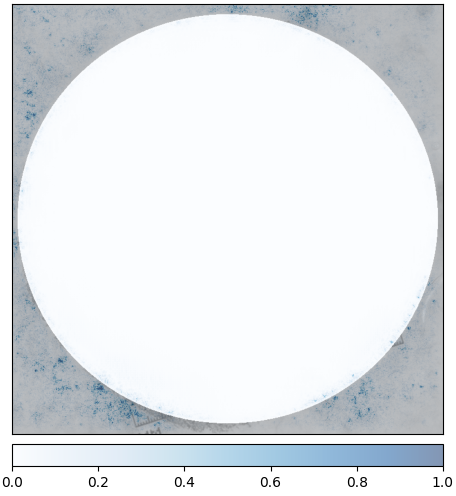}
\includegraphics[width=0.3\linewidth]{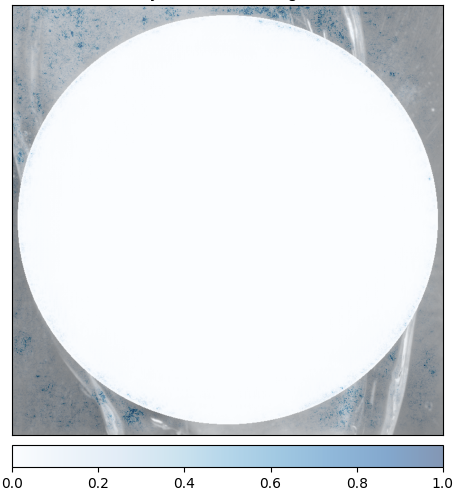}

  \caption{An example set of modified samples. Each lesion is covered with round white patches of exact size. Modified samples are used to train CNN to differentiate between benign and malignant lesions.}
  \label{fig.artifacts.debias}
\end{figure*}

 Hence, I used local explainability methods to shed some light on that problem. Integrated Gradient \cite{kokhlikyan2020captum} method was used to analyze relevance maps of skin lesions. I produced heatmaps using a model trained on standard data and compared it with the model trained on modified data. 
First, the heatmaps for original images with a model trained on standard data are generated. Then, I created heatmaps for covered images with the same model. I repeated the process with a model trained on covered data: generated heatmaps for original images and, once again, hidden part of the data. 

The example heatmaps are presented in Figure~\ref{fig.artifacts.debias}. The difference is very subtle. First, the samples trained on original images occur on the circle border, which cannot be observed on modified samples. This behavior is expected, as it suggests that the model trained on real data focuses more on edges -- the model trained on modified data learned to ignore it. Next, both heatmaps show blue areas on light reflections and gel borders. The model trained on modified samples pays greater attention to very subtle shadow changes visible on the skin. In general, both heatmaps are very similar and difficult to analyze. Another reason might be that model trained on modified samples has lower prediction accuracy, making heatmap generation difficult (heatmap generation methods work best with high accuracy models).

\section{Discussion}
Skin lesion datasets seem to be influenced by many types of artifacts. Some of them can have a substantial impact on the model's performance. Models fueled with biased data can use wrong features to predict malignancy. As it is evident for the dermatologists that a type of dermatoscope used (e.g., the kind that adds a black frame) does not change the malignancy of the lesion, it might not be evident for the model. Statistics show that 30\% more malignant lesions have black frames. Does that mean that black frames are an essential feature of malignancy? No, they are not. But the data suggests that.
Hence, there is a need to either make data cleaner from artifacts or mitigate unwanted effects on the models. 
Moreover, the skin diagnostic methods show that the image texture is an important feature. The shape, however, can still be a deciding factor when evaluating the lesion. The experiments show potential in using conflicting shapes and textures for data augmentation towards better robustness in the inference. Finally, I examined how well the model will predicted without any information about the skin lesion. The results showed a drop in predictions, on average from $F_1 = 0.8$ to  $F_1 = 0.65$, proving that the model can learn to distinguish malignant skin lesions from benign based only on the scare information from the modified images.

The manual inspection of biases in data included three major steps: manual identification of possibly biasing features (selecting biases to annotate), manual bias annotation, and descriptive statistics. Additionally, I examined the shape and texture bias in terms of clinical importance and trained the model with and without the object of interest to compare how the results change. The approach successfully detected and confirmed biases in data but required lots of resources and time to conduct. This showed the need for an automated process that would significantly lower manual labor needs. In the next chapter, I focused on detecting bias in data without tiring manual inspection.


\clearpage  
\lhead{\emph{Chapter 5: Identifying bias with global explanations}}  
\chapter{Identifying bias with global explanations} 
\label{chapter.gebi}
\section{Introduction}
As mentioned earlier, one of the methods of bias detection is a manual inspection of data and predictions (see \textit{Chapter \ref{chapter.skin_lesion_bias}: Identifying bias with manual data inspection}). This time-consuming approach requires a lot of technical and domain knowledge. Manual exploration can be boosted with local explainability methods such as attribution maps visualized as heatmaps or based on prediction perturbation. But is it possible to manually inspect every vast dataset used to fuel deep learning algorithms? ImageNet currently has over 14 mln. images, Amazon Reviews over 82 mln. of text reviews, Common Voice over 1000 hours of speech. Inspecting those files might be Sisyphean labor. 

The goal of this work is to propose a method of improving the process of bias identification. Instead of manual analysis of the sample by sample, or prediction by prediction, I present a methodological approach in using global explainability methods for a broader inspection of the model, and as a result, also the data. Preliminary experiments showed that existing global explainers were not fit to help bias discovery in dermoscopy datasets. Hence, I proposed further improvements to the existing methods, creating a much more robust method. Additionally, I offered a testing pipeline proposition that includes a proposal of metrics and evaluation. The \textit{global explanations for bias identification} method is tested on the skin lesion dataset, and the discovered biases correspond to the manually found ones.

\section{Methodology}
\subsection{Global explanations}
One of the global explainability methods is SpRAy \cite{lapuschkin2019unmasking} described in detail in \textit{Chapter~\ref{chapter.xai}: Explainable Artificial Intelligence}. The idea of the method is to generate local explanations for all data instances and then cluster them to reveal patterns in the model's reasoning. Global explanation helps a user avoid a time-consuming manual analysis of individual attribution maps. It still requires a manual review of the resulting clusters instead. There is one significant flaw of the method. The method uses only the attribution maps for prediction strategies identification. The analysis of attribution maps, without the actual inputs, makes it strongly attribution-dependent. Moreover, the attribution maps are usually tricky to analyze without the input image, as they tend to be noisy. I propose \textit{Global Explanations for Bias Identification} (GEBI) to overcome those problems, which is a proposition of improvement over SpRAy.

The main contribution is a proposition to cluster both attributions and corresponding input instances simultaneously. Additionally, I proposed a different dimensionality reduction method: I reduced dimensionality with the isomap algorithm \cite{balasubramanian2002isomap} instead of simple geometrical image resize.
Each instance and attribution is reduced with an isomap to the selected dimension number $n$ and saved as a new representation vector.
Those representations are later concatenated, creating a large pair-representation of both attribution and input. Achieved clusters represent common prediction patterns. Those two seemingly minor improvements ended in a considerably more robust global explainer.

\begin{figure}[!htb]
\centering


 \includegraphics[width=1\textwidth]{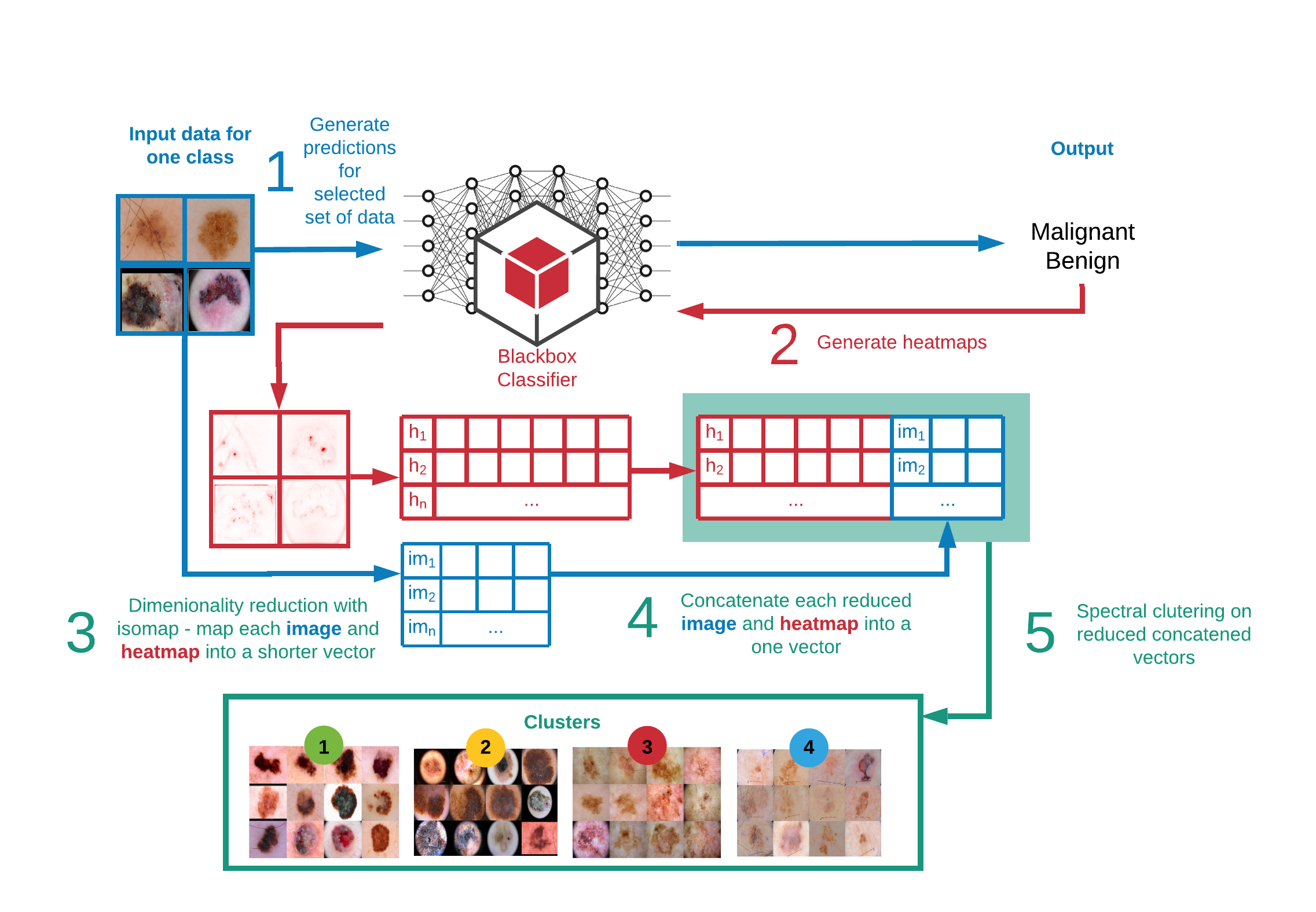}%
 \caption{The pipeline of  Global Explanations for Bias Identification (GEBI)} \label{fig.gebi}
\end{figure}

The steps of global explanations for bias identification are as follows.
\begin{enumerate}[\itshape   Step 1:]
\setlength{\itemsep}{1pt}
 \setcounter{enumi}{-1}
 \item Select images and a class to explain
 \item Compute attribution maps for samples of the selected class.
 \item Normalize and preprocess both input samples and attribution maps in the same manner
 \item Reduce the dimension of each input sample and relevance map with a dimension reduction algorithm.
 \item Concatenate each reduced sample with a relevant reduced attribution map
 \item Perform spectral clustering on reduced vectors
 \item Visualize and analyze the obtained clusters. Formulate and test the hypothesis with the bias insertion algorithm to test bias.
\end{enumerate}
The pipeline of GEBI is presented in Figure \ref{fig.gebi}

Images and class selection is a necessary preliminary step for the analysis. In most cases, it is crucial to cluster instances only from the same single class. Analyzing more than one class simultaneously should only be executed when looking for possible biases that may be common for all categories, as in the case of backdoor attacks \cite{wang2019neural}. 

When the instances and a class of interest are selected, it is time to compute attribution maps for all selected input samples (step 1). Here, I have used the LRP method to calculate attribution maps. However, it is possible to use other techniques such as perturbance-based heatmaps or gradient-based visualizations.
Next, all samples and attribution maps are normalized: I enhance contrast to enhance important clinical attributes of skin lesions and equalize white balance through histogram equalization to prevent dummy color-based clustering (step 2).

In the third step, the dimensionality of images and attribution maps is reduced. 
There are two significant factors here: the method of dimensionality reduction and target dimensionality size. 
I decided to downsize instances with the Isomap algorithm \cite{balasubramanian2002isomap} instead of image-downsizing. Standard downsizing provides a good scaling focusing on good visual results. However, it does not operate on the spectral dimension, which is needed to project multidimensional pixels values into another space creating meaningful embeddings. 
Most skin lesion images look similar: a light or dark brown lesion in the middle of the lighter skin. Slight differences between images must be found, as most images look the same after drastic downsizing.
What makes skin lesions interesting in the case of cancer prediction are small visual patterns like blotches, regression areas, blue-white veils -- or medically not meaningful artifacts that could cause bias like gel bubbles, hair, or ruler marks. 
Those patterns and artifacts can disappear from the image after a simple downscaling, whereas the general colors for every lesion would remain.
On the contrary, when using nonlinear dimensional reduction, most algorithms ignore common patterns for every instance, which is a desired outcome. 
The target dimensionality size should be individually selected for the problem. I achieved the best results when the input instances vector was around two times smaller than the attribution vector.

The fourth step is concatenating attribution and instance vectors (reduced in size attribution map and input image pairs). The vector's concatenation is a new and novel step that extends the SpRAy method. As mentioned, the clustering can be improved by concatenating both images and attribution maps. Clustering only attribution maps lead to a biased explainer that clusters attribution maps based on the attribution localization. For instance, it would cluster attribution maps with higher attributing scores in the top left corner together, even if features "under" attribution maps are entirely different. Adding information from images prevents such behavior. On the other hand, clustering only images does not bring any meaningful data other than which images are similar in terms of selected distance measures. Providing both information at the same time is the proposed solution. 

Finally, clustering of representation vectors is performed, similarly to in SpRAy. Users can apply spectral clustering as in SpRAy or any other clustering method. In performed experiments, the k-means gave similar results as spectral clustering. Users can use, e.g., an Elbow method, the Silhouette, or others \cite{kodinariya2013review} to estimate the optimal number of clusters needed. 

The resulting clusters can be visualized in a 3-D space using Isomap embeddings reduced to three dimensions. It is worth analyzing what kind of elements ended up in the same groups and what made them similar. If a cluster is well-separated from others, it is probably significantly different from others, which should be noticed by visual analysis of attribution maps or images. If some clusters slightly overlap, it is safer to observe the central instances rather than evaluate those at the border. To additionally test bias, users can formulate and test the hypothesis with a counterfactual bias insertion algorithm.

\subsection{Counterfactual bias insertion}
\begin{figure}[!htb]
\centering


 \begin{subfigure}{
      \includegraphics[width=\textwidth]{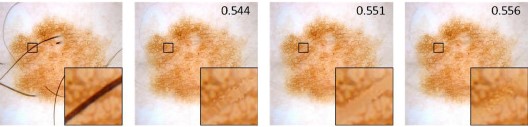}%
      }
    \end{subfigure}
    
     \begin{subfigure}{
      \includegraphics[width=\textwidth]{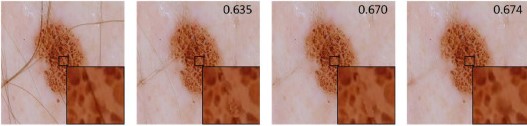}%
      }
    \end{subfigure}
    
{(a) Input images \quad (b) DullRazor \cite{lee1997dullrazor} \quad (c) Huang et al. \cite{huang2013robust} \quad (d) Li et al.  \cite{li2021digital} } 
    
 \caption{Artifacts removal example: hair removal case. Images are from Li et al. paper \cite{li2021digital}. The inset in each image shows the zoomed-in enlargement of the hair gap in inpainting results. The value of the Intra-SSIM of each image is also denoted.} \label{fig.counterfactual}
\end{figure}

Once possible biases are identified, we must confirm them and evaluate how they influence the model. In the past, some methods were proposed, such as measuring the difference in positive proportions in predicted labels or accuracy/recall differences, and others \cite{hardt2021amazon}. It can help determine if a model has a higher recall for one group of instances than another, especially if a feature should not influence the result. In most cases,  bias is tested by removing a potentially biasing feature from inputs and checking whether the results changed (e.g., each age group has a similar recall now). But in computer vision, removing the bias is challenging, especially when we have to remove an object from an image. For instance, removing an artifact from the image is difficult without leaving traces of such prepossessing on it. Even if we pull it using advanced image processing or some generative models, it might generate other regional artifacts. For instance, Li et al. \cite{li2021digital} proposed to use image inpainting for hair removal. The method shows great potential, yet new artifacts remain (see example in Figure \ref{fig.counterfactual}). Those artifacts might still bias the output and affect the results. 

\begin{figure}[!htb]
\centering 


 \includegraphics[width=0.6\textwidth]{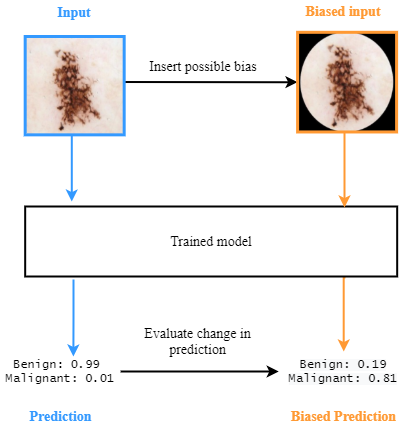}%

 \caption{Counterfactual bias insertion on the skin lesion classification example.} \label{Figure.counter-bias-image}
 
\end{figure}

Instead, I proposed to \textit{insert} the bias into every dataset sample and measure how the prediction changed. The method does not require advanced prepossessing, is easy to apply, and the results are straightforward to interpret. I presented the \textit{Counterfactual Bias Insertion (CBI)} method that checks how the prediction will change after purposely adding a bias to data. An example would be adding a black frame to a skin lesion image. Such modified images would be fed to the model, and the prediction would be checked. The result can be compared to the prediction score with the original input, without the frame. Ideally, the prediction should remain the same after inserting minor artifacts or small data shifts. The example with black frames is presented in Figure~\ref{Figure.counter-bias-image}. 

In the Figure, the potential bias was identified as a black frame. Input is modified by adding a black frame, and then the prediction is measured again. The prediction significantly changed after adding a frame to an image, meaning that data might be biased with black frames.

\begin{figure}[!htb]
\centering 


    \includegraphics[width=0.6\textwidth]{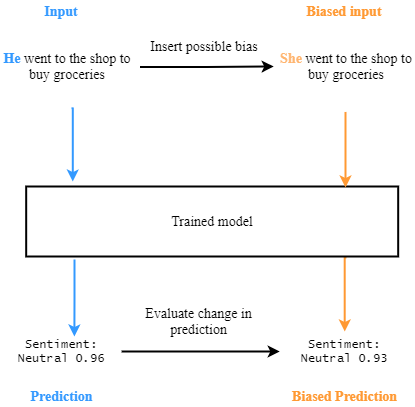}%

 \caption{Counterfactual bias insertion on sentiment classification example.} \label{Figure.counter-bias-text}
\end{figure}

The method is not restricted to images only, as it can be easily applied to text or audio classification. For text, for instance, we could similarly, as in LIME, modify an input sentence by adding/removing words or replacing them with others. However, we cannot apply any modification, as we must carefully modify it towards potential bias. For instance, if we suspect that the model is too sensitive to pronouns, we could change the pronoun in multiple sentences and compare the predictions. The text example is shown in Figure~\ref{Figure.counter-bias-text}. For this Figure, there was a hypothesis of gender bias in the data. The input was modified by pronoun change, and then the prediction was measured again. The prediction did not significantly change after pronoun change meaning that pronouns did not bias the sentiment's results in that case.

The steps of Counterfactual Bias Insertion are as follows:
\begin{enumerate}[\itshape   Step 1:]
\setlength{\itemsep}{1pt}
 \setcounter{enumi}{-1}
 \item (Preliminary) Identify possible biases
 \item Compute predictions for all samples from the dataset and store them.
 \item Insert examined bias to every sample.
 \item Compute predictions for all biased samples.
 \item Compare original predictions with biased predictions.
\end{enumerate}

The changes in prediction $p^{change}$ after bias insertion can be measured with mean and median change and by counting the number of instances that changed the predicted class.

The change in prediction is defined as the difference in the model's outputs to the original and biased input. The change in prediction is presented below (\ref{eq.change}).

\begin{equation}\label{eq.change} p^{change}_k = (p_k - p^{biased}_k) \end{equation} 
where $p^{change}_k$ is a change in prediction for input $k$, $p_k$ is a prediction for input $k$, and $p^{biased}_k$ is a prediction for biased input $k$.

The mean and median change is defined as a $mean_{change}$ (\ref{eq.mean}) and $median_{change}$ (\ref{eq.median}) change of all prediction changes. 

\begin{equation} \label{eq.mean} mean_{change}(p^{change}) =\frac{1}{n}\sum _{i=1}^{n}p^{change}_{i}= \frac{p^{change}_1 + p^{change}_2 + \cdots + p^{change}_n}{n} \end{equation}
\begin{equation} \label{eq.median} median_{change}(p^{change}) = \frac{p^{change}_{n/2} + p^{change}_{(n/2)+1}}{2} \end{equation}
where $p^{change}_n$ is a change in prediction for input $n$, and $n$ is number of all examined predictions.
Both measures will be close to zero when examined change is not a bias. Higher values mean higher chances of data being biased.

An additional measure is the number of switched classes. Prediction is considered as $switched$ (\ref{eq.switched}) when the predicted class changed when the input was biased. Ideally, the number of reversed classes should stay zero if the dataset is not biased with the examined artifact. Several switched classes do not examine the accuracy or correctness of the predicted category. It discusses only the influence of bias on the prediction.
Switched class is defined as follows (\ref{eq.switched}):
\begin{equation} \label{eq.switched}
{
    switched(k) = \begin{cases}
        1,& \text{if } c^{out}_{p_k} \neq c^{biased}_{p_k}\\
        0,              & \text{otherwise}
    \end{cases}
}
\end{equation}
where $c^{out}_{p_k}$ is a $class$ (based on prediction $p_k$) for input $k$, $c^{biased}_{p_k}$ is predicted $class$ for biased input $k$, and $n$ is number of all examined predictions.

The described method allows for extensive data and model regarding possible bias. The method directly shows how strongly the model was affected by potential biases in data. I used the proposed method in other chapters as well.

\section{Experiments}

\subsection{Data and training details}
To present the bias identification process with global explanations and their evaluation with counterfactual bias insertion, I used the dermoscopy dataset example again. I used data downloaded from International Skin Imaging Collaboration Archive \footnote{\href{https://www.isic-archive.com/}{www.isic-archive.com } -- an academia and industry partnership designed to facilitate the application of digital skin imaging to help reduce melanoma mortality}. The model was trained on 1088 malignant skin lesions and 12433 benign images. As for the test set,  200 images per class were used. A randomly selected subset of 4245 benign and 884 malignant cases was used for explainability tests. 

I finetuned DenseNet121 architecture following the style transfer procedure with pseudo-labeling as described in the \textit{Chapter \ref{chapter.neural-style})}. The basic augmentation techniques and early stopping were used. The final model achieved an AUC score of 0.869. I use Deep Taylor Decomposition for generating attribution maps \cite{montavonDTD}. Each attribution was preprocessed map before global explanations. Prepossessing included: resize, histogram equalization, and contrast-enhancing. Dimensionality reduction was performed with the Isomap algorithm \cite{balasubramanian2002isomap}: each image was reduced to a 10-dimensional vector; each attribution map to a 20-dimensional vector. I have tested DBSCAN, k-means, spectral clustering, OPTICS, affinity propagation, mean shift, and birch \cite{xu2015comprehensive} clustering algorithms for clustering. The best results were achieved for spectral clustering and k-means. I used the Elbow method \cite{kodinariya2013review} to determine the number of clusters. 

\subsection{Bias identification}
Those experiments investigated the model's prediction strategies, e.g., \textit{how the whole model worked}, and what caused each prediction. Prediction strategies were identified with SpRAy, IsoSpRAy, and with GEBI. I focused mainly on method GEBI, as this is my main contribution. The results are presented, commented and compared.

\begin{figure}[!htb]
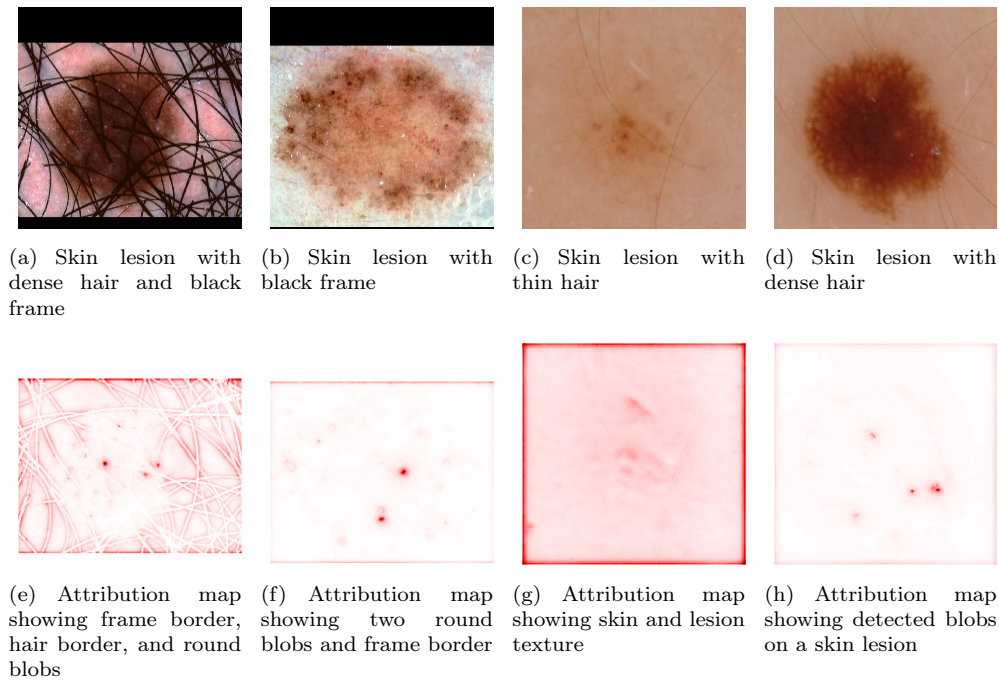

\centering

 \begin{subfigure}[Skin lesion with dense hair and black frame]{
      \includegraphics[width=0.2\linewidth]{Figures/xai-examples/dtd/im1000_original.png}%
      }
    \end{subfigure}
    \begin{subfigure}[Skin lesion with black frame]{
      \includegraphics[width=0.2\linewidth]{Figures/xai-examples/dtd/im1001_original.png}%
      }
    \end{subfigure}
     \begin{subfigure}[Skin lesion with thin hair]{
      \includegraphics[width=0.2\linewidth]{Figures/xai-examples/dtd/im1002_original.png}%
      }
    \end{subfigure}
     \begin{subfigure}[Skin lesion with dense hair]{
      \includegraphics[width=0.2\linewidth]{Figures/xai-examples/dtd/im1003_original.png}%
      }
    \end{subfigure}
    

 \begin{subfigure}[Attribution map showing frame border, hair border, and round blobs]{
      \includegraphics[width=0.2\linewidth]{Figures/xai-examples/dtd/im1000_deeptaylor.png}%
      }
    \end{subfigure}
    \begin{subfigure}[Attribution map showing two round blobs and frame border]{
      \includegraphics[width=0.2\linewidth]{Figures/xai-examples/dtd/im1001_deeptaylor.png}%
      }
    \end{subfigure}
     \begin{subfigure}[Attribution map showing skin and lesion texture]{
      \includegraphics[width=0.2\linewidth]{Figures/xai-examples/dtd/im1002_deeptaylor.png}%
      }
    \end{subfigure}
     \begin{subfigure}[Attribution map showing detected blobs on a skin lesion]{
      \includegraphics[width=0.2\linewidth]{Figures/xai-examples/dtd/im1003_deeptaylor.png}%
      }
    \end{subfigure}

 \caption{Counterfactual bias insertion on the image and text classification example.} \label{fig.dtd}
\end{figure}
\textit{GEBI. } Firstly, a local explainability method has to be selected. The local explanation method will be used to aggregate them globally with GEBI. I used the Deep Taylor Decomposition method, which allows for generating attribution maps. Contrary to gradient-based explanations, DTD attribution maps are relatively easy to analyze and significantly less noisy. Example DTD attribution maps are presented in Figure~\ref{fig.dtd}.

I investigated several preprocessing methods, including white balance, contrast enhancement, heatmap blurring, and smoothing. The best results were achieved with heatmap smoothing and contrast enhancement.
Moreover, several different images size were also exploited. Finally, I ended up with the matrix size 45x45x1, which gave the best results.
Next, additional dimensionality reduction with Isomap is made. I experimented with different dimensionality reduction methods, including PCA and kernel PCA, TSNE, and LDA.  
Additionally, my experiments tested different relations of attribution map sizes to image sizes. Experiments showed that the reduced attribution vector should be around two times greater than the image vector. In the case of the skin lesion dataset, the best results were achieved with vector sizes equal to twenty for an attribution map and ten for an image. 
Finally, I experimented with different methods of clustering, as well as various numbers of clusters. In the end, I selected the number of clusters with the Elbow method (four) \cite{kodinariya2013review}.

\begin{figure}[!htb]
\centering

\textbf{Attribution-based and input-based clustering}

 \begin{subfigure}[Attribution maps in the embedding space. Each dot represent different attribution map; each color different cluster.]{
      \includegraphics[width=0.45\linewidth]{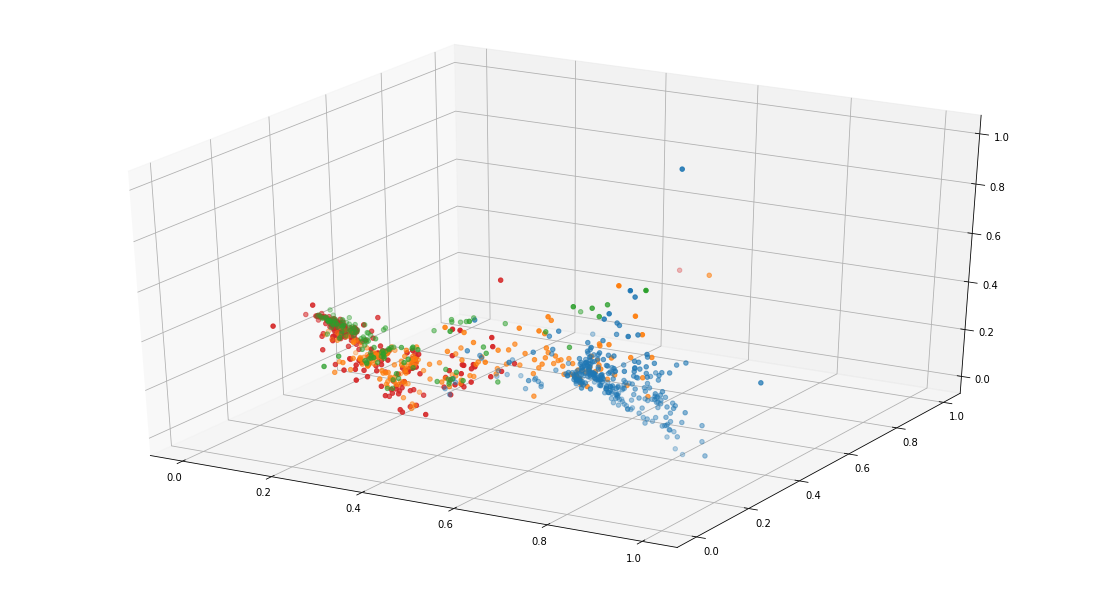}%
      }
    \end{subfigure}
    \begin{subfigure}[Images in the embedding space. Each dot represents different image, each color represents different cluster.]{
      \includegraphics[width=0.45\linewidth]{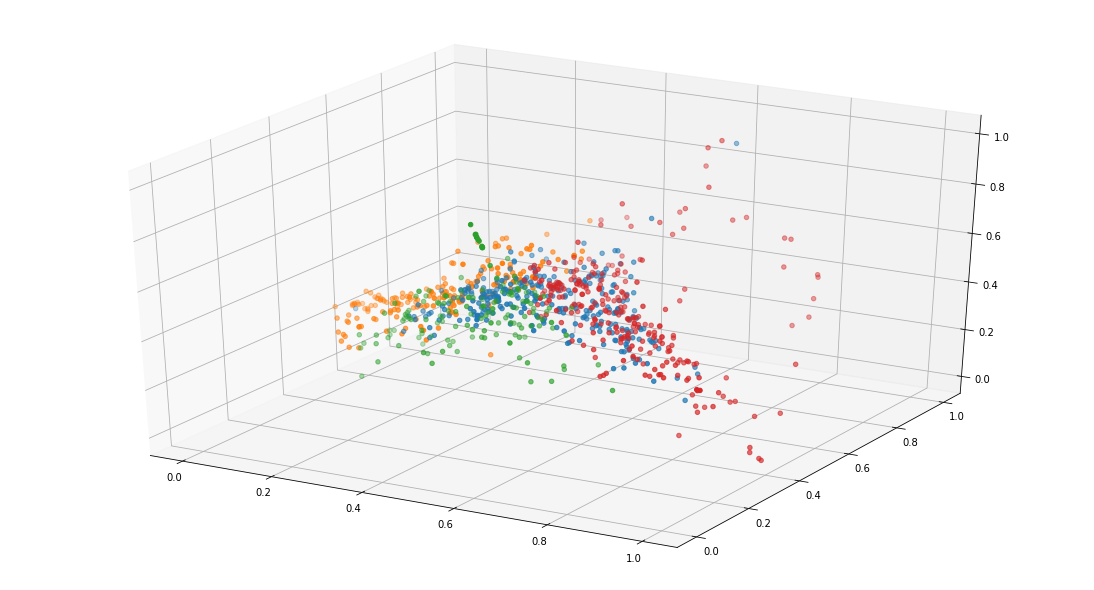}%
      }
    \end{subfigure}
  
 \caption{Global Explanations for Bias identification applied on concatenated images and attribution vectors.} \label{fig.embedding-gebi}
\end{figure}

The attribution- and input-based clustering was performed: I clustered the concatenated attribution map embeddings with input embeddings. Then, I mapped both images and heatmaps into 3D space and presented the results for visualization. This way, I could compare how well images and attribution maps have been clustered. 
I present images and attribution maps separately in their 3D embedding space in Figure~\ref{fig.embedding-gebi}.

The boundaries of clusters are quite well visible for both images and attribution maps. This suggests that values for both attribution and input embeddings were used during clustering.

\begin{figure}[!htb]
\centering
\textbf{Clustering results GEBI}

 \begin{subfigure}[Cluster 1.]{
      \includegraphics[width=0.45\linewidth]{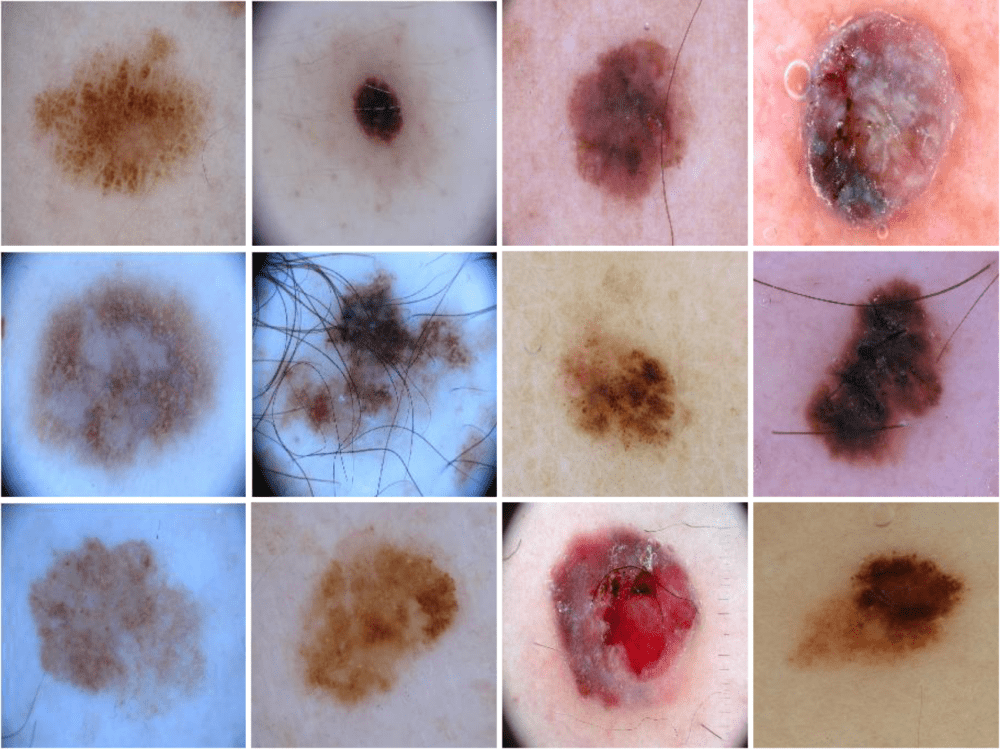}%
      }
    \end{subfigure}
    \begin{subfigure}[Cluster 2.]{
      \includegraphics[width=0.45\linewidth]{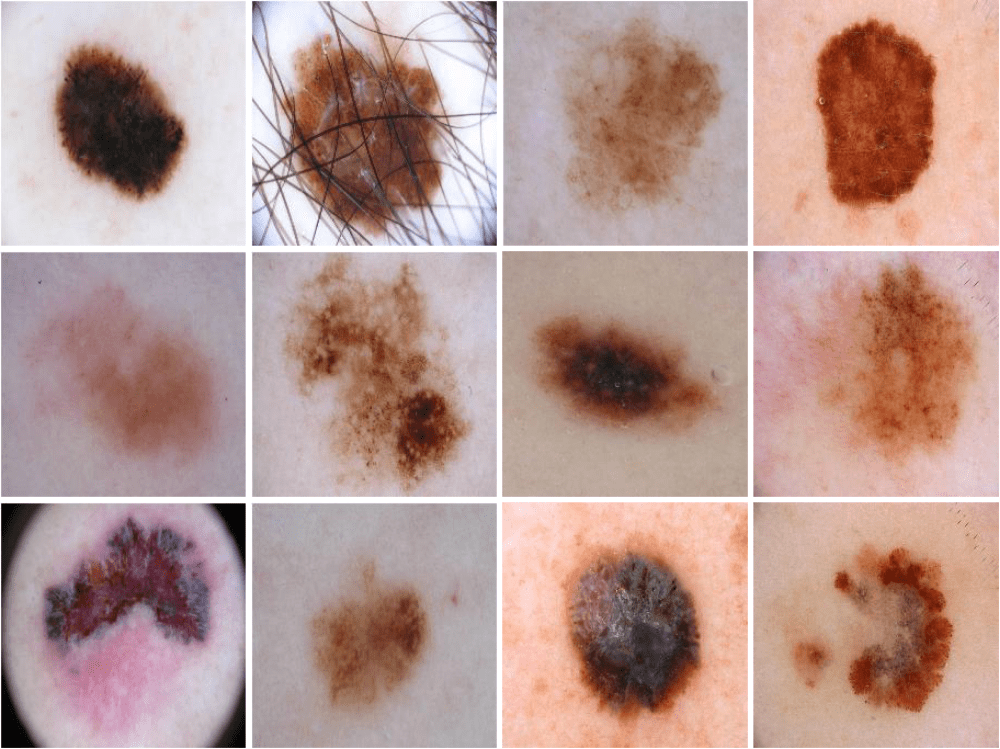}%
      }
    \end{subfigure}

     \begin{subfigure}[Cluster 3.]{
      \includegraphics[width=0.45\linewidth]{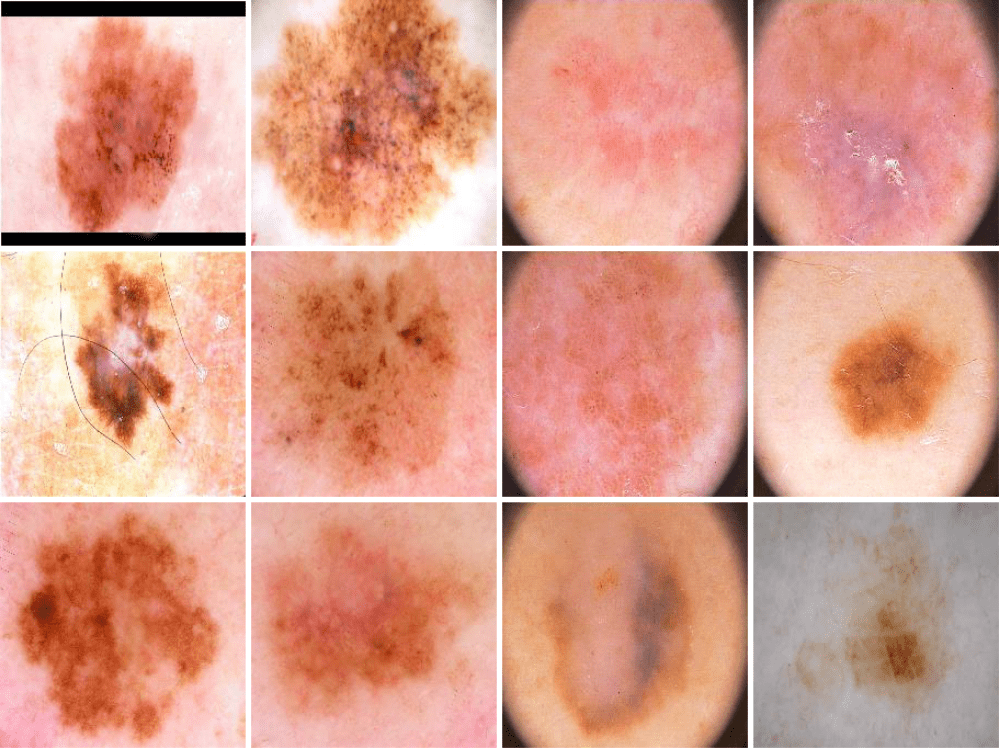}%
      }
    \end{subfigure}
      \begin{subfigure}[Cluster 4.]{
      \includegraphics[width=0.45\linewidth]{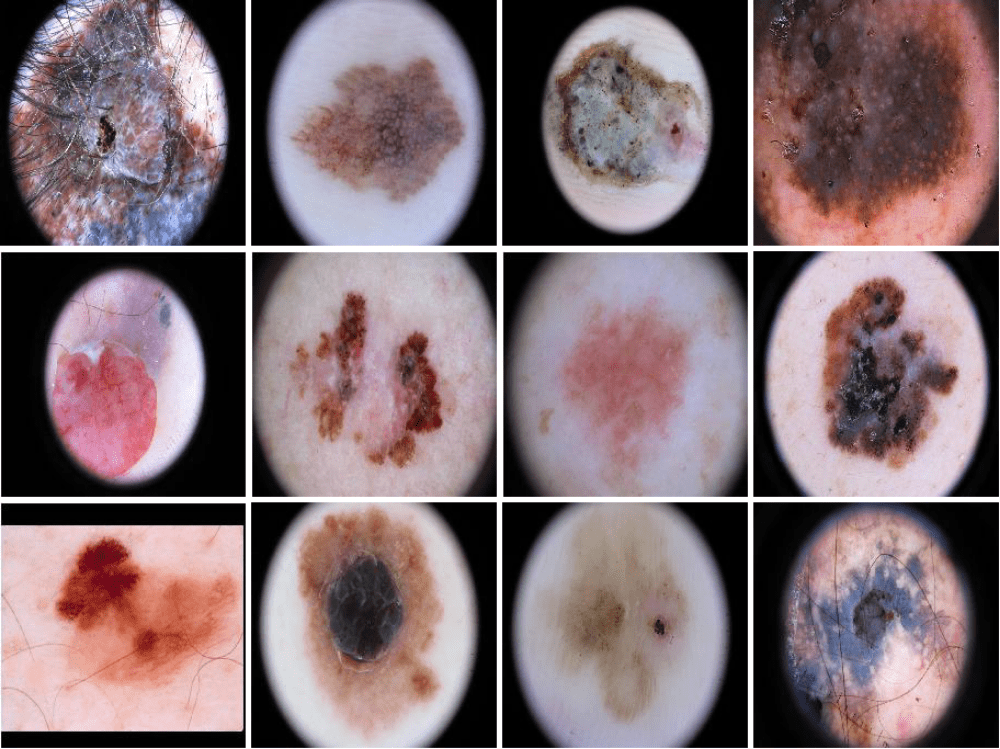}%
      }
    \end{subfigure}
  
 \caption{Resulting clusters achieved with GEBI - 12 first images from each cluster due to alphabetical filename order.} \label{fig.results-gebi}
\end{figure}

The resulted clusters are illustrated in Figure~\ref{fig.results-gebi}. It is worth noticing how cluster 4 stands out from others. It is easy to notice that most images have strong black frames.

\textbf{SpRAy.} Here, I followed the SpRAy methodology to identify potential prediction strategies for skin lesion classification. I used the Deep Taylor Decomposition method. All experiments have been done with Spectral clustering as in the original paper \cite{lapuschkin2019unmasking}. Using the Elbow method, I selected four clusters of interest. 

Firstly, an attribution-based clustering was performed: I clustered resized attribution maps with spectral clustering. Then, both images and heatmaps were mapped into 3d space for visualization purposes, allowing me to compare how well images and attribution maps are clustered. 

\begin{figure}[!htb]
\centering

\normalsize

 \begin{subfigure}[Attribution maps in the embedding space. Each dot represents different attribution map; each color different cluster.]{
      \includegraphics[width=0.45\linewidth]{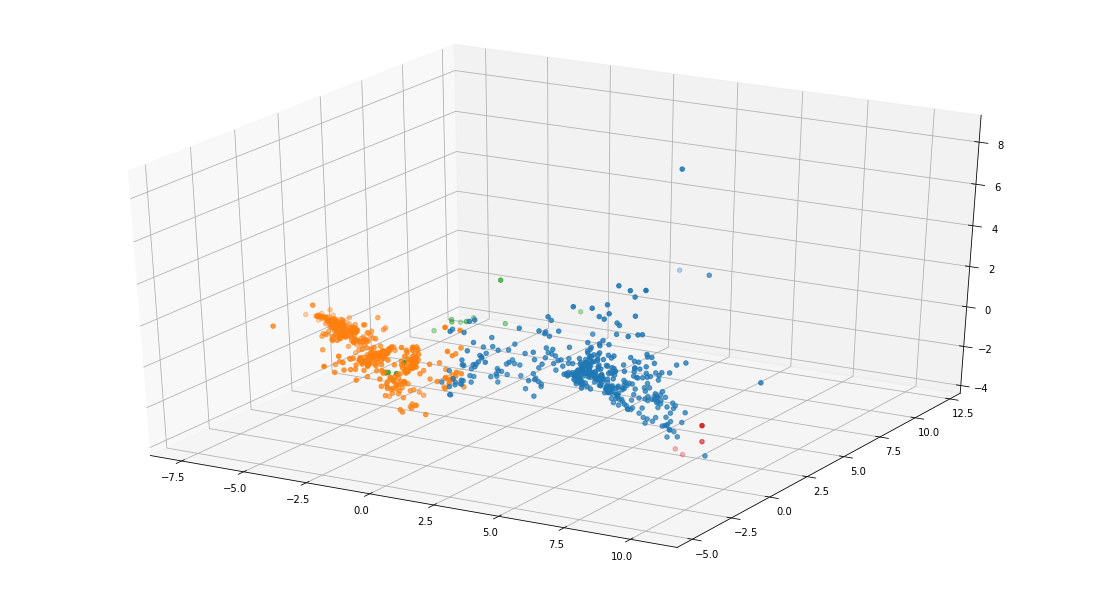}%
      }
    \end{subfigure}
    \begin{subfigure}[Images in the embedding space. Each dot represents different image; each color different cluster.]{
      \includegraphics[width=0.45\linewidth]{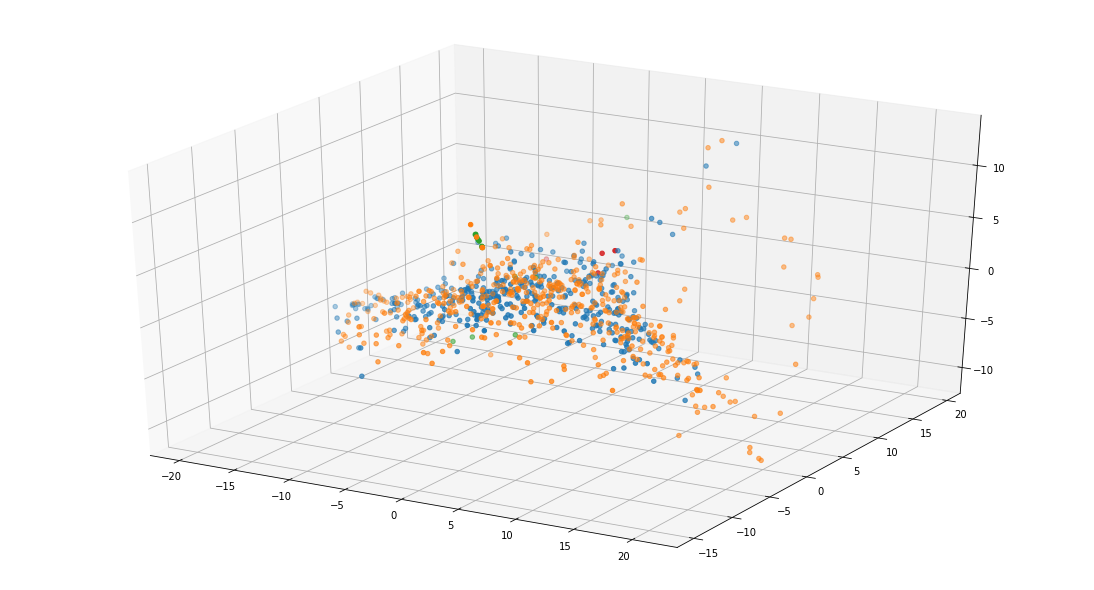}%
      }
    \end{subfigure}
    
    
     \begin{subfigure}[Attribution maps in the embedding space. Each dot represents different attribution map; each color different cluster.]{
      \includegraphics[width=0.45\linewidth]{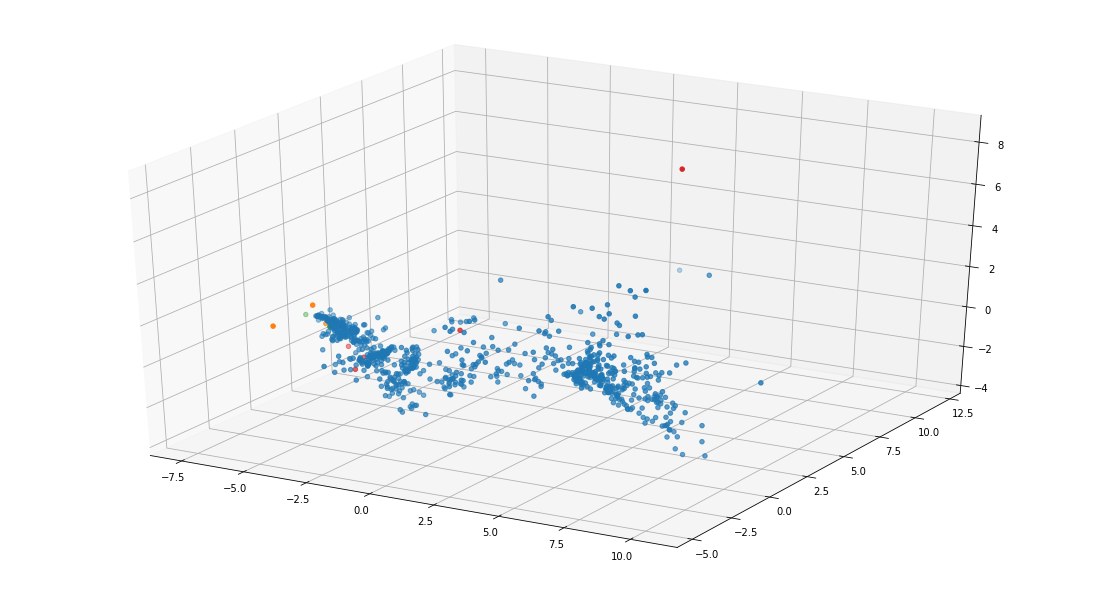}%
      }
    \end{subfigure}
     \begin{subfigure}[Images in the embedding space. Each dot represents different image; each color different cluster.]{
      \includegraphics[width=0.45\linewidth]{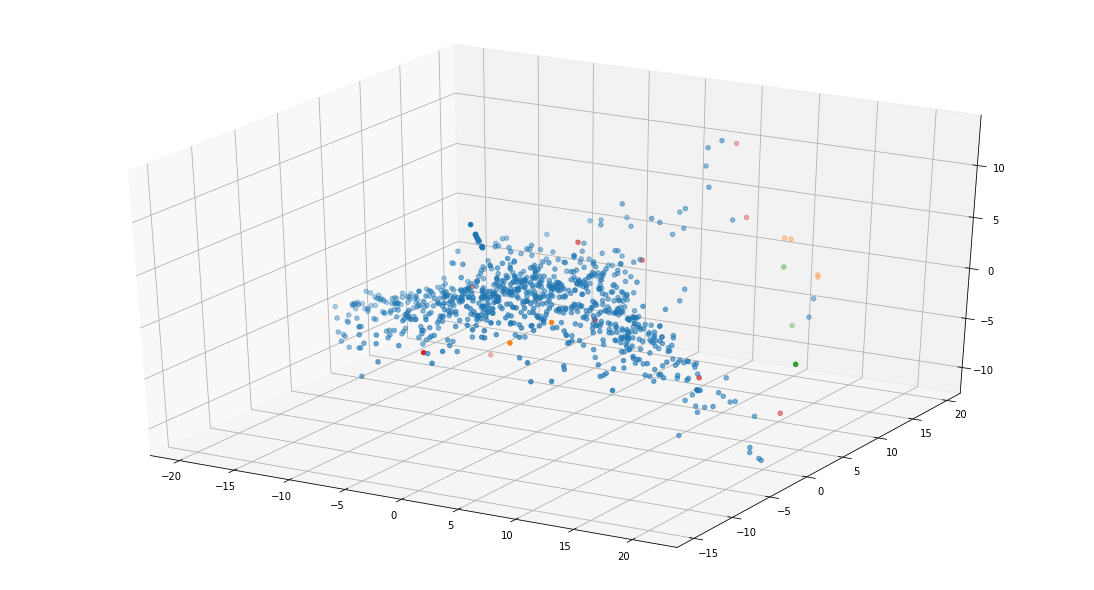}%
      }
    \end{subfigure}

 \caption{Spectral Relevance Attribution is applied solely to images or attribution maps.} \label{fig.embedding-SpRAy}
\end{figure}

Then the clustering process is repeated on input instances instead of attribution maps: I clustered resized images and followed the visualization process (image-based clustering).
Here, I present images and attribution maps in their 3d embedding space separately in Figure~\ref{fig.embedding-SpRAy}.

\begin{figure}[!htb]
\centering

\textbf{Attribution-based clustering}

\raggedleft
 \begin{subfigure}[Cluster 1.]{
      \includegraphics[width=0.45\linewidth]{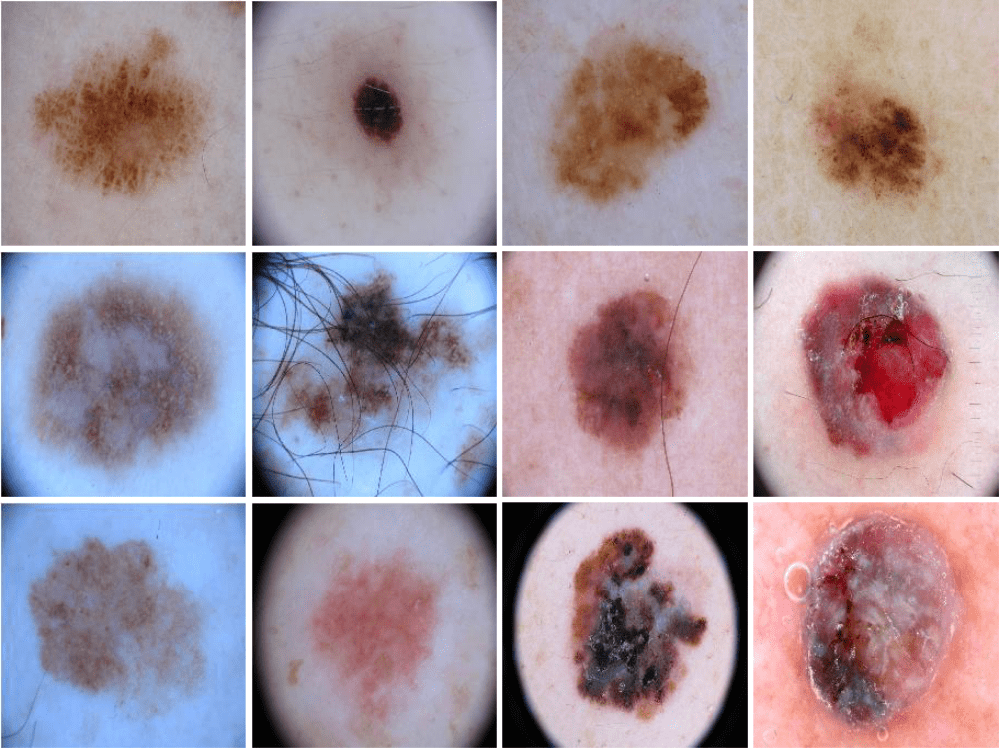}%
      }
    \end{subfigure}
    \begin{subfigure}[Cluster 2.]{
    \vspace{1cm}
      \includegraphics[width=0.225\linewidth]{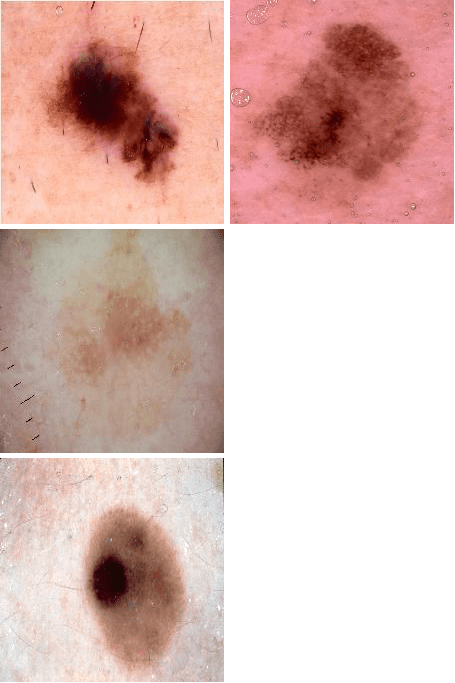}%
      \hspace{0.225\linewidth}
      }
    \end{subfigure}
    \centering
     \begin{subfigure}[Cluster 3.]{
      \includegraphics[width=0.3375\linewidth]{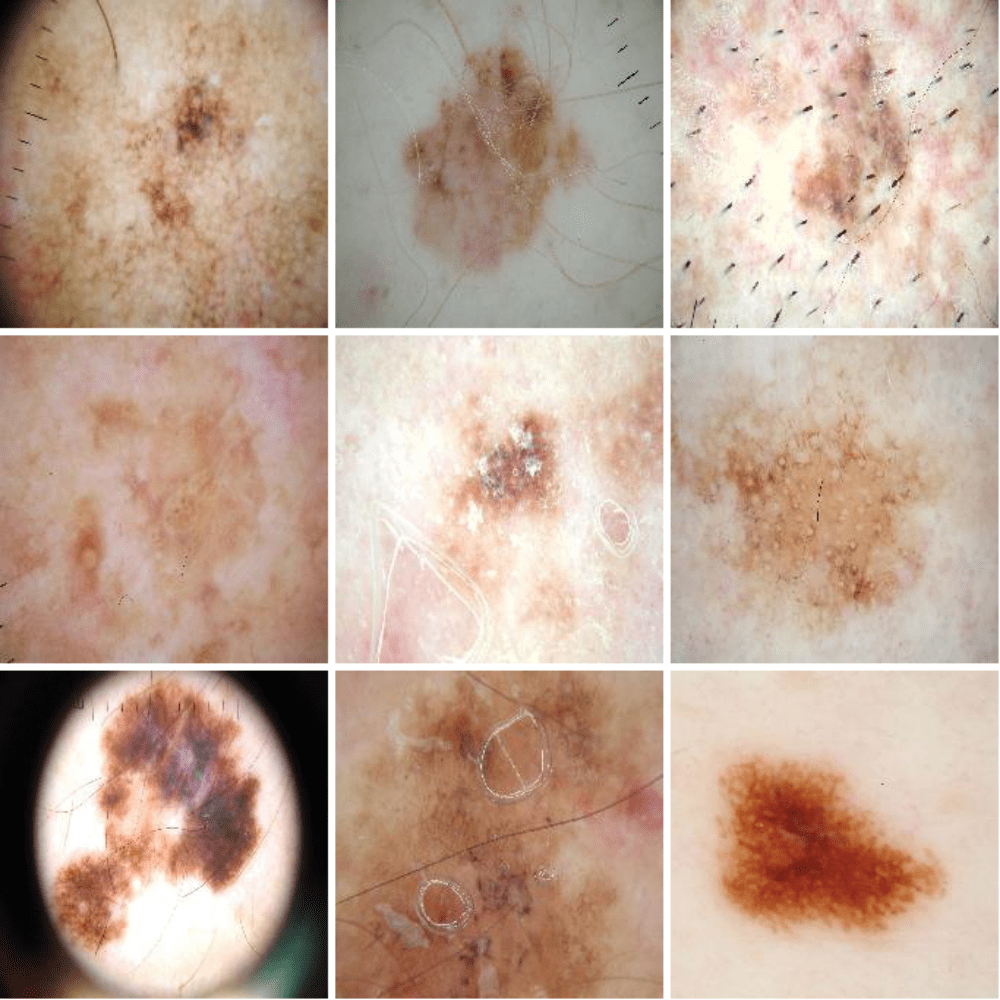}%
      \hspace{0.125\linewidth}
      }
    \end{subfigure}
     \begin{subfigure}[Cluster 4.]{
      \includegraphics[width=0.45\linewidth]{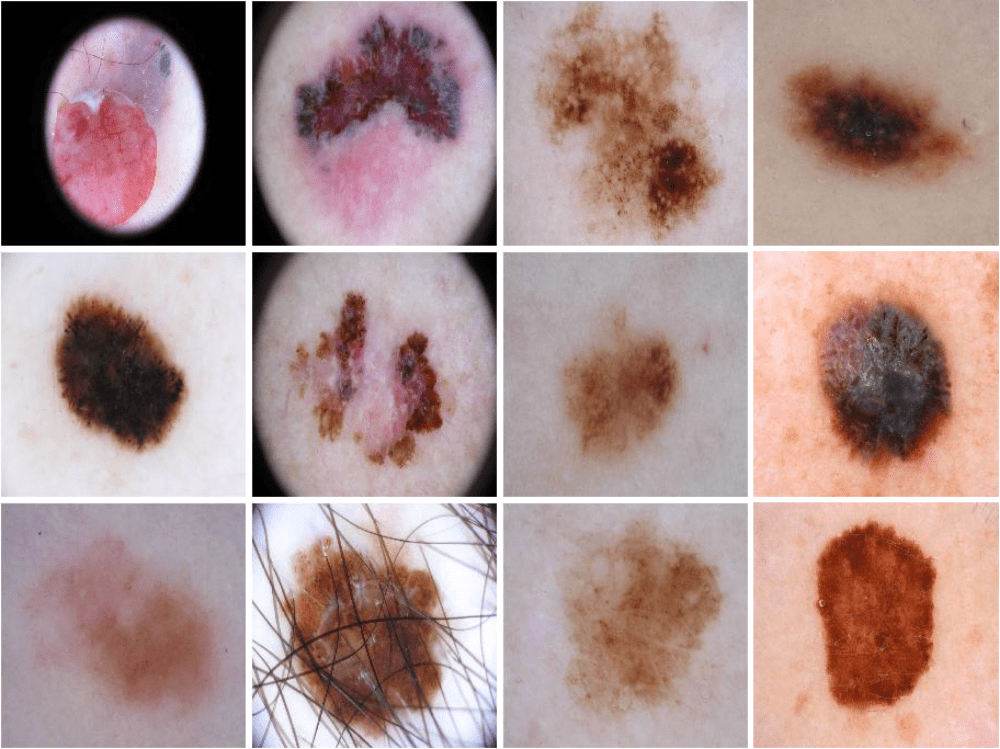}%
      }
    \end{subfigure}
    
 \textbf{Input-based clustering}  
  
 \begin{subfigure}[Cluster 1.]{
      \includegraphics[width=0.45\linewidth]{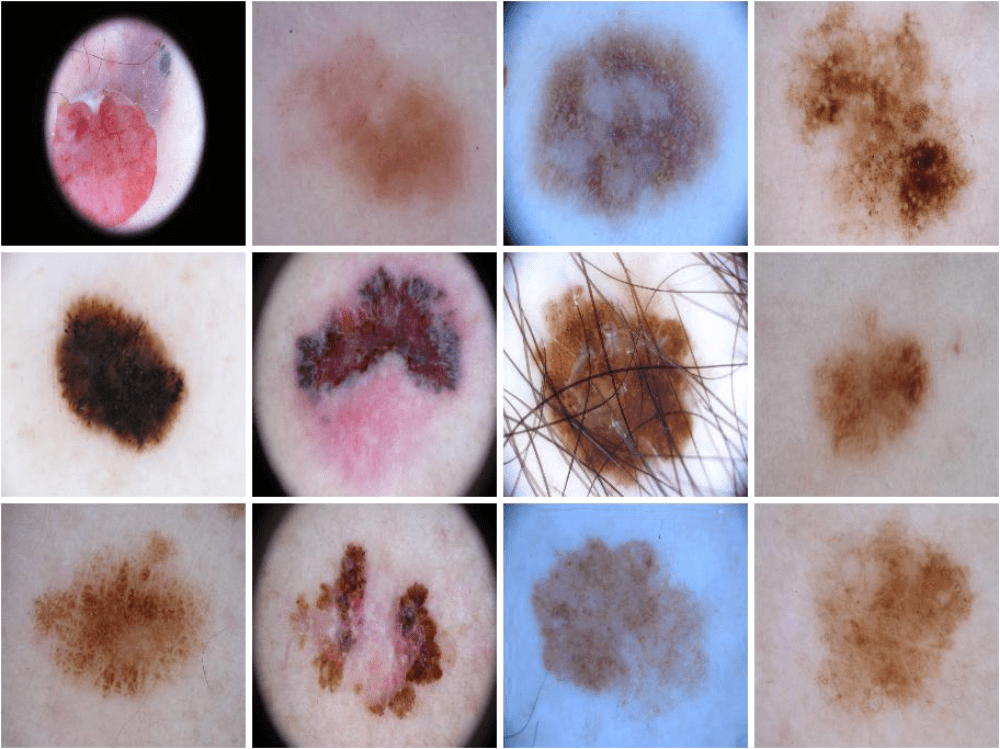}%
      }
    \end{subfigure}
    \begin{subfigure}[Cluster 2.]{
      \includegraphics[width=0.1125\linewidth]{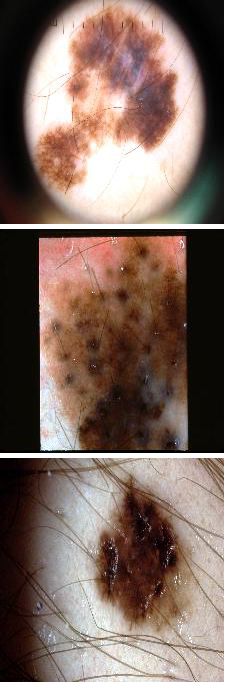}%
      }
    \end{subfigure}
     \begin{subfigure}[Cluster 3.]{
      \includegraphics[width=0.1125\linewidth]{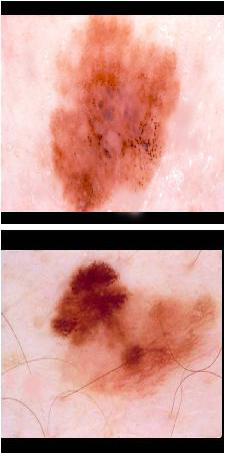}%
      }
    \end{subfigure}
      \begin{subfigure}[width=0.1125\linewidth][Cluster 4.]{
      \newline
      (empty)
      \newline
      \hspace{0.125\linewidth}
      }
    \end{subfigure}
 \caption{Resulting clusters achieved with SpRAy - 12 first images (or less) from each cluster due to filename alphabetical order.} \label{fig.results-SpRAy}
\end{figure}

The best results were achieved when clustering on heatmaps. As reported in the original SpRAy paper, \cite{lapuschkin2019unmasking} clustering only on images did not give any additional insight into the model -- most images have been assigned to only one cluster. Spectral clustering on attribution maps showed better results, yet most instances were divided into two groups.
Unfortunately, visual inspection of corresponding clusters did not give meaningful insight, as presented in Figure ~ \ref{fig.results-SpRAy}.

The source of the problem lies in the clustered representation, i.e., attribution maps that have been red in the middle have been clustered together. Images with most attention focused on the right side -- together again. Clustering based on the attribution localization might be an advantage in the case of analyzing images that have only one proper perspective to analyze, i.e., portraits, videos from antonymous cars, and landscape images. This is a disadvantage in issues where image rotation does not matter (microscopic images, skin images, aerial images).

\textit{IsoSpRAy. } As previously mentioned, I followed the SpRAy methodology \cite{lapuschkin2019unmasking} to identify potential prediction strategies for skin lesion classification. I modified the single step in SpRAy: reduced the dimensionality of images with the Isomap algorithm instead of an image downscaling. IsoSpRAy can be interpreted as the midway between GEBI and SpRAy. 
I used the deep Taylor decomposition method to generate attribution maps again. All experiments have been done with Spectral clustering as in the previous subsection. I selected 4 clusters of interest. 

\begin{figure}[!htb]
\centering
\Large

\normalsize
\textbf{Attribution-based clustering}

 \begin{subfigure}[Attribution maps in the embedding space. Each dot represent different attribution map; each color different cluster.]{
      \includegraphics[width=0.45\linewidth]{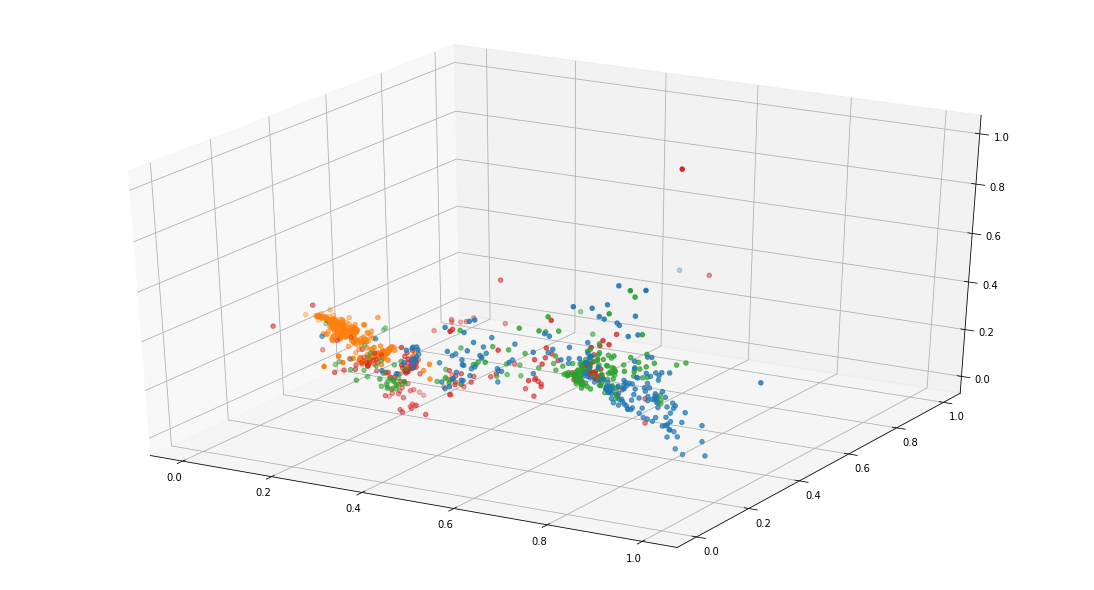}%
      }
    \end{subfigure}
    \begin{subfigure}[Images in the embedding space. Each dot represent different input image; each color different cluster.]{
      \includegraphics[width=0.45\linewidth]{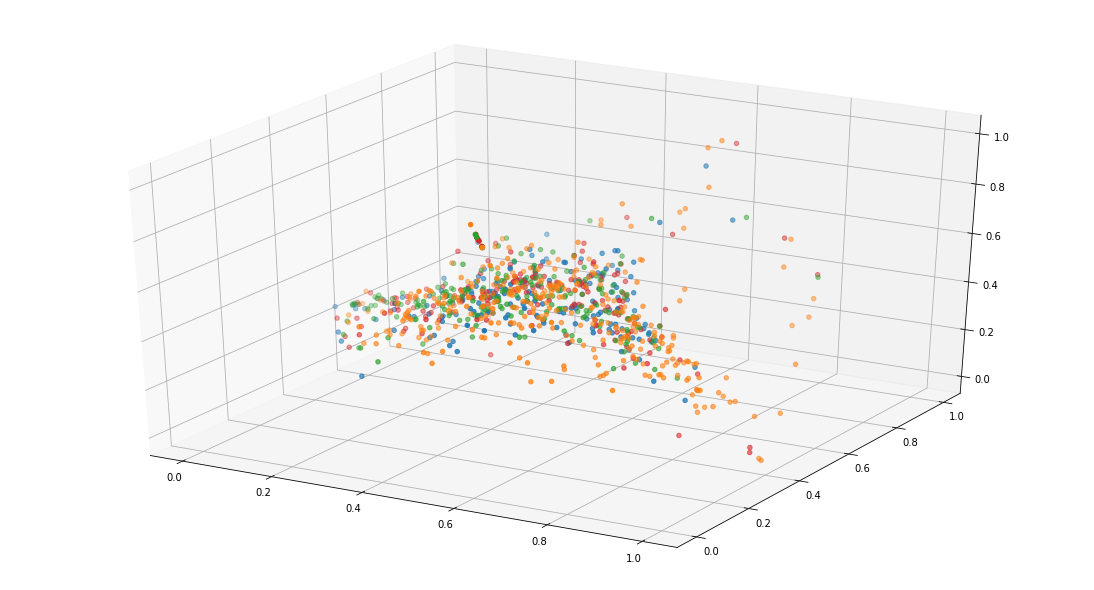}%
      }
    \end{subfigure}
    
    \textbf{Input-based clustering}  
    
     \begin{subfigure}[Attribution maps in the embedding space. Each dot represent different attribution map; each color different cluster.]{
      \includegraphics[width=0.45\linewidth]{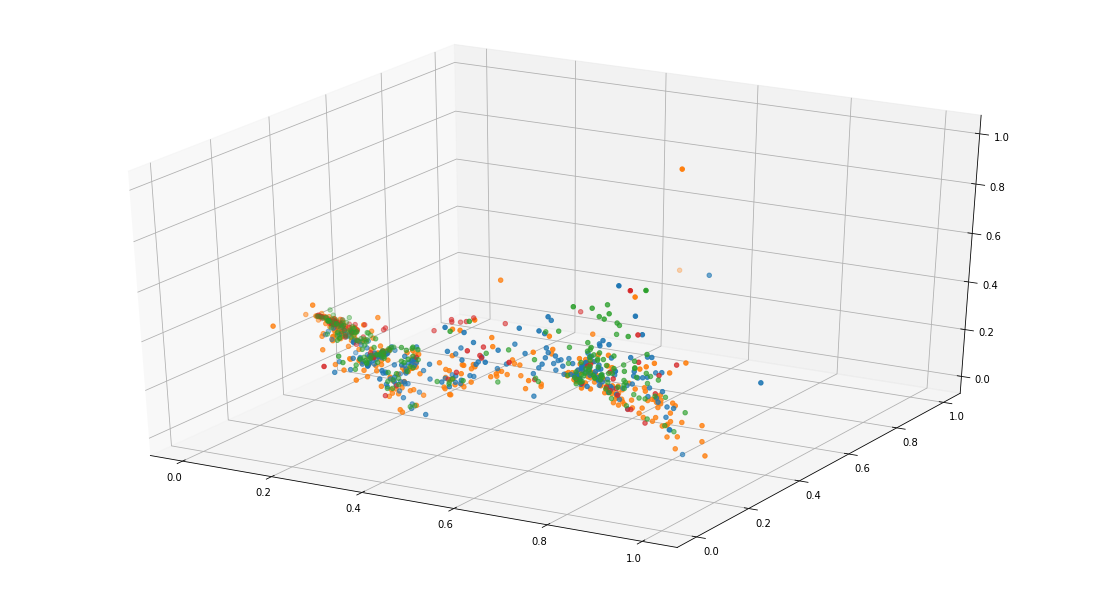}%
      }
    \end{subfigure}
     \begin{subfigure}[Images in the embedding space. Each dot represent different input image; each color different cluster.]{
      \includegraphics[width=0.45\linewidth]{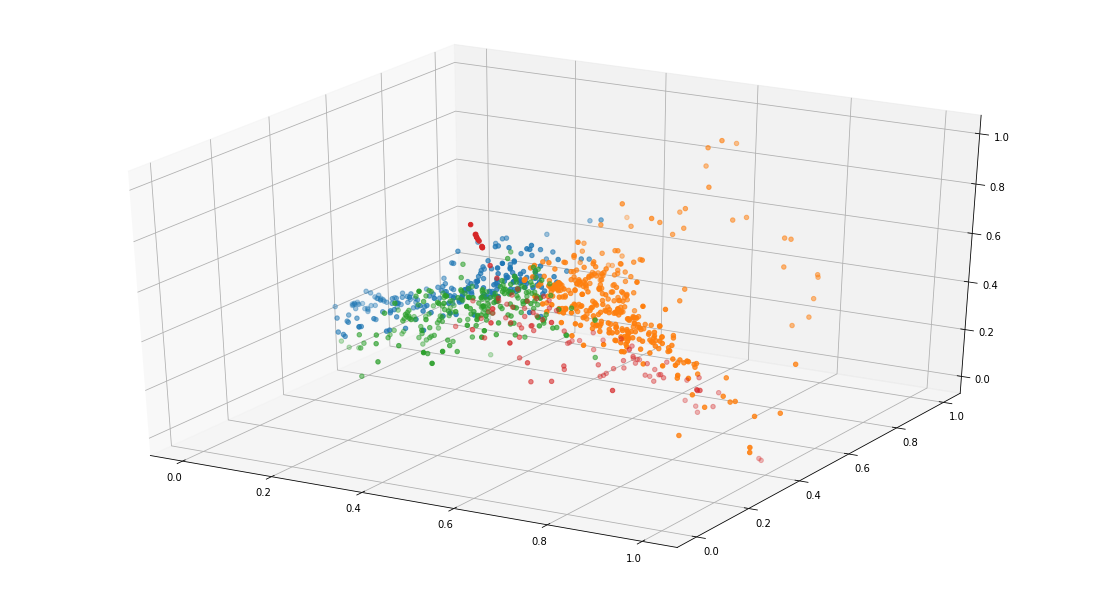}%
      }
    \end{subfigure}

 \caption{Spectral Relevance Attribution with Isomap reduction applied solely on images or attribution maps.} \label{fig.embedding-isoSpRAy}
\end{figure}

I present images and attribution maps separately in their 3d embedding space in Figure~\ref{fig.embedding-isoSpRAy}. The process of visualization is described in the previous subsection.

\begin{figure}[!htb]
\centering
 \textbf{ Input-based clustering}  
  
 \begin{subfigure}[Cluster 1.]{
      \includegraphics[width=0.45\linewidth]{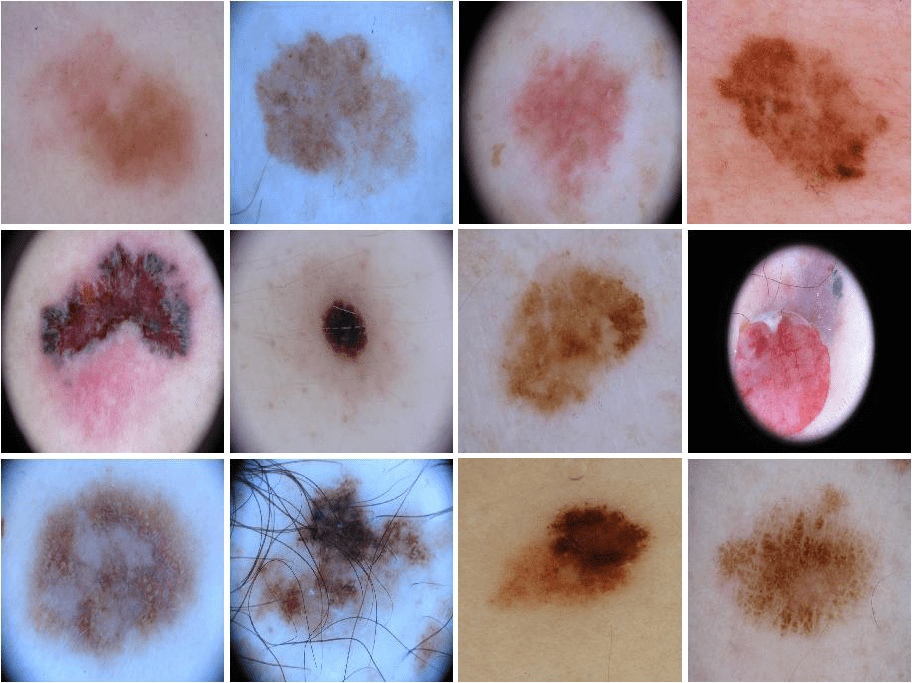}%
      }
    \end{subfigure}
    \begin{subfigure}[Cluster 2.]{
      \includegraphics[width=0.45\linewidth]{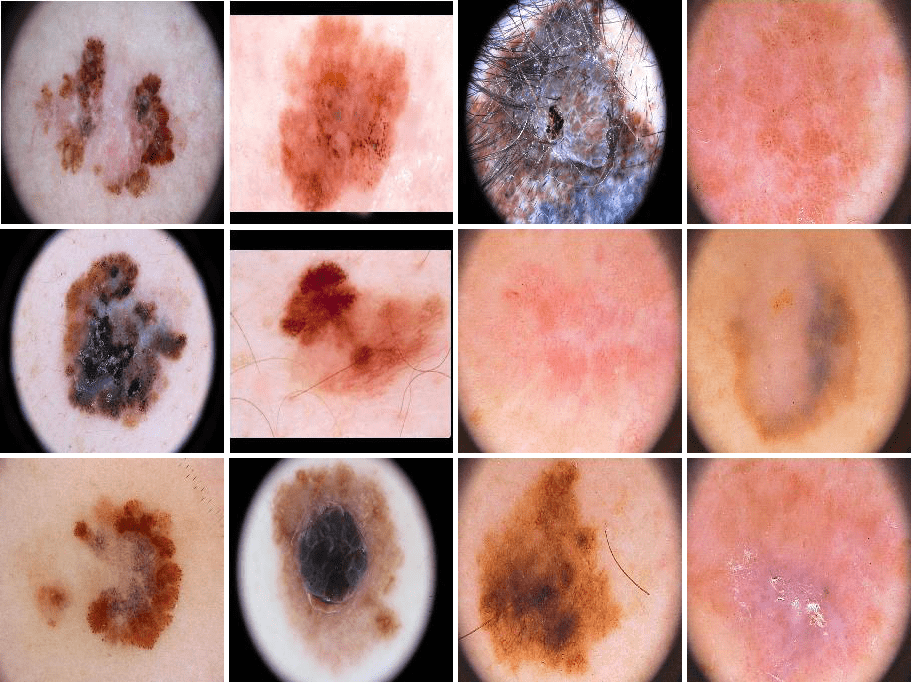}%
      }
    \end{subfigure}

     \begin{subfigure}[Cluster 3.]{
      \includegraphics[width=0.45\linewidth]{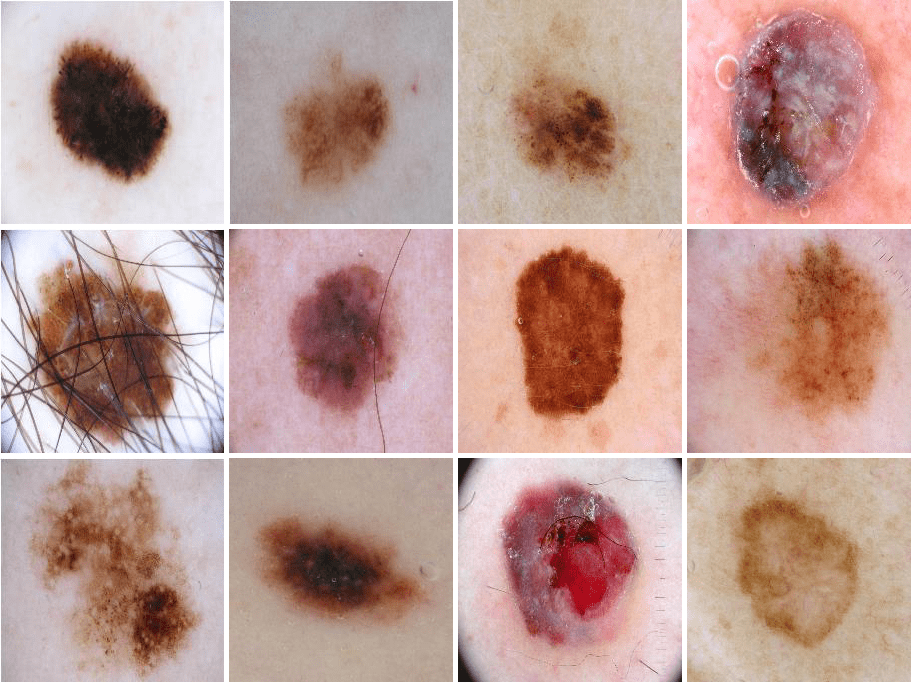}%
      }
    \end{subfigure}
      \begin{subfigure}[Cluster 4.]{
      \includegraphics[width=0.45\linewidth]{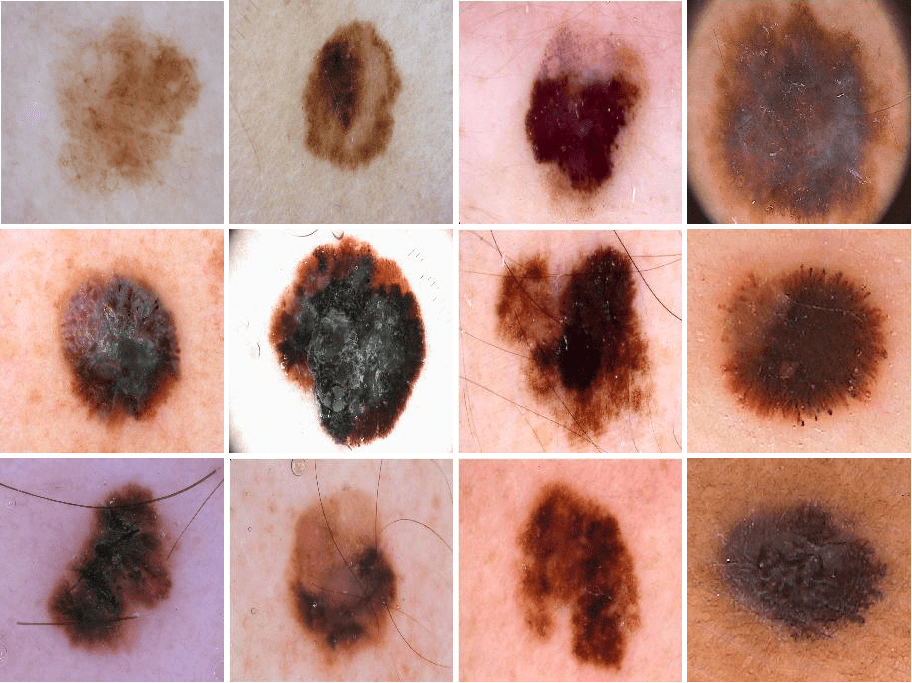}%
      }
    \end{subfigure}
 \caption{Input-based clustering. Resulting clusters achieved with IsoSpRAy - 12 first images from each cluster due to alphabetical filename order.} \label{fig.results-isoSpRAy-color}
\end{figure}

Again, the best results were achieved when clustering on heatmaps. However, contrary to SpRAy, this time, clustering only on images resulted in more than one cluster. When I applied input-based clustering on embeddings from the isomap algorithm, I achieved color-focused clusters, as presented in Figure~\ref{fig.results-isoSpRAy-color}. As a result, each cluster showed light or dark and textured or clean lesions/skin. Also, the two first clusters showed some examples with frames, whereas the other did not. Overall, it is hard to tell major differences between the clusters.

\begin{figure}[!htb]
\centering
\textbf{Attribution-based clustering}  

 \begin{subfigure}[Cluster 1.]{
      \includegraphics[width=0.45\linewidth]{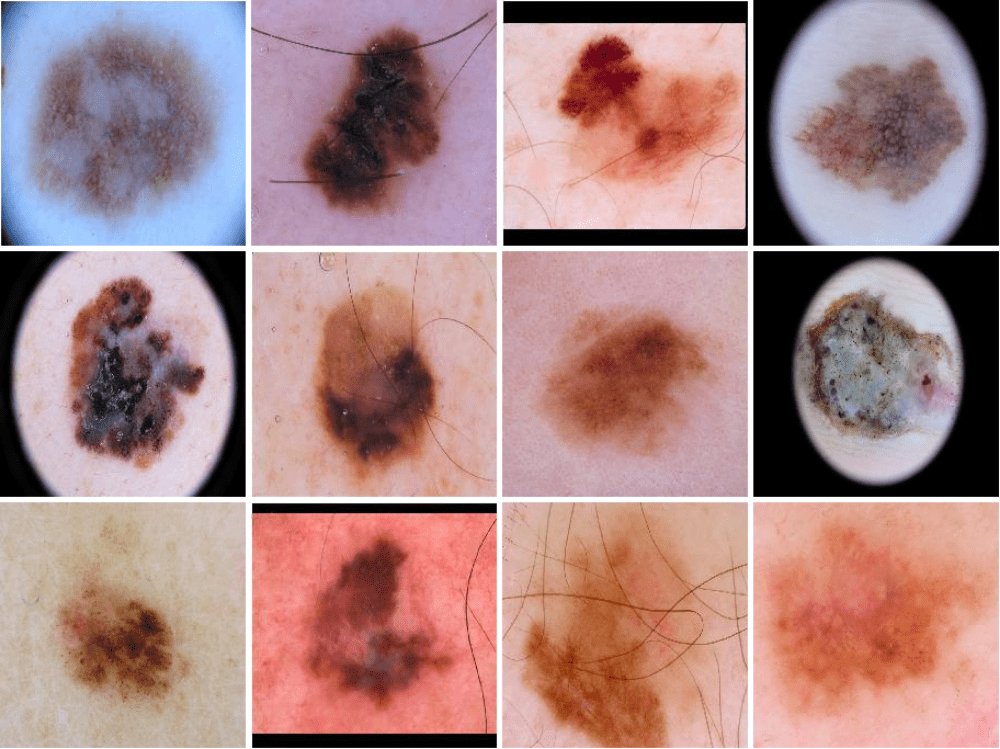}%
      }
    \end{subfigure}
    \begin{subfigure}[Cluster 2.]{
      \includegraphics[width=0.45\linewidth]{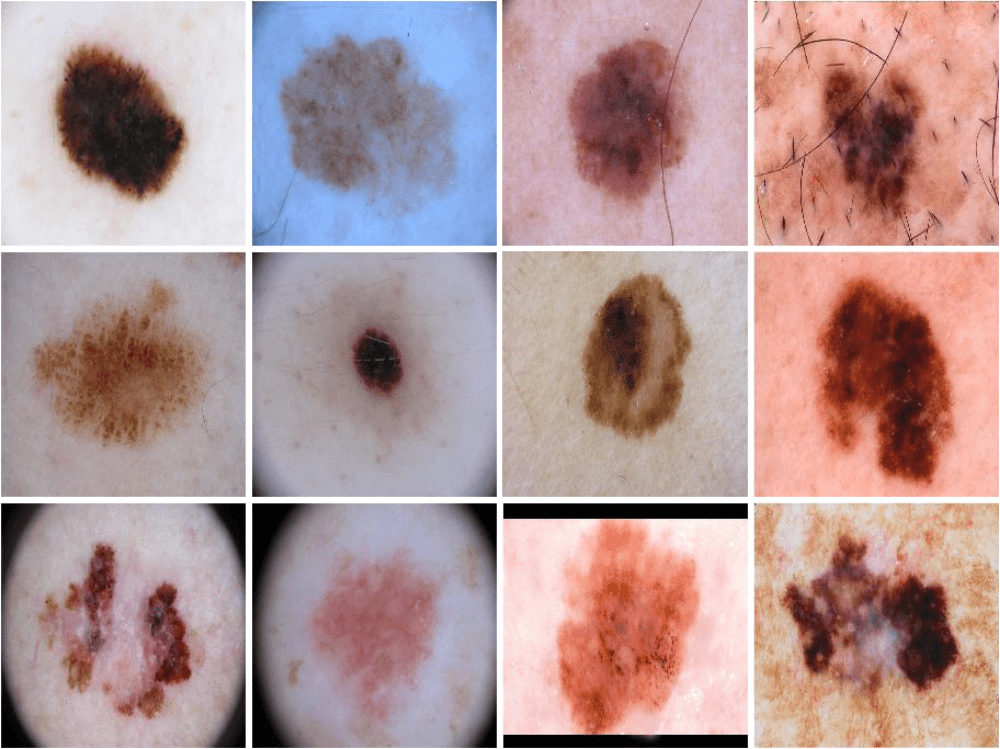}%
      }
    \end{subfigure}
    
     \begin{subfigure}[Cluster 3.]{
      \includegraphics[width=0.45\linewidth]{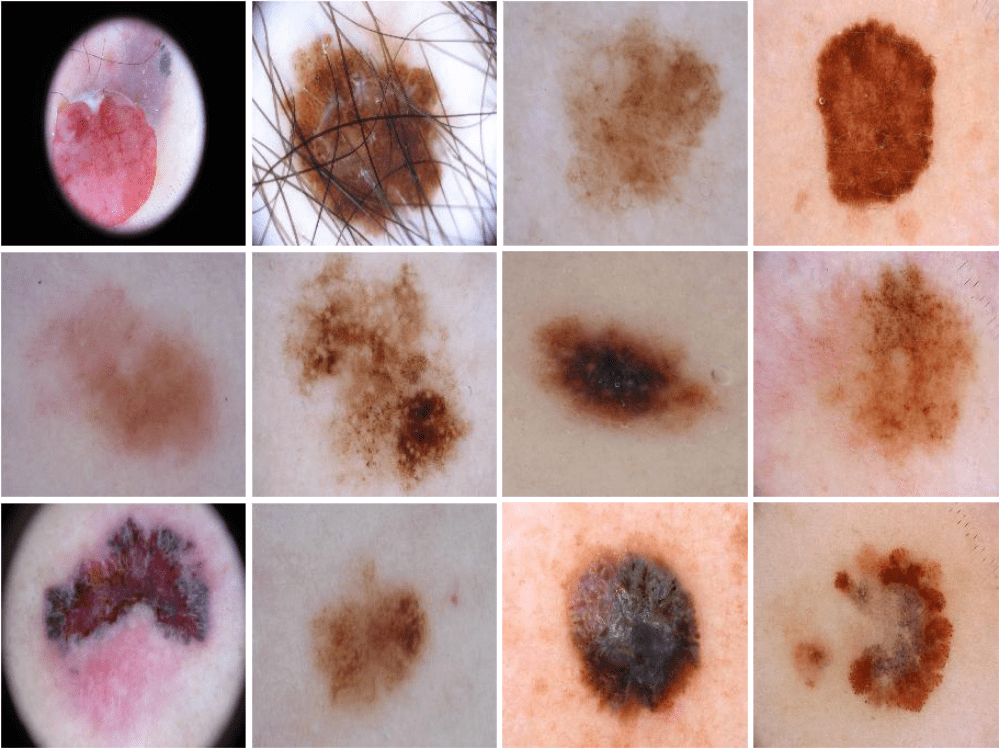}%
      }
    \end{subfigure}
     \begin{subfigure}[Cluster 4.]{
      \includegraphics[width=0.45\linewidth]{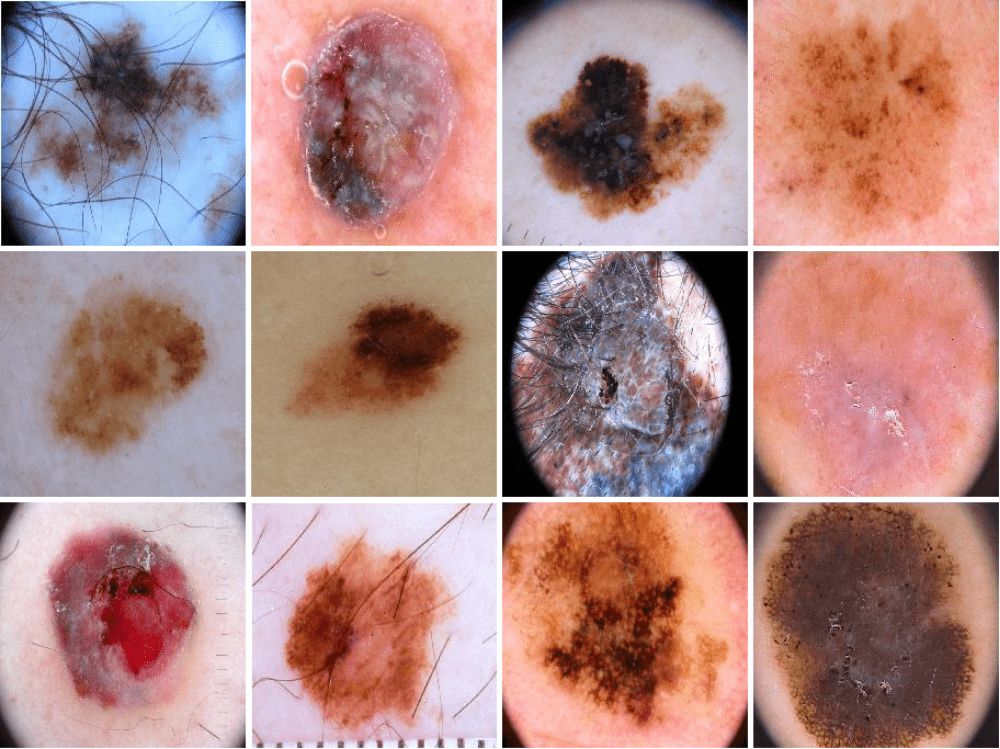}%
      }
    \end{subfigure}

 \caption{Attribution-based clustering. Resulting clusters achieved with IsoSpRAy - 12 first images from each cluster due to alphabetical filename order.}\label{fig.results-isoSpRAy-attribution}
\end{figure}

I achieved four distinctive clusters numbering from 180 to 260 images per cluster when clustering attribution maps. The resulting clusters showed huge variations in lesions colors and shapes. The artifacts also seemed not to play a significant factor. Clusters achieved with attribution-bassed IsoSpRAy are still difficult to interpret and hence, not valuable. Example clusters can be observed in Figure~\ref{fig.results-isoSpRAy-attribution}. 

As the experiments showed, there is a great difference between the results obtained with GEBI, SpRAy, and IsoSpRAy. The method GEBI overcomes the challenges connected to running clustering based only on the attribution maps or inputs. Input-based clustering ended up in grouping together primarily based on the color or white balance. On the other hand, clustering solely on the attribution maps was a shape-based clustering that focused on grouping together images with the attribution concentrate on specific places of the heatmap -- concatenating both of those led to much better results. This would be a preferable method where both shape and texture plays a vital role in decision-making. 

\subsection{Testing bias influence with Counterfactual Bias Insertion}

I experimented with evaluating how each predicted strategy impacted the model's decision. Two possible biases were identified: black frames and ruler marks. 
I prepared and tested how inserting an artifact into the image would change the prediction for this stage. 
The following artifacts were tested: black frames, ruler marks, and red circles.

\textbf{Black frames} were added to all images with no variations in size and position. Black frames are common in a skin lesion dataset and are often referred to as unwanted artifacts. They can be observed depending on the dermatoscope type.

\textbf{Ruler marks} are inserted with random size, angle, and position. Medical specialists often use rulers to show the skin lesion's size on the dermoscopic image. Additionally, certain dermatoscopes have built-in ruler marks that will appear in every image.

\textbf{Red circles} cannot be naturally found in standard skin lesion datasets; hence the difference in the predictions should be scarce. However, I use this example as a baseline result: inserted features that play any role in the classification process should show similar properties. Single red circles were placed randomly in the images, both within the skin and lesion areas. 

\begin{figure}[!htb]
\centering

\textbf{Ruler marks}  

 \begin{subfigure}[Inputs with inserted ruler mark. Ruler marks were randomly placed at the bottom half on the image.]{
      \includegraphics[width=0.5\linewidth]{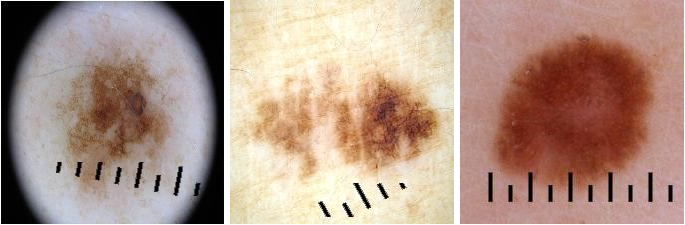}%
      }
    \end{subfigure}

\textbf{Black frames}  

 \begin{subfigure}[Inputs with inserted black frame. Black frames were placed in each image.]{
      \includegraphics[width=0.5\linewidth]{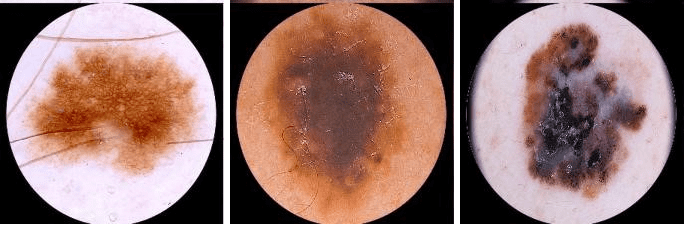}%
      }
    \end{subfigure}

\textbf{Red circles}  

 \begin{subfigure}[Inputs with inserted circle. As red circles are not a natural part of skin lesion datasets, I decided to use it as a baseline result.]{
      \includegraphics[width=0.5\linewidth]{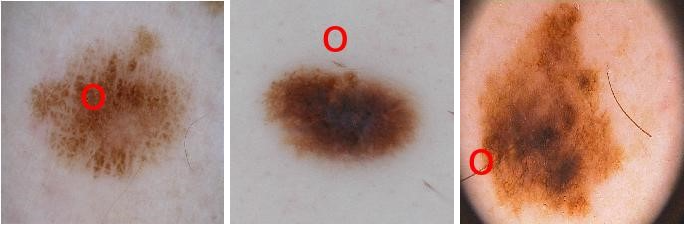}%
      }
    \end{subfigure}
   
 \caption{Example inputs with inserted bias.} \label{fig.example-bias-insertion}
\end{figure}

Modified samples are presented in Figure~\ref{fig.example-bias-insertion}.

I prepared modified samples with inserted artifacts for all benign and malignant skin lesions. 
Then predictions for 884 randomly selected malignant and benign samples were computed. I compared original predictions with predictions made for all biased samples. The results are presented in Table~\ref{tab:bias_insertion_results}. Table presents the maximum and mean change in prediction: lower values are desirable as they mean that inserting the bias did not significantly affected the output.

\begin{table}[]
\centering
\caption{Bias insertion experiments results -- maximum and mean change in prediction.}
\label{tab:bias_insertion_results}
\begin{tabular}{@{}llll@{}}
\toprule
\textbf{Bias type} & \textbf{Class} & \textbf{Mean} & \textbf{Max} \\ \midrule
Ruler mark    & Malignant       & 2.21                                           & 22.01                                          \\
              & Benign       & 1.23                                           & 19.91                          \\
Black frame   & Malignant       & 30.77                                          & 62.43                                          \\
\textbf{}     & Benign       & 32.04                                          & 63.66                                          \\
Red circle    & Malignant       & 2.27                                           & 15.51                                          \\ 
              & Benign       & 1.50                                           & 12.78  \\ \bottomrule
\end{tabular}%

\end{table}

The highest recorded prediction differences existed for the black frame insertion. On average, every output changed by 30.77\%. Adding frames biases the model towards the malignant class; 197 out of 884 skin lesions switched prediction to malignant after frame insertion (about 22.29\% of samples switched the predicted class). It is a significant finding to consider when training new models. Although black frames did not alter the skin lesion's appearance, predictions changes were considerable for malignant and benign skin lesions. Such a notable shift in prediction score should arouse some distrust in models' behavior. 

In contrast, introducing a ruler mark or a red circle did not change prediction scores. Ruler marks caused, on average, only a slight shift in predictions, of about 1.23 and 2.21 pp. However, still, it might be a dangerous reaction in some cases, as I observed a few instances that changed the predicted class from malignant to benign. 

Adding a red circle did not affect the output, but surprisingly, it was similar to the average change for ruler placement. A small number of approximately 1.5\% of images switched prediction from benign to malignant. A possible reason is that the red circle might be similar to dermatological attributes (e.g., atypical structures like blobs, dots, or streaks).

\section{Discussion}

I introduced and described a new method to detect flaws in datasets that influence the model's prediction. I conducted the necessary experiments and showed that SpRAy standalone might be difficult to use in datasets where image orientation does not matter. However, after only a few additional steps, the results get more precise and easier to interpret. I presented IsoSpRAy and GEBI, two methods for global explainability. The methods allow for detecting biases in data in a semi-supervised manner. 

Additionally, I introduced the counterfactual bias insertion method that measures bias influence on the prediction. As a result, I confirmed with a bias insertion measure that black frames, commonly found in skin lesion datasets, significantly affect model predictions. This finding shows the urgent need to reexamine the training dataset or reevaluate the training procedure. 

This part identified and explained what biases could be found in a machine learning pipeline and presented examples that might help understand the problem. As biases are usually very well hidden, I proposed to use explainable AI to help find them. Hence, in the \textit{Chapter \ref{chapter.xai}: Explainable Artificial Intelligence} I introduced the most common explainable AI algorithms. I showed how and when explainable AI could help investigate the reasons behind the model's predictions. Next, in Chapters \ref{chapter.skin_lesion_bias} and \ref{chapter.gebi}, two different approaches to bias identification were proposed: manual and semi-automated. Manual annotation is a time-consuming and reliable method of investigating the dataset. It gives a great insight into the problem showing what features are correlated with each other and whether class instances are correlated. The semi-automated method does not need annotations. Still, it requires more machine learning and XAI knowledge, as it requires analyzing the achieved clusters that include attribution maps and inputs. Both approaches require expert knowledge: when analyzing data, we must know what features are causing the prediction and what only creates a spurious correlation. My research is supported by the thorough experiments run on the public and widely used dataset of dermoscopic lesions. The results approached through different means (manual annotation vs. GEBI) agreed in terms of the results: the examined skin lesions dataset appears biased towards certain features. In the next part of the thesis, I focused on mitigating bias in data during the model's training.


\clearpage  
\part{Mitigating bias in data and models}

\clearpage  
\lhead{\emph{Chapter 6: Mitigating shape and texture bias with data augmentation}}  
\chapter{Mitigating shape and texture bias with data augmentation} \label{chapter.neural-style}
\section{Introduction}
Data augmentation finds its applicability in any deep learning-based method. In computer vision, images are usually augmented by simple, linear alterations and color augmentations, including sometimes more advanced augmentations like a cutout or neural transformations \cite{shorten2019survey}. In NLP, it is not uncommon to augment data with contextualized word embeddings, back translation, or the thesaurus \cite{wei2019eda}. Similarly, numerous methods were proposed when processing audio, such as pitch modification, time-stretching, or noise injection \cite{ko2015audio}. The input instances (images, text, audio, or others) are slightly modified during the training to increase input data variability without gathering new samples. It is especially useful in areas where the data is difficult to collect, such as fault diagnostics \cite{wang2019time}, law enforcement \cite{wang2018random}, and medicine-related studies \cite{costa2017adversarial,yuan2018regularized,peng2019structured}.

The problem of limited data and class imbalance raises other challenges, such as overfitting and underfitting \cite{goodfellow2016deep} leading to poor generalization abilities \cite{lawrence1997lessons}, or to the so-called accuracy paradox when the accuracy seems to be high, but the model is not robust. 
 Problems with scarce datasets and overfitting seem to be slightly less significant now, shifting the researchers attention to biased data. Datasets gradually grow over time, models are better and more robust, and their knowledge is more transferable. Today, commonly used models are already pretrained and publicly available to download. Models quickly tune to the new problem, and as shown in studies, larger models are usually easier and quicker to finetune \cite{sun2019meta}. However, if data is biased and we falsely believe classification metrics, we might end up with a model that works only on the data from our training distribution. Fortunately, awareness about this problem is rising. Researchers and engineers developed guidelines for testing models before releasing them in production \cite{breck2017ml}. Some of them include testing the models on the out-of-distribution data, i.e., training and testing on different datasets but the same problem. In many cases, using different test sets is enough to detect some difficulties in the model or data, but in some, it is not. The rapid development of the XAI method also sheds new light on the issue of biased models. 

In the past, neural style transfer was usually used to create eye-pleasing artworks. One of the first approaches that considered using style transfer as a data augmentation method for better robustness was style augmentation \cite{jackson2019style}. They defined the problem of a domain bias (domain shift) which is a problem that occurs when the model trained on data from a particular domain does not generalize to other datasets not seen during training. To mitigate the problem, Jackson et al. \cite{jackson2019style} introduced different artistic styles to training images and used them to enhance office cross-domain classification. Those were probably one of the first approaches that intentionally encouraged models to be less biased. 

Recently, a new problem connected to CNN's appeared. The seemingly unrelated issue called a \textit{shape and texture bias}~\cite{hermann2020origins}. Researchers discovered that CNNs tend to be biased towards texture, often ignoring the object's shape, leading to lower performance. The perfect tool for debiasing such models became neural style transfer. 
However, neural style transfer was used to augment datasets even before that. According to my knowledge, I proposed the first approach to neural style transfer data augmentation that used conflicting shapes and textures from two different classes~\cite{mikolajczyk2018data}, which resulted in realistic-looking (not artistic) images. I presented that using conflicting shapes and textures of benign and malignant skin lesions resulted in much more robust models~\cite{mikolajczyk2018data}.

Hence, I proposed one of the first approaches to use \textit{Neural Style Transfer (NST)} to generate more training examples to increase the model's robustness via debiasing model. I proposed to transfer the style from malignant skin lesions to benign ones to mix the shape of benign skin lesions with the texture of the malignant ones. This approach also significantly increased a small dataset's size, allowing for the generation of many new samples. 
I addressed two problems: the problem of scarce datasets and the problem of shape/texture bias. Both are approached by generating more training examples with neural style transfer by using conflicting shapes and textures:

\begin{enumerate}
    \item I introduced the taxonomy behind data augmentation methods. It is a part of my long-standing work that includes collecting and describing new data augmentation methods and repositories.
    \item I described the characteristics of data augmentation with style transfer. I presented the methodology behind the experiments and the results of the experiments.
    \item I inspected the predictions made by the model focusing on the shape and texture using local explanations generated using XAI.
\end{enumerate}


\section{Taxonomy}
\begin{figure}[!htb]
\centering
  \includegraphics[width=1\textwidth]{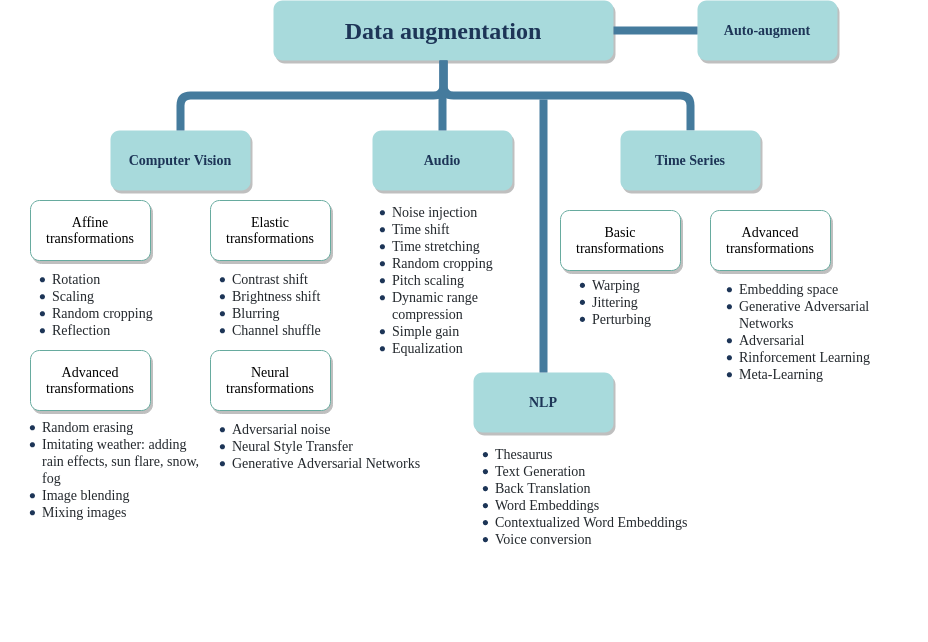}%
  \caption{Data augmentation methods diagram. \cite{Mikolajczyk_AgaMiko_data-augmentation-review_A_comprehensive_2021}} \label{fig.aug}
\end{figure}

\textit{Data augmentation} can be defined as any method of transforming a dataset that makes it larger; or, in other words, any method that slightly modifies the input data during the training to gain more variability. I summarized existing methods of data augmentation in the different applications in the diagram presented in Figure~\ref{fig.aug}.

Since data augmentation is a very broad subject, I restricted the taxonomy section to the most commonly used computer vision augmentation methods. The additional materials can be found online in my GitHub repository\footnote{\href{https://github.com/AgaMiko/data-augmentation-review}{AgaMiko/data-augmentation-review}: List of useful data augmentation resources}. 

\begin{figure*}[!htb]
\centering

    \begin{subfigure}[Original images]{
      \includegraphics[width=0.8\linewidth]{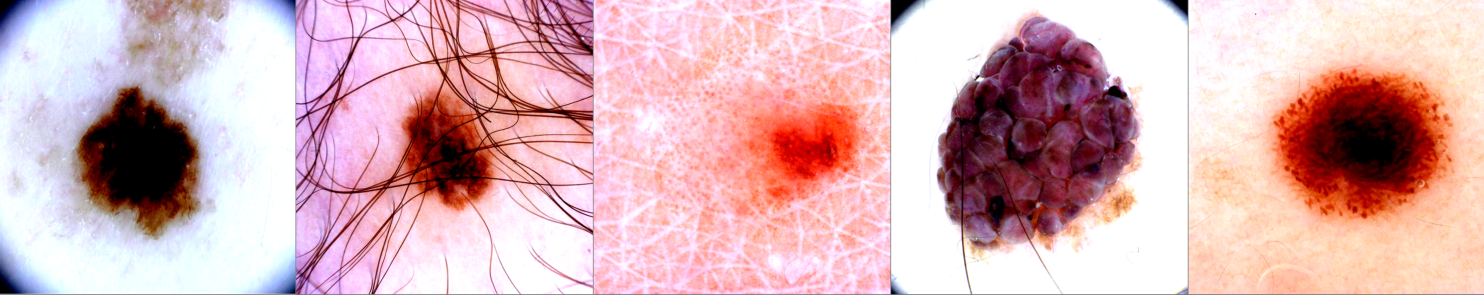}}
    \end{subfigure}
    \qquad

    \begin{subfigure}[Images after random shift, scale and rotate]{
      \includegraphics[width=0.8\linewidth]{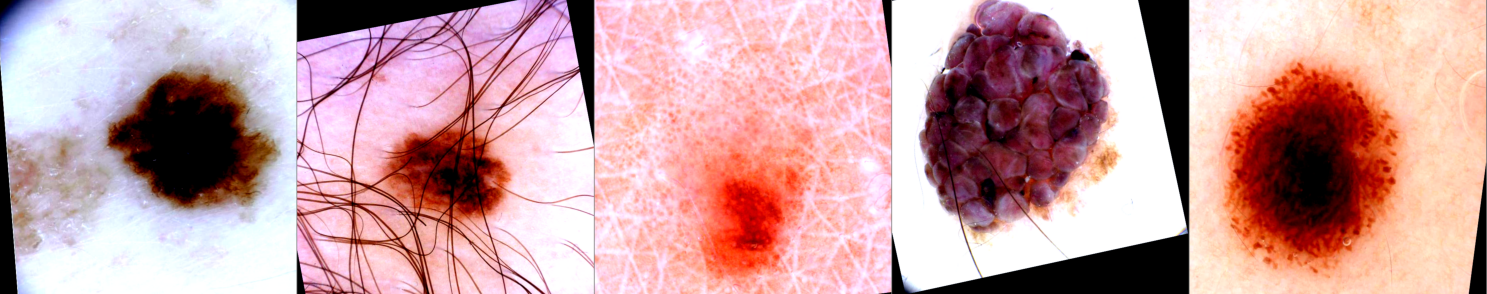}}
     \end{subfigure}

  \caption{Data augmentation with affine transformations.}
  \label{fig.augmentation.affine}
\end{figure*}

Image data augmentation methods can be divided into affine (linear), elastic, advanced, and neural transformations \cite{mikolajczyk2018data}. 
The most popular and proven effective current practice for data augmentation is to use the first group of transformations: affine transformations. They include random rotation, shift, reflection, scaling, and shearing. Example linear transformations are presented in Figure~\ref{fig.augmentation.affine}.

\begin{figure*}[!htb]
\centering


    \begin{subfigure}[Images after hue and brightness modification]{
      \includegraphics[width=0.8\linewidth]{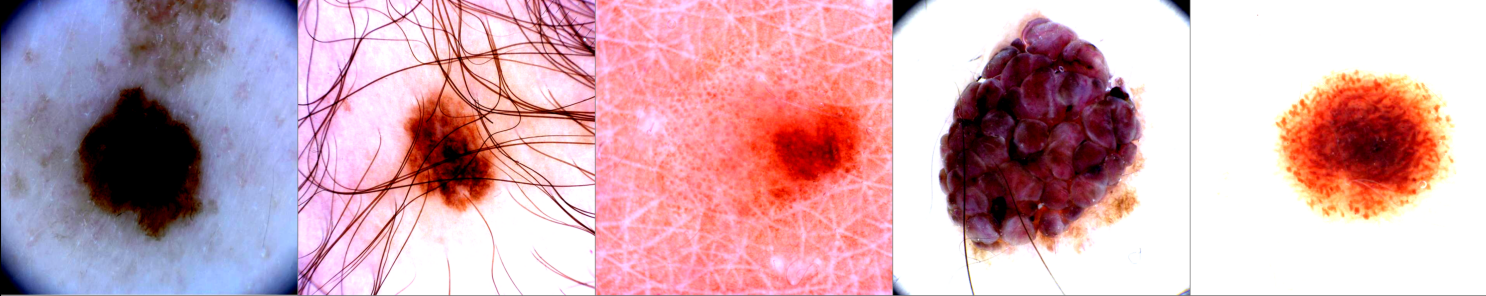}}
     \end{subfigure}
    
    \begin{subfigure}[Images after random hue and saturation modifications]{
      \includegraphics[width=0.8\linewidth]{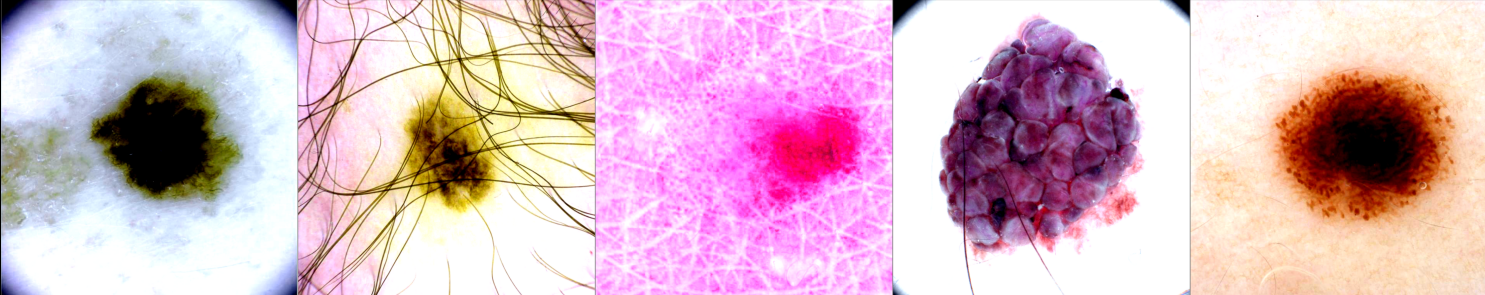}}
     \end{subfigure}
    
  \caption{Data augmentation with elastic transformations} \label{fig.augmentation.elastic}
\end{figure*}

Another common category are \textit{elastic transformations}. They are primarily based on color modification such as histogram equalization, randomly enhancing contrast or brightness, white-balancing, sharpening, and blurring. Blurring can be used to imitate the object or camera movement. Sometimes, traditional transformations are applied only to one channel of the image, or to the part of the image, for example, to create an RGB shift. Example elastic transformations are presented in Figure~\ref{fig.augmentation.elastic}. 

Despite their many advantages, in some cases, simple classical operations are not enough to significantly improve the neural network accuracy or overcome overfitting or domain shift problems \cite{mikolajczyk2018data}. The main reason is that these methods do not enrich the training data with new and task-specific information.

\begin{figure*}[!htb]
\centering


    \begin{subfigure}[Images after random optical and grid distortion]{
      \includegraphics[width=0.8\linewidth]{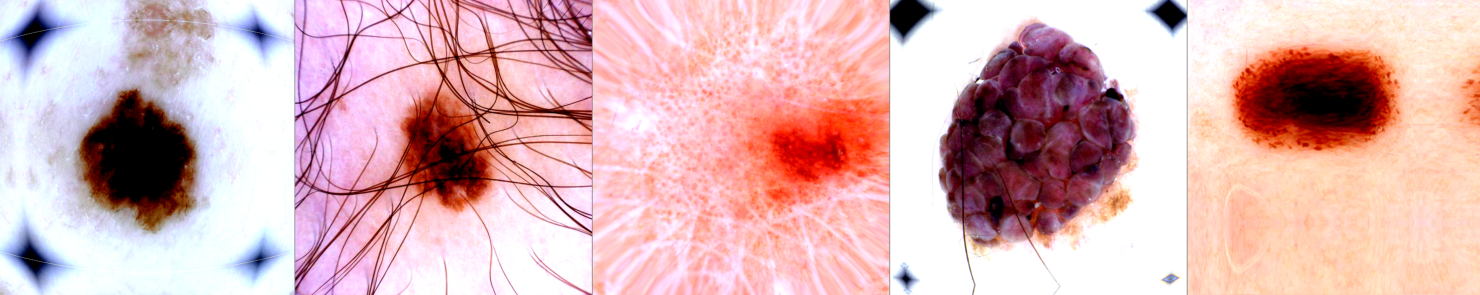}}
     \end{subfigure}
     
    \begin{subfigure}[Images after random cutout]{
      \includegraphics[width=0.8\linewidth]{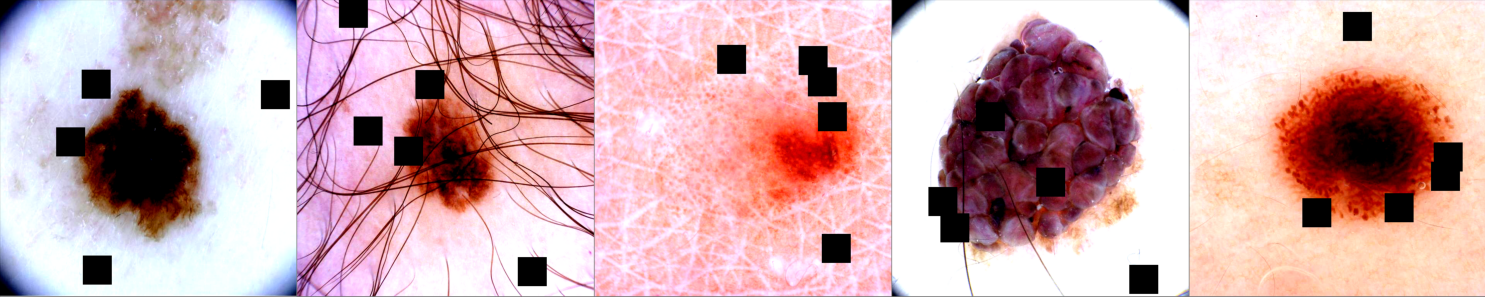}}
     \end{subfigure}

  \caption{Data augmentation with advanced transformations}
  \label{fig.augmnetation.advanced}
\end{figure*}

Exciting but less common approaches are called \textit{advanced transformations}. They include, for instance, optical and grid distortion, the \textit{random erasing technique} (random cutout) \cite{zhong2020random}, which randomly erases selected parts of an image, and \textit{stochastic additive noise augmentation}, which increases the robustness of the model. Next, \textit{domain randomization} \cite{tremblay2018training} and \textit{texture transfer} \cite{jackson2019style} are sometimes applied. In some specific cases, anyone can augment a dataset by randomly applying weather effects like adding fog to the image, rain effects, snow, or even sun flare. The examples are presented in Figure~\ref{fig.augmnetation.advanced}.

Finally, there is a category of \textit{neural-based transformations} that includes adversarial transformations, neural style transfer, and Generative Adversarial Networks (GANs). Those methods are still not as widespread as other methods but are gaining popularity. For instance, GANs can be used for conditional and unconditional image generation that might increase the dataset size. GANs are also used for image text-to-image synthesis, which allows generating images from given text descriptions for image-to-image translation (similar to style transfer, e.g., converting sketches to images) \cite{jabbar2021survey}. There are other exciting approaches, such as super-resolution, where the resolution of an image can be increased by generating relevant details with GANs, or image restoration (image inpainting), which restores missing pieces of an image. Even image blending, based on mixing selected parts of two shots to get a new one, can be performed with GANs \cite{jabbar2021survey}. GANs started gaining attention in medical image processing to solve tasks such as de-noising, segmentation, detection, and classification. Despite the rising popularity of GANs and their numerous applications, they are still hard to train and evaluate. They also seem to have a problem with the \textit{vanishing gradient}, which happens when the discriminator is poorly trained and does not provide appropriate feedback to the generator, or \textit{mode collapse}, which emerges during the training when the generator learns only a tiny distribution from data. The generator can get stuck in the small subspace and produce the same outputs despite changing input. GANs showed excellent results in image generation; however, they must be appropriately trained on vast amounts of data. It is hard to use them as a data augmentation tool, especially if we augment a very small or unbalanced dataset. In one of my very recent works, I also showed that GANs used for data augmentation could even increase biases in models and make them less robust.

\begin{figure}[!htb]
\centering
  \includegraphics[width=\textwidth]{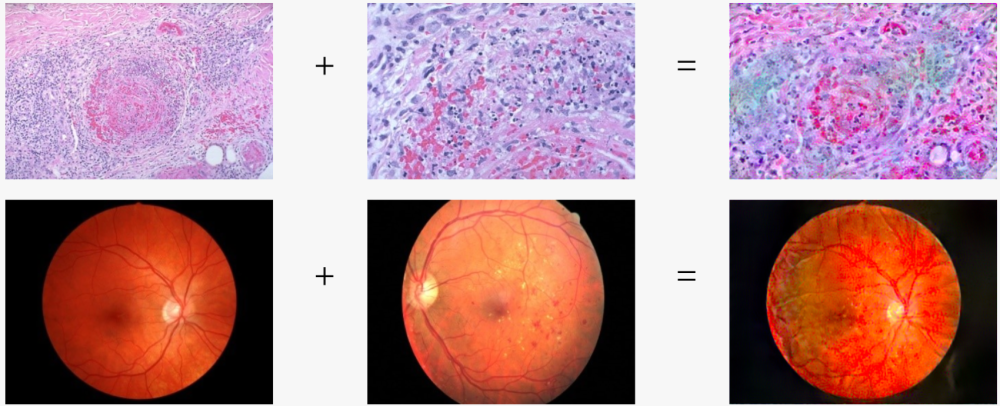}%
  \caption{Image data augmentation – neural transformations. }
  \label{fig.aug3}
\end{figure}

Another neural-based data augmentation method is called neural style transfer \cite{gatys2015neural}. The significant advantage of using style transfer is that even a few samples are enough to achieve high-quality results (Figure~\ref{fig.aug3}).  

Style transfer (ST) can be defined as any method used to mimic and copy the style of one image to another. Artistic stylization is a long-standing research topic developed for over 20 years under various names, such as image analogies \cite{hertzmann2001image}, image stylization \cite{shen2016automatic}, or non-photorealistic rendering \cite{selim2016painting}. Researchers tried to solve the problem by image processing and filtering or by stroke-based rendering, e.g.lacing virtual strokes upon the digital canvas to render a photograph with chosen style. Sometimes they tried region-based techniques such as placing strokes on selected image regions or example-based rendering: learning and recreating the mapping between the exemplar pair. Some of those ST algorithms could recreate given styles. However, they had limited flexibility, style diversity, and effective image structure extractions. Hence, I focused on neural style transfer instead of analyzing traditional style transfer algorithms.  

The first proposition of using neural networks to mimic an artistic style of one image while maintaining the content was published by Gatys et al. in 2015 \cite{gatys2015neural}. Since the first neural style transfer, many modifications to this method emerged, such as:
\begin{itemize}
\item \textit{semantic style transfer,} which is a style transfer performed by applying semantic masks, 
\item \textit{ arbitrary style transfer} that uses encoder-decoder networks without the need for pre-defining styles \cite{huang2017arbitrary}, 
\item \textit{doodle style transfer}, which allows for painting a new image based on the content doodle mask with the help of semantic masks of the style picture, 
\item \textit{stereoscopic style transfer}, which is a style transfer for 3D imaging and AR/VR \cite{chen2018stereoscopic}, 
\item and \textit{video style transfer }\cite{huang2017real}.  
\end{itemize}

As the details behind style transfer methods are not entirely relevant to this thesis, the information about the style transfer methods was moved to the \textit{Appendix \ref{appendix.a}: Neural Style Transfer methods for data augmentation.} Additionally, the Appendix briefly introduced the state-of-the-art NST algorithms, along with the corresponding taxonomy proposed by Jing et al. in \cite{jing2019neural}. Their strengths and weaknesses are discussed in terms of dataset generation.  

\section{Methodology}
It is well-known to the field experts that the malignancy of skin lesions is decided by multiple factors, including the shape of the skin lesion and its texture. The analysis of dermoscopy methods performed in Chapter~\ref{chapter.skin_lesion_bias} showed that a structure plays a much more significant role than in other image classification tasks. In all scenarios, the structure in the image was given higher importance scores than the shape of the lesion. Yet, the texture alone was not enough to consider a skin lesion as malignant in two out of three examined methods. This allows us to assume that \textit{both shape and texture play a vital role in classification}. Changes in the shape do not necessarily mean a suspicious lesion, but shape asymmetry and texture irregularities almost certainly make the skin lesion suspicious. 

\begin{figure}[!htb]
\centering
  \includegraphics[width=0.9\textwidth]{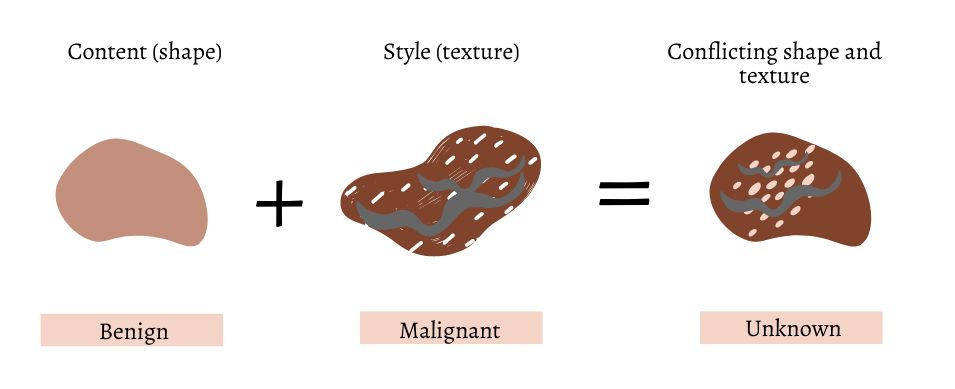}%
  \caption{Synthetic image created from content and style image.}
  \label{fig.style}
\end{figure}

Hence, the general idea of the approach proposed is to use unlabeled images generated with neural-based data augmentation to increase the amount of data by mixing two conflicting shapes and textures: benign shape with malignant texture (Figure~\ref{fig.style}). 

\begin{figure}
\centering
  \includegraphics[width=\textwidth]{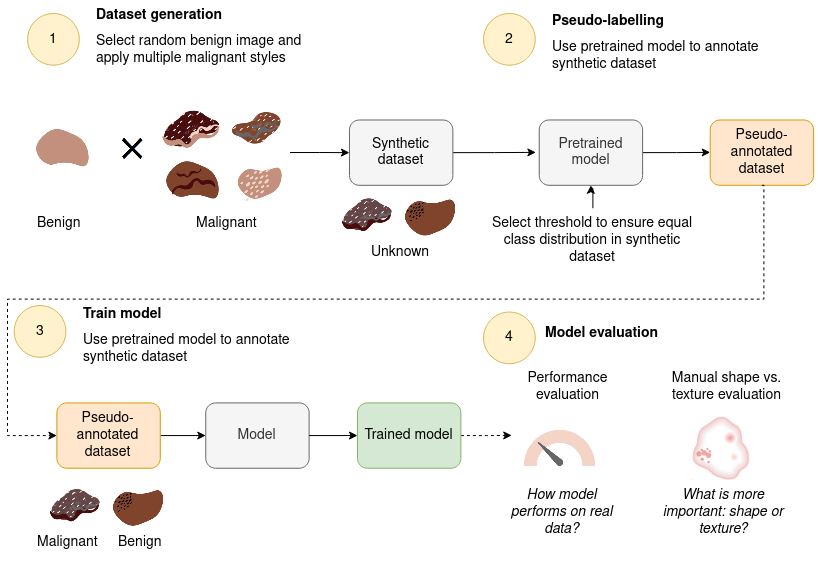}%
  \caption{The pipeline behind style transfer data augmentation.}
  \label{fig.style-pipeline}
\end{figure}

The method can be divided into three steps: data generation, synthetic data labeling (pseudo-labeling), and model training. The steps behind the method are presented in Figure~\ref{fig.style-pipeline}.

\textit{Dataset generation.} The first step is to generate new images by merging two different styles and contents. I applied the neural style transfer method \cite{gatys2015neural}, but any method that allows for style transfer could be used. Most benign skin lesions have similar mild textures. They usually have a solid color, are small,  have regular borders, and are less likely to have different structures. Hence, the synthetic database was generated by randomly selecting a benign skin lesion and repeatedly applying one hundred different malignant styles. 

\begin{figure}
\centering
  \includegraphics[width=0.8\textwidth]{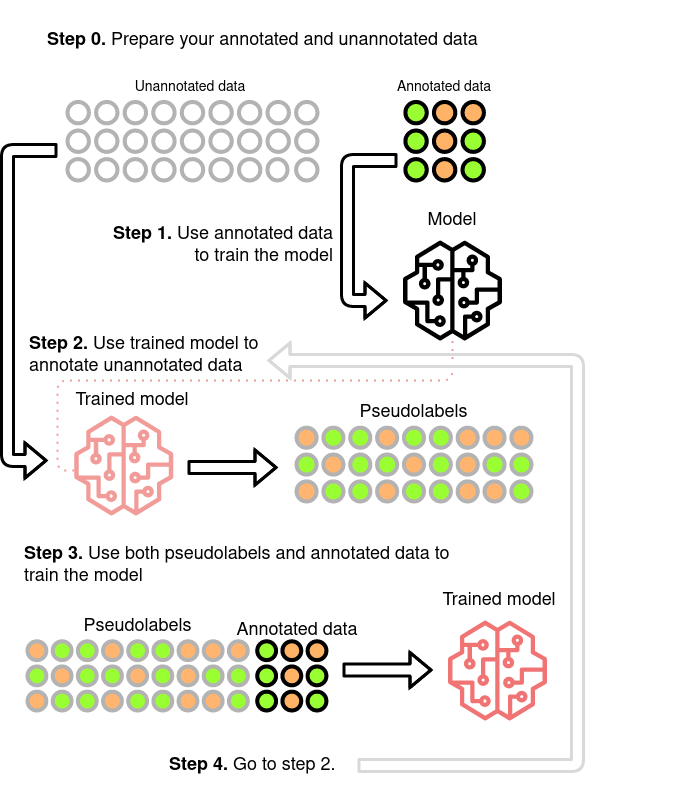}%
  \caption{Inforgraphic on how the pseudolabelling works.}
  \label{fig.pseudo}
\end{figure}

\textit{Synthetic data labeling.} Pseudo-labeling is a technique used in semi-supervised learning that allows using unlabeled and labeled data together to train the model. The unlabeled data is "pseudo"-annotated with predictions for every weight update. Those labels are then used as a ground-truth label during the next training step \cite{amini2015semi}. The process of pseudo-labeling is depicted in Figure~\ref{fig.pseudo}. This approach has been modified here: the classified and labeled images were generated with another pretrained model before training on a target dataset. However, I selected a different classification threshold than during model training, so the number of synthesized images in all classes remains equal. Contrary to classic pseudo-labeling, those labels did not change during the training. All generated images were included in the training set and all real images in the validation and test set. I paid particular attention to avoiding data leakage. I created a well-balanced dataset - the number of samples of each class in the validation and test sets was down-sampled to become equal. 

\textit{Model's training.} The final step is training with augmented data. Apart from the preliminary step (pseudo-labeling), there is not much different than in the standard training. The noisy pseudo-labels are treated as a ground truth during the training, and the model is evaluated on the validation set (only real images) every epoch.

\section{Experiments}

\subsection{Data and training details}
The original dataset consisted of 1088 malignant skin lesions and 12433 benign images. In particular, only 418 malignant lesions were used from the original dataset to create the database. Taking full advantage of the neural style transfer approach, 248 489 synthesized images were generated. The images were captured after every five iterations of the neural style transfer algorithm and lasted 30 iterations for each pair of images. I trained the networks on 54 030 synthesized images using only each pair's final, 30th iteration. It was validated on 1976 real images and tested on 200 images per five testing folds. Then, the performance was evaluated on four popular and commonly used deep neural network architectures (at a time).

I applied regularization techniques to each tested network: traditional data augmentation (rotation, zoom, shear, reflection), dropout, and early stopping. The experiments were performed with the same hyperparameters, regularization techniques, and architectures to make the comparison reliable. The proposed approach was compared with other networks regarding performance, repeatability, and the amount of new information carried in the generated dataset.


\subsection{Performance evaluation}

To assess the performance of the proposed approach, I considered the problem of skin lesions classification again. The experiments were conducted using four commonly used the experiments time architectures: VGG8, VGG11, VGG16, and DenseNet121. To evaluate the networks trained with the use of augmented data, I used standard statistical measures, such as the area under the ROC curve (AUC), accuracy (ACC), recall for each class ($R^{ben}$ – sensitivity, and $R^{mal}$ - specificity), and $F_1$ score.  

Five validation subsets were prepared to provide more reliable results for the experiments, and 5-fold cross-validation was applied. Particular attention to avoiding data leakage was paid, i.e., the test set did not contain the images used to generate the augmented training database in the validation and test sets \cite{shabtai2012survey}. Table~\ref{tab:stda-results} presents the classification results for each fold.

\begin{table}[!htb]
\caption{A comparison for standard training and a trainin with Style Transfer Data Augmentation method.}
\label{tab:stda-results}
\resizebox{\textwidth}{!}{%
\begin{tabular}{llllllllllllll}
\hline
\textbf{Model} & \textbf{Standard} &  &  &  &  &  &  & \textbf{ST-DA} &  &  &  &  &  \\
 & \multicolumn{1}{c}{\textbf{fold}} & \multicolumn{1}{c}{\textbf{0}} & \multicolumn{1}{c}{\textbf{1}} & \multicolumn{1}{c}{\textbf{2}} & \multicolumn{1}{c}{\textbf{3}} & \multicolumn{1}{c}{\textbf{4}} & \multicolumn{1}{c}{\textbf{Mean}} & \textbf{0} & \textbf{1} & \textbf{2} & \textbf{3} & \textbf{4} & \multicolumn{1}{c}{\textbf{Mean}} \\  \hline
\textbf{VGG8} & \textbf{ACC} & 0.840 & 0.820 & 0.850 & 0.880 & 0.850 & \textbf{0.848} & 0.840 & 0.880 & 0.850 & 0.880 & 0.850 & \textbf{0.860} \\
 & \textbf{AUC} & 0.805 & 0.820 & 0.845 & 0.845 & 0.805 & \textbf{0.824} & 0.835 & 0.880 & 0.845 & 0.875 & 0.850 & \textbf{0.857} \\
 & \textbf{$F_1^{ben}$} & 0.730 & 0.800 & 0.810 & 0.760 & 0.720 & \textbf{0.764} & 0.850 & 0.870 & 0.830 & 0.850 & 0.860 & \textbf{0.852} \\
 & \textbf{$F_1^{mal}$} & 0.960 & 0.840 & 0.890 & 1.000 & 0.980 & \textbf{0.934} & 0.830 & 0.890 & 0.860 & 0.900 & 0.840 & \textbf{0.864} \\
 & \textbf{$R^{ben}$} & 0.970 & 0.850 & 0.900 & 1.000 & 0.810 & \textbf{0.906} & 0.820 & 0.890 & 0.860 & 0.910 & 0.850 & \textbf{0.866} \\
 & \textbf{$R^{mal}$} & 0.640 & 0.790 & 0.790 & 0.690 & 0.840 & \textbf{0.750} & 0.850 & 0.870 & 0.830 & 0.830 & 0.850 & \textbf{0.846} \\ \hline
\textbf{VGG11} & \textbf{ACC} & 0.840 & 0.850 & 0.850 & 0.860 & 0.870 & \textbf{0.854} & 0.820 & 0.900 & 0.870 & 0.890 & 0.840 & \textbf{0.864} \\
 & \textbf{AUC} & 0.820 & 0.800 & 0.805 & 0.840 & 0.845 & \textbf{0.822} & 0.815 & 0.895 & 0.870 & 0.885 & 0.840 & \textbf{0.861} \\
 & \textbf{$F_1^{ben}$} & 0.760 & 0.720 & 0.730 & 0.770 & 0.770 & \textbf{0.750} & 0.810 & 0.890 & 0.860 & 0.870 & 0.850 & \textbf{0.856} \\
 & \textbf{$F_1^{mal}$} & 0.910 & 0.980 & 0.970 & 0.950 & 0.970 & \textbf{0.956} & 0.820 & 0.900 & 0.890 & 0.910 & 0.830 & \textbf{0.870} \\
 & \textbf{$R^{ben}$} & 0.930 & 0.990 & 0.980 & 0.960 & 0.980 & \textbf{0.968} & 0.830 & 0.900 & 0.890 & 0.910 & 0.820 & \textbf{0.870} \\
 & \textbf{$R^{mal}$} & 0.710 & 0.610 & 0.630 & 0.720 & 0.710 & \textbf{0.676} & 0.800 & 0.890 & 0.850 & 0.860 & 0.860 & \textbf{0.852} \\ \hline
\textbf{VGG16} & \textbf{ACC} & 0.830 & 0.850 & 0.840 & 0.830 & 0.820 & \textbf{0.834} & 0.830 & 0.850 & 0.810 & 0.880 & 0.840 & \textbf{0.842} \\
 & \textbf{AUC} & 0.825 & 0.850 & 0.835 & 0.830 & 0.815 & \textbf{0.831} & 0.830 & 0.850 & 0.815 & 0.875 & 0.840 & \textbf{0.842} \\
 & \textbf{$F_1^{ben}$} & 0.830 & 0.880 & 0.860 & 0.830 & 0.800 & \textbf{0.840} & 0.770 & 0.770 & 0.740 & 0.810 & 0.770 & \textbf{0.772} \\
 & \textbf{$F_1^{mal}$} & 0.820 & 0.820 & 0.810 & 0.830 & 0.830 & \textbf{0.822} & 0.930 & 0.990 & 0.940 & 0.970 & 0.950 & \textbf{0.956} \\
 & \textbf{$R^{ben}$} & 0.820 & 0.810 & 0.800 & 0.830 & 0.840 & \textbf{0.820} & 0.950 & 0.990 & 0.960 & 0.980 & 0.960 & \textbf{0.968} \\
 & \textbf{$R^{mal}$} & 0.830 & 0.890 & 0.870 & 0.830 & 0.790 & \textbf{0.842} & 0.710 & 0.710 & 0.670 & 0.770 & 0.720 & \textbf{0.716} \\ \hline
\textbf{\begin{tabular}[c]{@{}l@{}}VGG16* \end{tabular}} & \textbf{ACC} & 0.910 & 0.910 & 0.880 & 0.890 & 0.870 & \textbf{0.892} & 0.920 & 0.900 & 0.890 & 0.900 & 0.900 & \textbf{0.902} \\
 & \textbf{AUC} & 0.900 & 0.890 & 0.875 & 0.870 & 0.855 & \textbf{0.878} & 0.910 & 0.900 & 0.890 & 0.895 & 0.900 & \textbf{0.899} \\
 & \textbf{$F_1^{ben}$} & 0.850 & 0.820 & 0.840 & 0.800 & 0.800 & \textbf{0.822} & 0.970 & 0.900 & 0.910 & 0.890 & 0.930 & \textbf{0.920} \\
 & \textbf{$F_1^{mal}$} & 0.970 & 1.000 & 0.910 & 0.970 & 0.930 & \textbf{0.956} & 0.870 & 0.900 & 0.870 & 0.900 & 0.880 & \textbf{0.884} \\
 & \textbf{$R^{ben}$} & 0.970 & 1.000 & 0.920 & 0.980 & 0.940 & \textbf{0.962} & 0.850 & 0.900 & 0.860 & 0.900 & 0.870 & \textbf{0.876} \\
 & \textbf{$R^{mal}$} & 0.830 & 0.780 & 0.830 & 0.760 & 0.770 & \textbf{0.794} & 0.970 & 0.900 & 0.920 & 0.890 & 0.930 & \textbf{0.922} \\ \hline
\textbf{DenseNet} & \textbf{ACC} & 0.820 & 0.820 & 0.810 & 0.850 & 0.810 & \textbf{0.822} & 0.870 & 0.900 & 0.870 & 0.850 & 0.860 & \textbf{0.870} \\
 & \textbf{AUC} & 0.820 & 0.825 & 0.810 & 0.850 & 0.810 & \textbf{0.823} & 0.870 & 0.895 & 0.845 & 0.855 & 0.860 & \textbf{0.865} \\
 & \textbf{$F_1^{ben}$} & 0.750 & 0.740 & 0.730 & 0.780 & 0.730 & \textbf{0.746} & 0.890 & 0.870 & 0.770 & 0.840 & 0.880 & \textbf{0.850} \\
 & \textbf{$F_1^{mal}$} & 0.960 & 1.000 & 0.960 & 0.970 & 0.970 & \textbf{0.972} & 0.860 & 0.920 & 0.970 & 0.870 & 0.840 & \textbf{0.892} \\
 & \textbf{$R^{ben}$} & 0.970 & 1.000 & 0.970 & 0.980 & 0.980 & \textbf{0.980} & 0.850 & 0.930 & 0.980 & 0.870 & 0.830 & \textbf{0.892} \\
 & \textbf{$R^{mal}$} & 0.670 & 0.650 & 0.650 & 0.720 & 0.640 & \textbf{0.666} & 0.890 & 0.860 & 0.710 & 0.840 & 0.890 & \textbf{0.838} \\ \hline
\textbf{\begin{tabular}[c]{@{}l@{}}DenseNet*\end{tabular}} & \textbf{ACC} & 0.910 & 0.930 & 0.860 & 0.870 & 0.880 & \textbf{0.880} & 0.910 & 0.880 & 0.870 & 0.930 & 0.890 & \textbf{0.896} \\
 & \textbf{AUC} & 0.900 & 0.930 & 0.860 & 0.870 & 0.880 & \textbf{0.888} & 0.900 & 0.900 & 0.845 & 0.925 & 0.875 & \textbf{0.889} \\
 & \textbf{$F_1^{ben}$} & 0.950 & 0.925 & 0.830 & 0.840 & 0.870 & \textbf{0.883} & 0.860 & 0.875 & 0.770 & 0.870 & 0.810 & \textbf{0.837} \\
 & \textbf{$F_1^{mal}$} & 0.860 & 0.890 & 0.900 & 0.910 & 0.890 & \textbf{0.890} & 0.950 & 0.800 & 0.970 & 1.000 & 0.970 & \textbf{0.938} \\
 & \textbf{$R^{ben}$} & 0.840 & 0.970 & 0.910 & 0.920 & 0.890 & \textbf{0.906} & 0.960 & 0.990 & 0.980 & 1.000 & 0.980 & \textbf{0.982} \\
 & \textbf{$R^{mal}$} & 0.960 & 0.880 & 0.810 & 0.820 & 0.870 & \textbf{0.886} & 0.840 & 0.760 & 0.710 & 0.850 & 0.770 & \textbf{0.832} \\
\end{tabular}%
}
* with Transfer Learning
\end{table}

For all models, the average AUC was higher by 2.45 percentage points. When training with ST-DA, smaller models showed better improvement. On average, AUC increased by 3.30 percentage points for VGG8 and even more (by 3.90 pp) for VGG11. The most profound tested architecture, VGG16, trained with transfer learning, improved AUC by 2.10 pp on average. Pretrained VGG16 improved by 1.1 pp. DenseNet121 without transfer learning showed an AUC improvement of 4.1 pp, the greatest among all tested architectures. However, the results were almost identical to the standard approach when using pretrained models (increase by 0.1 pp). Interestingly, in contrast to classical approaches, the sensitivity and specificity were quite balanced. In ST-DA, the difference between sensitivity and specificity was a few times smaller than the training results without ST-DA. 
The performed experiments have shown that the network trained with synthesized images provides higher accuracy, repeatability, and reduced overfitting. Moreover, it is easy to implement. For instance, the image synthesis can be conducted with other methods, such as GANs, or even entirely replaced by introducing additional real but unlabeled data.

\subsection{Manual inspection of shape and texture}

One of the goals of the proposed data augmentation method is to generate new images and add new information to the classifier to facilitate training. As the experiments show, the new data successfully helped improve the training. However, I performed an additional manual investigation due to the new research that shed light on shape and texture bias problems. 

A typical shape and texture bias investigation is based on mixing those conflicting properties and checking the resulting prediction. However, in the case of skin lesion classification, it is hard to tell whether the change in prediction should occur. As mentioned earlier, both texture and shapes are crucial for the diagnosis. The shape or texture alone is not enough for classification, as they have to be examined altogether. Hence, instead of measuring predictions (which are already proven to be better), I manually analyzed which features were crucial for the prediction.

\begin{figure}[!htb]
\centering
  \includegraphics[width=\textwidth]{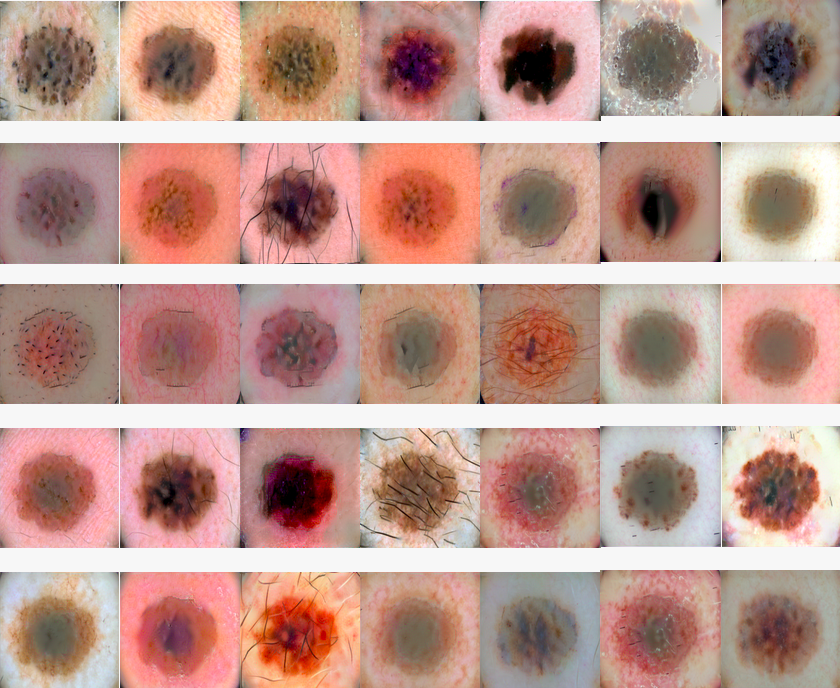}%
  \caption{Artificial images of skin lesions generated from the same content image and different styles.}
  \label{fig.nst}
\end{figure}

For this purpose, as in the typical approach, several images with contrasting shapes and textures were mixed. In style transfer, the style is usually connected to the image texture. On the other hand, the content is usually connected to the shapes. Hence, during my experiment, I generated images with the same content but different styles and styles but different contents. Generated images showed variations in regularity of border, along with new visible structures and new colors, but also copied some artifacts (such as dense hair, water bubbles, or markings) or sometimes changed the white balance of the image. Figure~\ref{fig.nst} presents examples of generated images.  

\begin{figure}
    \centering
    \includegraphics[width=0.8\textwidth]{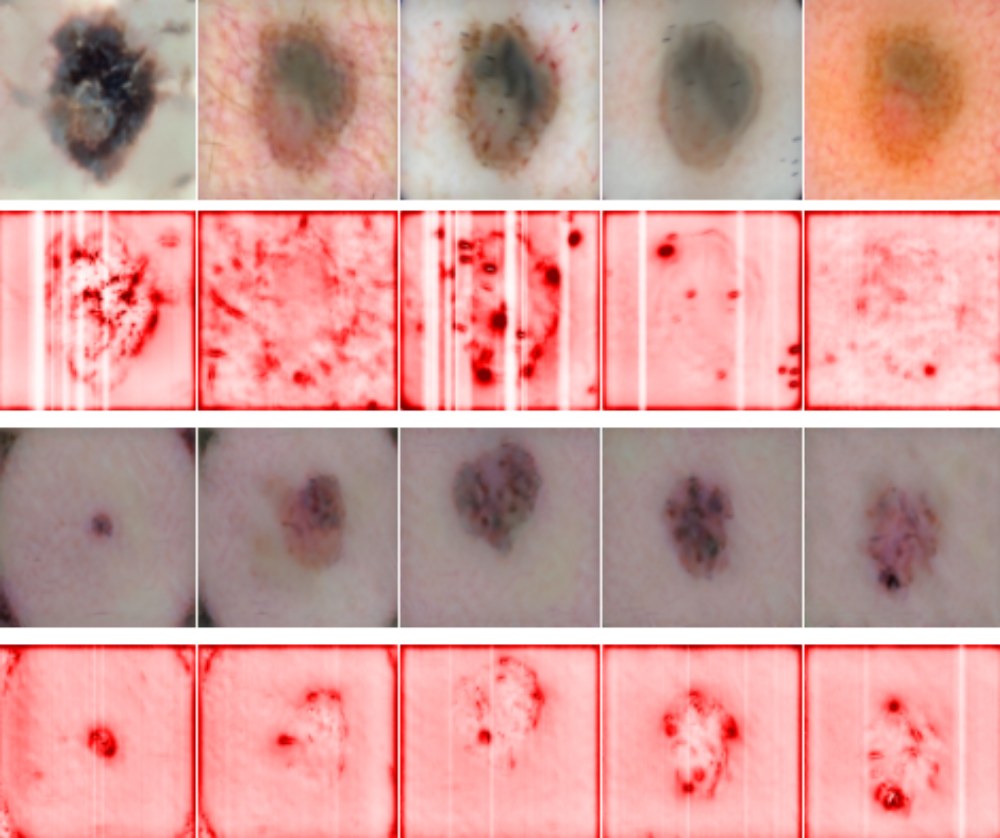}
    \caption{Ten skin lesion examples with accompanying heatmaps generated with Deep Taylor Decomposition. }
    \label{fig:stda-attribution}
\end{figure}

To compare shape vs. texture, I generated the images with the same content but different styles (the first row in Figure~\ref{fig:stda-attribution}). To retrieve the information on which parts of an image had the most significant impact on the prediction, I generated relevance maps with a local explainability method called Deep Taylor Decomposition \cite{montavonDTD}. The heatmaps were very diversified; each map focused on a different image part. The next five heatmaps were generated with the same styles (texture) but different contents (shapes). Some similarities between the heatmaps can be observed; even though the heatmaps show different regions, the highlighted features seem similar. In the case of the same texture but different shapes, some similarities between the relevance maps can be observed: even though the relevance highlight different regions, the highlighted features are similar. For instance, it is quite easy to notice that some globules and other textural patterns are strongly emphasized. However, we can also see that the border of the skin lesion is also highlighted, which suggests that the predictor might analyze the skin lesion's shape. The heatmaps of rotated, reflected, scaled, and deformed images share many similarities and often look almost the same (Figure~\ref{fig:stda-attribution}).

\section{Characteristics of data augmentation with style transfer }
Using the taxonomy and categorization proposed by Kukvcka et al. \cite{kukavcka2017regularization} I discuss the universality of style transfer data augmentation, its effect on data representation, the stochasticity of transformation parameters, the structure of the transformation space, the dependence of distribution, and the phase of ST-DA application.  

\textit{Universality.} The style transfer-data augmentation does not apply to all data domains, so it should be defined as a domain-specific transformation. Expressly, ST-DA can be limited when the classifier must recognize specific structures, colors, and patterns instead of shapes (detect parts of objects). For example, structure plays a vital role in mammography images. The benign and malignant areas in the breast look differently, while the shape in scene classification, i.e., detecting parts of cars and buildings, classifies the scene as a city. The method requires expert knowledge to be successfully applied. 

\textit{{Effect on data representation.}} ST-DA alters the distribution of low-level visual features in the image while preserving the semantic content. That means that the method falls into the category of representation-modifying transformations. In this category, contrary to representation-preserving algorithms, the distribution created by the transformation does not attempt to mimic the ground truth data distribution. The goal of data augmentation based on representation-modifying transformations is to map the data to a different distribution (or even a new feature space) to change the original representation and enrich the dataset with further information (by adding new augmented samples). The idea is that adding new samples with modified internal representation should make learning easier. 

\textit{Stochasticity of transformation parameters.} 
The stochasticity of transformation parameters is responsible for generating new samples \cite{kukavcka2017regularization}. ST-DA allows generating larger, possibly infinite datasets because the transformation parameters are stochastic. Hence, for example, we can randomly select which image will be the style and content image and choose hyperparameters in a specific range. 

\textit{The transformation space.} 
The transformation space can be divided into the input, hidden feature, and target spaces. In ST-DA, only the input (data augmentation) is changed.  

\textit{Dependence of distribution.} 
ST-DA disentangles the representations of two selected images and then mixes them to create a new image. That means that the data distribution of transformation parameters depends on images that were the content and the style of a newly generated image. The generated image can have the same class as the content, style, or even none of the above depending on the dataset. 

\textit{Phase.} 
Regularization can be applied before, during, or after network training \cite{kukavcka2017regularization} – during the phase of network usage. Unlike input normalization, for instance, ST-DA is used only during the neural network training and does not have to be used after the training.

\section{Discussion}
I proposed using Neural Style Transfer to augment the dataset, increase robustness, and fight shape and texture bias. In the past, style transfer has been mainly used to synthesize artistic style to create new visually pleasing artworks. I showed that the proposed method of data augmentation based on the neural style transfer could be effectively used as a new regularization technique.

The method was tested on the skin lesion dataset by producing almost 250 000 images. The results showed that style transfer-based augmentation is a promising way to enrich the dataset. The performed tests have revealed that the method has excellent perspectives for the future, as a higher AUC score was achieved, along with higher repeatability. The AUC rose on average by 2.45 pp and was accompanied by a better balance between sensitivity and specificity (on average, the difference between them was ten times smaller). Moreover, thanks to ST-DA, the standard deviation inside the folds was twice as small, which shows that higher repeatability of the training process has been achieved. 

ST-DA is an efficient regularization method that can be applied with other techniques. This method is especially valuable when there are no pretrained models available. Moreover, it can help with the texture and shape bias, a common problem among CNN-based architectures.

However, using the style transfer to generate new images is domain-specific. The usage of this method may be limited to images where the texture plays a significant role, such as skin lesions, breast mammography, and histopathology imaging. Its effectiveness in shape-dominated tasks, e.g., detecting the general shape of the cat, including the contours of the cat's whiskers, ears, and eyes, requires further testing.  

\clearpage  
\lhead{\emph{Chapter 7: Bias-targeted data augmentation}}  
\chapter{Bias-targeted data augmentation} \label{chapter.targeted}

\section{Introduction}


In the first part of the thesis I mentioned that bias is prevalent in all types of data and models. Certain kinds of biases in data strongly affect the models leading to lower performance and mistaking correlation with causation, e.g., black frames in skin lesion dataset (described in Chapter \ref{chapter.skin_lesion_bias} and Chapter \ref{chapter.gebi}). Biased models are dangerous to use in production, especially in decision making. The question is: \textit{How to mitigate biases in models?} 
In the previous \textit{Chapter \ref{chapter.neural-style}: Mitigating shape and texture bias with data augmentation}  I proposed to use neural style transfer data augmentation to mitigate a shape/texture bias prevalent in convolutional models. This helps only partially, as the shapes and texture bias is an algorithmic bias that has its source in the model's architecture. But what about bias that is already in the data? 

I imagine two main ways to deal with the problem: (1) removing biases in the prepossessing or (2) forcing the model to ignore biases. Removing those biases is a difficult task. Even very advanced methods of image inpainting still leave new artifacts behind them. Here, I focused on forcing the model to ignore the biases. If we randomly add biases to input during the training, the model will start ignoring it, as such a feature will seem irrelevant. This approach will break the cycle of mistaking correlation with causation by destroying the spurious correlations.

In this chapter of the thesis I showed that using data augmentation might be a valuable tool that helps mitigate the most common biases in data:
\begin{itemize}
\item \textit{observer bias} \cite{mahtani2018catalogue} which might appear when annotators use personal opinion to label data; 
\item  \textit{sampling bias} when data is acquired in such a way that not all samples have the same sampling probability \cite{mehrabi2019survey};
\item  \textit{data handling bias} when how data is handled distort the classifier's output; 
\item  \textit{instrument bias} meaning imperfections in the instrument or method used to collect the data \cite{he2012bias}.
\end{itemize}

The main contribution is a proposition on coupling augmentation with explainable artificial intelligence, resulting in a synergistic effect in reducing flows in machine learning algorithms.

\section{Methodology}

The proposed \textit{Targeted Data Augmentation} experiments (TDA) are divided into: (1) \textit{bias identification}, (2) \textit{augmentation policy design}, (3) \textit{training with targeted data augmentation}, and  (4) \textit{model evaluation}. The pipeline is presented in Figure \ref{fig.bias-targeted-aug-pipeline}.

\hspace{\linewidth}

\begin{minipage}{\linewidth}
    \centering
  \includegraphics[width=0.9\textwidth]{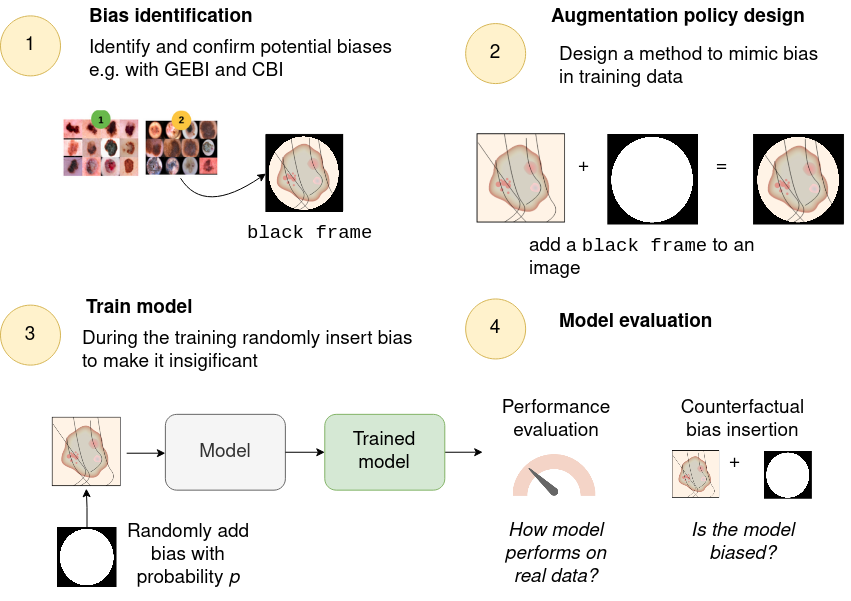}%
  \captionof{figure}{Targeted data augmentation steps.}
  \label{fig.bias-targeted-aug-pipeline}
\end{minipage}

\clearpage
A bias identification is a preliminary, supervised step aiming to detect unwanted data biases. To detect bias, I used the global explanations for bias identification, and measure it with the counterfactual bias insertion (Chapter \ref{chapter.gebi}). Identified biases can be used with targeted data augmentation methods. Augmentation policy design is a step necessary to perform augmentation: it should explain how the data should be augmented (how the bias should be inserted). 
Next, there is a training with targeted data augmentations. During the training, selected biases were randomly added to the input, according to the designed augmentation policy (i.e. adding a black frame to an image). This process made the bias presence more random, and as a result, less correlated with certain classes. Finally, I compared the performance and counterfactual bias insertion measures for model trained with different  bias augmentation probabilities. 

\subsection{Bias identification}

The key to successful bias-targeted (or artifact-targeted) data augmentation is the well-defined source of possible bias in data. This can be done in various ways: 
\begin{itemize}
\item by identifying bias by manual inspection of explanations,
\item by identifying bias with the help of global explanation methods (e.g., GEBI \cite{mikolajczyk2021towards}),
bias is identified with GEBI or another method of choice, e.g., SpRAy, manual inspection with local explanations, manual inspection, 
\item by utilizing field expert's knowledge.
\end{itemize}

Once the bias is identified, a user might move on the process of mitigation. 
Potential biases are selected with both GEBI (Chapter~\ref{chapter.gebi}) and statistical artifact analysis (Chapter~\ref{chapter.skin_lesion_bias}).

\subsection{Augmentation policy}
\begin{figure}[!htb]
\centering
\textbf{Original images}

  \includegraphics[width=0.8\textwidth]{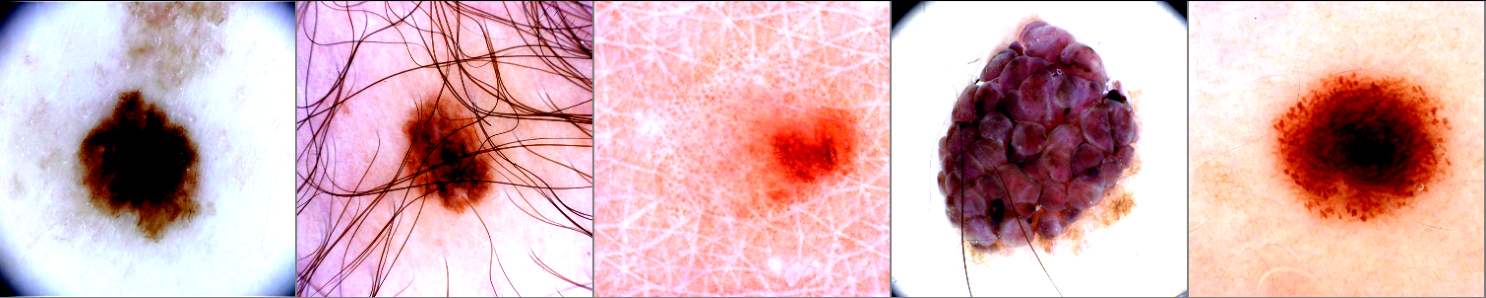}%
  
  \textbf{Augmented images with medium density hair}
  
  \includegraphics[width=0.8\textwidth]{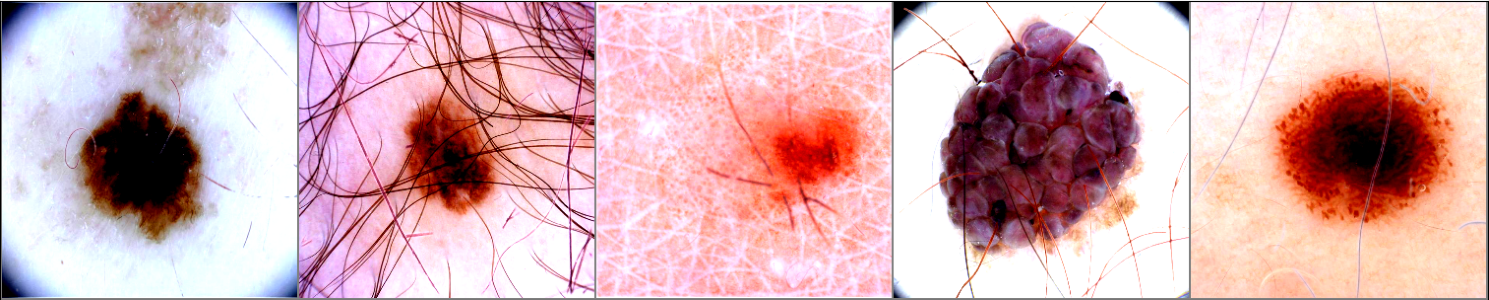}%
  
    \textbf{Medium hair masks}
  
  \includegraphics[width=0.16\textwidth]{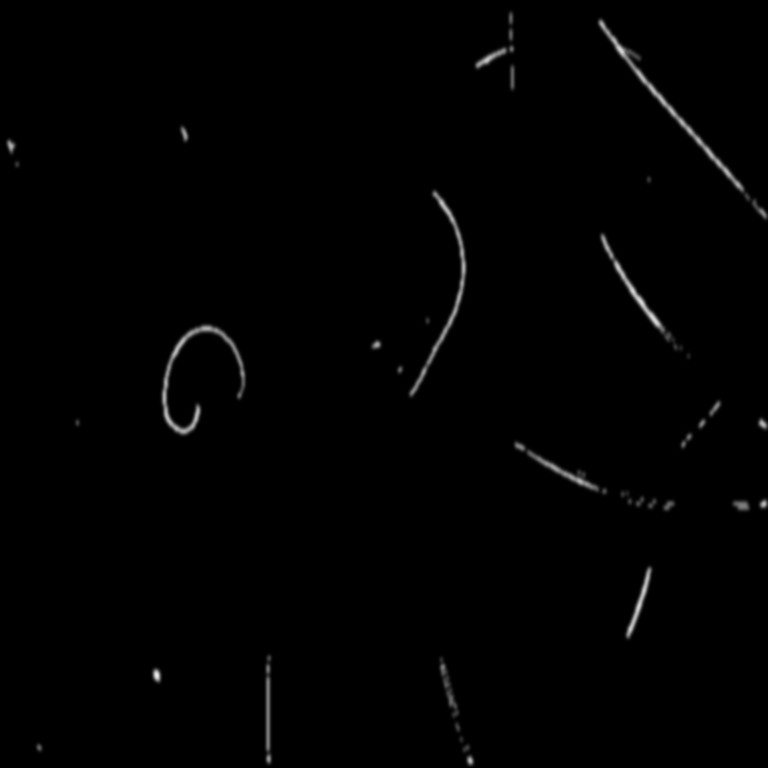}%
  \includegraphics[width=0.16\textwidth]{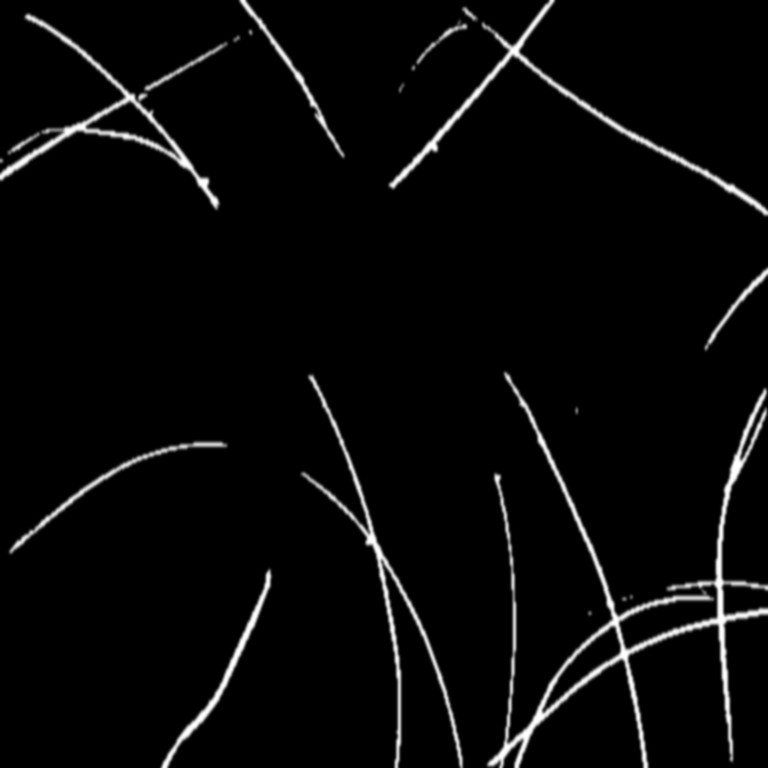}%
    \includegraphics[width=0.16\textwidth]{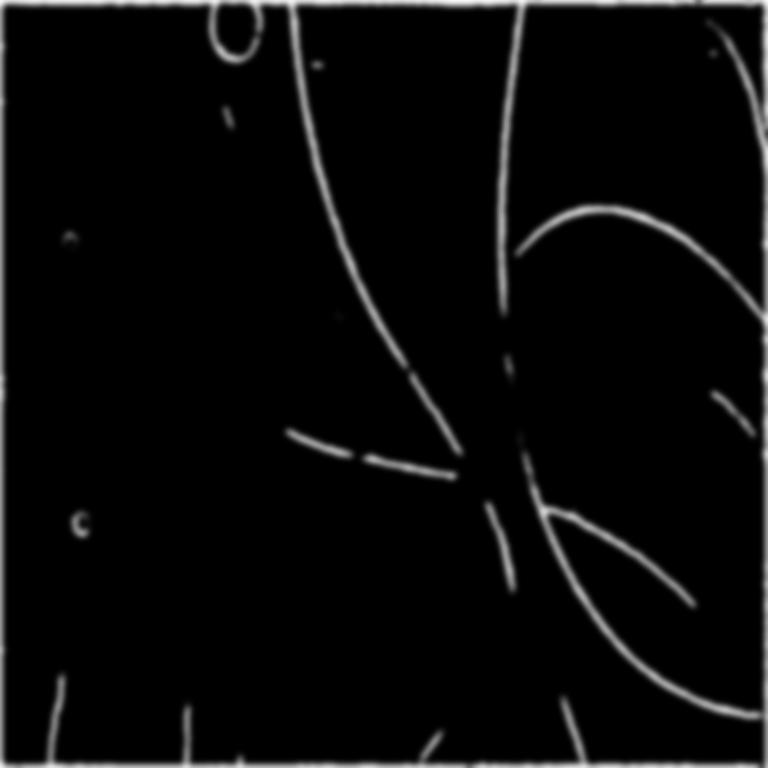}%
  \includegraphics[width=0.16\textwidth]{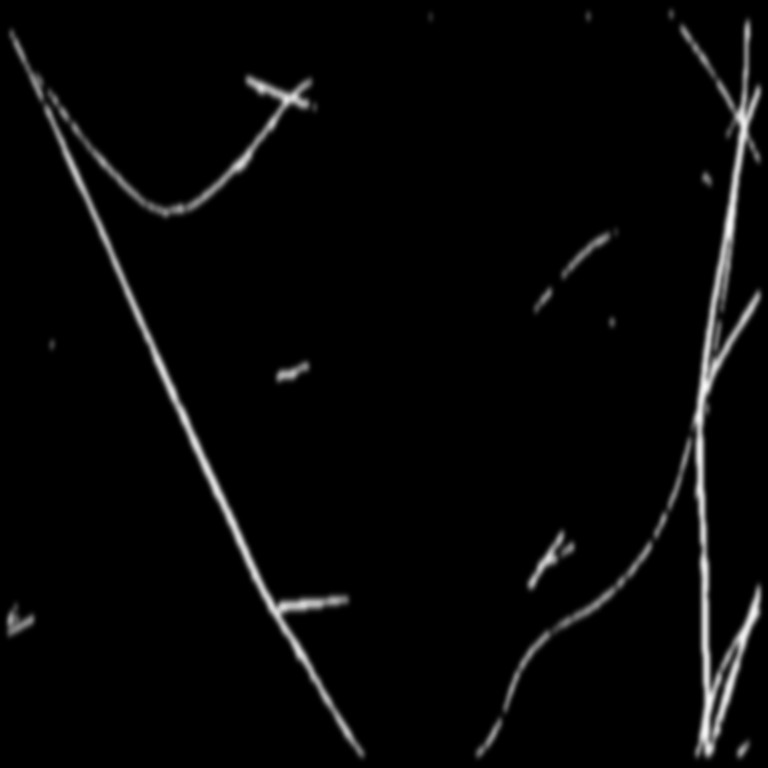}%
  \includegraphics[width=0.16\textwidth]{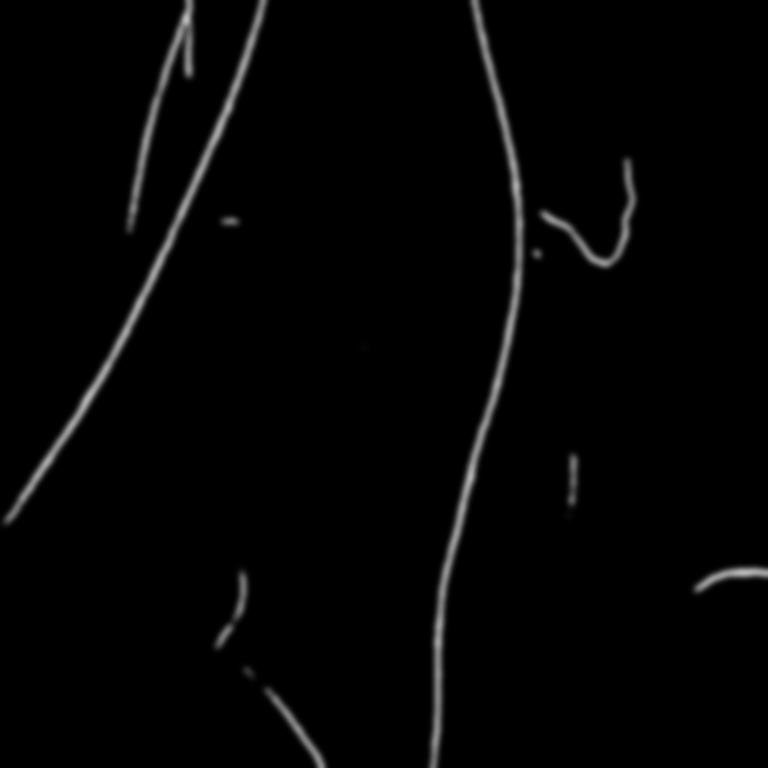}%
  
  \textbf{Medium hair source images}
  
  \includegraphics[width=0.16\textwidth]{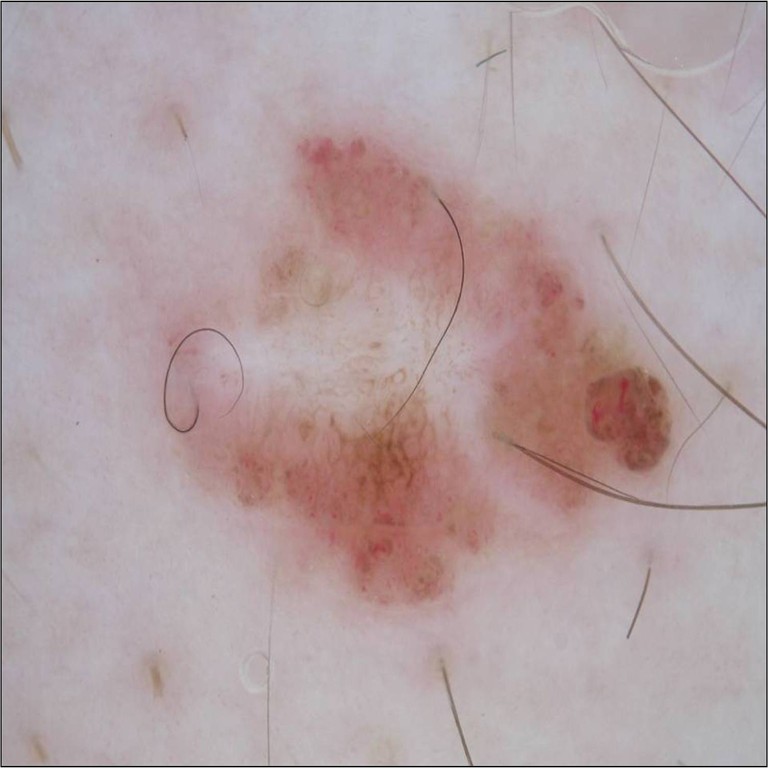}%
  \includegraphics[width=0.16\textwidth]{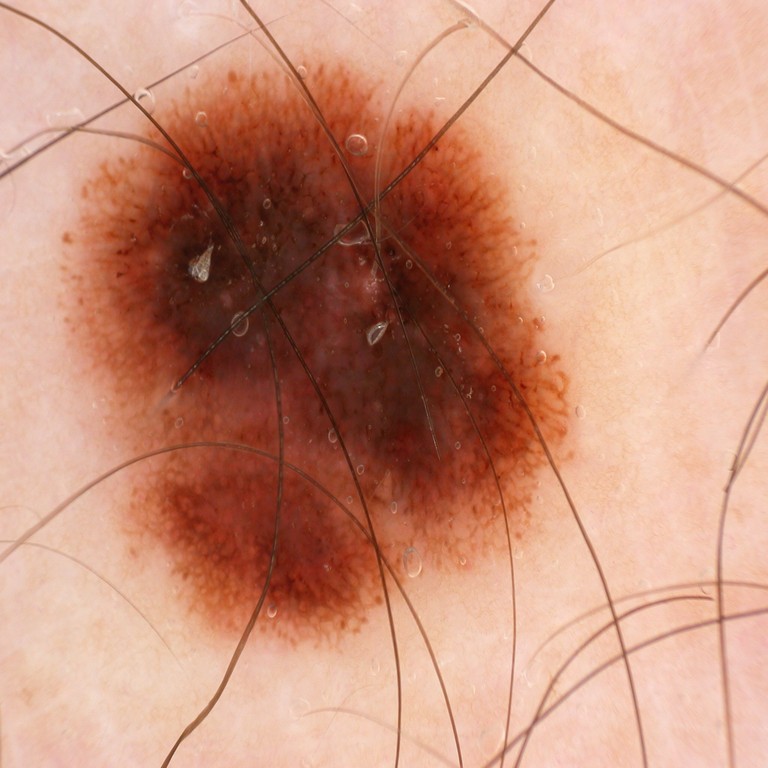}%
    \includegraphics[width=0.16\textwidth]{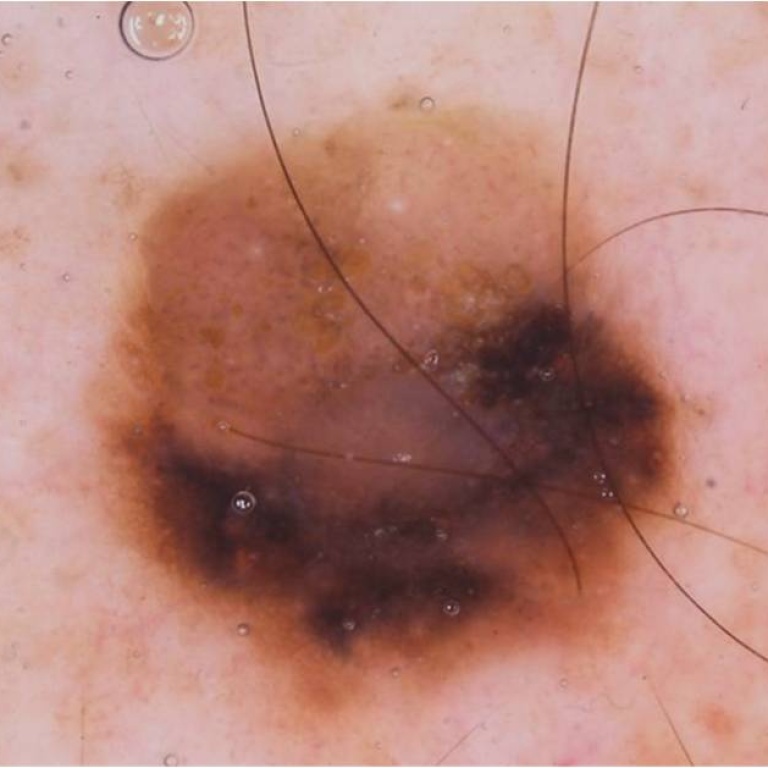}%
  \includegraphics[width=0.16\textwidth]{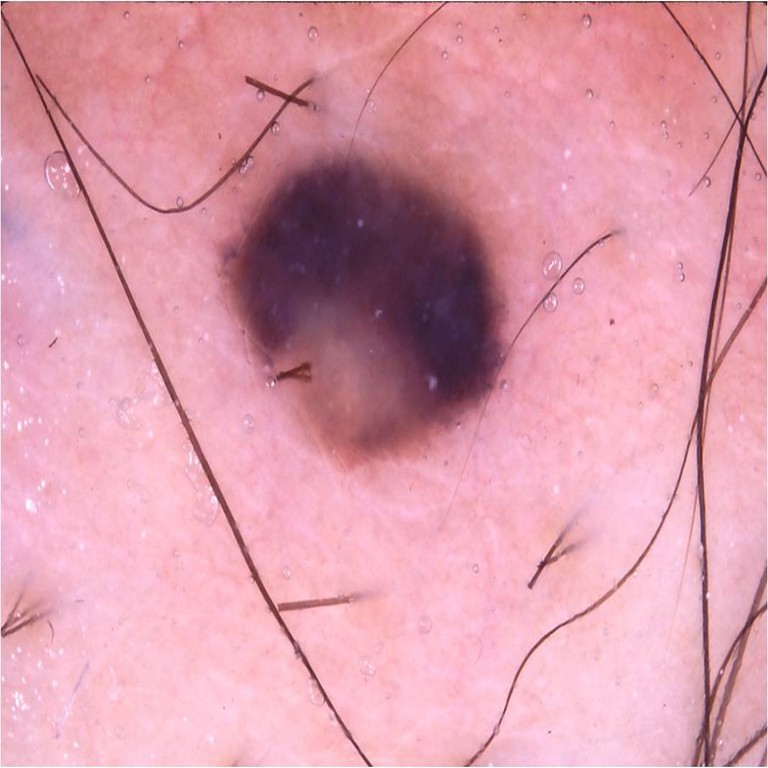}%
  \includegraphics[width=0.16\textwidth]{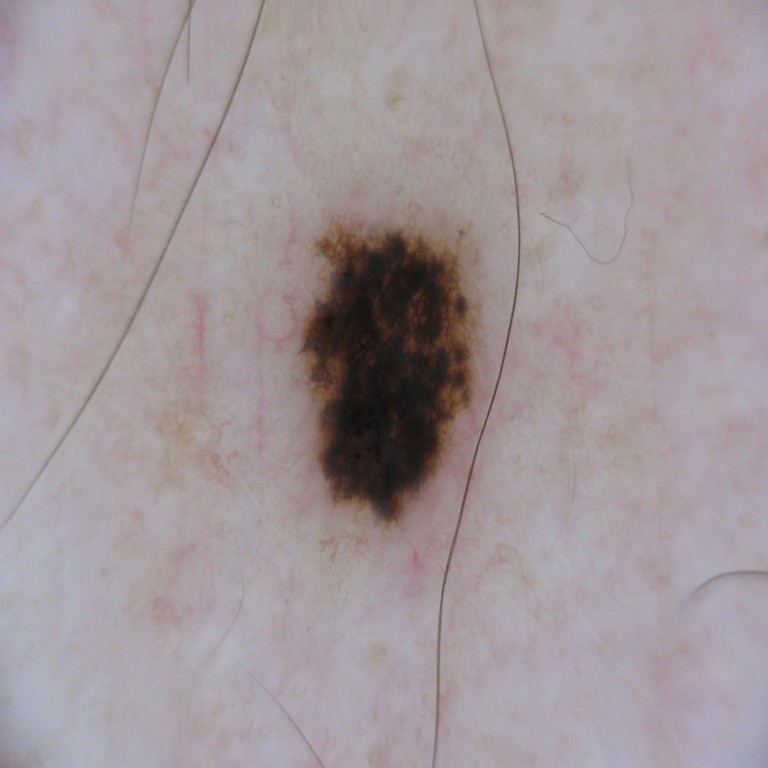}%
  
  \caption{Example of medium density hair augmentation. Input images are randomly augmented with the medium-density hair extracted from the training images.}
  \label{fig.bias-aug.medium}
\end{figure}

Augmentation policy depends mostly on the examined problem. It describes \textit{what} feature is being modified and \textit{how} to modify it. This can include feature modification, adding elements, categorical features values swapping, or others. For instance, if we recognize a potential bias based as a feature called \textit{country of origin}, then we can randomly modify the \textit{country of origin} value during the training (i.e. randomly switch \textit{Poland} to \textit{China}). When we suppose that the bias in \textit{bird song classification} comes from the sounds of the city when classifying doves songs (because doves are often recorded in cities in contrary to other bird species), then we might randomly add city noises to samples. If the algorithm is biased towards an \textit{age} (i.e. works well on people aged 16--18 but poorly on 19 years old) we can modify \textit{age values} in certain range. 
In this case, I focused on examining skin lesion classification problem. I have decided to augment the dataset with hair, ruler marks and black frames, as they are possibly biasing factors. 

\begin{figure}[!htb]
\centering
\textbf{Original images}

  \includegraphics[width=0.8\textwidth]{Figures/aug/clean_aug.png}%
  
  \textbf{Augmented images with short hair}
  
  \includegraphics[width=0.8\textwidth]{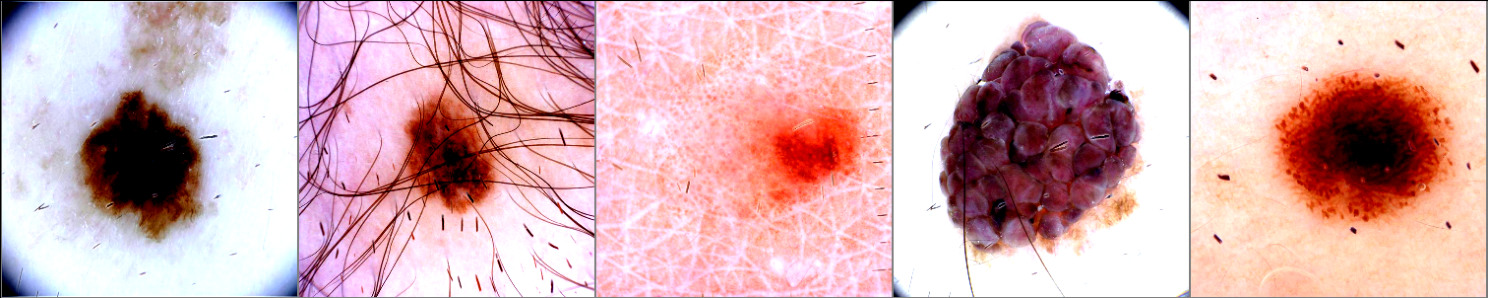}%
  
  \textbf{Short hair masks}
  
  \includegraphics[width=0.16\textwidth]{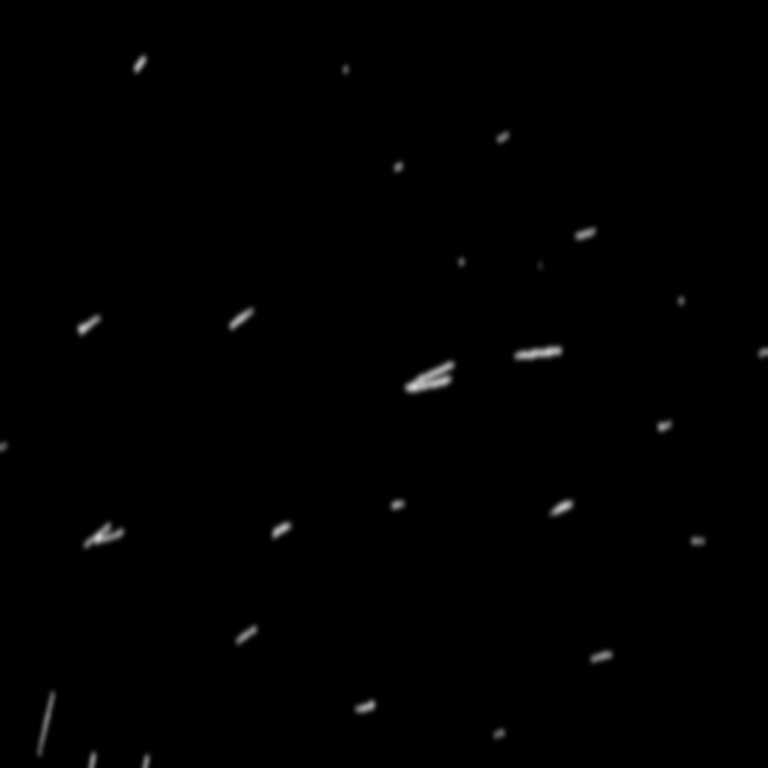}%
  \includegraphics[width=0.16\textwidth]{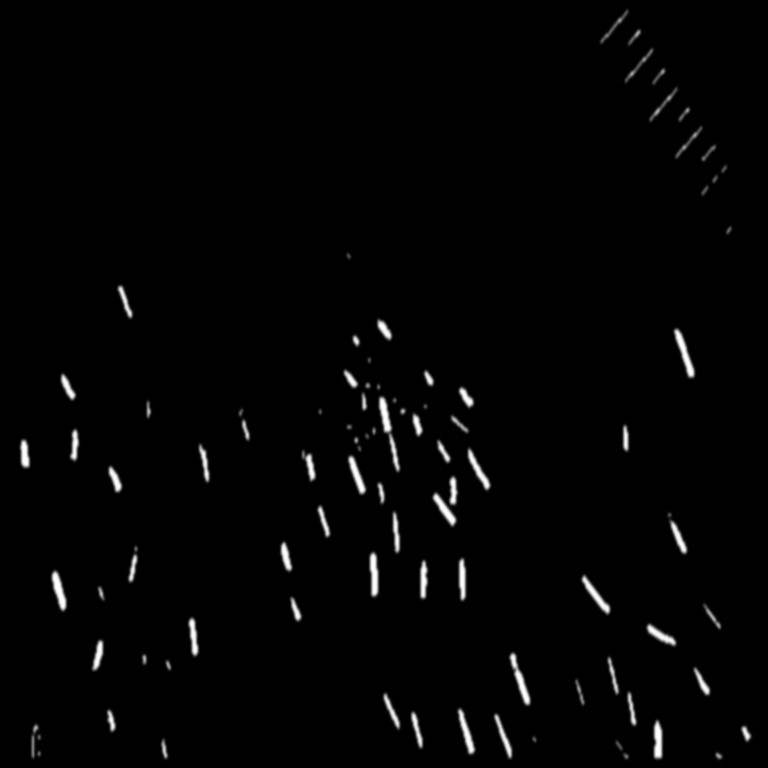}%
    \includegraphics[width=0.16\textwidth]{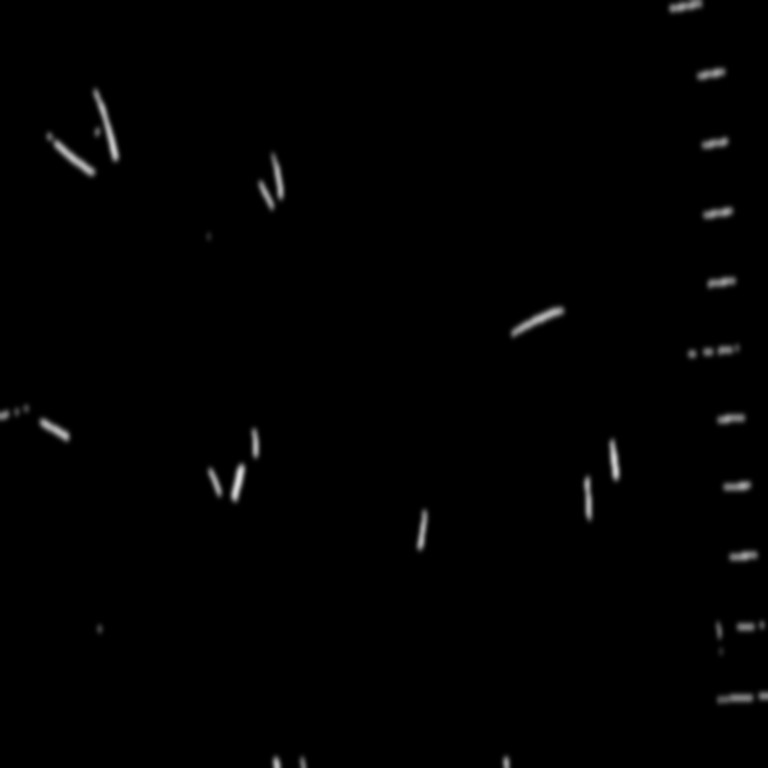}%
  \includegraphics[width=0.16\textwidth]{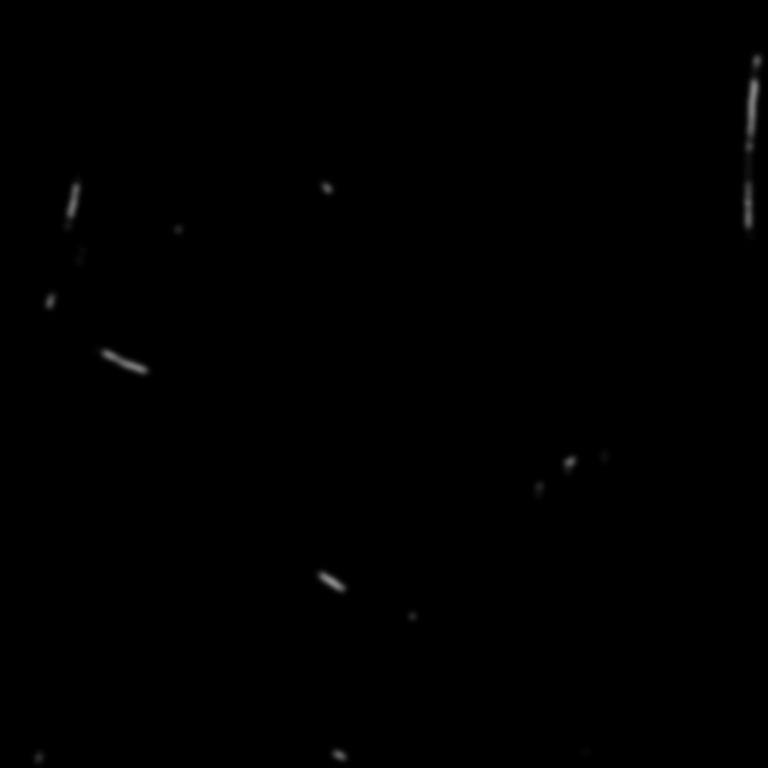}%
  \includegraphics[width=0.16\textwidth]{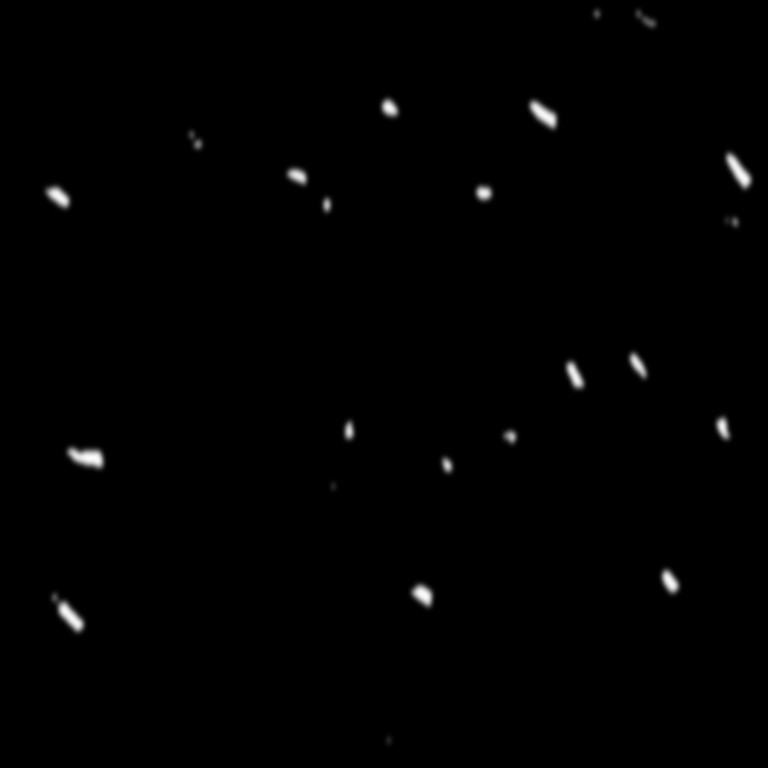}%
  
  \textbf{Short hair source images}
  
  \includegraphics[width=0.16\textwidth]{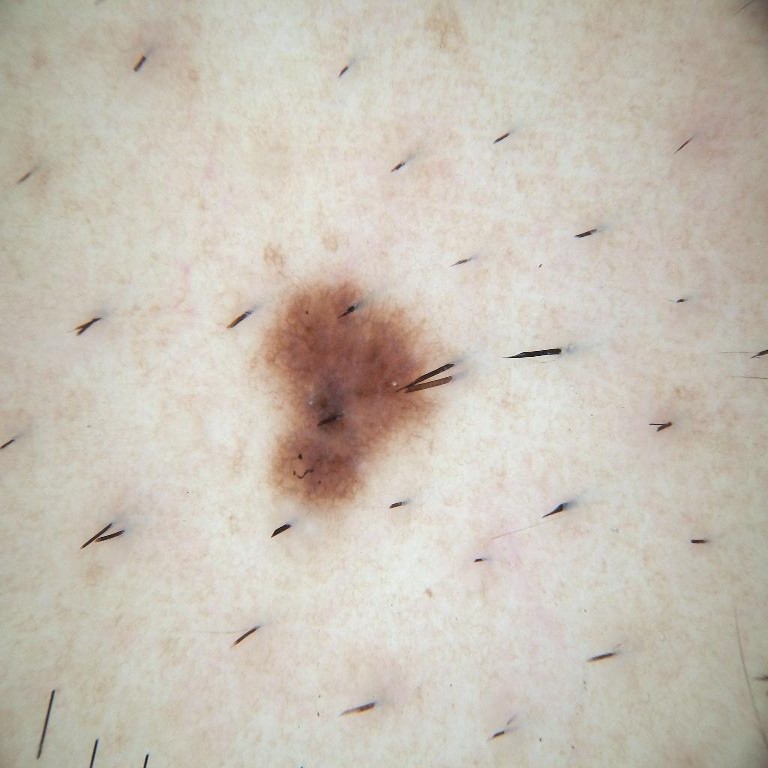}%
  \includegraphics[width=0.16\textwidth]{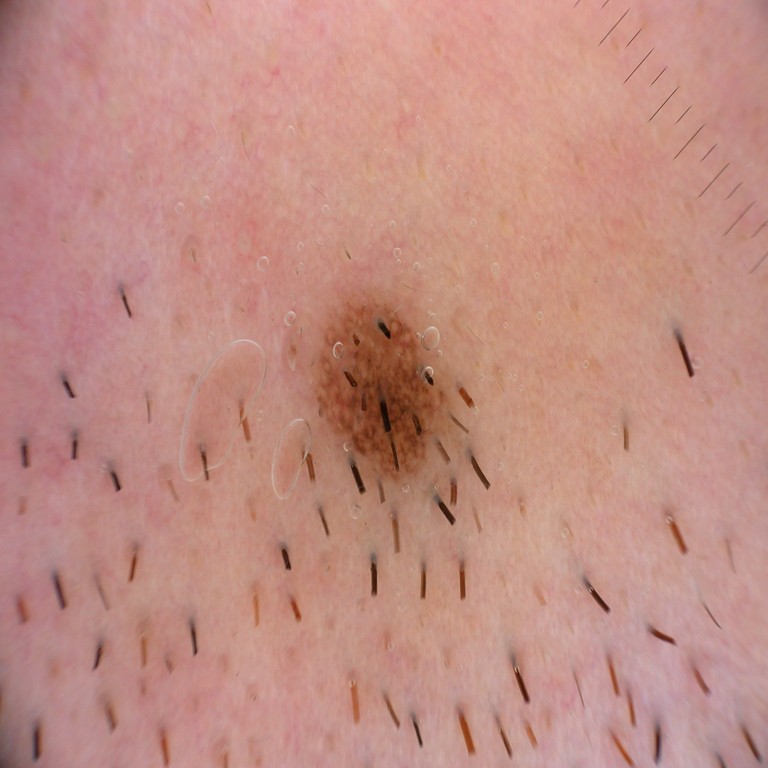}%
    \includegraphics[width=0.16\textwidth]{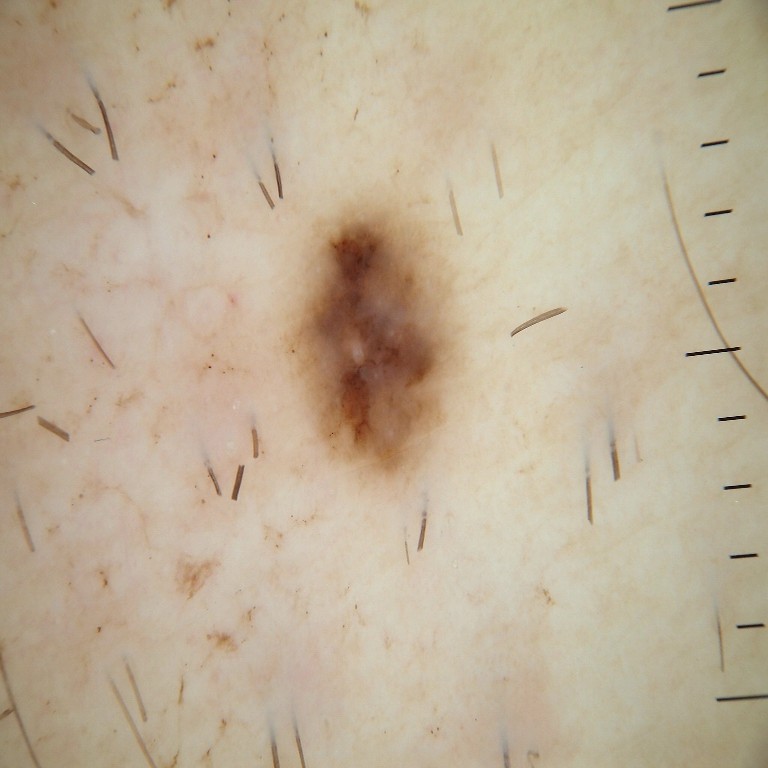}%
  \includegraphics[width=0.16\textwidth]{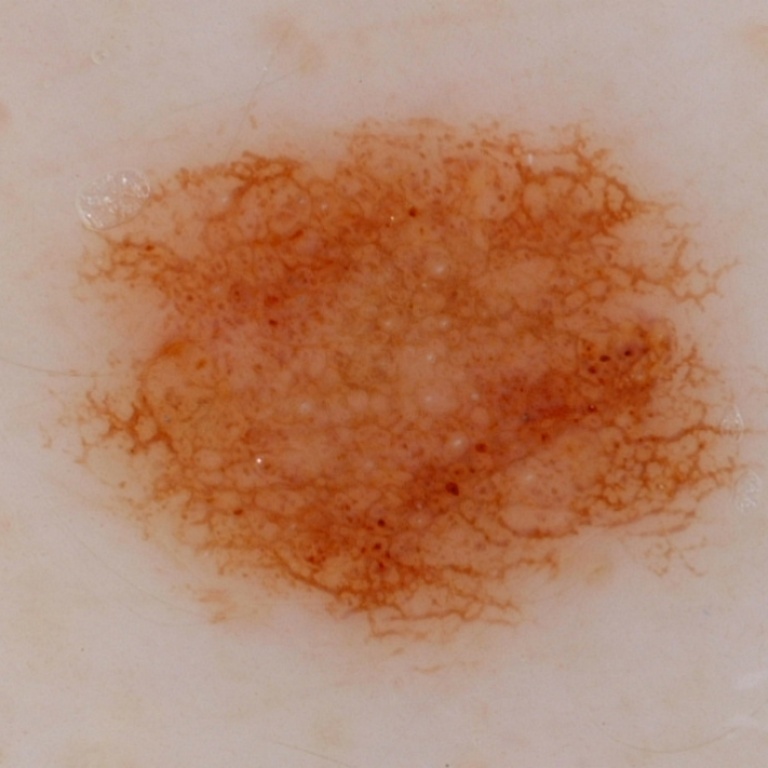}%
  \includegraphics[width=0.16\textwidth]{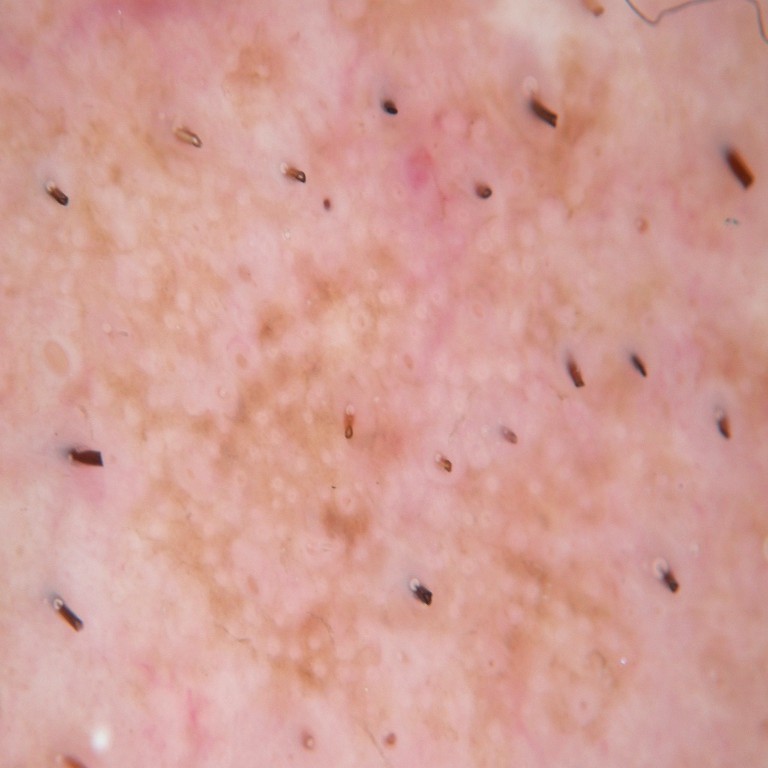}%
  
  \caption{Example of short hair augmentation.}
  \label{fig.bias-aug.short}
\end{figure}
\textbf{Hair augmentation.} To augment hair on the skin lesion images I used hair segmentation masks with corresponding hair images. Each segmentation mask was used to cut real hair from images with skin lesions. Then, those hair were randomly placed on the augmented image. Additionally, I randomly rotated each segmentation mask before placing it to additionally randomize the process. The process of hair augmentation was done before any color modifications of the image. Hair are randomly placed on the image, even if other hair are already visible. Ruler and hair augmentation are done in a single step, as the same segmentation masks are used to insert those features. As hair density is also an important factor, it was decided to test it with dense, short and normal hair. \textit{Medium hair} are a typical hair of various colors and thickness. Examples of medium hair augmentation are presented in Figure \ref{fig.bias-aug.medium}.

\textit{Short hair} are short or shaved hair, usually dark, that can be easily mistaken with some atypical, differential structures of skin lesions. Original images with augmented images with short hair are presented in Figure \ref{fig.bias-aug.short}

\begin{figure}[!htb]
\centering
\textbf{Original images}
  \includegraphics[width=0.8\textwidth]{Figures/aug/clean_aug.png}%
  
  \textbf{Augmented images with dense hair}
  \includegraphics[width=0.8\textwidth]{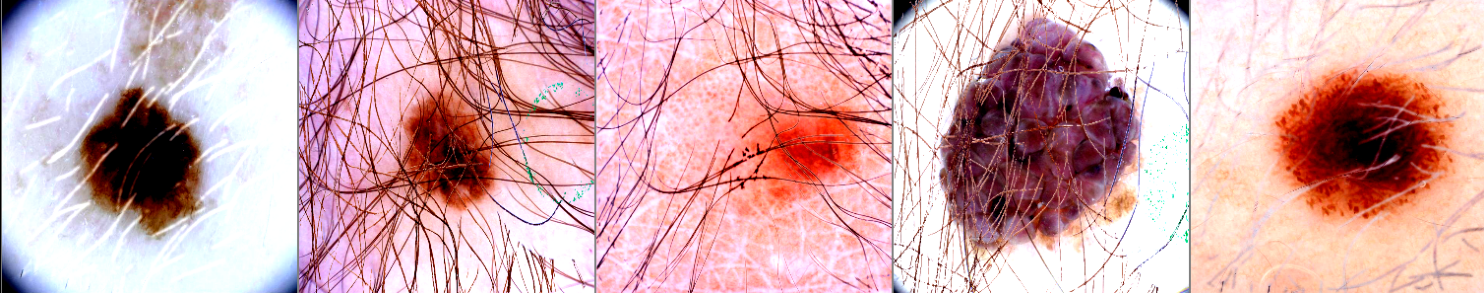}%
  
      \textbf{Dense hair masks}
  
  \includegraphics[width=0.16\textwidth]{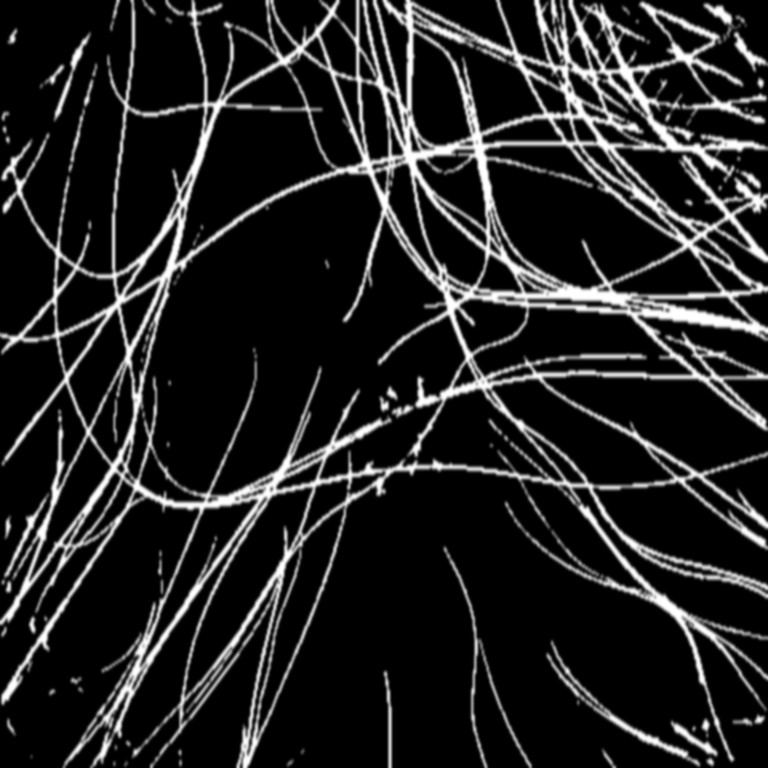}%
  \includegraphics[width=0.16\textwidth]{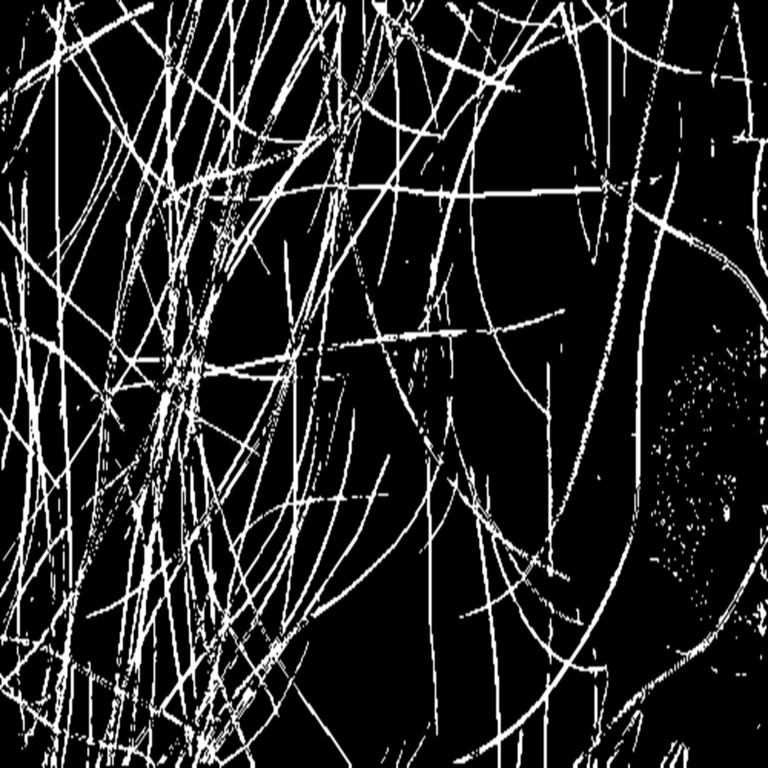}%
    \includegraphics[width=0.16\textwidth]{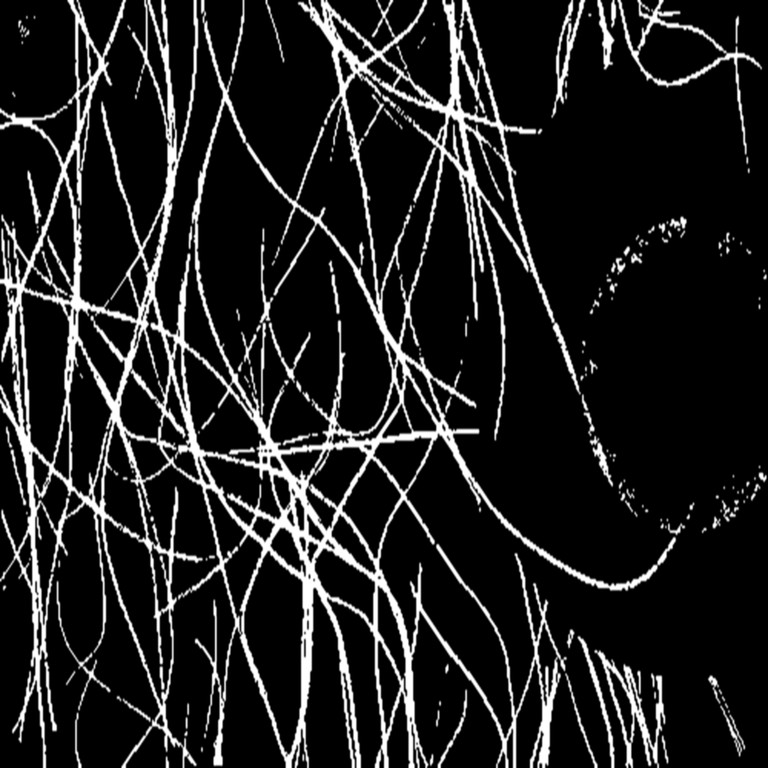}%
  \includegraphics[width=0.16\textwidth]{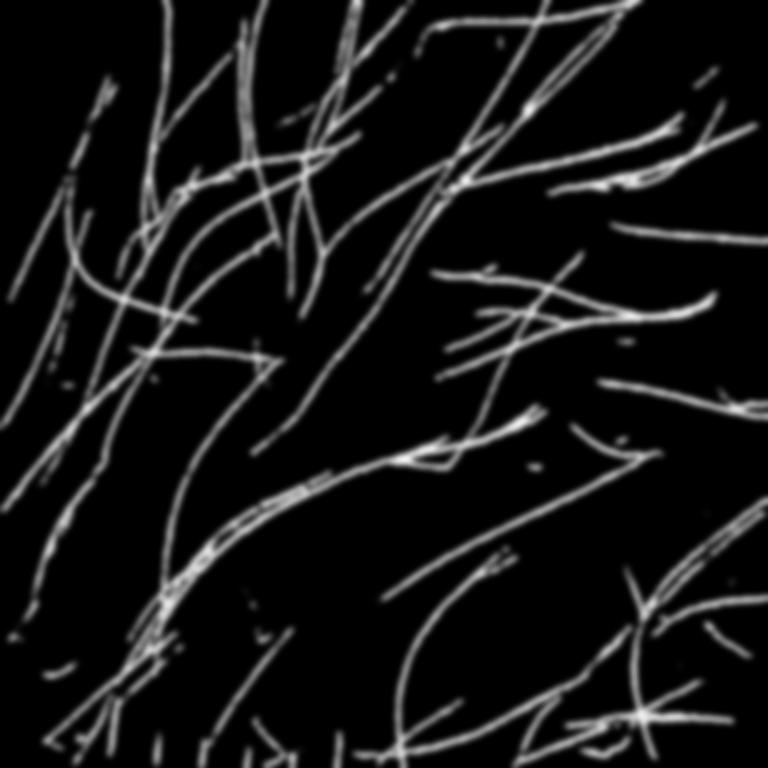}%
  \includegraphics[width=0.16\textwidth]{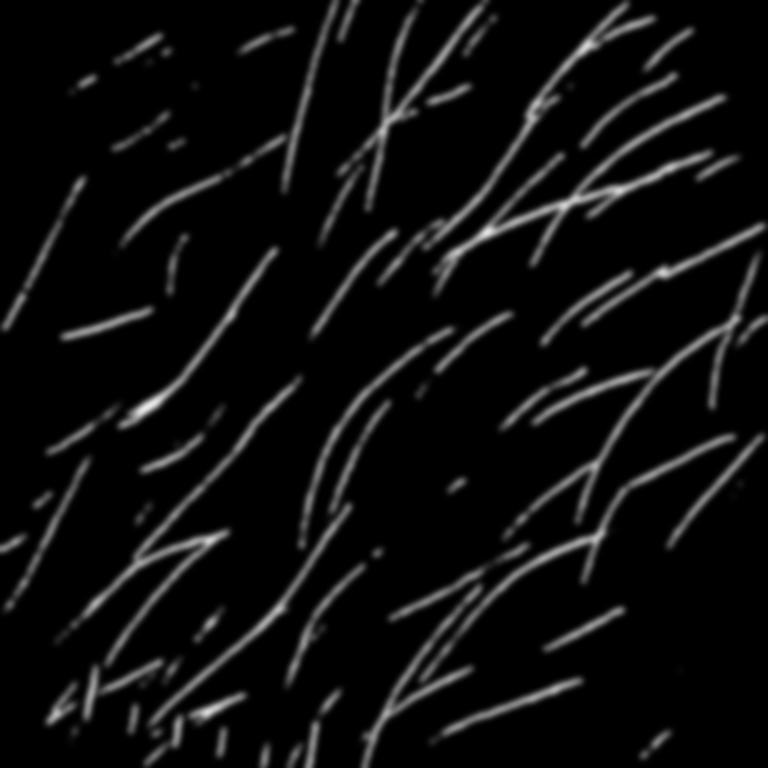}%
  
  \textbf{Dense hair source images}
  
   \includegraphics[width=0.16\textwidth]{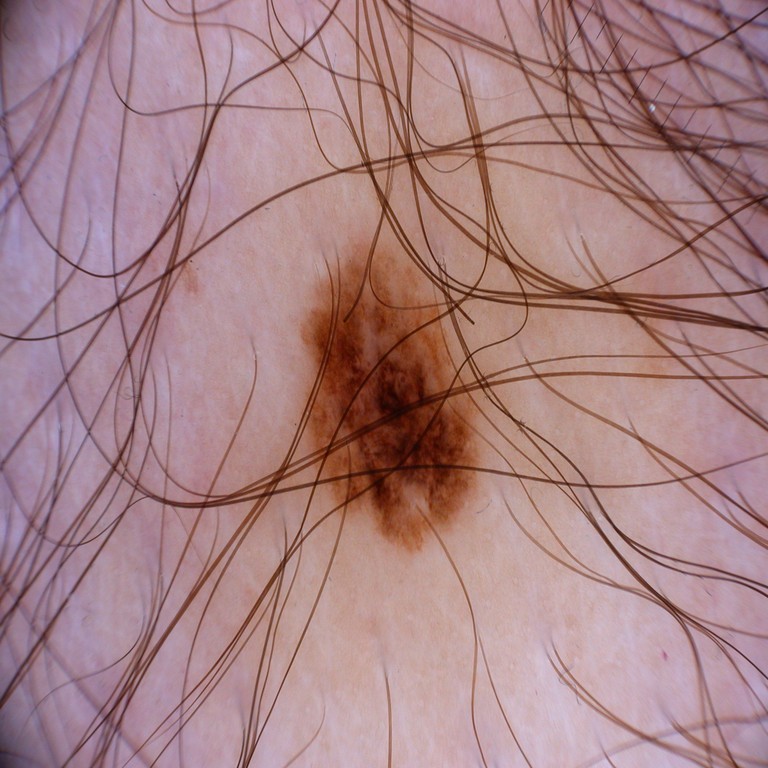}%
  \includegraphics[width=0.16\textwidth]{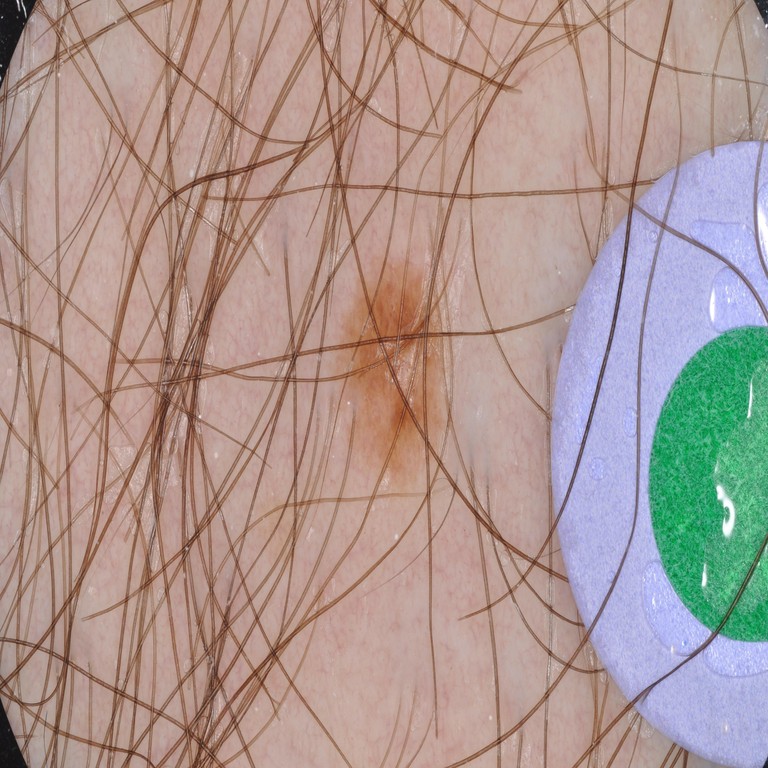}%
    \includegraphics[width=0.16\textwidth]{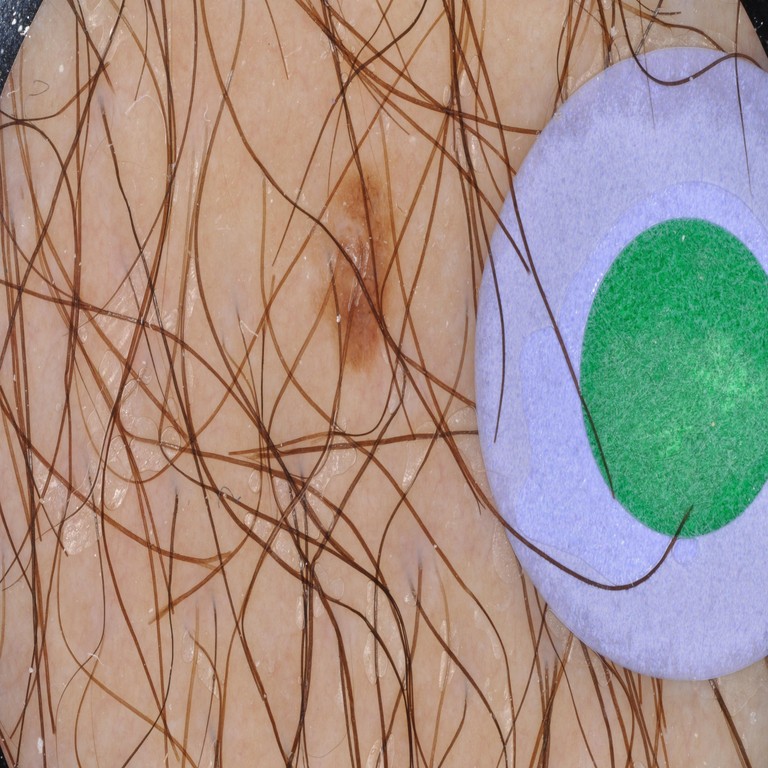}%
  \includegraphics[width=0.16\textwidth]{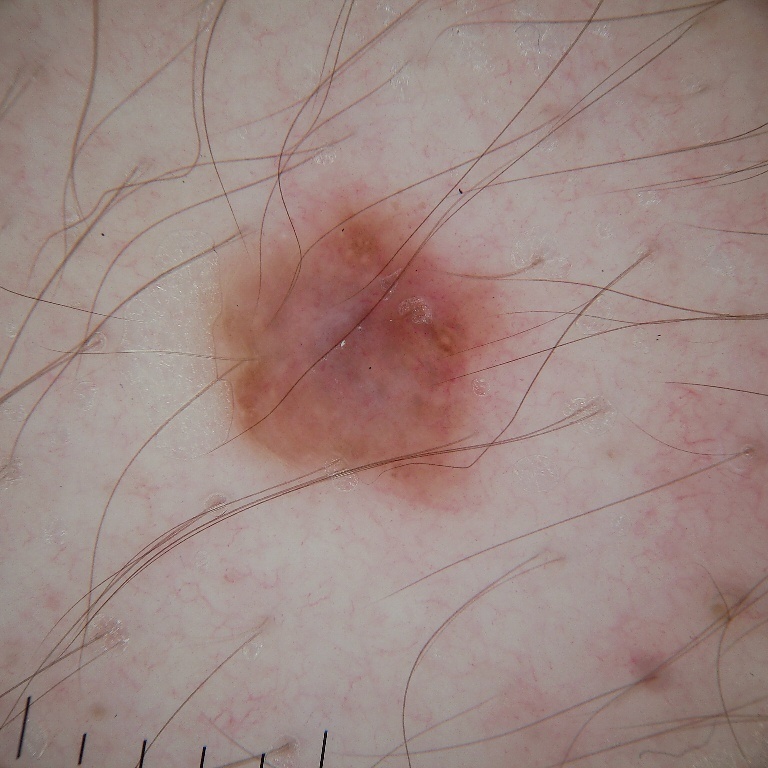}%
  \includegraphics[width=0.16\textwidth]{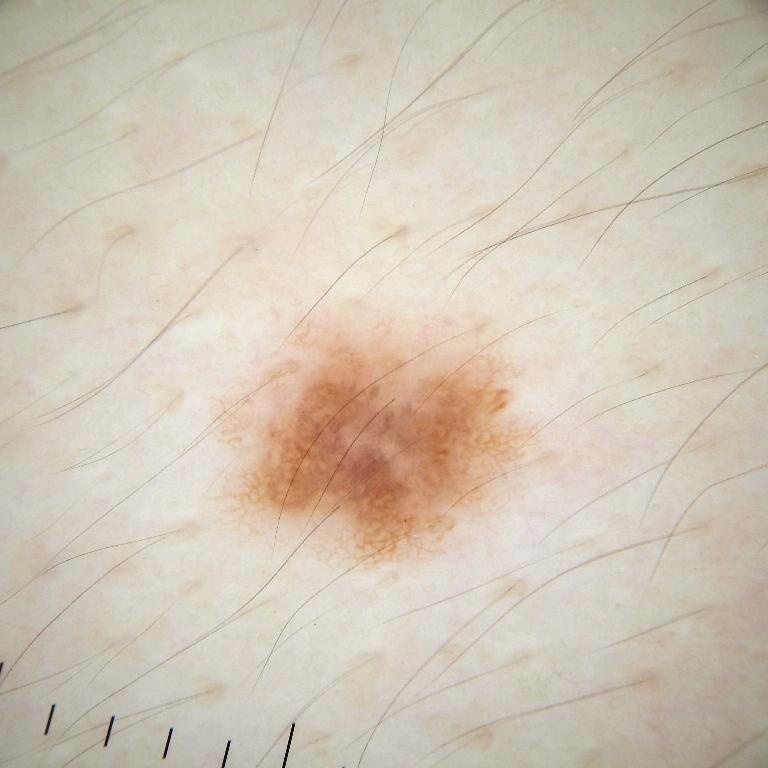}%
  
  \caption{Example of dense hair augmentation.}
  \label{fig.bias-aug.dense}
\end{figure}

\textit{Dense hair }are dense or very thick hair covering a significant part of an image. Usually, the lesion with dense hair is located on the patient’s head.  
Dense hair examples are presented in Figure \ref{fig.bias-aug.dense}.

\clearpage

\begin{figure}[!htb]
\centering
\textbf{Original images}
  \includegraphics[width=0.8\textwidth]{Figures/aug/clean_aug.png}%
  
  \textbf{Augmented images with frames}
  \includegraphics[width=0.8\textwidth]{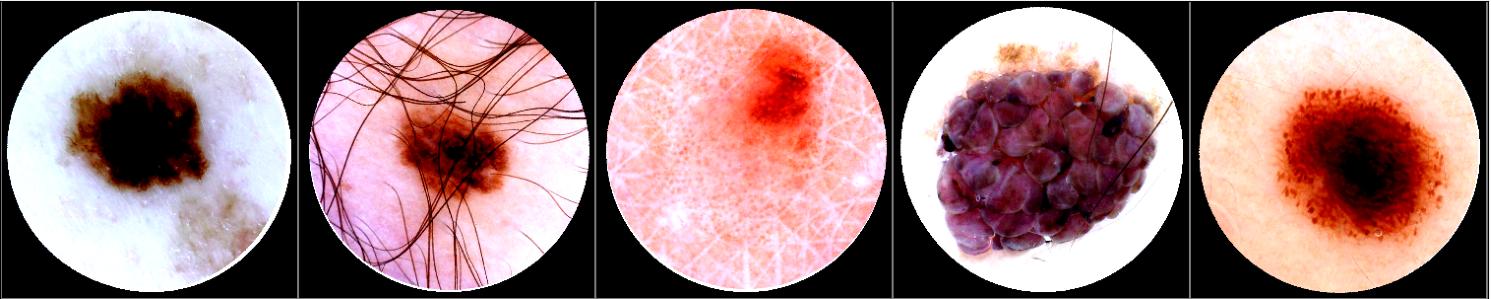}%
  
  \caption{Example of black frame augmentation. Input images are randomly augmented with the black frames related to the type of dermatoscope used to capture the image.}
  \label{fig.bias-aug.frame}
\end{figure}

\textbf{Frame augmentation.} Both black and white round markings around the skin lesion, black rectangle edges, vignette are defined the frames. However, I focused on the black round and rectangular markings of different sizes and shapes. Example black frame augmentations are presented in Figure~\ref{fig.bias-aug.frame}. Frame augmentation is done in a separate step.

\clearpage

\begin{figure}[!htb]
\centering
\textbf{Original images}
  \includegraphics[width=0.8\textwidth]{Figures/aug/clean_aug.png}%
  
  \textbf{Augmented images with ruler marks}
  \includegraphics[width=0.8\textwidth]{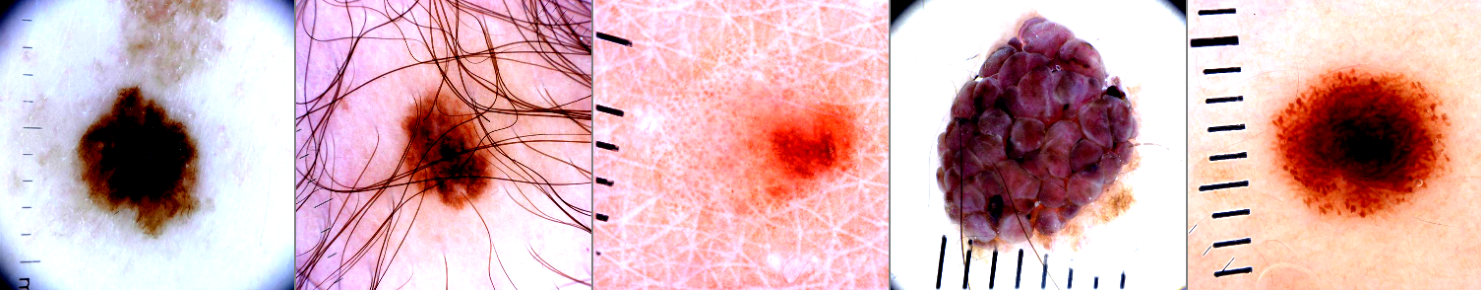}%
  
\textbf{Ruler marks masks}

  \includegraphics[width=0.16\textwidth]{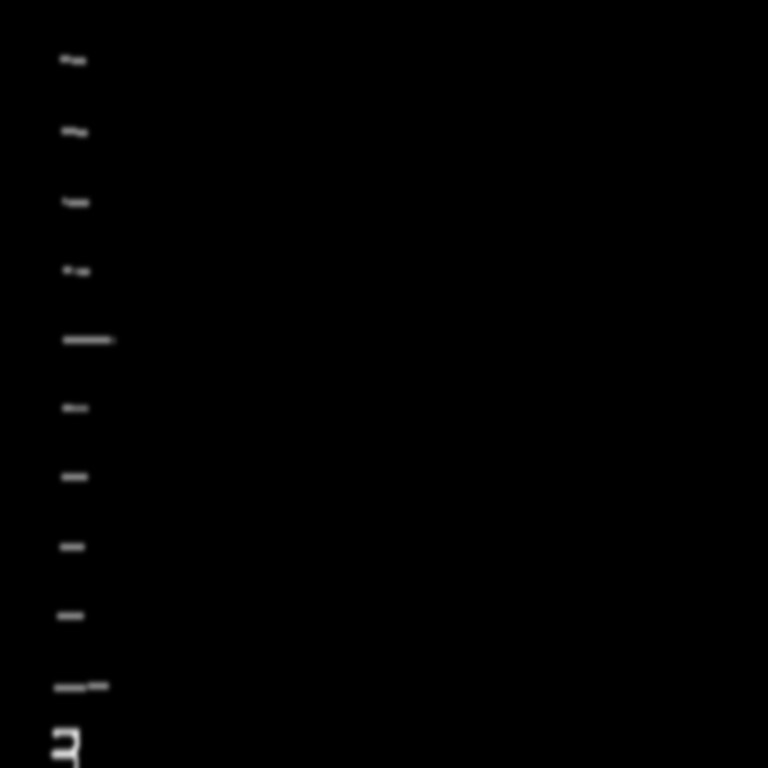}%
    \includegraphics[width=0.16\textwidth]{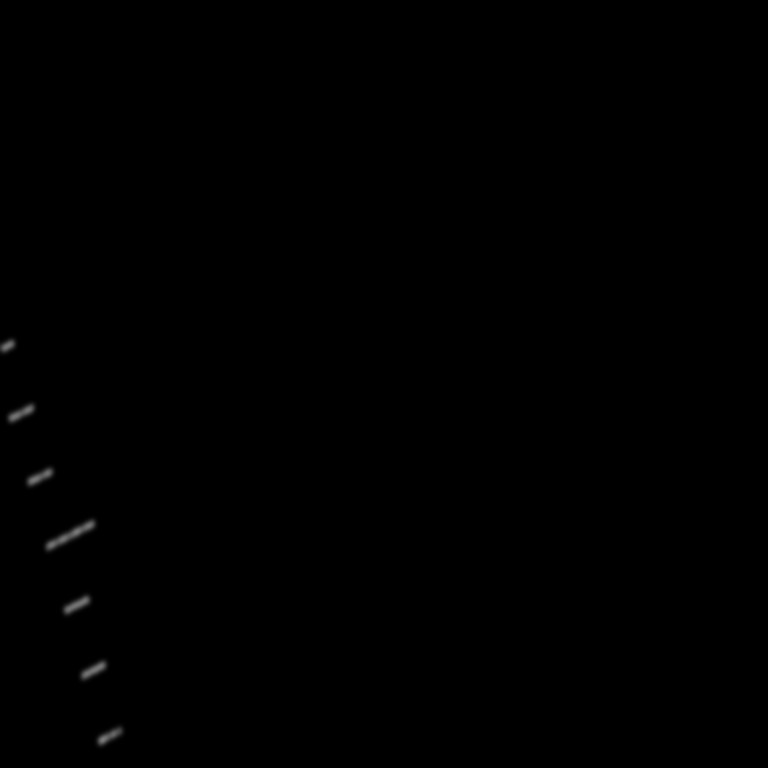}%
  \includegraphics[width=0.16\textwidth]{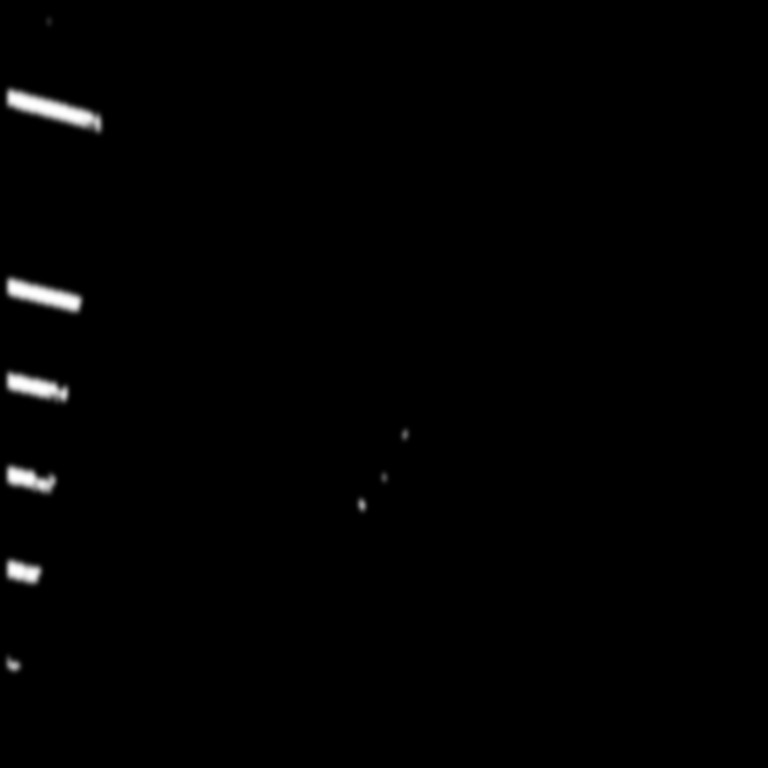}%
  \includegraphics[width=0.16\textwidth]{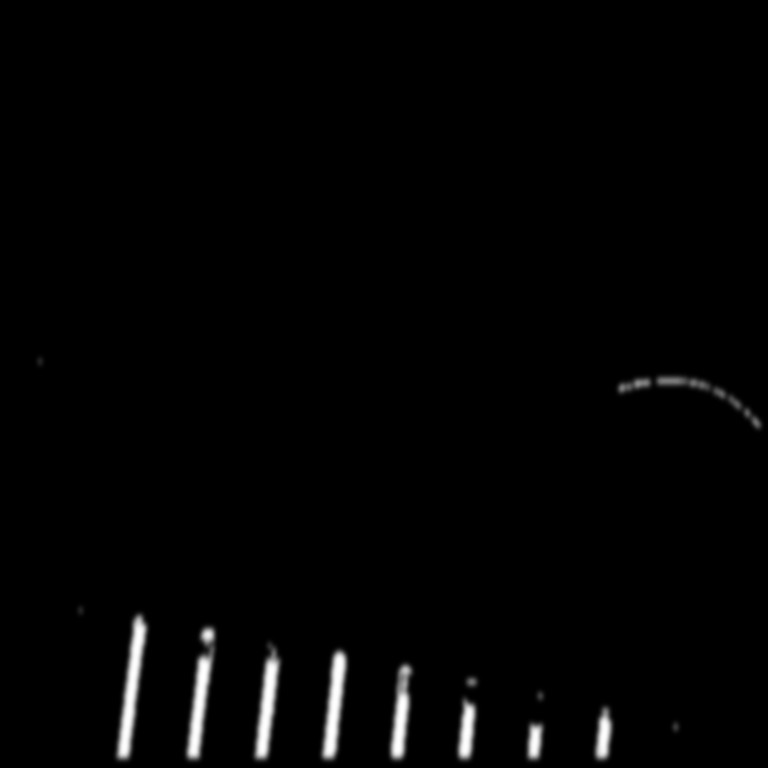}%
  \includegraphics[width=0.16\textwidth]{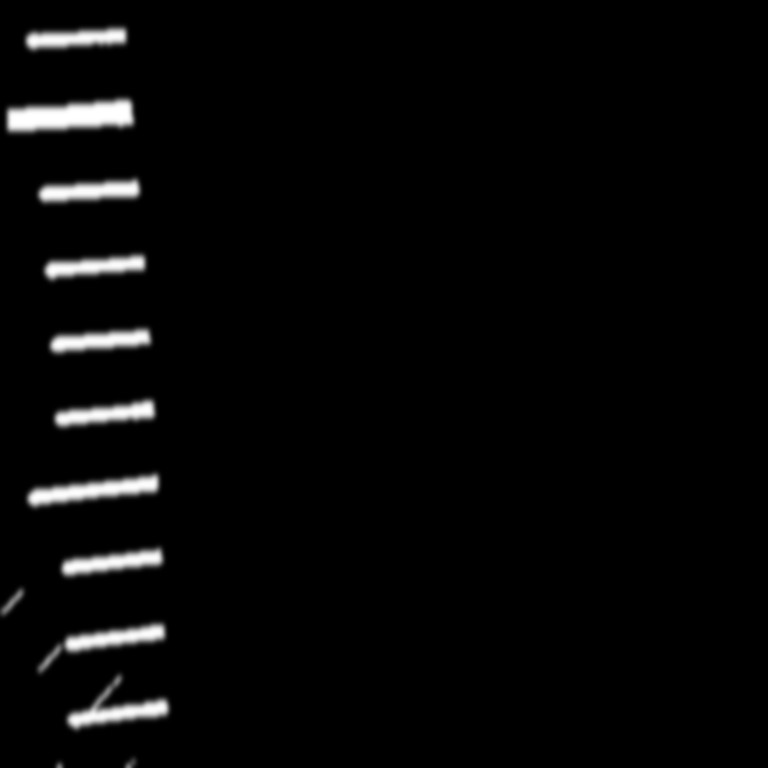}%
  
  \textbf{Ruler marks source images}

  \includegraphics[width=0.16\textwidth]{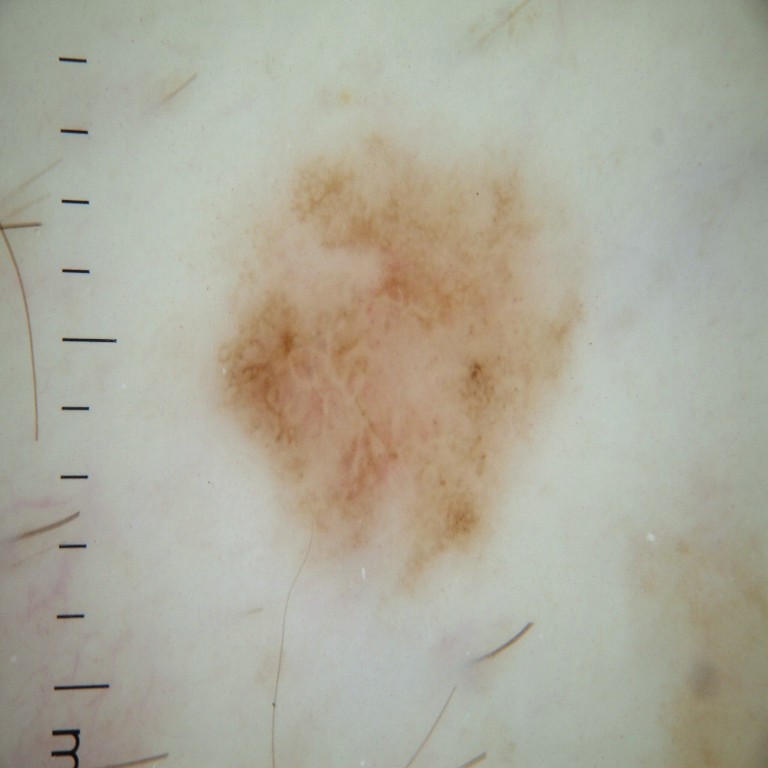}%
    \includegraphics[width=0.16\textwidth]{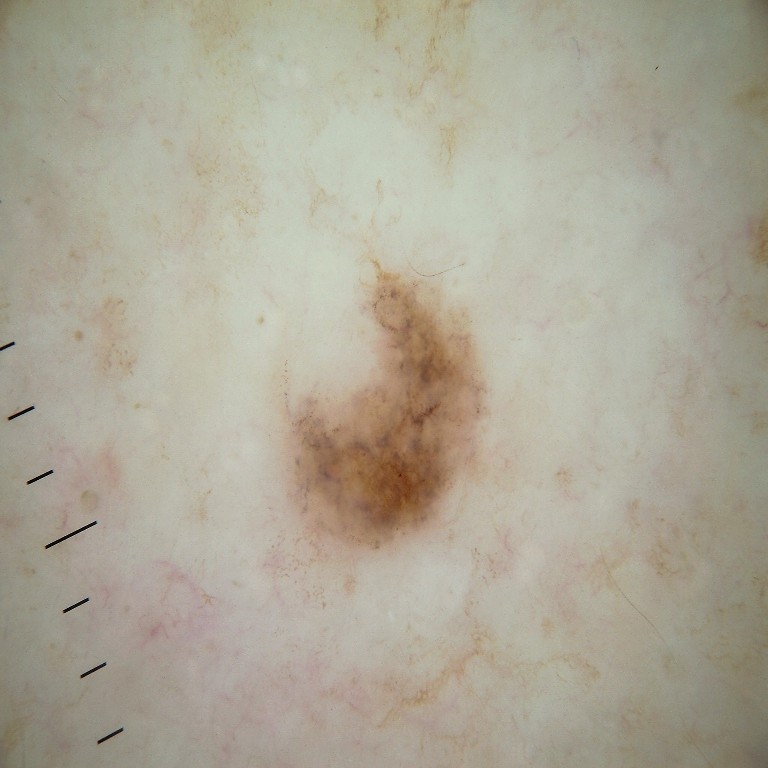}%
  \includegraphics[width=0.16\textwidth]{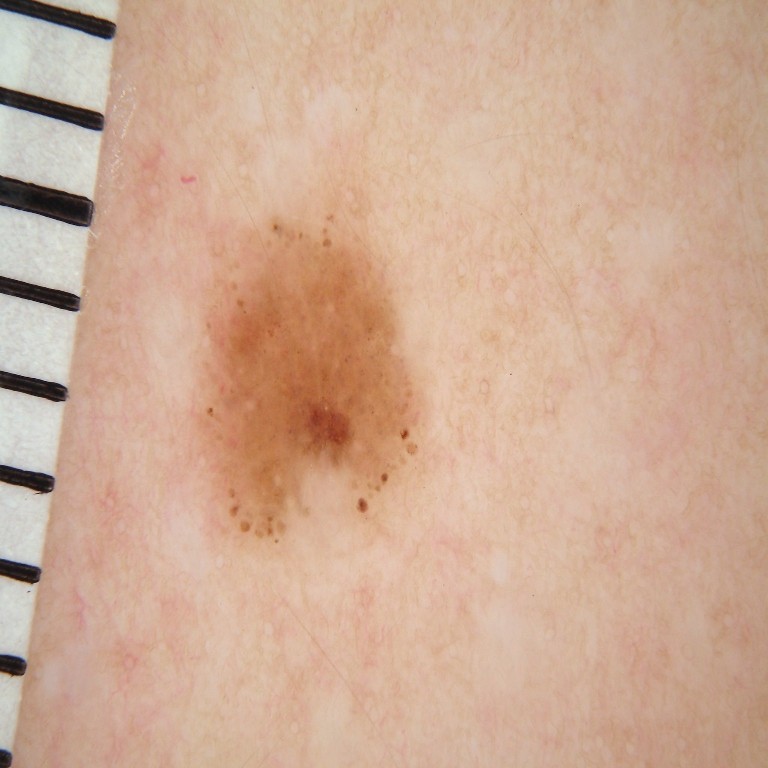}%
  \includegraphics[width=0.16\textwidth]{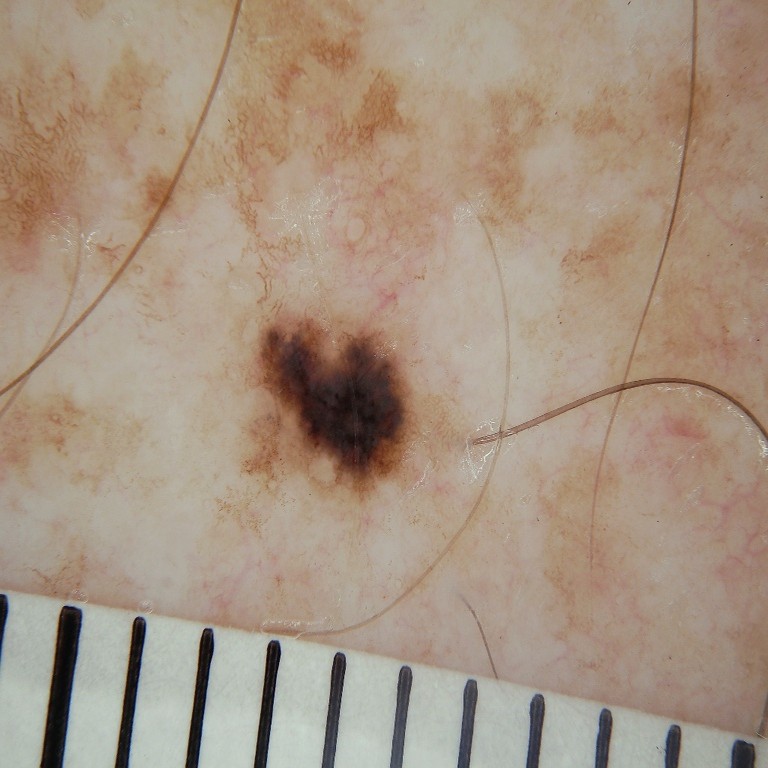}%
  \includegraphics[width=0.16\textwidth]{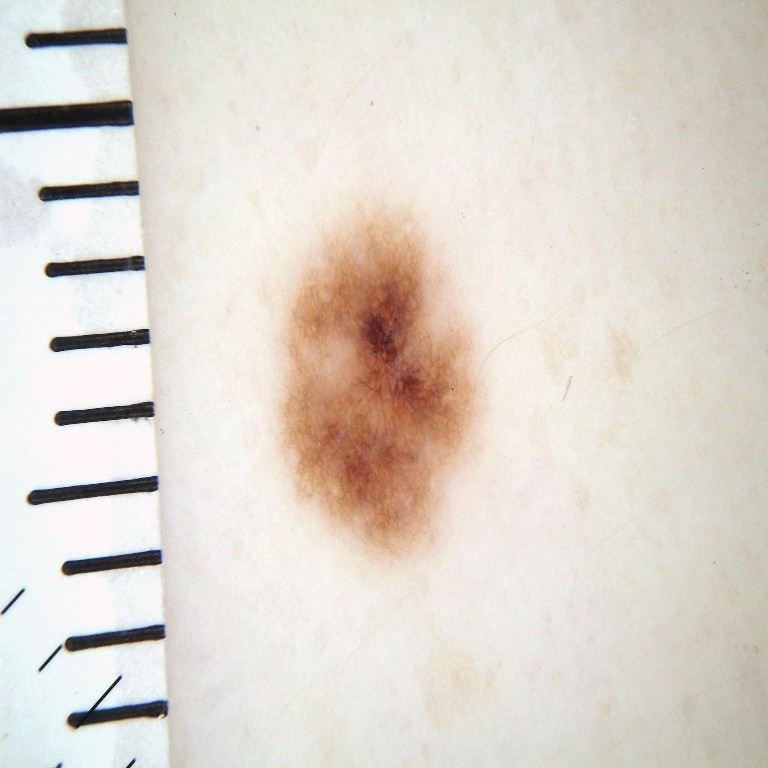}%
  
  \caption{Example of ruler marks augmentation.}
  \label{fig.bias-aug.ruler}
\end{figure}

\textbf{Ruler augmentation. }The ruler marks are a partially or fully visible ruler markings of different shapes and colors. They can be found throughout the whole dataset of skin lesions. The augmentation process of ruler marks was identical to hair augmentation. I used segmentation masks and corresponding images to insert ruler mark randomly into images. Examples of ruler augmentations are presented in Figure \ref{fig.bias-aug.ruler}. Ruler and hair augmentation are done in a single step, as for convenience the same segmentation masks are used to insert those features.

\clearpage
\section{Experiments}
\subsection{Data and training details}
For the following experiments, The data from ISIC Archive challenges 2019 \cite{tschandl2018ham10000, codella2018skin, combalia2019bcn20000} and 2020 \cite{isic2020} was used. As for the training set, 47,171 benign skin lesions and 4,753 malignant were used. For the test set, 6 180 benign and 366 malignant were selected. I manually prepared frame images, and used hair and ruler segmentation masks proposed by Ramella \cite{ramella2021hair}. 

For experiments, a family of EfficientNet \cite{tan2019EfficientNet} models were used. I used only EfficientNets B2 -- B5, where the naming denotes the size of neural net (B0 is the smallest architecture, B7 is the largest). All models were trained for ten epochs, as they tend to overfit in case of longer training. I used $lr = 5e-5$ with step scheduler multiplying $lr$ every epoch by *0.9. For EfficientNet-B2 I used batch size 20, EfficientNet-B3  batch size 12, and for EfficientNet-B4 use batch size $b=6$. 
The data augmentation pipeline was inspired by the winning model of ISIC challenge 2020 \cite{ha2020identifying}. For every training the following image transformations were used: random horizontal ($p=0.5$) and vertical flip ($p=0.5$), random bias insertion: hair and ruler together ($p=[0.0, 0.25, 0.5, 0.75, 1.0]$) or a frame ($p=[0.0, 0.25, 0.5, 0.75, 1.0]$), random brightness ($p=0.65$), random contrast ($p=0.65$), random hue/saturation ($p=0.5$), random shift/scale/rotate ($p=0.85$), resize, cutout ($p=0.1$) and normalization.

\subsection{Performance evaluation}
\begin{table*}[]
\centering
\caption{Comparison of different models trained with different augmentation policies. $F_1$, recall, and precision for org are calculated on images without added frames. and whereas aug for the images with frames.}
\label{tab:frame-aug-performance}
\resizebox{\textwidth}{!}{%
\begin{tabular}{lllllllll}
\hline
\textbf{model*} & \textbf{$p$} & \textbf{$F_1^{org}$} & \textbf{$F_1^{aug}$} & \textbf{$F_1^{mean}$} & \textbf{$recall^{org}$} & \textbf{$recall^{aug}$} & \textbf{$precision^{org}$} & \textbf{$precision^{aug}$} \\ \hline
B2 & 0 & 59.96\% & 52.99\% & 56.48\% & 46.45\% & 55.74\% & 84.58\% & 50.50\% \\
 & 0.25 & 60.14\% & 59.12\% & 59.63\% & 46.17\% & 44.26\% & 86.22\% & 89.01\% \\
 & \textbf{0.5} & \textbf{64.24\%} & \textbf{62.50\%} & \textbf{63.37\%} & \textbf{53.01\%} & \textbf{50.55\%} & \textbf{81.51\%} & \textbf{81.86\%} \\
 & 0.75 & 60.14\% & 59.49\% & 59.82\% & 46.17\% & 44.54\% & 86.22\% & 89.56\% \\
 & 1 & 58.42\% & 58.02\% & 58.22\% & 46.45\% & 46.45\% & 78.70\% & 77.27\% \\ \hline
B3 & 0 & 66.15\% & 54.69\% & 60.42\% & 58.47\% & 58.20\% & 76.16\% & 51.57\% \\
 & \textbf{0.25} & \textbf{64.93\%} & \textbf{61.73\%} & \textbf{63.33\%} & \textbf{51.09\%} & \textbf{46.72\%} & \textbf{89.05\%} & \textbf{90.96\%} \\
 & 0.5 & 62.13\% & 58.59\% & 60.36\% & 48.63\% & 44.26\% & 85.99\% & 86.63\% \\
 & 0.75 & 62.50\% & 61.54\% & 62.02\% & 49.18\% & 46.99\% & 85.71\% & 89.12\% \\
 & 1 & 62.44\% & 61.15\% & 61.80\% & 49.73\% & 49.45\% & 83.87\% & 80.09\% \\ \hline
B4 & 0 & 68.35\% & 61.60\% & 64.98\% & 55.46\% & 58.74\% & 89.04\% & 64.76\% \\
 & 0.25 & 64.93\% & 63.76\% & 64.35\% & 54.37\% & 51.91\% & 80.57\% & 82.61\% \\
 & 0.5 & 66.01\% & 64.40\% & 65.21\% & 54.64\% & 51.64\% & 83.33\% & 85.52\% \\
 & 0.75 & 66.99\% & 65.01\% & 66.00\% & 57.10\% & 53.55\% & 81.01\% & 82.70\% \\
 & \textbf{1} & \textbf{67.55\%} & \textbf{65.65\%} & \textbf{66.60\%} & \textbf{62.57\%} & \textbf{58.74\%} & \textbf{73.40\%} & \textbf{74.39\%} \\ \hline
B5 & 0 & 25.27\% & 12.05\% & 18.66\% & 53.83\% & 77.05\% & 16.51\% & 6.54\% \\
 & 0.25 & 65.55\% & 61.94\% & 63.74\% & 53.28\% & 48.91\% & 85.15\% & 84.43\% \\
 & 0.5 & 63.56\% & 61.51\% & 62.53\% & 50.27\% & 47.81\% & 86.38\% & 86.21\% \\
 & \textbf{0.75} & \textbf{65.56\%} & \textbf{63.05\%} & \textbf{64.31\%} & \textbf{54.10\%} & \textbf{52.46\%} & \textbf{83.19\%} & \textbf{79.01\%} \\
 & 1 & 61.46\% & 64.48\% & 62.97\% & 47.27\% & 51.09\% & 87.82\% & 87.38\% \\ \hline
 *EfficientNet
\end{tabular}%
}
\end{table*}
Three general metrics to evaluate the model's performance were used: $F_1$ score, $precision$ and $recall$. Each score is calculated for both original, unmodified data, and data with targeted data augmentation. Precision is the fraction of relevant instances among the retrieved instances, while recall (sensitivity) is the fraction of relevant instances that were retrieved. F-measure is the harmonic mean of precision and recall. F-score is commonly used to measure the performance on datasets with unbalanced class distribution. 

The $recall^{org}$, $precision^{org}$, and $F_1^{org}$ measures were used to evaluate how well algorithm performs on the original, unmodified data. However, the goal of Targeted Data Augmentation was not to achieve the best performance possible, but to be more robust to certain biases. Hence, the $recall^{aug}$, $precision^{aug}$,$F_1^{aug}$ is used to measure the performance on data with inserted bias. 
Ideally, $F_1^{aug}$ should be the same as $F_1^{org}$, meaning that adding a bias to data does not change its performance. Higher differences between $F_1^{aug}$ and $F_1^{org}$ mean higher vulnerability to inserted bias. The $F_1^{mean}$ is a mean of $F_1^{org}$ and $F_1^{aug}$, hence it shows how well model performs on both original and modified data.

In case of \textit{frame augmentation}, most models achieved the best performance without data augmentation ($p = 0.0$), with one exception of EfficientNet-B2. However, at the same time, the $F_1^{aug}$, a measure used to examine the robustness to bias, was the lowest in all cases when $p = 0.0$. On average, the F-measure dropped by around seven to eight percentage points in all cases after adding a bias to the test set. Those experiments show that adding only a black frame to an image drastically changes the outcome. Many instances were wrongly recognized after this slight modification. Each experiment ended with the highest $F_1^{mean}$ for experiments with augmentation probability higher than zero. This standalone already shows the advantages of applying targeted data augmentations. The results are presented in Table~\ref{tab:frame-aug-performance}.

\begin{table}
\centering
\caption{Comparison of different models trained with different augmentation policies. The results are reported for models trained with random hair augmentations altogether with selected probability p and later tested of selected images with hair.}
\label{tab:hair-aug-performance}

\begin{tabular}{
p{0.063\linewidth}
  p{0.069\linewidth}
  p{0.054\linewidth}
  p{0.13\linewidth}
  p{0.13\linewidth}
  p{0.13\linewidth}
  p{0.17\linewidth}
  p{0.17\linewidth}
} 
\toprule
\textbf{model*} & \textbf{type} & \textbf{$p$} & \textbf{$F_1^{org}$} & \textbf{$F_1^{aug}$} & \textbf{$F_1^{mean}$} & \textbf{$precision^{org}$} & \textbf{$precision^{aug}$} \\ 
\midrule
B2 & short & 0 & 59.96\% & 54.04\% & 57.00\% & 84.58\% & 82.58\% \\
 &  & 0.25 & 61.14\% & 57.96\% & 59.55\% & 83.10\% & 83.94\% \\
 &  & 0.5 & 62.02\% & 59.43\% & 60.73\% & 85.58\% & 85.20\% \\
 & \textbf{} & \textbf{0.75} & \textbf{63.59\%} & \textbf{61.40\%} & \textbf{62.50\%} & \textbf{84.93\%} & \textbf{85.78\%} \\
 &  & 1 & 58.09\% & 55.47\% & 56.78\% & 88.76\% & 89.63\% \\
 & medium & 0 & 59.96\% & 51.42\% & 55.69\% & 84.58\% & 83.44\% \\
 &  & 0.25 & 61.14\% & 58.20\% & 59.67\% & 83.10\% & 82.09\% \\
 &  & 0.5 & 62.02\% & 58.82\% & 60.42\% & 85.58\% & 80.19\% \\
 & \textbf{} & \textbf{0.75} & \textbf{63.59\%} & \textbf{64.38\%} & \textbf{63.99\%} & \textbf{84.93\%} & \textbf{86.24\%} \\
 &  & 1 & 58.09\% & 56.26\% & 57.18\% & 88.76\% & 83.78\% \\
 & dense & 0 & 59.96\% & 35.82\% & 47.89\% & 84.58\% & 81.55\% \\
 &  & 0.25 & 61.14\% & 50.00\% & 55.57\% & 83.10\% & 81.48\% \\
 &  & 0.5 & 62.02\% & 53.43\% & 57.73\% & 85.58\% & 78.72\% \\
\textbf{} & \textbf{} & \textbf{0.75} & \textbf{63.59\%} & \textbf{52.47\%} & \textbf{58.03\%} & \textbf{84.93\%} & \textbf{86.25\%} \\
 &  & 1 & 58.09\% & 48.92\% & 53.51\% & 88.76\% & 86.21\% \\ 
\midrule
B3 & short & 0 & 66.15\% & 62.89\% & 64.52\% & 76.16\% & 74.07\% \\
 &  & 0.25 & 65.02\% & 64.96\% & 64.99\% & 75.00\% & 78.38\% \\
 &  & 0.5 & 62.61\% & 59.80\% & 61.21\% & 82.22\% & 78.32\% \\
 &  & 0.75 & 66.35\% & 65.27\% & 65.81\% & 77.66\% & 79.84\% \\
\textbf{} & \textbf{} & \textbf{1} & \textbf{67.43\%} & \textbf{64.57\%} & \textbf{66.00\%} & \textbf{84.71\%} & \textbf{81.93\%} \\
 & medium & 0 & 66.15\% & 63.74\% & 64.95\% & 76.16\% & 78.71\% \\
 &  & 0.25 & 65.02\% & 63.76\% & 64.39\% & 75.00\% & 70.90\% \\
 &  & 0.5 & 62.61\% & 61.36\% & 61.99\% & 82.22\% & 80.80\% \\
 &  & 0.75 & 66.35\% & 63.97\% & 65.16\% & 77.66\% & 78.26\% \\
\textbf{} & \textbf{} & \textbf{1} & \textbf{67.43\%} & \textbf{65.36\%} & \textbf{66.40\%} & \textbf{84.71\%} & \textbf{81.30\%} \\
 & dense & 0 & 66.15\% & 50.83\% & 58.49\% & 76.16\% & 77.97\% \\
 &  & 0.25 & 65.02\% & 59.33\% & 62.18\% & 75.00\% & 67.36\% \\
 &  & 0.5 & 62.61\% & 52.89\% & 57.75\% & 82.22\% & 73.66\% \\
 &  & 0.75 & 66.35\% & 57.34\% & 61.85\% & 77.66\% & 77.93\% \\
\textbf{} & \textbf{} & \textbf{1} & \textbf{67.43\%} & \textbf{56.99\%} & \textbf{62.21\%} & \textbf{84.71\%} & \textbf{84.86\%} \\ 
\midrule
B4 & short & 0 & 68.35\% & 62.61\% & 65.48\% & 89.04\% & 86.12\% \\
 &  & 0.25 & 65.34\% & 62.89\% & 64.12\% & 83.12\% & 84.72\% \\
\textbf{} & \textbf{} & \textbf{0.5} & \textbf{66.44\%} & \textbf{66.20\%} & \textbf{66.32\%} & \textbf{88.24\%} & \textbf{90.52\%} \\
 &  & 0.75 & 61.32\% & 58.80\% & 60.06\% & 88.21\% & 87.57\% \\
 &  & 1 & 62.43\% & 60.81\% & 61.62\% & 88.06\% & 85.22\% \\
 & medium & 0 & 68.35\% & 61.37\% & 64.86\% & 89.04\% & 90.43\% \\
 &  & 0.25 & 65.34\% & 64.44\% & 64.89\% & 83.12\% & 82.83\% \\
\textbf{} & \textbf{} & \textbf{0.5} & \textbf{66.44\%} & \textbf{65.65\%} & \textbf{66.05\%} & \textbf{88.24\%} & \textbf{86.94\%} \\
 &  & 0.75 & 61.32\% & 60.18\% & 60.75\% & 88.21\% & 88.36\% \\
 &  & 1 & 62.43\% & 61.32\% & 61.88\% & 88.06\% & 88.21\% \\
 & dense & 0 & 68.35\% & 45.93\% & 57.14\% & 89.04\% & 89.68\% \\
 &  & 0.25 & 65.34\% & 54.95\% & 60.15\% & 83.12\% & 73.18\% \\
\textbf{} & \textbf{} & \textbf{0.5} & \textbf{66.44\%} & \textbf{57.50\%} & \textbf{61.97\%} & \textbf{88.24\%} & \textbf{82.99\%} \\
 &  & 0.75 & 61.32\% & 51.16\% & 56.24\% & 88.21\% & 88.00\% \\
 &  & 1 & 62.43\% & 55.87\% & 59.15\% & 88.06\% & 87.72\% \\
\bottomrule
* EfficientNet

\end{tabular}

\end{table}

Surprisingly, in experiments with \textit{hair augmentation}, most models achieved the best performance with data augmentation ($p > 0.0$). Moreover, the performance on original data was higher, but the performance on modified data was significantly greater. The $F_1^{aug}$, a measure used to examine the robustness to bias, was the lowest in all cases when data augmentation was turned off (augmentation probability $p = 0.0$). On average, the F-measure dropped by around five to even twenty-five percentage points in all cases after adding a bias to the test set. Those experiments show that adding hair input images change the outcome. Previous experiments showed that hair is not that strongly correlated with any class, as it is a common feature in all skin lesion types. However, many instances were wrongly recognized after inserting fake hair. 
This might suggest that the model learned to associate specific hair shapes with the malignancy of skin lesions.

The highest drop in performance was observed in case of dense hair. Adding dense hair resulted in performance that was lower 24.14 pp. in case of EfficientNet-B2, 15.32 pp.  EfficientNet-B3, 22.42 pp. EfficientNet-B4, and 20.9 pp. in case of EfficientNet-B5. The lowest differences observed were in case of short hair; from around four to six percentage points of difference. The results in medium hair were similar to results on black frame insertion. The difference varied around six to eight percentage points.
It is worth to notice, that difference between $F_1^{aug}$ and $F_1^{org}$ drastically lowered when training with $p > 0.0$. Again, each experiment ended with the highest $F_1^{mean}$ for experiments with augmentation probability higher than zero. The results are presented in Table~\ref{tab:hair-aug-performance}.

\begin{table}[]
\centering
\caption{Comparison of different models trained with different augmentation policies. $F_1$, recall, and precision for $org$ are calculated on images without added ruler marks, whereas $aug$ for the images with ruler marks.}
\label{tab:ruler-aug-performance}
\resizebox{\textwidth}{!}{%
\begin{tabular}{@{}lllllllll@{}}
\toprule
model* & \textbf{$p$} & \textbf{$F_1^{org}$} & \textbf{$F_1^{aug}$} & \textbf{$F_1^{mean}$} & \textbf{$recall^{org}$} & \textbf{$recall^{aug}$} & \textbf{$precision^{org}$} & \textbf{$precision^{aug}$} \\ \midrule
B2 & 0 & 59.96\% & 57.09\% & 58.53\% & 46.45\% & 42.90\% & 84.58\% & 85.33\% \\
 & 0.25 & 61.14\% & 60.22\% & 60.68\% & 48.36\% & 45.90\% & 83.10\% & 87.50\% \\
 & 0.5 & 62.02\% & 59.82\% & 60.92\% & 48.63\% & 46.17\% & 85.58\% & 84.92\% \\
 & \textbf{0.75} & \textbf{63.59\%} & \textbf{62.76\%} & \textbf{63.18\%} & \textbf{50.82\%} & \textbf{49.73\%} & \textbf{84.93\%} & \textbf{85.05\%} \\
 & 1 & 58.09\% & 55.76\% & 56.93\% & 43.17\% & 40.98\% & 88.76\% & 87.21\% \\ \midrule
B3 & 0 & 66.15\% & 61.25\% & 63.70\% & 58.47\% & 49.45\% & 76.16\% & 80.44\% \\
 & 0.25 & 65.02\% & 65.51\% & 65.27\% & 57.38\% & 56.83\% & 75.00\% & 77.32\% \\
 & 0.5 & 62.61\% & 61.41\% & 62.01\% & 50.55\% & 48.91\% & 82.22\% & 82.49\% \\
 & 0.75 & 66.35\% & 65.38\% & 65.87\% & 57.92\% & 55.46\% & 77.66\% & 79.61\% \\
 & \textbf{1} & \textbf{67.43\%} & \textbf{66.45\%} & \textbf{66.94\%} & \textbf{56.01\%} & \textbf{54.64\%} & \textbf{84.71\%} & \textbf{84.75\%} \\ \midrule
B4 & 1 & 66.01\% & 60.42\% & 63.22\% & 54.64\% & 46.72\% & 83.33\% & 85.50\% \\
 & 0.25 & 65.34\% & 66.22\% & 65.78\% & 53.83\% & 54.10\% & 83.12\% & 85.34\% \\
 & \textbf{0.5} & \textbf{66.44\%} &\textbf{ 66.78\%} & \textbf{66.61\%} & \textbf{53.28\% }& \textbf{53.28\%} & \textbf{88.24\%} & \textbf{89.45\% }\\
 & 0.75 & 61.32\% & 60.29\% & 60.81\% & 46.99\% & 45.63\% & 88.21\% & 88.83\% \\
 & 1 & 62.43\% & 60.96\% & 61.70\% & 48.36\% & 46.72\% & 88.06\% & 87.69\% \\ \midrule
B5 & 0 & 66.12\% & 56.03\% & 61.08\% & 55.19\% & 41.26\% & 82.45\% & 87.28\% \\
 & 0.25 & 64.34\% & 54.58\% & 59.46\% & 50.27\% & 39.07\% & 89.32\% & 90.51\% \\
 & \textbf{0.5} & \textbf{67.11\%} & \textbf{66.10\%} & \textbf{66.60\%} & \textbf{55.46\%} & \textbf{53.01\%} & \textbf{84.94\%} & \textbf{87.78\%} \\
 & 0.75 & 63.59\% & 61.28\% & 62.44\% & 48.91\% & 45.63\% & 90.86\% & 93.30\% \\
 & 1 & 58.80\% & 56.12\% & 57.46\% & 44.26\% & 40.71\% & 87.57\% & 90.30\% \\ \bottomrule
 * EfficientNet
\end{tabular}%
}
\end{table}

\textit{Ruler augmentation} was applied in one step altogether with hair augmentation. Surprisingly, almost all models had better performance when training with frame augmentation ($p > 0.0$). However, in contrary to other augmentations, rule augmentation show relatively low difference between $F_1^{aug}$ and $F_1^{org}$. The difference, though, seemed to be rising together with increasing models' size. In the case of EffiecientNet-B2, the difference was around two percentage points, whereas in B3 5 pp., B4 even more (six pp.), and B5 ended up with the highest values of 10 pp. 

After training with TDA, the differences lowered significantly, in many cases almost to zero. 
Additionally, the $F_1^{mean}$ was the highest for TDA, in all cases, again. The results are presented in Table~\ref{tab:ruler-aug-performance}.

\subsection{Counterfactual bias evaluation}

For more comprehensive evaluation of models' behaviour towards inserted biases the counterfactual bias evaluation was performed. I used three general metrics to evaluate the model's performance: number of switched predictions $s$, $mean^{change}$ and $median^{change}$.
With number of switched predictions $s$ metric, I evaluated how many instances changed predicted class after adding the bias to data. Additionally, $mean^{change}$ and $median^{change}$ was used, which mean measuring the difference in models' output after a bias insertion. Higher rates, mean a higher risk of giving predictions based on wrong premises. Such unwanted cases call for the need for targeted data augmentation.
The details behind those metrics are described in Chapter~\ref{chapter.gebi}, which describes GEBI. 
\begin{table}[]
\centering
\caption{Counterfactual bias insertion experiments - frame augmentation.}
\label{tab:frame-aug-bias}
\begin{tabular}{lllllll}
\hline
\textbf{model*} & \textbf{p} & \textbf{mean} & \textbf{median} & \textbf{$s_{sum}$} &  \textbf{$s_{ben \rightarrow mal}$} & \textbf{$s_{mal \rightarrow ben}$}\\
\hline
B2 & 0 & 5.43\% & 1.25\% & 241 & 19 & 222 \\
 & 0.25 & 1.07\% & 0.02\% & 54 & 34 & 20 \\
 & 0.5 & 1.13\% & 0.02\% & 50 & 31 & 19 \\
 & 0.75 & 1.07\% & 0.02\% & 48 & 31 & 17 \\
 & 1 & 1.49\% & 0.06\% & 72 & 34 & 38 \\ \hline
B3 & 0 & 4.66\% & 0.73\% & 208 & 38 & 170 \\
 & 0.25 & 0.96\% & 0.00\% & 44 & 33 & 11 \\
 & 0.5 & 0.92\% & 0.01\% & 48 & 34 & 14 \\
 & 0.75 & 1.05\% & 0.02\% & 39 & 28 & 11 \\
 & 1 & 1.44\% & 0.04\% & 59 & 25 & 34 \\ \hline
B4 & 0 & 4.16\% & 0.62\% & 138 & 17 & 121 \\
 & 0.25 & 1.24\% & 0.10\% & 51 & 34 & 17 \\
 & 0.5 & 0.86\% & 0.03\% & 47 & 33 & 14 \\
 & 0.75 & 1.34\% & 0.07\% & 53 & 37 & 16 \\
 & 1 & 1.63\% & 0.05\% & 77 & 50 & 27 \\ \hline
B5 & 0 & 4.16\% & 0.62\% & 138 & 17 & 121 \\
 & 0.25 & 1.17\% & 0.02\% & 85 & 51 & 34 \\
 & 0.5 & 1.38\% & 0.04\% & 62 & 36 & 26 \\
 & 0.75 & 1.49\% & 0.07\% & 77 & 36 & 41 \\
 & 1 & 1.35\% & 0.05\% & 71 & 27 & 44 \\ \bottomrule
 * EfficientNet
\end{tabular}%

\end{table}
Black frames insertion caused predicted class change in 241 instances (EfficientNet-B2). In most cases, a skin lesion once classified as a benign, changed prediction to a malignant after frame insertion. This corresponds to artifacts statistics as black frames are more often appear in malignant skin lesions (see Section~\ref{section:statistcis}).
Each pre-trained model was tested with frame augmentation probability $p$ to measure the model's sensitivity to a certain artifact. Results show mean, median, and maximum detected changes in prediction after adding a black frame to the image. Additionally, the number of instances that changed a prediction after adding a black frame is presented. The experiment is run on the whole test set: 6180 benign images and 366 malignant. The results are in Table~\ref{tab:frame-aug-bias}.
A significant drop in the number of switched predictions $s$ predictions was observed when raining in TDA. In all cases $p=0.5$ seems to give the best results e.g. the lowest number of switched predictions $s$ and $mean$ values. The results are presented in Table~\ref{tab:frame-aug-bias}.

\begin{table}[]
\centering
\caption{Counterfactual bias insertion results - hair augmentation.}
\label{tab:hair-aug-bias}
\resizebox{0.8\textwidth}{!}{%
\begin{tabular}{@{}llllllll@{}}
\toprule
\textbf{model*} & \textbf{type} & \textbf{p} & \textbf{mean} & \textbf{median} & \textbf{$s_{sum}$} &  \textbf{$s_{ben \rightarrow mal}$} & \textbf{$s_{mal \rightarrow ben}$}\\ \midrule
B2 & short & \textbf{0} & \textbf{0.94\%} & \textbf{0.02\%} & \textbf{45} & \textbf{34} & \textbf{11} \\
 &  & 0.25 & 0.91\% & 0.02\% & 42 & 31 & 11 \\
 &  & 0.5 & 0.97\% & 0.03\% & 40 & 26 & 14 \\
 &  & 0.75 & 0.80\% & 0.03\% & 43 & 29 & 14 \\
 &  & 1 & 0.72\% & 0.03\% & 26 & 20 & 6 \\
 & medium & 0 & 1.22\% & 0.02\% & 62 & 50 & 12 \\
\textbf{} & \textbf{} & \textbf{0.25} & \textbf{1.14\%} & \textbf{0.03\%} & \textbf{64} & \textbf{38} & \textbf{26} \\
 &  & 0.5 & 1.12\% & 0.04\% & 52 & 24 & 28 \\
 &  & 0.75 & 0.97\% & 0.04\% & 37 & 19 & 18 \\
 &  & 1 & 0.96\% & 0.05\% & 35 & 14 & 21 \\
\textbf{} & dense & \textbf{0} & \textbf{1.83\%} & \textbf{0.06\%} & \textbf{136} & \textbf{117} & \textbf{19} \\
 &  & 0.25 & 2.32\% & 0.16\% & 109 & 80 & 29 \\
 &  & 0.5 & 2.44\% & 0.22\% & 102 & 61 & 41 \\
 &  & 0.75 & 2.41\% & 0.26\% & 107 & 83 & 24 \\
 &  & 1 & 1.83\% & 0.16\% & 71 & 52 & 19 \\
 \hline
B3 & short & \textbf{0} & \textbf{1.39\%} & \textbf{0.03\%} & \textbf{71} & \textbf{41} & \textbf{30} \\
 &  & 0.25 & 1.21\% & 0.03\% & 59 & 40 & 19 \\
 &  & 0.5 & 0.87\% & 0.02\% & 41 & 20 & 21 \\
 &  & 0.75 & 1.00\% & 0.03\% & 44 & 32 & 12 \\
 &  & 1 & 0.68\% & 0.01\% & 40 & 22 & 18 \\
\textbf{} & medium & \textbf{0} & \textbf{1.75\%} & \textbf{0.03\%} & \textbf{86} & \textbf{59} & \textbf{27} \\
 &  & 0.25 & 1.53\% & 0.04\% & 85 & 33 & 52 \\
 &  & 0.5 & 1.03\% & 0.03\% & 43 & 22 & 21 \\
 &  & 0.75 & 1.19\% & 0.04\% & 60 & 40 & 20 \\
 &  & 1 & 0.84\% & 0.01\% & 38 & 17 & 21 \\
\textbf{} & dense & \textbf{0} & \textbf{2.69\%} & \textbf{0.09\%} & \textbf{180} & \textbf{142} & \textbf{38} \\
 &  & 0.25 & 3.46\% & 0.30\% & 164 & 78 & 86 \\
 &  & 0.5 & 2.34\% & 0.14\% & 124 & 72 & 52 \\
 &  & 0.75 & 2.46\% & 0.16\% & 124 & 92 & 32 \\
 &  & 1 & 1.80\% & 0.06\% & 85 & 71 & 14 \\
 \hline
B4 & short & \textbf{0} & \textbf{1.39\%} & \textbf{0.03\%} & \textbf{63} & \textbf{41} & \textbf{22} \\
 &  & 0.25 & 1.35\% & 0.07\% & 57 & 39 & 18 \\
 &  & 0.5 & 0.88\% & 0.04\% & 34 & 22 & 12 \\
 &  & 0.75 & 0.70\% & 0.02\% & 40 & 25 & 15 \\
 &  & 1 & 0.67\% & 0.03\% & 30 & 14 & 16 \\
\textbf{} & medium & \textbf{0} & \textbf{1.19\%} & \textbf{0.01\%} & \textbf{86} & \textbf{63} & \textbf{23} \\
 &  & 0.25 & 1.46\% & 0.08\% & 60 & 32 & 28 \\
 &  & 0.5 & 1.05\% & 0.04\% & 43 & 21 & 22 \\
 &  & 0.75 & 0.76\% & 0.01\% & 34 & 20 & 14 \\
 &  & 1 & 0.72\% & 0.02\% & 26 & 16 & 10 \\
 & dense & 0 & 1.84\% & 0.02\% & 134 & 118 & 16 \\
\textbf{} & \textbf{} & \textbf{0.25} & \textbf{3.14\%} & \textbf{0.27\%} & \textbf{145} & \textbf{81} & \textbf{64} \\
 &  & 0.5 & 2.22\% & 0.14\% & 109 & 68 & 41 \\
 &  & 0.75 & 1.67\% & 0.05\% & 75 & 60 & 15 \\
 &  & 1 & 1.59\% & 0.08\% & 74 & 52 & 22 \\
 \bottomrule
 * EfficientNet
\end{tabular}%
}
\end{table}
Each pretrained model was tested with hair augmentation probability $p$ to measure the model's sensitivity to hair. Results show detected changes in prediction after adding hair to the image. Additionally, the number of instances that changed prediction is presented. The experiment is run on the whole test set: 6180 benign images and 366 malignant. The results of CBI experiments for hair are in Table \ref{tab:hair-aug-bias}.
In case of performance evaluation, the largest accuracy gap was in case of dense hair. Here, the same pattern can be observed. Instances augmented with dense hair switched predicted classes few times more often than augmented with other type of hair. Again, in all cases the number of switched predictions $s$ measure dropped when trained with targeted data augmentations.

\begin{table}[]
\centering
\caption{Counterfactual bias insertion - ruler augmentation.}
\label{tab:ruler-aug-bias}
\begin{tabular}{@{}lllllll@{}}
\toprule
\textbf{model*} & \textbf{p} & \textbf{mean} & \textbf{median} & \textbf{$s_{sum}$} &  \textbf{$s_{ben \rightarrow mal}$} & \textbf{$s_{mal \rightarrow ben}$}\\ \midrule
B2 & 0 & 0.76\% & 0.02\% & 31 & 24 & 7 \\
 & 0.25 & 0.54\% & 0.01\% & 29 & 25 & 4 \\
 & 0.5 & 0.61\% & 0.01\% & 21 & 15 & 6 \\
 & 0.75 & 0.49\% & 0.01\% & 17 & 11 & 6 \\
 & 1 & 0.42\% & 0.02\% & 18 & 12 & 6 \\ \midrule
B3 & 0 & 1.37\% & 0.02\% & 78 & 67 & 11 \\
 & 0.25 & 0.70\% & 0.01\% & 31 & 21 & 10 \\
 & 0.5 & 0.47\% & 0.01\% & 18 & 13 & 5 \\
 & 0.75 & 0.58\% & 0.01\% & 30 & 24 & 6 \\
 & 1 & 0.40\% & 0.00\% & 18 & 12 & 6 \\ \midrule
B4 & 1 & 1.00\% & 0.00\% & 60 & 50 & 10 \\
 & 0.25 & 0.66\% & 0.03\% & 29 & 17 & 12 \\
 & 0.5 & 0.46\% & 0.02\% & 11 & 7 & 4 \\
 & 0.75 & 0.41\% & 0.01\% & 19 & 13 & 6 \\
 & 1 & 0.32\% & 0.01\% & 14 & 10 & 4 \\ \midrule
B5 & 0 & 1.09\% & 0.05\% & 76 & 74 & 2 \\
 & 0.25 & 0.83\% & 0.03\% & 58 & 53 & 5 \\
 & 0.5 & 0.57\% & 0.01\% & 28 & 23 & 5 \\
 & 0.75 & 0.68\% & 0.01\% & 34 & 26 & 8 \\
 & 1 & 0.47\% & 0.01\% & 26 & 23 & 3 \\ \bottomrule
\end{tabular}%
* EfficientNet
\end{table}

Each pretrained model was tested with ruler augmentation probability $p$ to measure the model's sensitivity to the artifact. Results show mean, median, and maximum detected changes in prediction after adding ruler mark to the image. Additionally, the number of instances that changed prediction is presented. The experiment is run on the whole test set: 6180 benign images and 366 malignant. The results are presented i Table \ref{tab:ruler-aug-bias}. The conclusions for ruler augmentations are similar. The bias influence lowered after training with TDA. The difference is not spectacular as in the case of hair and frame augmentation, but the ruler mark does not have such a strong influence on the prediction (see Chapter~\ref{chapter.gebi}).

Those experiments concluded that TDA successfully help in mitigating biases in data by inserting them randomly during the training. On average, the best results were achieved with augmentation probabilities between 0.25 and 0.75. Training with no augmentation gave the worst results in two out of three averaged cases, and augmenting always ($p=1.0$) was the worst only ion the ruler augmentation scenario. The summarized results with mean values calculated across all architectures are presented in Table \ref{tab:bias-summary}. For the presentation purpose, I averaged all the metrics across different architectures (EfficientNets B2-B5), for some subtypes (\textit{hair: short, medium, dense}), and and for some probabilities (averaged for probabilities $p \in \langle 0.25, 0.50, 0.75 \rangle$).

\begin{table}[]
\caption{Summary of the mean results with and without targeted data augmentation.}
\label{tab:bias-summary}
\resizebox{\textwidth}{!}{%
\begin{tabular}{rrrrrrrrr}
\hline
 type & $p$ & \textbf{$F_1^{org}$} & \textbf{$F_1^{aug}$} & \textbf{$F_1^{mean}$} & STD($F_1^{mean}$) & \textbf{$s_{sum}$} &  \textbf{$s_{ben \rightarrow mal}$}** & \textbf{$s_{mal \rightarrow ben}$}** \\ 
 \hline
frame & 0 & 64,90\% & 55,99\% & 60,45\% & 3,49\% & 181,25 & 12,55\% & 87,45\% \\
 & * & \textbf{63,89\%} & \textbf{61,89\%} & \textbf{62,89\%} & \textbf{2,28\%} & \textbf{54,83} & \textbf{63,53\%} & \textbf{36,47\%} \\
 & 1 & 62,47\% & 62,33\% & 62,40\% & 3,45\% & 69,75 & 48,75\% & 51,25\% \\
 \hline
hair & 0 & 64,82\% & 54,29\% & 59,56\% & 5,93\% & 95,89 & 77,06\% & 22,94\% \\
 & *  & \textbf{63,70\%} & \textbf{59,63\%} & \textbf{61,66\%} & \textbf{4,03\%} & \textbf{73,30} & \textbf{64,07\%} & \textbf{35,93\%} \\
 & 1 & 62,65\% & 58,40\% & 60,53\% & 4,28\% & 47,22 & 65,41\% & 34,59\% \\
 \hline
ruler & 0 & 64,56\% & 58,70\% & 61,63\% & 2,36\% & 61,25 & 87,76\% & 12,24\% \\
 & *  & \textbf{64,07\%} & \textbf{62,53\%} & \textbf{63,30\%} & \textbf{2,77\%} & \textbf{27,08} & \textbf{76,31\%} & \textbf{23,69\%} \\
 & 1 & 61,69\% & 59,82\% & 60,76\% & 4,64\% & 19,00 & 75,00\% & 25,00\% \\ 
 \hline
\end{tabular}
}
\\ 
* $p \in \langle 0.25, 0.50, 0.75 \rangle$, ** divided by $s_{sum}$
\end{table}

\section{Discussion}

The introduced \textit{Targeted Data Augmentation} method was proved to help mitigate biases by randomly inserting biases into data instances. 
Targeted data augmentation resulted in higher $F_1^{org}$ measures than without data augmentation in eighteen cases out of twenty examined. Still, the $F_1^{mean}$ value that measures the mean between how the model works on original data and data with inserted bias was always the best with TDA. Those results suggest that TDA made the model more robust against the biases in every case. 

A model insensitive to those biases should have the same $F_1^{aug}$  and $F_1^{org}$. Yet, in all cases, the $F_1^{aug}$ score was significantly lower than $F_1^{org}$, suggesting that the model was sensitive towards inserted bias in all cases. Applying the targeted data augmentation made the model more robust to such changes, even though training augmentations differed from the test ones -- I used different segmentation masks for the training and testing. The difference between $F_1^{org}$ and $F_1^{aug}$ was much smaller with augmentation probability $p > 0$, proving again that the method was working well. 

Additional examination with counterfactual bias insertion showed the desired outcome again. In every case, the number of switched instances was significantly lower with TDA. Targeted data augmentations resulted in lower mean and median prediction changes after inserting the bias.

Interestingly, the best results were achieved for frame insertion - the number of switched predictions $s_{sum}$ dropped several times (on average from 180 to 54 switched instances). Hair and ruler results depended on the type of augmentation, but on average, the best scores were achieved with $ p > 0 $. The worst scenario was observed for the visible dense hair. It can be observed that the drop in switched predictions was lower than in other scenarios yet higher than in the case of short hair. I speculated that short hair is similar to black and brown globules, which might indicate malignant skin lesions.

This study showed that a simple method of bias insertion to data can successfully make models more robust towards biases. It can be easily added to any ML pipeline as it works as any other data augmentation methods. The advantage of the method is that it is easy to use and design, and that it always get significantly higher results (with probabilities within certain range). The downside is that in some cases it might be difficult to prepare a proper data augmentation policy describing how to insert biases. Also, there the process is highly randomized which makes it difficult to control.


\clearpage  
\lhead{\emph{Chapter 8: Debiasing effect of training with attribution feedback}}  
\chapter{Debiasing effect of training with attribution feedback} \label{chapter.attribution}
\section{Introduction}
Previous chapters showed that inserting biases into the input might help ignore unwanted features. Targeted data augmentations help mitigate biases by destroying spurious correlation via random inserting. It indirectly forces the model to ignore them. However, the ultimate goal is not only to ignore certain biases but also to ensure that the model focuses on the right ones. Not many works have been published on improving the model by forcing it to focus on the \textit{good} features. The brief review on that problem was presented in the \textit{Section \ref{section.bias-mitigation}: Bias mitigation (Chapter \ref{chapter.bias_in_ML}: Bias in machine learning pipeline).}

To approach this problem, I proposed to use a \textit{training with attribution feedback} that uses a deep model with end-to-end trainable attribution. During the training, the model can be forced to focus its attribution on the correct region. The model uses \textit{classification loss} and an additional \textit{attribution loss}, which takes care of the correctness of the attribution maps. The goal of the training is to modify the model parameters so that attribution maps generated for each instance will focus on an essential part. The \textit{good features} or \textit{essential regions} should correlate with the result and, most importantly, have roots in causation. For example, the attribution should concentrate on the dog, not the grass, trees, or leash, when differentiating cats from dogs.

Here, I focused solely on forcing the model to ignore potential biases. I compared the results with Targeted Data Augmentation and used previous experiences in bias detection to select bias to mitigate. 

\section{Methodology}

The main idea is that the model should include both a classification module and an attribution module. The classification module is responsible for maintaining high classification results, whereas the attribution one is for acquiring current attribution maps and comparing them with desired ones. The attribution module is responsible for encouraging the model to use good features that have roots in causation and ignore the spurious correlations.

\begin{figure}
\centering
  \includegraphics[width=\textwidth]{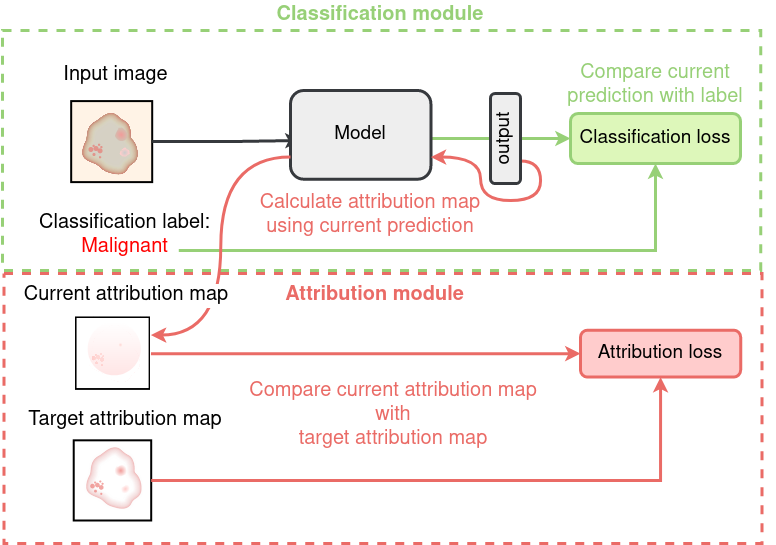}%
  \caption{Illustration of the idea of training with attribution feedback. } \label{fig.attribution.idea}
\end{figure}

The illustration of the proposed model with trainable attribution is presented in Figure~\ref{fig.attribution.idea}. The diagram shows the model with two modules: classification and trainable attribution module. After making a prediction, the attention (attribution) map is generated and compared with the desired attribution map. 

\subsection{Bias identification}

Similarly, as in \textit{Targeted Data Augmentation}, the key to successful training is the well-defined source of possible bias in data. Bias detection can be done in various ways: 
\begin{itemize}
\item by identifying bias by manual inspection of explanations,
\item by identifying bias with the help of global explanation methods (e.g., GEBI from Chapter~\ref{chapter.gebi}),
bias is identified with GEBI or another method of choice, e.g., SpRay, manual inspection with local explanations, manual inspection,
\item by utilizing expert's knowledge.
\end{itemize}

In this step, we need to define what features are important and which should be ignored. Once the bias is identified, a user might move on to the mitigation process. I selected potential biases with both statistical artifact analysis (Chapter~\ref{chapter.skin_lesion_bias}) and GEBI (Chapter~\ref{chapter.gebi}). The selected bias complies with the one selected for targeted data augmentation experiments: black frame.

\subsection{Attribution feedback}
The idea of the method is to fine-tune the model to make it ignore, enhance or adjust some parts of the input. Attribution feedback requires a current prediction, the desired classification label, and a current attribution map and desired attribution map for the weights optimization step. So, the first step is data preparation. In a standard classification, we need inputs and labels, and here similarly: we need to generate \textit{Desired Attribution Maps} first. I defined four methods of automatic or semiautomatic generation of Desired Attribution Maps: (1) \textit{forcing to ignore}, (2) \textit{forcing to remain}, (3) \textit{forcing to adjust}, and (4) \textit{forcing to focus}. During my PhD research, I experimented only with the forcing to ignore approach. The design, implementation, and analysis of the mentioned ideas' performance are beyond this dissertation's scope.

\begin{figure}[!htb]
\centering
  \includegraphics[width=0.7\textwidth]{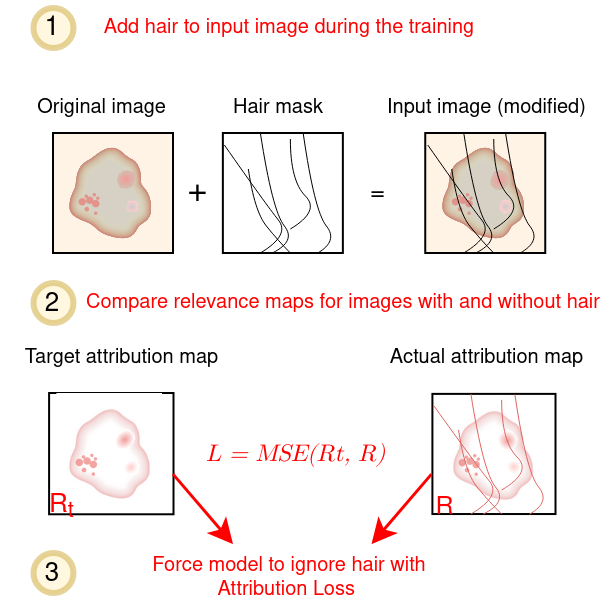}%
  \caption{\textit{Forcing to ignore} example. Illustrated example of skin classification training that forces the model to ignore hair. In ideal case the attribution map of image with and without hair should be the same.} \label{fig.attribution_loss}
\end{figure}

\textbf{Forcing to ignore.} In a skin lesions example, desired attribution maps should force the model to ignore the skin or artifacts during classification. We could use a segmentation mask of skin lesions to achieve that and apply them on the \textit{current attribution map} to hide the skin parts. Then, we would train the model to correctly predict the skin lesion class and, simultaneously, focus only on the skin lesion and ignore the skin itself. Another approach in \textit{forcing to ignore} could be using attribution feedback for bias mitigation. We could push the model to ignore the parts that correlate with the predicted class but are not necessarily a deciding feature. For example, black frames in skin lesions are strongly correlated with malignant skin lesions, but they are not causing the malignancy of skin lesions (see Chapter~\ref{chapter.skin_lesion_bias}). Hence we could force the model to ignore such misleading features. The force to ignore example is illustrated in Figure~\ref{fig.attribution_loss}.

\begin{figure}[!htb]
\centering
  \includegraphics[width=0.7\textwidth]{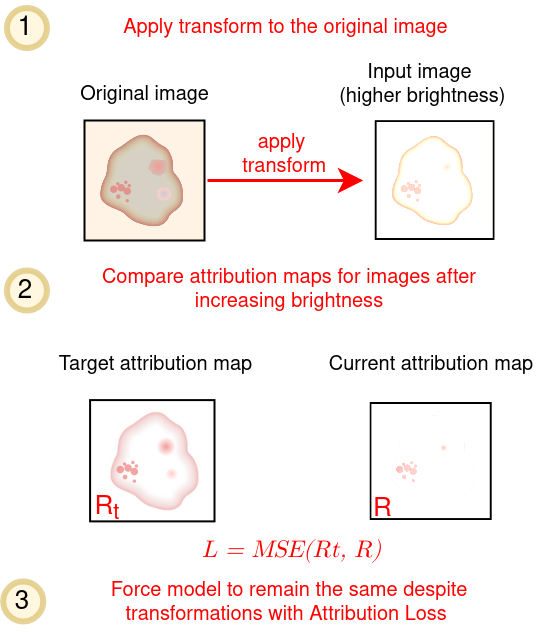}%
  \caption{\textit{Forcing to remain} example. Illustrated example of skin classification training that forces the model to generate the same attribution map after image increasing brightness.} \label{fig.force-to-remain}
\end{figure}

\textbf{Forcing to remain.} Another approach would be robustness enhancement, e.g., we could force the model to make the attribution map look the same despite light conditions. For instance, we could generate an attribution map for a particular input. Then, apply some color modification to the input image. Then, create an attribution map on a modified image and compare it with the previous one. The model could be rained to correctly predict the skin lesion class and forced to focus on the same parts of the image despite the lightning conditions. Chapter~\ref{chapter.skin_lesion_bias}). The forcing to remain example is presented in Figure~\ref{fig.force-to-remain}.

\begin{figure}[!htb]
\centering
  \includegraphics[width=0.7\textwidth]{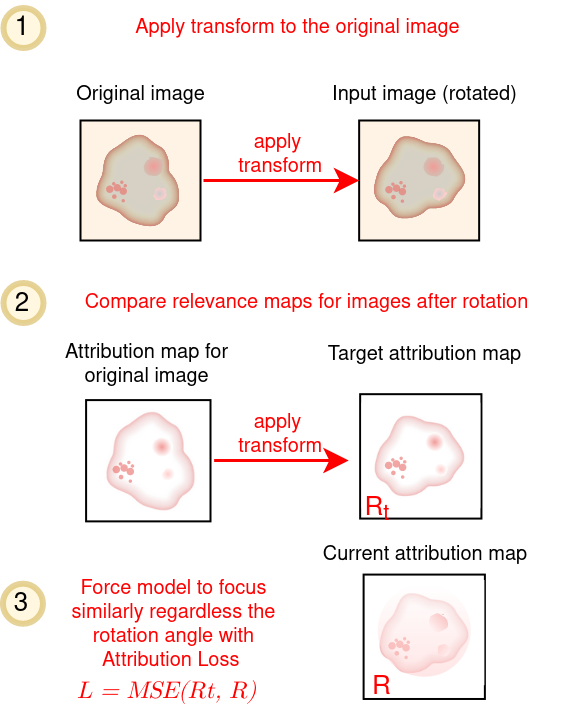}%
  \caption{\textit{Forcing to adjust} example. Illustrated example of skin classification training that forces the model to react the same after image rotation.} \label{fig.force-to-adjust}
\end{figure}

\textbf{Forcing to adjust.} Similarly, we could apply linear modifications to the input (e.g., shear, rotation, reflection). We would expect that the attribution map of the modified image would look the same as the map before the modification but be transformed in the same way. Hence here, we could take the original attribution map, change it with the same linear transformations, and compare it with the attribution map generated from modified input. Hence we could force the model to ignore such misleading features. The forcing to adjust example is illustrated in Figure~\ref{fig.force-to-adjust}.

\begin{figure}[!htb]
\centering
  \includegraphics[width=0.8\textwidth]{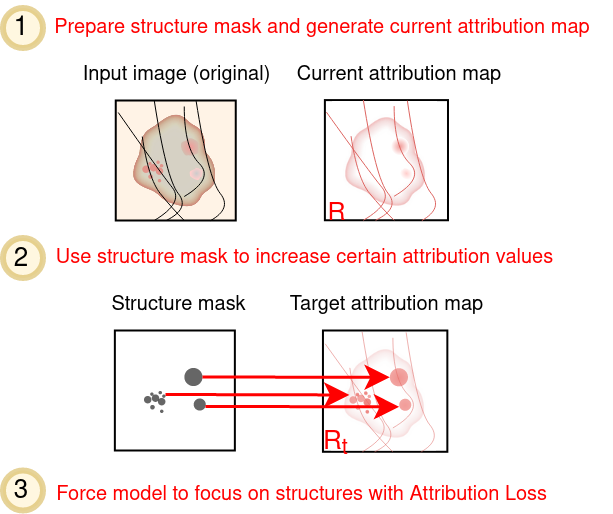}%
  \caption{\textit{Forcing to focus} example. Illustrated example of skin classification training that forces the model to focus of differential structures, that might be an early sign of melanoma.} \label{fig.force-to-focus}
\end{figure}

\textbf{Forcing to focus.} The last proposed approach includes forcing the model to focus on intended regions that we consider significant. The forcing can be done by training the model by generating desired attribution maps with enhanced parts of higher importance. For instance, the segmentation masks with lesion structures could be used to create desired attribution map by multiplying the current values of the attribution map or by adding weights to such a map. The force to focus example is illustrated in Figure~\ref{fig.force-to-focus}.

In a skin lesion classification, it is expected that attribution maps with inserted artifacts should look the same as those without them. That would prove that the model is robust against biases. However, that is not the case. As shown in the previous Chapter, the generated attribution maps often highlight various dermoscopic artifacts such as frames, gel bubbles, or hair. 

I used the training with attribution feedback to make the model ignore detected biases. First, attribution maps for original images and biased images are produced for each sample and batch. In the training and validation steps, each batch of attribution maps is compared to the desired attribution maps. I selected a saliency-based method for attribution map generation since it is relatively fast to generate and is differentiable. Other attribution map generation methods would require more computational resources. An example of forcing to ignore in a skin lesion case study would be forcing a model to ignore hair on the skin lesion. The hair could be artificially inserted into an image as in the targeted data augmentation. Here, instead of randomly adding hair during the training to \textit{indirectly} force it to make hair insignificant, I added it always, forcing the model to ignore it \textit{directly}. I generated attribution maps for an input image and the image with inserted hair. The attribution map should stay the same as before insertion; if it changes, then the weights need to be updated. The loss function used for comparing attribution maps is called an Attribution Loss and is introduced in the next section.

\subsection{Loss function}

Training with attribution feedback uses attribution and classification loss. Attribution loss forces the model to focus on or ignore some parts of the input — classification loss takes care of prediction accuracy. 

First, the Classification Loss needs to be selected. It will be used alone for the pretraining and later with the attribution loss for fine-tuning. It is responsible for classical prediction error minimization. For that purpose, basically, any classification loss will be acceptable. I used traditional cross-entropy loss for the classification task~(\ref{equation.loss-cls}).

\begin{equation} \label{equation.loss-cls}
    {L_{cls}} = -\frac{1}{n}\sum_{i=1}^n\sum_{c=1}^{m} y_{i,c}\, \log (p_{i,c})
\end{equation}
where:
\begin{conditions}
L_{cls} & Cross-Entropy loss function, \\
n & number of instances, \\
m & number of classes, \\
log & the natural log, \\
y & binary indicator if class label $c$ is the correct for observation $i$, \\
p & predicted probability observation $i$ is of class $c$. \\
\end{conditions}

The main job of attribution loss is to encourage the model to focus on certain regions/parts of an input. This is done by comparing the current attribution map with desired attribution map. The comparison can be made in different ways using, for instance, losses commonly adapted for embeddings comparison (i.e., cosine loss, Hinge loss, MSE loss). In those experiments, a mean squared difference between the attribution map and desired attribution map was used for the attribution loss. In that case, the desired attribution map is the heatmap generated before any modifications to the input (e.g., before bias insertion). The Attribution Loss is presented below~(\ref{equation.loss-atr}).

\begin{equation} \label{equation.loss-atr}
    {L_{atr}}=\frac{1}{n}\sum_{i=1}^n(R_i-\hat{R_i})^2.
\end{equation}
where:
\begin{conditions}
L_{atr} & Mean Square Error loss function, \\
R_i     &  Attribution map for unmodified image $I_i$, \\
\hat{R_i}     &  Attribution map for modified image $\hat{I_i}$.\\   
\end{conditions}

However, using only an attribution loss for fine-tuning might make the model ignore the biases (e.g., hair) but also lose its knowledge towards differentiating between classes (e.g., malignant and benign skin lesions). Hence, the weighted combination of Classification Loss and Attribution Loss is minimized during the training to minimize classification error and make the model focus on a proper region (i.e., ignore the black frame and skin around the lesion). The final loss function should weight $L_{atr}$ with $L_{cls}$.

The loss used during the attribution-feedback fine-tuning is a weighted sum of classification loss and cross-entropy loss. The definition is presented below~(\ref{equation.loss-total}).

\begin{equation} \label{equation.loss-total}
    {Loss} = (1-\alpha) L_{cls} + \alpha L_{atr} = \frac{1}{2n}\sum_{i=1}^n
    \left( - (1-\alpha) \sum_{c=1}^{m} y_{i,c}\, \log (p_{i,c}) + \alpha (R_i-\hat{R_i})^2 \right)
\end{equation}
where:
\begin{conditions}
L_{cls} & Classification loss function, \\
L_{atr} & Attribution loss function, \\
\alpha & Weight parameter. Larger values than 0.5 will encourage $L_{atr}$ to be more important. \\
\end{conditions}

\section{Experiments}
To measure the debiasing effect of training with attribution feedback, it is planned to test the \textit{forcing to ignore} approach. During the training, the model will be trained in classification and attribution losses to ignore black frames, the type of artifact that was proven to be the most biasing. The details behind the methodology are described in previous sections. The experiments will be evaluated with performance and counterfactual bias insertion metrics, similarly to the targeted augmentations.

\subsection{Data and training details}
For the following experiments, the same data from ISIC Archive challenges 2019 \cite{tschandl2018ham10000, codella2018skin, combalia2019bcn20000} and 2020 \cite{isic2020} was used, as in the Targeted Data Augmentation experiments. Hence, as for the training set, I used 47 171 benign skin lesions and 4 753 malignant. For the test set: 6 180 benign and 366 malignant. Additionally, for the bias insertion examination, I used personally prepared frame images. I trained all models for ten epochs. Adam optimizer with $lr = 5e-4$ was used with a step scheduler multiplying $lr$ every epoch by 0.9. For EfficientNet-B2 I used batch size 20, EfficientNet-B3, batch size=12, and for EfficientNet-B4 used batch size $b=6$. 
Then, I fine-tuned models with Attribution feedback for 2 epochs. I used $lr = 5e-5$ with step scheduler multiplying $lr$ every epoch by 0.9 with the same batch sizes as during the pretraining. The data augmentation pipeline was inspired by the winning model of ISIC challenge 2020 \cite{ha2020identifying}. For every training I used the following image transformations: random horizontal ($p=0.5$) and vertical flip ($p=0.5$), frame bias insertion ($p=1.0$), random brightness ($p=0.65$), random contrast ($p=0.65$), random hue/saturation ($p=0.5$), random shift/scale/rotate ($p=0.85$), resize, cutout ($p=0.1$) and normalization.

\subsection{Performance evaluation}

I use three general metric to evaluate the model's performance: $F_1$ score, $precision$ and $recall$. Each score is calculated for standard training and training with attribution feedback. Similarly, as in the Targeted Data Augmentations (Chapter~\ref{chapter.targeted}) I presented metrics calculated on the original data and data with inserted bias. Metrics calculated on original data have $org$ notation and biased data with $bias$.

Moreover, $recall$, $precision$, and $F_1$ measures are used to evaluate how well the algorithm performs on the original, unmodified data. 
However, the goal of Targeted Data Augmentation was not to achieve the best performance possible but to be more robust to certain biases. Hence, the $recall^{bias}$, $precision^{bias}$,$F_1^{bias}$ is used to measure the performance on data with inserted bias. Ideally, $F_1^{bias}$ should be the same as $F_1^{org}$, meaning that adding a bias to data does not change its performance. Higher differences between $F_1^{bias}$ and $F_1^{org}$ mean higher vulnerability to inserted bias. The $F_1^{mean}$ is a mean of $F_1^{org}$ and $F_1^{bias}$ hence it shows how well the model performs on both original and modified data. 

Regarding the attribution feedback's performance, I observed that, on average, a $F_1^{mean}$ score slightly dropped. The $F_1^{org}$ score was lowered when fine-tuning with feedback attribution from around one to three percentage points. The $F_1^{bias}$ score remained at a similar level: it dropped by around 0.5 pp for EfficientNet-B2, rose by 1.4 pp for EfficientNet-B3, and dropped again by 0.74 pp. for EfficientNet-B4. Such changes in F-measure come from the higher precision and lower recall in the case of attribution feedback. A model with higher precision scores but lower recall returns very few results, but most predictions are correct. It is, however, more likely to have false negatives. 

The performance results are presented in the Table~\ref{tab:attribution-performance}.  $F_1^{org}$, $recall^{org}$, and $precision^{org}$ are calculated on images without added frames, whereas $F_1^{bias}$, $recall^{bias}$, and $precision^{bias}$ for the images with frames.
\begin{table}[]
\centering
\caption{Comparison of different models trained with and without attribution feedback.}
\label{tab:attribution-performance}
\resizebox{\textwidth}{!}{%
\begin{tabular}{@{}lllllllll@{}}
\toprule
model* & type & \textbf{$F_1^{org}$} & \textbf{$F_1^{bias}$} & \textbf{$F_1^{mean}$} & \textbf{$recall^{org}$} & \textbf{$recall^{bias}$} & \textbf{$precision^{org}$} & \textbf{$precision^{bias}$} \\ \midrule
B2 & standard & 59.96\% & 52.99\% & 56.48\% & 46.45\% & 55.74\% & 84.58\% & 50.50\% \\
 & attribution & 56.72\% & 52.46\% & 54.59\% & 42.08\% & 43.72\% & 87.01\% & 65.57\% \\
B3 & standard & 66.15\% & 54.69\% & 60.42\% & 58.47\% & 58.20\% & 76.16\% & 51.57\% \\
 & attribution & 65.38\% & 56.01\% & 60.70\% & 55.46\% & 50.27\% & 79.61\% & 63.23\% \\
B4 & standard & 68.35\% & 61.60\% & 64.98\% & 55.46\% & 58.74\% & 89.04\% & 64.76\% \\
 & attribution & 65.63\% & 60.86\% & 63.24\% & 51.64\% & 50.55\% & 90.00\% & 76.45\% \\ \bottomrule
\end{tabular}%
}
\raggedright 
\small
* EfficientNet
\end{table}

\subsection{Counterfactual bias evaluation}

Counterfactual bias evaluation was performed to comprehensively evaluate models' behaviour towards inserted biases. I used three metrics to evaluate the model's performance: $switched$, $mean_{change}$ and $median_{change}$. Those metrics are introduced in details in the \textit{Chapter~\ref{chapter.gebi}:  Identifying bias with global explanations}.
With the $switched$ metric, I evaluated how many instances changed the predicted class after adding the bias to the data. Additionally, $mean_{change}$ and $median_{change}$ are used for measuring the difference in models' output after a bias insertion. Higher rates mean a higher risk of giving predictions based on wrong premises.

An interesting observation is a significantly lowered number of switched predictions. In case of EfficientNet-B2 $switched_{all}$ lowered from 241 to 121. After attribution retraining number of instances that switched predicted class after a frame insertion changed from $switched_{to mal} = 222$ to $switched_{to mal} = 94$. In contrast, the number of cases that switched from malignant to benign slightly rose from 19 to 27. 
EfficientNet-B3 and -B4 show generally similar behavior. The number of switched predictions lowered in both cases, but not as significantly as in the first one. The results of counterfactual bias insertion are presented in Table~\ref{tab:attribution-bias}. Each model with the inserted frame was tested to measure the model's sensitivity. Results show mean, median, and maximum detected changes in prediction after adding a black frame to the image. Additionally, the number of instances that changed a prediction after adding a black frame is presented. The experiment is run on the whole test set: 6180 benign images and 366 malignant.

\begin{table}[]
\centering
\caption{Comparison of different models trained with and without attribution feedback. }
\label{tab:attribution-bias}
\begin{tabular}{@{}lllllll@{}}
\toprule
\textbf{model} & \textbf{type} & \textbf{mean} & \textbf{median} & \textbf{$s_{sum}$} &  \textbf{$s_{ben \rightarrow mal}$} & \textbf{$s_{mal \rightarrow ben}$}\\ \midrule
EfficientNet-B2 & standard & 5.43\% & 1.25\% & 241 & 19 & 222 \\
 & attribution & 3.51\% & 0.54\% & 121 & 27 & 94 \\
EfficientNet-B3 & standard & 4.66\% & 0.73\% & 208 & 38 & 170 \\
 & attribution & 3.23\% & 0.20\% & 162 & 63 & 99 \\
EfficientNet-B4 & standard & 4.16\% & 0.62\% & 138 & 17 & 121 \\
 & attribution & 3.48\% & 1.00\% & 104 & 36 & 68 \\ \bottomrule
\end{tabular}%

\end{table}

\section{Discussion}
The attribution feedback method showed that it could be used for direct bias reduction in models. The technique resulted in reduced bias influence in all models in every case. The number of cases where prediction switched to another class after adding bias lowered significantly, although much less than in the case of Targeted Data Augmentations. The $F_1^{mean}$ scores were sometimes lower when training with attribution feedback, which leaves the field for improvement.

The advantage of this method is that it is a fine-tuning method that can work even only with one or two epochs of training, contrary to the required full-length training with TDA. Moreover, the method gives great possibilities regarding the applicability as it can be used to train the model to focus or ignore some features. The attribution task is similar to the self-supervised tasks, where one needs to design a pretext task to train.
The disadvantage is that the results are still worse than the TDA. The method required a user to select a desired attribution map generation method. This also made the attribution method much slower, as it has to generate two attribution maps for every input in a batch during the training. 
Although it helped to reduce the number of switched samples, it still showed a slight drop in accuracy. The accuracy drop is also visible in the case of biased examples. 

Those observations indicated that training with Attribution Feedback is an interesting alternative to the problem of bias mitigation but requires more research. Future experiments could include different heatmap generation methods like forcing to remain, focus or adjust. Other designs of fine-tuning pipelines and hyperparameters could be investigated. Moreover, it could be interesting to examine different approaches to attribution feedback, such as training with human knowledge feedback, e.g., forcing the model to focus on differential structures visible on the skin lesion's surface.   
\clearpage  
\lhead{\emph{Chapter 9: Summary}}  

\chapter{Summary}
In the doctoral dissertation, I comprehensively presented the current state-of-the-art research regarding bias and explainable AI and proposed novel methods for detecting bias and minimizing its impact on machine learning models. Biases are common in machine learning pipelines, and very often, they create spurious solid correlations that negatively affect the models. 

In the thesis, I hypothesized that identification and further mitigation of biases in data with explainable AI techniques, combined with proper data augmentation, increase the model's accuracy and robustness against certain tendencies. To prove it, I presented the background, proposed appropriate methods, conducted experiments, and finally analyzed the results demonstrating that they allow for bias detection and mitigation. 

First, I presented a broad literature review regarding bias in the machine learning pipeline. 
The pipeline was divided into six main stages: (1) literature review, (2) data collection, (3) data analysis, (4) model selection and training, (5) interpretation of the results, and (6) publication. With over forty described biases with examples, this is the first such extensive analysis of biases focused on deep learning. Additionally, I provided well-commented illustrations and presented numerous computer vision, natural language processing, and statistical examples. The chapter ends with a brief review of a few existing bias detection and mitigation approaches.

As the explainable AI is an essential part of the thesis, I also provided a taxonomy with an analytical commentary, and a description of local and global explainability methods. Each method was analyzed regarding the type, the efficiency, usability, differentiability, advantages, and disadvantages. Finally, I presented a literature review on XAI evaluation methods and how the humans-in-the-loop approaches are used in explainability.

Then, I proposed a methodology of manual data exploration toward bias identification. It was illustrated in the selected case study of skin lesion classification. Through an extensive manual data annotation, I explained how to discover spurious correlations of artifacts in a dataset that might introduce biases into the model. I found that some artifacts (black frames) are correlated with one of the classes (malignant skin lesions), whereas the other (ruler marks) were associated with the other class (benign). Continuing the dataset analysis, I analyzed the problem of shape and texture bias: I examined the importance of texture and shape in skin lesion classification, which required clinical knowledge. I concluded that skin lesions usually must be irregular in shape and texture to be malignant. Hence in contrast to most classification problems, the texture is equally crucial as shape. 
Moreover, I examined whether discovered biases might affect the model. I trained the classification model on images with the covered object of interest. The results proved that the drop in accuracy exists but is inadequate because all casual features were hidden. The accuracy drop suggests that the model learned spurious correlations to classify biased images based on clinically insignificant features. Such an approach to manual data exploration can also be applied to other studies, although it demands a solid scientific background and fields expert knowledge.

Next, I presented my proposition of an alternative approach to manual bias identification - GEBI. I designed, described, and conducted extensive experiments with GEBI: a semi-automated global explainability method that correctly detected possible biases in the data without manual annotation. I described and proposed new methods and metrics to assess the bias influence on the model numerically. The experiments proved that trained models are sensitive to previously discovered biases. There was a noticeable shift in prediction after inserting a particular biasing factor into the input. The method allows to screen through thousands of images in seconds semi-automatically but still requires human assistance and parameter tuning.

Following the results and conclusions of conducted research, I proposed three methods to mitigate possible biases in the second part of the thesis. I used a neural style transfer data augmentation to reduce texture and shape bias, which is prevalent in convolutional neural networks. By transferring one class's style (representing texture) to the conflicting content (representing shape) of the other, I generated a whole new dataset of images. This resulted in over 200k synthetic images with varying shapes and textures. I proved that the model trained with neural style transfer data augmentation achieved much higher results than the original training set. Additional tests that included analysis of predictions with local explainability methods showed that it increased the model's robustness and forced it to recognize shapes and textures. 

I used two alternative methods: mitigation via targeted data augmentation and mitigation via training with attribution feedback, for the problem of observer bias, sampling bias, data handling and instrument bias. In the targeted data augmentations, I randomly inserted biases into the model to modify correlations between biases and classes. I designed, proposed, and verified with experiments the whole targeted data augmentation pipeline. As the feature removal is challenging, I proposed inserting the possibly biasing features instead. In the form of data augmentation, random artifact insertion successfully affected the distribution of artifacts in data and classes, removing unwanted correlations. This approach to augmentation significantly lowered the influence of certain features and resulted in more robust models. The results showed that targeted data augmentations could reduce the effect of bias several times in comparison to standard training. Moreover, the method is model-agnostic and can be easily used in any machine learning pipeline. However, to successfully use it, the user must first identify bias and prepare a bias insertion algorithm, which can be difficult.

Finally, I proposed, conducted experiments, and commented results of novel training with attribution feedback. With attribution feedback, I forced and guided the model to make a prediction ignoring the undesirable features. The model was trained by comparing current and target attribution maps generated with explainability methods. A training procedure simultaneously minimized classification loss and attribution loss, leading to the generation of less-biased heatmaps and thus predictions. The major drawback is the attribution map generation method, which is noisy and generally insensitive to model parameter changes. Changing or improving might lead to much better performance and results.

Within the dissertation, I comprehensively approached the bias in data and models problem and provided the suitable tools to mitigate it.
The proposed approaches and methods are generic, but their effectiveness has been analyzed on the bases of the skin lesion classification case study. Moreover, the used example of the dermoscopic dataset allowed me to discover, explore and understand better biases and weaknesses in those datasets.

The key contributions of the dissertation include:
\begin{itemize}
    \item comprehensive, synthesized overview of the current state of knowledge regarding bias in machine learning pipeline, as well as methods of explainable artificial intelligence with analytical commentary,
    \item manual exploration, description, and analysis of open skin lesions datasets. Manual annotation of skin lesion dataset and sharing it, so it can serve as a benchmark for future studies,
    \item the proposition of method for manual identification of spurious correlations in datasets, carrying out extensive experiments, and elaborate analysis of the results,
    \item the proposition of \textit{GEBI}, which is a semi-automated global explainability method for bias detection, carrying out extensive experiments, and elaborate analysis of the results,
    \item the proposition of \textit{Style Transfer Data Augmentation} to increase robustness and mitigate shape and texture bias, carrying out extensive experiments, and elaborate analysis of the results,
    \item the proposition of \textit{Targeted Data Augmentations} which significantly reduces the bias influence and increases the model's performance and robustness, carrying out extensive experiments, and elaborate analysis of the results,
    \item the proposition of \textit{Training with Attribution Feedback} method, including novel \textit{Attribution Loss} function that can be used to mitigate biases, carrying out extensive experiments, and elaborate analysis of the results.
\end{itemize}


The above-indicated research has allowed me to conclude that using explainable AI and data augmentation for bias identification and mitigation increases the model's accuracy and robustness, which in my opinion, proves the truth of the thesis formulated in the dissertation.  
\clearpage  
\lhead{\emph{Appendix A: Neural Style Transfer methods
for data augmentation}}  
\addtocontents{toc}{\vspace{2em}} 

\appendix 

\chapter{Neural Style Transfer methods for data augmentation} \label{appendix.a}

Current NST algorithms are approaching the transfer of the style either by iterative optimization of an image to fit the content and style of chosen images (Image-Optimization-Based Online Neural Methods) or by optimizing the generative model, which can produce stylized images (Model-Optimization-Based Offline Neural Methods).  

\section{Image-Optimization-Based Online Neural Methods}

This branch of methods focuses on a fundamental property of deep convolutional neural networks, which says that each layer in the CNN has a non-linear filter bank, the complexity of which increases with layer position in the network. The first layers in the CNN focus on only pixel values, while more complicated features can be extracted deeper into the network. For example, in the case of dog vs. cat recognition, in one of the last layers, CNN learned to detect a dog's tongue as an essential feature for classification. 

In the first image-optimization-based NST, \textit{the neural algorithm of artistic style}, Gatys et al. \cite{gatys2015neural}  used the abovementioned property of CNNs to extract different levels of features at each convolutional layer to disentangle from an image its style and content. They noticed that the representation of features in the first layers of the CNN could be described as the content of the image, while the last layers' representations, with more advanced features, can be understood as a style. According to that observation, it is possible to extract the style or content from a given image and apply a new style. To achieve that, they optimized an image with the loss function equal to the difference between the current style feature representation and the goal style feature representation (representation from the style image). 

The details of this algorithm are as follows. Prepare any pretrained convolutional neural network without fully connected layers and select images for the style and the content. Given the content image, $I_{c}$ and the style image $I_{s}$, the algorithm in \cite{gatys2015neural} optimizes the base image $I$ to minimize the following loss function
$L_{total}$ (\ref{eq.gatys}).

\begin{equation} \label{eq.gatys}
 L_{total},I_{c},I_{s},I= \alpha L_{c},I_{c},I+ \beta L_{s},I_{s},I
\end{equation}

where $\alpha$ and $\beta$ are the factors that weight the content loss function $L_{c}$ with the style loss function $L_{s}$ for the eye-pleasant final result $I$. $L_{c}$ is defined as the mean square error between the content representation $F_{l}(I_{c})$ of the given content image and the content representation of the base image $F_{l}(I)$ (\ref{eq.alpha}). 

\begin{equation} \label{eq.alpha}
 L_{c},I_{c},I= \frac{1}{2}\sum_{i,j}^{} F_{i,j}^{l}I_{c} - F_{i,j}^{l}I^{2}
\end{equation}

where $F_{i,j}^{l}$ is the activation of the $i$-th filter at the position $j$ in the layer $l$. The style loss $L_{s}$, which is equal to the mean-squared distance between the entries of the Gram matrix from the style image $G(F^{l}(I_{s}))$
and the Gram matrix from the base image $GF^{l}I$, is minimized (\ref{eq.gram}).

\begin{equation} \label{eq.gram}
 L_{s},I_{s},I= \sum_{i,j}^{} GF_{i,j}^{l} I_{c} - GF_{i,j}^{l} I^{2} 
\end{equation}

The Gram matrix is calculated as an inner product between the vectorized feature maps $i$ and $j$ in the layer $l$ (\ref{eq.gram2}). 

\begin{equation} \label{eq.gram2}
 G_{i,j}^{l} = \sum_{k}^{} F_{i,k}^{l} F_{j,k}
\end{equation}

After calculating the total loss function, the base image $I$ is updated with the selected optimization algorithm. The process of calculating the loss function $L_{total}$ and image updates are repeated until the style and content of the base image match the selected images or until the chosen iteration number is met. 

\begin{figure}
\centering
  \includegraphics[width=0.6\textwidth]{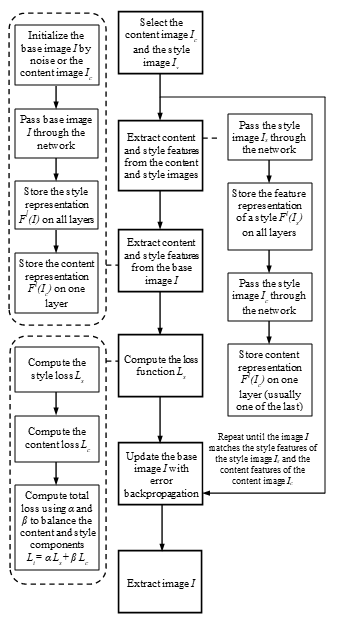}%
  \caption{Neural Algorithm of artistic style – flowchart } \label{fig.style-transfer}
\end{figure}

Since the method's first publication, numerous algorithm improvements were published. The flowchart of the process is shown in Figure \ref{fig.style-transfer}. 

\section{Model-Optimization-Based Offline Neural Methods}

This family of methods is an answer to the efficiency issue in the \textit{Image-Optimization-Based Online Neural Methods}. Instead of optimizing the base image (i.e., the image we want to stylize), the model that can generate stylized images is optimized. This section briefly introduces three types of Model-Optimization-Based Offline Neural Methods, categorized by the number of styles that a single model can produce: \textit{One-Style-Per-Model}, \textit{Multiple-Style-Per-Model}, and \textit{Arbitrary-Style-Per-Model}. 

\subsection{One-Style-Per-Model}

The general approach in One-Style-Per-Model methods is to optimize a model to transfer a style, using the use of two components: Image Transformation Network $f_w$ and Loss Network $\phi$ \cite{johnson2016perceptual}. At first, the deep fully-convolutional Transform Network transforms the base (input) image $I$ into the output image $I_{out}$. Given the input image $I$ and the target image $I_{out}$, the algorithm optimizes the model to minimize $L_{total}$, being the following combination of weights $\delta_i$ with loss function $L_i$: 

\begin{equation}
 L_{total} I^{out},I = \sum_{i,j} \lambda_i L_i f_w I, I^{out}
\end{equation}

where each loss function $L_i$ measures the difference between the target output image $I_{out}$ and the actual output image $f_wI$.  

Then the output image from the Image Transformation Network $f_wI$ becomes the input for the Loss Network, along with two other inputs: the target content Image $I_c$ and the target style image $I_s$. The Loss Network $\phi$ defines the content reconstruction loss $L_c$ and several style reconstruction losses $L_s$, which, similarly to the loss functions described in Image-Optimization-Based Online Neural Methods, measure the differences between the style and content. However, contrary to the approach presented in Gatys et al. \cite{gatys2015neural}, the Content Loss Function (also called the Feature Reconstruction Loss) encourages pixels to have similar feature representations rather than exactly match them. The loss function is defined as the mean square error between content (feature) representations: 

\begin{equation}
 L_{c}^{\phi,j} I_{c},I_{\phi}^{out} = \frac{1}{C_j H_j W_j} F^l I_c - F_l I_\phi^{out}
\end{equation}
 
where the output image $I_\phi^{out}$ equals the output of the Image Transformation Network $f_wI$, and $F^l(I_\phi^{out})$ are the activations in layer $l$ (content feature representation) of the network $\phi$ with the shape of $Cj \times H_j \times W_j$. 

The style reconstruction loss is defined in the same manner as in Gatys et al. with the Gram Matrix.

\begin{equation}
 G_{j}^{\phi} I_{\phi}^{out} = \frac{1}{C_j H_j W_j} \sum_{h=1}^{H_j} \sum_{w=1}^{W_j} F^l {I_{\phi}^{out}}_{h,w,c,\phi_j} F^l {I_{\phi}^{out}}_{h,w,c'}
\end{equation}

where, like in the previous equation,
$F^l(I_{\phi}^{out})$ are the activations of the $l$
layer of the network $\phi$ (with the shape of
of $Cj \times H_j \times W_j$). Given $C_j$ - the dimensional features, the Gram Matrix $G^\phi_j(I_{\phi}^{out})$ calculates scalar values for each point on the $H_j \times W_j$ grid, where each grid location is an independent sample. The value of the Gram Matrix is minimized during the training. 

Johnson et al. \cite{johnson2016perceptual} picked a few styles and trained corresponding models to be able to transfer them. Experiments showed great potential in the time needed to generate a stylized image. A single image achieves an impressive speed of two magnitudes faster than the approach by Gatys et al. \cite{gatys2015neural} However, their calculation does not include the time needed to train the style network, which is very time-consuming.

\subsection{Multiple-Style-Per-Model}

The One-style-per-model method can achieve high performance in producing a single stylized image. However, the time necessary to train the model for each style makes it inefficient for many types of data augmentation. This section focuses on the Multiple-Style-Per-Model approach, which attempts to find a compromise between the time of single stylization and the time of training.  

Dumoulin et al. \cite{dumoulin2016learned} proposed a solution to train the model to produce n styles, which does not merge many types of style-networks but only tunes the network to another style with affine transformations of parameters (by scaling or shifting) after the normalization. The idea was based on the interesting observation that the weights within the convolutional layers are shared across many styles. Instead of training n-style networks, they train one network with additional tuning parameters $\gamma_s$ and $\beta_s$ for each style, where $\gamma_s$ is the scaling parameter, and $\beta_s$ is the shifting parameter. The authors call their approach conditional instance normalization (CIN) \cite{dumoulin2016learned} and define it as follows: 

\begin{equation}
CINFI_{c,s}=\gamma_s \frac{FI_c-\mu F I_c}{\sigma F I_c} + \beta_s
\end{equation}

where $F$ is the feature map, $s$ defines the style, and $\mu$ and $\sigma$ are the mean and standard deviations, respectively, between spatial axes.  

Inspired by the idea of \textit{texton} \cite{julesz1983human} (an essential element of texture), Chen at al. \cite{chen2017stylebank} proposed an analogous solution to Dumoulin et al. \cite{dumoulin2016learned} but instead of adding tuning parameters, they used properties of an auto-encoder. Like in the previous approach, this idea uses a small number of parameters to represent the style rather than training an individual network for each style. The concept is to use an auto-encoder with \textit{StyleBank}, which has convolution filters for each style. In this approach, each channel of the style bank can be understood as a filtering dictionary or a look-up table (a bank of filters is defined in the feature embedding space produced by the auto-encoder). The proposed network architecture has three corresponding modules: the image encoder module consisting of convolutional layers, the StyleBank layer with a fixed number of represented styles, and the convolutional decoder. The auto-encoder is trained with the standard identity loss equal to the mean squared error between the input image $I$ and the output image $I_{AE^{out}}$ (output from the auto-encoder): 

\begin{equation}
L_I I _{AE^{out}},I = I_{AE^{out}} - I^2
\end{equation}

The style loss and the content loss are calculated separately and are the same as in the previously described paper by Gatys et al. \cite{gatys2015neural}. 

Another interesting approach, different from the abovementioned methods, is based on using a combination of style and content, and not only the style alone, as the input. This branch of algorithms does not change its size with the growing number of styles. It is an important feature for data augmentation, as a larger number of styles often provides better results. Li et al. focused on synthesizing multiple textures (styles) in one network. They proposed to use the deep generative feed-forward network, which can synthesize diverse results for each style. Their multi-style transfer network uses an auto-encoder, but unlike in \cite{chen2017stylebank}, instead of using an additional StyleBank layer, they concatenate the output result from the encoder with their selection unit. The selection unit is responsible for transferring different styles to the selected image. As the authors explained, for each bit in the selection unit, a corresponding noise map is generated from the chosen distribution and then concatenated with the results encoded from the content. The concatenated result of features from the style and the content is then decoded to the final synthesis image. Their solution uses three other losses: the style and content losses, just like in \cite{gatys2015neural}, and the additional diversity loss proposed by them. The diversity loss measures the visual difference of the final result with the same style but initialized with a different noise. In other words, it measures variation in visual appearance between any pair of outputs $I_{1}^{out}, I_{2}^{out}, ..., I_{n}^{out}$ and randomly reordered outputs
$J_{1}^{out}, J_{2}^{out}, ... , J_{n}^{out}$ satisfying that $I_{i}^{out} \neq J_{i}^{out}$. The diversity loss is defined as follows: 

\begin{equation}
 L_{diversity} I_{out},I = \frac{1}{n} \sum_{n=1}^{N} (FI_{i}^{out} - FJ_{j}^{out})^2
\end{equation}
 
where $F(I_{i}^{out})$ and $F(J_{i}^{out})$ are feature representations of output images. The total loss function is the weighted sum of the style, content, and diversity loss functions. 

\subsection{Arbitrary-Style-Per-Model}

This category of methods allows generating all styles with only one model. One approach that helps create any style with just one model is \textit{StyleSwap} \cite{chen2016fast}. The authors proposed to develop a stylized image by replacing the content image with the best matching style counterpart, patch-by-patch. They offered to use the architecture similar to auto-encoders, with the StyleSwap layer between the encoder and the decoder. However, the proposed new layer creates a computational bottleneck and drastically increases computation time. 



\addtocontents{toc}{\vspace{2em}}  
\backmatter

\label{Bibliography}
\lhead{\emph{Bibliography}}  
\bibliographystyle{unsrt}
\bibliography{Bibliography}

\begin{thebibliography}{100}

\bibitem{mikolajczyk2018data}
Agnieszka Mikołajczyk and Michał Grochowski.
\newblock Data augmentation for improving deep learning in image classification
  problem.
\newblock In {\em 2018 International Interdisciplinary PhD Workshop (IIPhDW)},
  pages 117--122. IEEE, 2018.

\bibitem{k2014bias}
Bernard~C K.~Choi and Anita~W P.~Pak.
\newblock Bias, overview.
\newblock {\em Wiley StatsRef: Statistics Reference Online}, 2014.

\bibitem{mehrabi2021survey}
Ninareh Mehrabi, Fred Morstatter, Nripsuta Saxena, Kristina Lerman, and Aram
  Galstyan.
\newblock A survey on bias and fairness in machine learning.
\newblock {\em ACM Computing Surveys (CSUR)}, 54(6):1--35, 2021.

\bibitem{aleu2020assessing}
Flores~F. Gonzalez~Aleu, Gonzalez~R. Perez~J., and Garza-Reyes J.A.
\newblock Assessing systematic literature review bias: kaizen events in
  hospitals case study.
\newblock {\em Proceedings of the 10th International Conference on Industrial
  Engineering and Operations Management (IEOM)}, 2020.

\bibitem{dubben2005systematic}
Hans-Hermann Dubben and Hans-Peter Beck-Bornholdt.
\newblock Systematic review of publication bias in studies on publication bias.
\newblock {\em Bmj}, 331(7514):433--434, 2005.

\bibitem{gotzsche1987reference}
Peter~C G{\o}tzsche.
\newblock Reference bias in reports of drug trials.
\newblock {\em Br Med J (Clin Res Ed)}, 295(6599):654--656, 1987.

\bibitem{oswald2012confirmation}
Margit~E Oswald and Stefan Grosjean.
\newblock Confirmation bias.
\newblock In {\em Cognitive illusions}, pages 91--108. Psychology Press, 2012.

\bibitem{mellin1957work}
W~Mellin.
\newblock Work with new electronic ‘brains’ opens field for army math
  experts.
\newblock {\em The Hammond Times}, 10:66, 1957.

\bibitem{zheng2017truth}
Yudian Zheng, Guoliang Li, Yuanbing Li, Caihua Shan, and Reynold Cheng.
\newblock Truth inference in crowdsourcing: Is the problem solved?
\newblock {\em Proceedings of the VLDB Endowment}, 10(5):541--552, 2017.

\bibitem{gregory2012research}
Katherine~E Gregory and Lucy Radovinsky.
\newblock Research strategies that result in optimal data collection from the
  patient medical record.
\newblock {\em Applied Nursing Research}, 25(2):108--116, 2012.

\bibitem{lynch2018openlittermap}
Se{\'a}n Lynch.
\newblock Openlittermap. com--open data on plastic pollution with blockchain
  rewards (littercoin).
\newblock {\em Open Geospatial Data, Software and Standards}, 3(1):1--10, 2018.

\bibitem{mehrabi2019survey}
Ninareh Mehrabi, Fred Morstatter, Nripsuta Saxena, Kristina Lerman, and Aram
  Galstyan.
\newblock A survey on bias and fairness in machine learning.
\newblock {\em arXiv preprint arXiv:1908.09635}, 2019.

\bibitem{wolter1986some}
Kirk~M Wolter.
\newblock Some coverage error models for census data.
\newblock {\em Journal of the American Statistical Association},
  81(394):337--346, 1986.

\bibitem{mclachlan1984method}
Andrew~D McLachlan, Rodger Staden, and D~Ross Boswell.
\newblock A method for measuring the non-random bias of a codon usage table.
\newblock {\em Nucleic acids research}, 12(24):9567--9575, 1984.

\bibitem{mikolajczyk2021towards}
Agnieszka Mikołajczyk, Michał Grochowski, and Arkadiusz Kwasigroch.
\newblock Towards explainable classifiers using the counterfactual
  approach-global explanations for discovering bias in data.
\newblock {\em Journal of Artificial Intelligence and Soft Computing Research},
  11(1):51--67, 2021.

\bibitem{hamm2011skin}
Henning Hamm and Peter~H H{\"o}ger.
\newblock Skin tumors in childhood.
\newblock {\em Deutsches {\"A}rzteblatt International}, 108(20):347, 2011.

\bibitem{huff1993lie}
Darrell Huff.
\newblock {\em How to lie with statistics}.
\newblock WW Norton \& Company, 1993.

\bibitem{he2012bias}
Jia He and Fons van~de Vijver.
\newblock Bias and equivalence in cross-cultural research.
\newblock {\em Online readings in psychology and culture}, 2(2):2307--0919,
  2012.

\bibitem{nirmal2017dermatoscopy}
Balakrishnan Nirmal.
\newblock Dermatoscopy image characteristics and differences among commonly
  used standard dermatoscopes.
\newblock {\em Indian dermatology online journal}, 8(3):233, 2017.

\bibitem{olteanu2019social}
Alexandra Olteanu, Carlos Castillo, Fernando Diaz, and Emre K{\i}c{\i}man.
\newblock Social data: Biases, methodological pitfalls, and ethical boundaries.
\newblock {\em Frontiers in Big Data}, 2:13, 2019.

\bibitem{ciampaglia2018algorithmic}
Giovanni~Luca Ciampaglia, Azadeh Nematzadeh, Filippo Menczer, and Alessandro
  Flammini.
\newblock How algorithmic popularity bias hinders or promotes quality.
\newblock {\em Scientific reports}, 8(1):1--7, 2018.

\bibitem{abdollahpouri2019unfairness}
Himan Abdollahpouri, Masoud Mansoury, Robin Burke, and Bamshad Mobasher.
\newblock The unfairness of popularity bias in recommendation.
\newblock {\em arXiv preprint arXiv:1907.13286}, 2019.

\bibitem{abdollahpouri2019managing}
Himan Abdollahpouri, Robin Burke, and Bamshad Mobasher.
\newblock Managing popularity bias in recommender systems with personalized
  re-ranking.
\newblock In {\em The thirty-second international flairs conference}, 2019.

\bibitem{mahtani2018catalogue}
Kamal Mahtani, Elizabeth~A Spencer, Jon Brassey, and Carl Heneghan.
\newblock Catalogue of bias: observer bias.
\newblock {\em BMJ evidence-based medicine}, 23(1):23, 2018.

\bibitem{lamb1992biased}
Kevin Lamb.
\newblock Biased tidings: The media and the cyril burt controversy.
\newblock {\em Mankind Quarterly}, 33(2):203, 1992.

\bibitem{fletcher1991science}
Ronald Fletcher.
\newblock {\em Science, ideology, and the media: The Cyril Burt scandal}.
\newblock Transaction Publishers, 1991.

\bibitem{jensen1980bias}
Arthur~R Jensen.
\newblock Bias in mental testing.
\newblock {\em Applied Psychological Measurement}, 4(3):403--410, 1980.

\bibitem{burt1943ability}
Cyril Burt.
\newblock Ability and income.
\newblock {\em British Journal of Educational Psychology}, 13(2):83--98, 1943.

\bibitem{kiritchenko2018examining}
Svetlana Kiritchenko and Saif~M Mohammad.
\newblock Examining gender and race bias in two hundred sentiment analysis
  systems.
\newblock {\em arXiv preprint arXiv:1805.04508}, 2018.

\bibitem{koolagudi2012emotion}
Shashidhar~G Koolagudi and K~Sreenivasa Rao.
\newblock Emotion recognition from speech: a review.
\newblock {\em International journal of speech technology}, 15(2):99--117,
  2012.

\bibitem{yi2020focal}
Jiangyan Yi, Jianhua Tao, Zhengkun Tian, Ye~Bai, and Cunhang Fan.
\newblock Focal loss for punctuation prediction.
\newblock {\em Proc. Interspeech 2020}, pages 721--725, 2020.

\bibitem{moro2017prosody}
Anna Mor{\'o} and Gy{\"o}rgy Szasz{\'a}k.
\newblock A prosody inspired rnn approach for punctuation of machine produced
  speech transcripts to improve human readability.
\newblock In {\em 2017 8th IEEE International Conference on Cognitive
  Infocommunications (CogInfoCom)}, pages 000219--000224. IEEE, 2017.

\bibitem{bohavc2017text}
Marek Boh{\'a}{\v{c}}, Michal Rott, and Vojt{\v{e}}ch Kov{\'a}{\v{r}}.
\newblock Text punctuation: an inter-annotator agreement study.
\newblock In {\em International Conference on Text, Speech, and Dialogue},
  pages 120--128. Springer, 2017.

\bibitem{hellstrom2020bias}
Thomas Hellstr{\"o}m, Virginia Dignum, and Suna Bensch.
\newblock Bias in machine learning--what is it good for?
\newblock {\em arXiv preprint arXiv:2004.00686}, 2020.

\bibitem{artstein2017inter}
Ron Artstein.
\newblock Inter-annotator agreement.
\newblock In {\em Handbook of linguistic annotation}, pages 297--313. Springer,
  2017.

\bibitem{kaneko2022gender}
Masahiro Kaneko, Aizhan Imankulova, Danushka Bollegala, and Naoaki Okazaki.
\newblock Gender bias in masked language models for multiple languages.
\newblock {\em arXiv preprint arXiv:2205.00551}, 2022.

\bibitem{nangia-etal-2020-crows}
Nikita Nangia, Clara Vania, Rasika Bhalerao, and Samuel~R. Bowman.
\newblock {C}row{S}-pairs: A challenge dataset for measuring social biases in
  masked language models.
\newblock In {\em Proceedings of the 2020 Conference on Empirical Methods in
  Natural Language Processing (EMNLP)}, pages 1953--1967, Online, November
  2020. Association for Computational Linguistics.

\bibitem{goddard2012automation}
Kate Goddard, Abdul Roudsari, and Jeremy~C Wyatt.
\newblock Automation bias: a systematic review of frequency, effect mediators,
  and mitigators.
\newblock {\em Journal of the American Medical Informatics Association},
  19(1):121--127, 2012.

\bibitem{yanasak2014mr}
Nathan~E Yanasak and Michael~J Kelly.
\newblock Mr imaging artifacts and parallel imaging techniques with calibration
  scanning: a new twist on old problems.
\newblock {\em Radiographics}, 34(2):532--548, 2014.

\bibitem{mccroskey1966ethos}
James~C. McCroskey and Robert~E. Dunham.
\newblock Ethos: A confounding element in communication research.
\newblock {\em Speech Monographs}, 33(4):456--463, 1966.

\bibitem{axelson1978aspects}
Olav Axelson.
\newblock Aspects on confounding in occupational health epidemiology.
\newblock {\em Scandinavian journal of work, environment \& health}, pages
  98--102, 1978.

\bibitem{greenland1980control}
Sander Greenland and Raymond Neutra.
\newblock Control of confounding in the assessment of medical technology.
\newblock {\em International journal of epidemiology}, 9(4):361--367, 1980.

\bibitem{miettinen1974confounding}
Olli Miettinen.
\newblock Confounding and effect-modification.
\newblock {\em American Journal of Epidemiology}, 100(5):350--353, 1974.

\bibitem{pearl2009causal}
Judea Pearl et~al.
\newblock Causal inference in statistics: An overview.
\newblock {\em Statistics surveys}, 3:96--146, 2009.

\bibitem{vanderweele2013definition}
Tyler~J VanderWeele and Ilya Shpitser.
\newblock On the definition of a confounder.
\newblock {\em Annals of statistics}, 41(1):196, 2013.

\bibitem{tulchinsky2014measuring}
Theodore~H Tulchinsky and Elena~A Varavikova.
\newblock Measuring, monitoring, and evaluating the health of a population.
\newblock {\em The New Public Health}, page~91, 2014.

\bibitem{bidel2013emerging}
Siamak Bidel and Jaakko Tuomilehto.
\newblock The emerging health benefits of coffee with an emphasis on type 2
  diabetes and cardiovascular disease.
\newblock {\em European endocrinology}, 9(2):99, 2013.

\bibitem{mcnamee2003confounding}
Roseanne McNamee.
\newblock Confounding and confounders.
\newblock {\em Occupational and environmental medicine}, 60(3):227--234, 2003.

\bibitem{althubaiti2016information}
Alaa Althubaiti.
\newblock Information bias in health research: definition, pitfalls, and
  adjustment methods.
\newblock {\em Journal of multidisciplinary healthcare}, 9:211, 2016.

\bibitem{pourhoseingholi2012control}
Mohamad~Amin Pourhoseingholi, Ahmad~Reza Baghestani, and Mohsen Vahedi.
\newblock How to control confounding effects by statistical analysis.
\newblock {\em Gastroenterology and hepatology from bed to bench}, 5(2):79,
  2012.

\bibitem{alexander2015confounding}
Lorraine~K Alexander, Brettania Lopes, Kristen Ricchetti-Masterson, and Karin~B
  Yeatts.
\newblock Confounding bias.
\newblock {\em UNC CH Department of Epidemiology}, pages 1--5, 2015.

\bibitem{landeiro2017controlling}
Virgile Landeiro and Aron Culotta.
\newblock Controlling for unobserved confounds in classification using
  correlational constraints.
\newblock In {\em Proceedings of the International AAAI Conference on Web and
  Social Media}, volume~11, 2017.

\bibitem{10.2307/3703850}
Sander Greenland.
\newblock Quantifying biases in causal models: Classical confounding vs
  collider-stratification bias.
\newblock {\em Epidemiology}, 14(3):300--306, 2003.

\bibitem{10.1093/aje/kwi002}
Yu-Kang Tu, Robert West, George T.~H. Ellison, and Mark~S. Gilthorpe.
\newblock {Why Evidence for the Fetal Origins of Adult Disease Might Be a
  Statistical Artifact: The “Reversal Paradox” for the Relation between
  Birth Weight and Blood Pressure in Later Life}.
\newblock {\em American Journal of Epidemiology}, 161(1):27--32, 01 2005.

\bibitem{day2016robust}
Felix~R Day, Po-Ru Loh, Robert~A Scott, Ken~K Ong, and John~RB Perry.
\newblock A robust example of collider bias in a genetic association study.
\newblock {\em The American Journal of Human Genetics}, 98(2):392--393, 2016.

\bibitem{viallon2016can}
Vivian Viallon and Marine Dufournet.
\newblock Can collider bias fully explain the obesity paradox?
\newblock {\em arXiv preprint arXiv:1612.06547}, 2016.

\bibitem{messick1981reversal}
David~M Messick and John~P Van~de Geer.
\newblock A reversal paradox.
\newblock {\em Psychological Bulletin}, 90(3):582, 1981.

\bibitem{tu2008simpson}
Yu-Kang Tu, David Gunnell, and Mark~S Gilthorpe.
\newblock Simpson's paradox, lord's paradox, and suppression effects are the
  same phenomenon--the reversal paradox.
\newblock {\em Emerging themes in epidemiology}, 5(1):1--9, 2008.

\bibitem{baeza2018bias}
Ricardo Baeza-Yates.
\newblock Bias on the web.
\newblock {\em Communications of the ACM}, 61(6):54--61, 2018.

\bibitem{panch2019artificial}
Trishan Panch, Heather Mattie, and Rifat Atun.
\newblock Artificial intelligence and algorithmic bias: implications for health
  systems.
\newblock {\em Journal of global health}, 9(2), 2019.

\bibitem{nisan2001algorithmic}
Noam Nisan and Amir Ronen.
\newblock Algorithmic mechanism design.
\newblock {\em Games and Economic behavior}, 35(1-2):166--196, 2001.

\bibitem{lum2016statistical}
Kristian Lum and James Johndrow.
\newblock A statistical framework for fair predictive algorithms.
\newblock {\em arXiv preprint arXiv:1610.08077}, 2016.

\bibitem{edelman2017racial}
Benjamin Edelman, Michael Luca, and Dan Svirsky.
\newblock Racial discrimination in the sharing economy: Evidence from a field
  experiment.
\newblock {\em American Economic Journal: Applied Economics}, 9(2):1--22, 2017.

\bibitem{lloyd2018bias}
Kirsten Lloyd.
\newblock Bias amplification in artificial intelligence systems.
\newblock {\em arXiv preprint arXiv:1809.07842}, 2018.

\bibitem{bolukbasi2016man}
Tolga Bolukbasi, Kai-Wei Chang, James~Y Zou, Venkatesh Saligrama, and Adam~T
  Kalai.
\newblock Man is to computer programmer as woman is to homemaker? debiasing
  word embeddings.
\newblock {\em Advances in neural information processing systems},
  29:4349--4357, 2016.

\bibitem{mayson2018bias}
Sandra~G Mayson.
\newblock Bias in, bias out.
\newblock {\em The Yale Law Journal}, 128:2218, 2018.

\bibitem{zhao2017men}
Jieyu Zhao, Tianlu Wang, Mark Yatskar, Vicente Ordonez, and Kai-Wei Chang.
\newblock Men also like shopping: Reducing gender bias amplification using
  corpus-level constraints.
\newblock {\em arXiv preprint arXiv:1707.09457}, 2017.

\bibitem{cowgill2019economics}
Bo~Cowgill and Catherine~E Tucker.
\newblock Economics, fairness and algorithmic bias.
\newblock {\em preparation for: Journal of Economic Perspectives}, 2019.

\bibitem{wong2019democratizing}
Pak-Hang Wong.
\newblock Democratizing algorithmic fairness.
\newblock {\em Philosophy \& Technology}, pages 1--20, 2019.

\bibitem{corbett2017algorithmic}
Sam Corbett-Davies, Emma Pierson, Avi Feller, Sharad Goel, and Aziz Huq.
\newblock Algorithmic decision making and the cost of fairness.
\newblock In {\em Proceedings of the 23rd acm sigkdd international conference
  on knowledge discovery and data mining}, pages 797--806, 2017.

\bibitem{juliaangwin_2016}
Julia Angwin, Jeff Larson, Surya Mattu, and Lauren Kirchner.
\newblock Machine bias.
\newblock In {\em Ethics of Data and Analytics}, pages 254--264. Auerbach
  Publications, 2016.

\bibitem{dieterich2016compas}
William Dieterich, Christina Mendoza, and Tim Brennan.
\newblock Compas risk scales: Demonstrating accuracy equity and predictive
  parity.
\newblock {\em Northpoint Inc}, 7(7.4):1, 2016.

\bibitem{majchrowska2022deep}
Sylwia Majchrowska, Agnieszka Miko{\l}ajczyk, Maria Ferlin, Zuzanna
  Klawikowska, Marta~A Plantykow, Arkadiusz Kwasigroch, and Karol Majek.
\newblock Deep learning-based waste detection in natural and urban
  environments.
\newblock {\em Waste Management}, 138:274--284, 2022.

\bibitem{sun2019mitigating}
Tony Sun, Andrew Gaut, Shirlyn Tang, Yuxin Huang, Mai ElSherief, Jieyu Zhao,
  Diba Mirza, Elizabeth Belding, Kai-Wei Chang, and William~Yang Wang.
\newblock Mitigating gender bias in natural language processing: Literature
  review.
\newblock {\em arXiv preprint arXiv:1906.08976}, 2019.

\bibitem{mikolajczyk2022debiasing}
Agnieszka Mikołajczyk, Sylwia Majchrowska, and Sandra Carrasco~Limeros.
\newblock The (de)biasing effect of gan-based augmentation methods on skin
  lesion images.
\newblock In {\em International Conference on Medical Image Computing and
  Computer-Assisted Intervention}. Springer, 2022.

\bibitem{rahaman2019spectral}
Nasim Rahaman, Aristide Baratin, Devansh Arpit, Felix Draxler, Min Lin, Fred
  Hamprecht, Yoshua Bengio, and Aaron Courville.
\newblock On the spectral bias of neural networks.
\newblock In {\em International Conference on Machine Learning}, pages
  5301--5310. PMLR, 2019.

\bibitem{hermann2020origins}
Katherine Hermann, Ting Chen, and Simon Kornblith.
\newblock The origins and prevalence of texture bias in convolutional neural
  networks.
\newblock {\em Advances in Neural Information Processing Systems},
  33:19000--19015, 2020.

\bibitem{schwarz2021frequency}
Katja Schwarz, Yiyi Liao, and Andreas Geiger.
\newblock On the frequency bias of generative models.
\newblock {\em Advances in Neural Information Processing Systems}, 34, 2021.

\bibitem{khayatkhoei2020spatial}
Mahyar Khayatkhoei and Ahmed Elgammal.
\newblock Spatial frequency bias in convolutional generative adversarial
  networks.
\newblock {\em arXiv preprint arXiv:2010.01473}, 2020.

\bibitem{cinelli2020making}
Carlos Cinelli and Chad Hazlett.
\newblock Making sense of sensitivity: Extending omitted variable bias.
\newblock {\em Journal of the Royal Statistical Society: Series B (Statistical
  Methodology)}, 82(1):39--67, 2020.

\bibitem{cakra2015stock}
Yahya~Eru Cakra and Bayu~Distiawan Trisedya.
\newblock Stock price prediction using linear regression based on sentiment
  analysis.
\newblock In {\em 2015 international conference on advanced computer science
  and information systems (ICACSIS)}, pages 147--154. IEEE, 2015.

\bibitem{friedman1996bias}
Batya Friedman and Helen Nissenbaum.
\newblock Bias in computer systems.
\newblock {\em ACM Transactions on Information Systems (TOIS)}, 14(3):330--347,
  1996.

\bibitem{gordon1995evaluation}
Diana~F Gordon and Marie Desjardins.
\newblock Evaluation and selection of biases in machine learning.
\newblock {\em Machine learning}, 20(1-2):5--22, 1995.

\bibitem{suresh2019framework}
Harini Suresh and John~V Guttag.
\newblock A framework for understanding unintended consequences of machine
  learning.
\newblock {\em arXiv preprint arXiv:1901.10002}, 2019.

\bibitem{masis2021interpretable}
Serg Mas{\'\i}s.
\newblock {\em Interpretable Machine Learning with Python: Learn to build
  interpretable high-performance models with hands-on real-world examples}.
\newblock Packt Publishing Ltd, 2021.

\bibitem{baker2021algorithmic}
Ryan~S Baker and Aaron Hawn.
\newblock Algorithmic bias in education.
\newblock {\em EdArXiv preprint: 10.35542/osf.io/pbmvz}, Mar 2021.

\bibitem{altman2015association}
Naomi Altman and Martin Krzywinski.
\newblock Points of significance: Association, correlation and causation.
\newblock {\em Nature methods}, 12(10), 2015.

\bibitem{markovits1989belief}
Henry Markovits and Guilaine Nantel.
\newblock The belief-bias effect in the production and evaluation of logical
  conclusions.
\newblock {\em Memory \& cognition}, 17(1):11--17, 1989.

\bibitem{stelmakh2021prior}
Ivan Stelmakh, Nihar~B Shah, Aarti Singh, and Hal Daum{\'e}~III.
\newblock Prior and prejudice: The novice reviewers' bias against resubmissions
  in conference peer review.
\newblock {\em Proceedings of the ACM on Human-Computer Interaction},
  5(CSCW1):1--17, 2021.

\bibitem{lerman2014leveraging}
Kristina Lerman and Tad Hogg.
\newblock Leveraging position bias to improve peer recommendation.
\newblock {\em PloS one}, 9(6):e98914, 2014.

\bibitem{balakrishnan2021towards}
Guha Balakrishnan, Yuanjun Xiong, Wei Xia, and Pietro Perona.
\newblock Towards causal benchmarking of biasin face analysis algorithms.
\newblock In {\em Deep Learning-Based Face Analytics}, pages 327--359.
  Springer, 2021.

\bibitem{schaaf2021towards}
Nina Schaaf, Omar~de Mitri, Hang~Beom Kim, Alexander Windberger, and Marco~F
  Huber.
\newblock Towards measuring bias in image classification.
\newblock In {\em International Conference on Artificial Neural Networks},
  pages 433--445. Springer, 2021.

\bibitem{lapuschkin2019unmasking}
Sebastian Lapuschkin, Stephan W{\"a}ldchen, Alexander Binder, Gr{\'e}goire
  Montavon, Wojciech Samek, and Klaus-Robert M{\"u}ller.
\newblock Unmasking clever hans predictors and assessing what machines really
  learn.
\newblock {\em Nature communications}, 10(1):1--8, 2019.

\bibitem{kokhlikyan2020captum}
Narine Kokhlikyan, Vivek Miglani, Miguel Martin, Edward Wang, Bilal Alsallakh,
  Jonathan Reynolds, Alexander Melnikov, Natalia Kliushkina, Carlos Araya, Siqi
  Yan, and Orion Reblitz-Richardson.
\newblock Captum: A unified and generic model interpretability library for
  pytorch, 2020.

\bibitem{stock2018convnets}
Pierre Stock and Moustapha Cisse.
\newblock Convnets and imagenet beyond accuracy: Understanding mistakes and
  uncovering biases.
\newblock In {\em Proceedings of the European Conference on Computer Vision
  (ECCV)}, pages 498--512, 2018.

\bibitem{kim2016examples}
Been Kim, Rajiv Khanna, and Oluwasanmi~O Koyejo.
\newblock Examples are not enough, learn to criticize! criticism for
  interpretability.
\newblock {\em Advances in neural information processing systems}, 29, 2016.

\bibitem{serna2021ifbid}
Ignacio Serna, Aythami Morales, Julian Fierrez, and Javier Ortega-Garcia.
\newblock Ifbid: Inference-free bias detection.
\newblock {\em arXiv preprint arXiv:2109.04374}, 2021.

\bibitem{wang2020towards}
Zeyu Wang, Klint Qinami, Ioannis~Christos Karakozis, Kyle Genova, Prem Nair,
  Kenji Hata, and Olga Russakovsky.
\newblock Towards fairness in visual recognition: Effective strategies for bias
  mitigation.
\newblock In {\em Proceedings of the IEEE/CVF conference on computer vision and
  pattern recognition}, pages 8919--8928, 2020.

\bibitem{dwork2018decoupled}
Cynthia Dwork, Nicole Immorlica, Adam~Tauman Kalai, and Max Leiserson.
\newblock Decoupled classifiers for group-fair and efficient machine learning.
\newblock In {\em Conference on fairness, accountability and transparency},
  pages 119--133. PMLR, 2018.

\bibitem{zhang2018mitigating}
Brian~Hu Zhang, Blake Lemoine, and Margaret Mitchell.
\newblock Mitigating unwanted biases with adversarial learning.
\newblock In {\em Proceedings of the 2018 AAAI/ACM Conference on AI, Ethics,
  and Society}, pages 335--340, 2018.

\bibitem{le2020adversarial}
Ronan Le~Bras, Swabha Swayamdipta, Chandra Bhagavatula, Rowan Zellers, Matthew
  Peters, Ashish Sabharwal, and Yejin Choi.
\newblock Adversarial filters of dataset biases.
\newblock In {\em International Conference on Machine Learning}, pages
  1078--1088. PMLR, 2020.

\bibitem{huang2019brain}
Qiaoying Huang, Xiao Chen, Dimitris Metaxas, and Mariappan~S Nadar.
\newblock Brain segmentation from k-space with end-to-end recurrent attention
  network.
\newblock In {\em International Conference on Medical Image Computing and
  Computer-Assisted Intervention}, pages 275--283. Springer, 2019.

\bibitem{barata2019deep}
Catarina Barata, Jorge~S Marques, and M~Emre~Celebi.
\newblock Deep attention model for the hierarchical diagnosis of skin lesions.
\newblock In {\em Proceedings of the IEEE/CVF Conference on Computer Vision and
  Pattern Recognition Workshops}, pages 0--0, 2019.

\bibitem{li2019zoom}
Hongyang Li, Yu~Liu, Wanli Ouyang, and Xiaogang Wang.
\newblock Zoom out-and-in network with map attention decision for region
  proposal and object detection.
\newblock {\em International Journal of Computer Vision}, 127(3):225--238,
  2019.

\bibitem{jain2019attention}
Sarthak Jain and Byron~C. Wallace.
\newblock {A}ttention is not {E}xplanation.
\newblock In {\em Proceedings of the 2019 Conference of the North {A}merican
  Chapter of the Association for Computational Linguistics: Human Language
  Technologies, Volume 1 (Long and Short Papers)}, pages 3543--3556,
  Minneapolis, Minnesota, June 2019. Association for Computational Linguistics.

\bibitem{wiegreffe2019attention}
Sarah Wiegreffe and Yuval Pinter.
\newblock Attention is not not explanation.
\newblock In {\em Proceedings of the 2019 Conference on Empirical Methods in
  Natural Language Processing and the 9th International Joint Conference on
  Natural Language Processing (EMNLP-IJCNLP)}, pages 11--20, Hong Kong, China,
  November 2019. Association for Computational Linguistics.

\bibitem{grimsley2020attention}
Christopher Grimsley, Elijah Mayfield, and Julia R.S.~Bursten.
\newblock Why attention is not explanation: Surgical intervention and causal
  reasoning about neural models.
\newblock In {\em Proceedings of the 12th Language Resources and Evaluation
  Conference}, pages 1780--1790, Marseille, France, May 2020. European Language
  Resources Association.

\bibitem{tutek2020staying}
Martin Tutek and Jan Snajder.
\newblock Staying true to your word: (how) can attention become explanation?
\newblock In {\em Proceedings of the 5th Workshop on Representation Learning
  for NLP}, pages 131--142, Online, July 2020. Association for Computational
  Linguistics.

\bibitem{woody2004more}
Andrea~I Woody.
\newblock More telltale signs: what attention to representation reveals about
  scientific explanation.
\newblock {\em Philosophy of Science}, 71(5):780--793, 2004.

\bibitem{tang2018attention}
Yuxing Tang, Xiaosong Wang, Adam~P Harrison, Le~Lu, Jing Xiao, and Ronald~M
  Summers.
\newblock Attention-guided curriculum learning for weakly supervised
  classification and localization of thoracic diseases on chest radiographs.
\newblock In {\em International Workshop on Machine Learning in Medical
  Imaging}, pages 249--258. Springer, 2018.

\bibitem{hou2018self}
Qibin Hou, PengTao Jiang, Yunchao Wei, and Ming-Ming Cheng.
\newblock Self-erasing network for integral object attention.
\newblock {\em Advances in Neural Information Processing Systems}, 31, 2018.

\bibitem{chai2020human}
Chengliang Chai and Guoliang Li.
\newblock Human-in-the-loop techniques in machine learning.
\newblock {\em Data Engineering}, 37:16, 2020.

\bibitem{das2020opportunities}
Arun Das and Paul Rad.
\newblock Opportunities and challenges in explainable artificial intelligence
  {(XAI):} {A} survey.
\newblock {\em CoRR}, abs/2006.11371, 2020.

\bibitem{gilpin2018explaining}
Leilani~H Gilpin, David Bau, Ben~Z Yuan, Ayesha Bajwa, Michael Specter, and
  Lalana Kagal.
\newblock Explaining explanations: An overview of interpretability of machine
  learning.
\newblock In {\em 2018 IEEE 5th International Conference on data science and
  advanced analytics (DSAA)}, pages 80--89. IEEE, 2018.

\bibitem{molnar2020interpretable}
Christoph Molnar.
\newblock {\em Interpretable machine learning}.
\newblock [Online; accessed September-2021], 2020.

\bibitem{pintelas2020grey}
Emmanuel Pintelas, Ioannis~E Livieris, and Panagiotis Pintelas.
\newblock A grey-box ensemble model exploiting black-box accuracy and white-box
  intrinsic interpretability.
\newblock {\em Algorithms}, 13(1):17, 2020.

\bibitem{arrieta2020explainable}
Alejandro~Barredo Arrieta, Natalia D{\'\i}az-Rodr{\'\i}guez, Javier Del~Ser,
  Adrien Bennetot, Siham Tabik, Alberto Barbado, Salvador Garc{\'\i}a, Sergio
  Gil-L{\'o}pez, Daniel Molina, Richard Benjamins, et~al.
\newblock Explainable artificial intelligence (xai): Concepts, taxonomies,
  opportunities and challenges toward responsible ai.
\newblock {\em Information Fusion}, 58:82--115, 2020.

\bibitem{queipo2005surrogate}
Nestor~V Queipo, Raphael~T Haftka, Wei Shyy, Tushar Goel, Rajkumar
  Vaidyanathan, and P~Kevin Tucker.
\newblock Surrogate-based analysis and optimization.
\newblock {\em Progress in aerospace sciences}, 41(1):1--28, 2005.

\bibitem{messalas2019model}
Andreas Messalas, Yiannis Kanellopoulos, and Christos Makris.
\newblock Model-agnostic interpretability with shapley values.
\newblock In {\em 2019 10th International Conference on Information,
  Intelligence, Systems and Applications (IISA)}, pages 1--7. IEEE, 2019.

\bibitem{mohseni2018multidisciplinary}
Sina Mohseni, Niloofar Zarei, and Eric~D Ragan.
\newblock A multidisciplinary survey and framework for design and evaluation of
  explainable ai systems.
\newblock {\em ACM Transactions on Interactive Intelligent Systems (TiiS)},
  11(3-4):1--45, 2021.

\bibitem{van2008visualizing}
Laurens Van~der Maaten and Geoffrey Hinton.
\newblock Visualizing data using t-sne.
\newblock {\em Journal of machine learning research}, 9(11), 2008.

\bibitem{lakkaraju2016interpretable}
Himabindu Lakkaraju, Stephen~H Bach, and Jure Leskovec.
\newblock Interpretable decision sets: A joint framework for description and
  prediction.
\newblock In {\em Proceedings of the 22nd ACM SIGKDD international conference
  on knowledge discovery and data mining}, pages 1675--1684, 2016.

\bibitem{cai2019human}
Carrie~J Cai, Emily Reif, Narayan Hegde, Jason Hipp, Been Kim, Daniel Smilkov,
  Martin Wattenberg, Fernanda Viegas, Greg~S Corrado, Martin~C Stumpe, et~al.
\newblock Human-centered tools for coping with imperfect algorithms during
  medical decision-making.
\newblock In {\em Proceedings of the 2019 CHI Conference on Human Factors in
  Computing Systems}, pages 1--14, 2019.

\bibitem{wilkinson2009history}
Leland Wilkinson and Michael Friendly.
\newblock The history of the cluster heat map.
\newblock {\em The American Statistician}, 63(2):179--184, 2009.

\bibitem{guo2009novel}
Chenlei Guo and Liming Zhang.
\newblock A novel multiresolution spatiotemporal saliency detection model and
  its applications in image and video compression.
\newblock {\em IEEE transactions on image processing}, 19(1):185--198, 2009.

\bibitem{selvaraju2017grad}
Ramprasaath~R Selvaraju, Michael Cogswell, Abhishek Das, Ramakrishna Vedantam,
  Devi Parikh, and Dhruv Batra.
\newblock Grad-cam: Visual explanations from deep networks via gradient-based
  localization.
\newblock In {\em Proceedings of the IEEE international conference on computer
  vision}, pages 618--626, 2017.

\bibitem{zeiler2014visualizing}
Matthew~D Zeiler and Rob Fergus.
\newblock Visualizing and understanding convolutional networks.
\newblock In {\em European conference on computer vision}, pages 818--833.
  Springer, 2014.

\bibitem{springenberg2014striving}
Jost~Tobias Springenberg, Alexey Dosovitskiy, Thomas Brox, and Martin
  Riedmiller.
\newblock Striving for simplicity: The all convolutional net.
\newblock {\em arXiv preprint arXiv:1412.6806}, 2014.

\bibitem{laurent2018multilinear}
Thomas Laurent and James Brecht.
\newblock The multilinear structure of relu networks.
\newblock In {\em International conference on machine learning}, pages
  2908--2916. PMLR, 2018.

\bibitem{adebayo2018local}
Julius Adebayo, Justin Gilmer, Ian Goodfellow, and Been Kim.
\newblock Local explanation methods for deep neural networks lack sensitivity
  to parameter values.
\newblock {\em arXiv preprint arXiv:1810.03307}, 2018.

\bibitem{adebayo2018sanity}
Julius Adebayo, Justin Gilmer, Michael Muelly, Ian Goodfellow, Moritz Hardt,
  and Been Kim.
\newblock Sanity checks for saliency maps.
\newblock In {\em Proceedings of the 32nd International Conference on Neural
  Information Processing Systems}, NIPS'18, page 9525–9536, Red Hook, NY,
  USA, 2018. Curran Associates Inc.

\bibitem{nguyen2016multifaceted}
Anh~Mai Nguyen, Jason Yosinski, and Jeff Clune.
\newblock Multifaceted feature visualization: Uncovering the different types of
  features learned by each neuron in deep neural networks.
\newblock {\em CoRR}, abs/1602.03616, 2016.

\bibitem{lime}
Marco~Tulio Ribeiro, Sameer Singh, and Carlos Guestrin.
\newblock "why should {I} trust you?": Explaining the predictions of any
  classifier.
\newblock In {\em Proceedings of the 22nd {ACM} {SIGKDD} International
  Conference on Knowledge Discovery and Data Mining, San Francisco, CA, USA,
  August 13-17, 2016}, pages 1135--1144, 2016.

\bibitem{rahnama2019study}
Amir Hossein~Akhavan Rahnama and Henrik Bostr{\"o}m.
\newblock A study of data and label shift in the lime framework.
\newblock {\em arXiv preprint arXiv:1910.14421}, 2019.

\bibitem{slack2020fooling}
Dylan Slack, Sophie Hilgard, Emily Jia, Sameer Singh, and Himabindu Lakkaraju.
\newblock Fooling lime and shap: Adversarial attacks on post hoc explanation
  methods.
\newblock In {\em Proceedings of the AAAI/ACM Conference on AI, Ethics, and
  Society}, pages 180--186, 2020.

\bibitem{lundberg2017unified}
Scott~M Lundberg and Su-In Lee.
\newblock A unified approach to interpreting model predictions.
\newblock {\em Advances in neural information processing systems}, 30, 2017.

\bibitem{montavon2019layer}
Gr{\'e}goire Montavon, Alexander Binder, Sebastian Lapuschkin, Wojciech Samek,
  and Klaus-Robert M{\"u}ller.
\newblock Layer-wise relevance propagation: an overview.
\newblock {\em Explainable AI: interpreting, explaining and visualizing deep
  learning}, pages 193--209, 2019.

\bibitem{montavonDTD}
Gr{\'e}goire Montavon, Sebastian Lapuschkin, Alexander Binder, Wojciech Samek,
  and Klaus-Robert M{\"u}ller.
\newblock Explaining nonlinear classification decisions with deep taylor
  decomposition.
\newblock {\em Pattern Recognition}, 65:211--222, 2017.

\bibitem{kim2018interpretability}
Been Kim, Martin Wattenberg, Justin Gilmer, Carrie Cai, James Wexler, Fernanda
  Viegas, et~al.
\newblock Interpretability beyond feature attribution: Quantitative testing
  with concept activation vectors (tcav).
\newblock In {\em International conference on machine learning}, pages
  2668--2677. PMLR, 2018.

\bibitem{ribeiro2016why}
Marco~Tulio Ribeiro, Sameer Singh, and Carlos Guestrin.
\newblock " why should i trust you?" explaining the predictions of any
  classifier.
\newblock In {\em Proceedings of the 22nd ACM SIGKDD international conference
  on knowledge discovery and data mining}, pages 1135--1144, 2016.

\bibitem{van2019global}
Ilse van~der Linden, Hinda Haned, and Evangelos Kanoulas.
\newblock Global aggregations of local explanations for black box models.
\newblock {\em arXiv preprint arXiv:1907.03039}, 2019.

\bibitem{ibrahim2019global}
Mark Ibrahim, Melissa Louie, Ceena Modarres, and John Paisley.
\newblock Global explanations of neural networks: Mapping the landscape of
  predictions.
\newblock In {\em Proceedings of the 2019 AAAI/ACM Conference on AI, Ethics,
  and Society}, pages 279--287, 2019.

\bibitem{lin2020you}
Yi-Shan Lin, Wen-Chuan Lee, and Z~Berkay Celik.
\newblock What do you see? evaluation of explainable artificial intelligence
  (xai) interpretability through neural backdoors.
\newblock In {\em Proceedings of the 27th ACM SIGKDD Conference on Knowledge
  Discovery \& Data Mining}, pages 1027--1035, 2021.

\bibitem{hase2020evaluating}
Peter Hase and Mohit Bansal.
\newblock Evaluating explainable {AI}: Which algorithmic explanations help
  users predict model behavior?
\newblock In {\em Proceedings of the 58th Annual Meeting of the Association for
  Computational Linguistics}, pages 5540--5552, Online, July 2020. Association
  for Computational Linguistics.

\bibitem{simonyan2013deep}
K~Simonyan, A~Vedaldi, and A~Zisserman.
\newblock Deep inside convolutional networks: visualising image classification
  models and saliency maps.
\newblock In {\em Proceedings of the International Conference on Learning
  Representations (ICLR)}. ICLR, 2014.

\bibitem{buccinca2020proxy}
Zana Bu{\c{c}}inca, Phoebe Lin, Krzysztof~Z Gajos, and Elena~L Glassman.
\newblock Proxy tasks and subjective measures can be misleading in evaluating
  explainable ai systems.
\newblock In {\em Proceedings of the 25th International Conference on
  Intelligent User Interfaces}, pages 454--464, 2020.

\bibitem{fong2017interpretable}
Ruth~C Fong and Andrea Vedaldi.
\newblock Interpretable explanations of black boxes by meaningful perturbation.
\newblock In {\em Proceedings of the IEEE International Conference on Computer
  Vision}, pages 3429--3437, 2017.

\bibitem{samek2016evaluating}
Wojciech Samek, Alexander Binder, Gr{\'e}goire Montavon, Sebastian Lapuschkin,
  and Klaus-Robert M{\"u}ller.
\newblock Evaluating the visualization of what a deep neural network has
  learned.
\newblock {\em IEEE transactions on neural networks and learning systems},
  28(11):2660--2673, 2016.

\bibitem{yeh2019fidelity}
Chih-Kuan Yeh, Cheng-Yu Hsieh, Arun Suggala, David~I Inouye, and Pradeep~K
  Ravikumar.
\newblock On the (in) fidelity and sensitivity of explanations.
\newblock {\em Advances in Neural Information Processing Systems}, 32, 2019.

\bibitem{hooker2018benchmark}
Sara Hooker, Dumitru Erhan, Pieter-Jan Kindermans, and Been Kim.
\newblock A benchmark for interpretability methods in deep neural networks.
\newblock {\em Advances in neural information processing systems}, 32, 2019.

\bibitem{yang2019benchmarking}
Mengjiao Yang and Been Kim.
\newblock Benchmarking attribution methods with relative feature importance.
\newblock {\em arXiv e-prints}, pages arXiv--1907, 2019.

\bibitem{ghorbani2019interpretation}
Amirata Ghorbani, Abubakar Abid, and James Zou.
\newblock Interpretation of neural networks is fragile.
\newblock In {\em Proceedings of the AAAI Conference on Artificial
  Intelligence}, volume~33, pages 3681--3688, 2019.

\bibitem{bach2015pixel}
Sebastian Bach, Alexander Binder, Gr{\'e}goire Montavon, Frederick Klauschen,
  Klaus-Robert M{\"u}ller, and Wojciech Samek.
\newblock On pixel-wise explanations for non-linear classifier decisions by
  layer-wise relevance propagation.
\newblock {\em PloS one}, 10(7):e0130140, 2015.

\bibitem{pmlrv70sundararajan17a}
Mukund Sundararajan, Ankur Taly, and Qiqi Yan.
\newblock Axiomatic attribution for deep networks.
\newblock In Doina Precup and Yee~Whye Teh, editors, {\em Proceedings of the
  34th International Conference on Machine Learning}, volume~70 of {\em
  Proceedings of Machine Learning Research}, pages 3319--3328. PMLR, 06--11 Aug
  2017.

\bibitem{alvarezmelis2018robust}
David Alvarez~Melis and Tommi Jaakkola.
\newblock Towards robust interpretability with self-explaining neural networks.
\newblock {\em Advances in neural information processing systems}, 31, 2018.

\bibitem{luss2021leveraging}
Ronny Luss, Pin-Yu Chen, Amit Dhurandhar, Prasanna Sattigeri, Yunfeng Zhang,
  Karthikeyan Shanmugam, and Chun-Chen Tu.
\newblock Leveraging latent features for local explanations.
\newblock {\em Proceedings of the 27th ACM SIGKDD Conference on Knowledge
  Discovery and Data Mining}, page 1139–1149, 2021.

\bibitem{arya2019one}
Vijay Arya, Rachel~KE Bellamy, Pin-Yu Chen, Amit Dhurandhar, Michael Hind,
  Samuel~C Hoffman, Stephanie Houde, Q~Vera Liao, Ronny Luss, Aleksandra
  Mojsilovi{\'c}, et~al.
\newblock One explanation does not fit all: A toolkit and taxonomy of ai
  explainability techniques.
\newblock {\em KI - Künstliche Intelligenz}, 34:235–250, 2020.

\bibitem{biessmann2021quality}
Felix Biessmann and Dionysius Refiano.
\newblock Quality metrics for transparent machine learning with and without
  humans in the loop are not correlated.
\newblock {\em arXiv preprint arXiv:2107.02033}, 2021.

\bibitem{pleiss2020identifying}
Geoff Pleiss, Tianyi Zhang, Ethan Elenberg, and Kilian~Q Weinberger.
\newblock Identifying mislabeled data using the area under the margin ranking.
\newblock In H.~Larochelle, M.~Ranzato, R.~Hadsell, M.F. Balcan, and H.~Lin,
  editors, {\em Advances in Neural Information Processing Systems}, volume~33,
  pages 17044--17056. Curran Associates, Inc., 2020.

\bibitem{li2017webvision}
Wen Li, Limin Wang, Wei Li, Eirikur Agustsson, and Luc Van~Gool.
\newblock Webvision database: Visual learning and understanding from web data.
\newblock {\em arXiv preprint arXiv:1708.02862}, 2017.

\bibitem{zeng2021validating}
Qingkai Zeng, Mengxia Yu, Wenhao Yu, Tianwen Jiang, and Meng Jiang.
\newblock Validating label consistency in {NER} data annotation.
\newblock In {\em Proceedings of the 2nd Workshop on Evaluation and Comparison
  of NLP Systems}, pages 11--15, Punta Cana, Dominican Republic, November 2021.
  Association for Computational Linguistics.

\bibitem{proencca2020taco}
Pedro~F Proen{\c{c}}a and Pedro Sim{\~o}es.
\newblock Taco: Trash annotations in context for litter detection.
\newblock {\em arXiv preprint arXiv:2003.06975}, 2020.

\bibitem{isic2019BCN20000}
Marc Combalia, Noel~CF Codella, Veronica Rotemberg, Brian Helba, Veronica
  Vilaplana, Ofer Reiter, Cristina Carrera, Alicia Barreiro, Allan~C Halpern,
  Susana Puig, et~al.
\newblock Bcn20000: Dermoscopic lesions in the wild.
\newblock {\em arXiv preprint arXiv:1908.02288}, 2019.

\bibitem{isic2019codella}
Noel~CF Codella, David Gutman, M~Emre Celebi, Brian Helba, Michael~A Marchetti,
  Stephen~W Dusza, Aadi Kalloo, Konstantinos Liopyris, Nabin Mishra, Harald
  Kittler, et~al.
\newblock Skin lesion analysis toward melanoma detection: A challenge at the
  2017 international symposium on biomedical imaging (isbi), hosted by the
  international skin imaging collaboration (isic).
\newblock In {\em 2018 IEEE 15th international symposium on biomedical imaging
  (ISBI 2018)}, pages 168--172. IEEE, 2018.

\bibitem{isic2019ham10000}
Philipp Tschandl, Cliff Rosendahl, and Harald Kittler.
\newblock The ham10000 dataset, a large collection of multi-source
  dermatoscopic images of common pigmented skin lesions.
\newblock {\em Scientific data}, 5(1):1--9, 2018.

\bibitem{isic2020}
Veronica Rotemberg, Nicholas Kurtansky, Brigid Betz-Stablein, Liam Caffery,
  Emmanouil Chousakos, Noel Codella, Marc Combalia, Stephen Dusza, Pascale
  Guitera, David Gutman, et~al.
\newblock A patient-centric dataset of images and metadata for identifying
  melanomas using clinical context.
\newblock {\em Scientific data}, 8(1):1--8, 2021.

\bibitem{gewirtzman2003evaluation}
AJ~Gewirtzman, J-H Saurat, and RP~Braun.
\newblock An evaluation of dermoscopy fluids and application techniques.
\newblock {\em British Journal of Dermatology}, 149(1):59--63, 2003.

\bibitem{navarrete2019ink}
Cristi{\'a}n Navarrete-Dechent, Pablo Uribe, and Ashfaq Marghoob.
\newblock Ink-enhanced dermoscopy is a useful tool to differentiate acquired
  solitary plaque porokeratosis from other scaly lesions.
\newblock {\em Journal of the American Academy of Dermatology},
  80(6):e137--e138, 2019.

\bibitem{bissoto19deconstructing}
Alceu Bissoto, Eduardo Valle, and Sandra Avila.
\newblock Debiasing skin lesion datasets and models? not so fast.
\newblock In {\em ISIC Skin Image Anaylsis Workshop, 2020 {IEEE} Conference on
  Computer Vision and Pattern Recognition Workshops (CVPRW)}, 2020.

\bibitem{mchugh2012interrater}
Mary~L McHugh.
\newblock Interrater reliability: the kappa statistic.
\newblock {\em Biochemia medica}, 22(3):276--282, 2012.

\bibitem{argenziano2003dermoscopy}
Giuseppe Argenziano, H~Peter Soyer, Sergio Chimenti, Renato Talamini, Rosamaria
  Corona, Francesco Sera, Michael Binder, Lorenzo Cerroni, Gaetano De~Rosa,
  Gerardo Ferrara, et~al.
\newblock Dermoscopy of pigmented skin lesions: results of a consensus meeting
  via the internet.
\newblock {\em Journal of the American Academy of Dermatology}, 48(5):679--693,
  2003.

\bibitem{braun2021mmd}
Ralph~Braun dermoscopedia – Scott~Menzies.
\newblock Menzies method --- dermoscopedia, 2021.
\newblock [Online; accessed 7-December-2021].

\bibitem{argenziano2021spcd}
Giulycalabrese Alina De~Rosa, Teresa~Russo and Giuseppe Argenziano.
\newblock Seven point checklist --- dermoscopedia, 2021.
\newblock [Online; accessed 7-December-2021].

\bibitem{Bissoto_2019_CVPR_Workshops}
Alceu Bissoto, Michel Fornaciali, Eduardo Valle, and Sandra Avila.
\newblock (de)constructing bias on skin lesion datasets.
\newblock In {\em Proceedings of the IEEE/CVF Conference on Computer Vision and
  Pattern Recognition (CVPR) Workshops}, June 2019.

\bibitem{balasubramanian2002isomap}
Mukund Balasubramanian, Eric~L Schwartz, Joshua~B Tenenbaum, Vin de~Silva, and
  John~C Langford.
\newblock The isomap algorithm and topological stability.
\newblock {\em Science}, 295(5552):7--7, 2002.

\bibitem{wang2019neural}
Bolun Wang, Yuanshun Yao, Shawn Shan, Huiying Li, Bimal Viswanath, Haitao
  Zheng, and Ben~Y Zhao.
\newblock Neural cleanse: Identifying and mitigating backdoor attacks in neural
  networks.
\newblock In {\em 2019 IEEE Symposium on Security and Privacy (SP)}, pages
  707--723. IEEE, 2019.

\bibitem{kodinariya2013review}
Trupti~M Kodinariya and Prashant~R Makwana.
\newblock Review on determining number of cluster in k-means clustering.
\newblock {\em International Journal}, 1(6):90--95, 2013.

\bibitem{hardt2021amazon}
Michaela Hardt, Xiaoguang Chen, Xiaoyi Cheng, Michele Donini, Jason Gelman,
  Satish Gollaprolu, John He, Pedro Larroy, Xinyu Liu, Nick McCarthy, et~al.
\newblock Amazon sagemaker clarify: Machine learning bias detection and
  explainability in the cloud.
\newblock {\em arXiv preprint arXiv:2109.03285}, 2021.

\bibitem{li2021digital}
Wei Li, Alex Noel~Joseph Raj, Tardi Tjahjadi, and Zhemin Zhuang.
\newblock Digital hair removal by deep learning for skin lesion segmentation.
\newblock {\em Pattern Recognition}, 117:107994, 2021.

\bibitem{lee1997dullrazor}
Tim Lee, Vincent Ng, Richard Gallagher, Andrew Coldman, and David McLean.
\newblock Dullrazor{\textregistered}: A software approach to hair removal from
  images.
\newblock {\em Computers in biology and medicine}, 27(6):533--543, 1997.

\bibitem{huang2013robust}
Adam Huang, Shun-Yuen Kwan, Wen-Yu Chang, Min-Yin Liu, Min-Hsiu Chi, and
  Gwo-Shing Chen.
\newblock A robust hair segmentation and removal approach for clinical images
  of skin lesions.
\newblock In {\em 2013 35th Annual International Conference of the IEEE
  Engineering in Medicine and Biology Society (EMBC)}, pages 3315--3318. IEEE,
  2013.

\bibitem{xu2015comprehensive}
Dongkuan Xu and Yingjie Tian.
\newblock A comprehensive survey of clustering algorithms.
\newblock {\em Annals of Data Science}, 2(2):165--193, 2015.

\bibitem{shorten2019survey}
Connor Shorten and Taghi~M Khoshgoftaar.
\newblock A survey on image data augmentation for deep learning.
\newblock {\em Journal of Big Data}, 6(1):1--48, 2019.

\bibitem{wei2019eda}
Jason Wei and Kai Zou.
\newblock Eda: Easy data augmentation techniques for boosting performance on
  text classification tasks.
\newblock {\em arXiv preprint arXiv:1901.11196}, 2019.

\bibitem{ko2015audio}
Tom Ko, Vijayaditya Peddinti, Daniel Povey, and Sanjeev Khudanpur.
\newblock Audio augmentation for speech recognition.
\newblock In {\em Sixteenth annual conference of the international speech
  communication association}, 2015.

\bibitem{wang2019time}
Haishuai Wang, Qin Zhang, Jia Wu, Shirui Pan, and Yixin Chen.
\newblock Time series feature learning with labeled and unlabeled data.
\newblock {\em Pattern Recognition}, 89:55--66, 2019.

\bibitem{wang2018random}
Nannan Wang, Xinbo Gao, and Jie Li.
\newblock Random sampling for fast face sketch synthesis.
\newblock {\em Pattern Recognition}, 76:215--227, 2018.

\bibitem{costa2017adversarial}
Pedro Costa, Adrian Galdran, Maria~Ines Meyer, Ana~Maria Mendon{\c{c}}a, and
  Aur{\'e}lio Campilho.
\newblock Adversarial synthesis of retinal images from vessel trees.
\newblock In {\em International Conference Image Analysis and Recognition},
  pages 516--523. Springer, 2017.

\bibitem{yuan2018regularized}
Xiaohui Yuan, Lijun Xie, and Mohamed Abouelenien.
\newblock A regularized ensemble framework of deep learning for cancer
  detection from multi-class, imbalanced training data.
\newblock {\em Pattern Recognition}, 77:160--172, 2018.

\bibitem{peng2019structured}
Jialin Peng, Xiaofeng Zhu, Ye~Wang, Le~An, and Dinggang Shen.
\newblock Structured sparsity regularized multiple kernel learning for
  alzheimer’s disease diagnosis.
\newblock {\em Pattern recognition}, 88:370--382, 2019.

\bibitem{goodfellow2016deep}
Ian Goodfellow, Yoshua Bengio, Aaron Courville, and Yoshua Bengio.
\newblock {\em Deep learning}, volume~1.
\newblock MIT press Cambridge, 2016.

\bibitem{lawrence1997lessons}
Steve Lawrence, C~Lee Giles, and Ah~Chung Tsoi.
\newblock Lessons in neural network training: Overfitting may be harder than
  expected.
\newblock In {\em AAAI/IAAI}, pages 540--545. Citeseer, 1997.

\bibitem{sun2019meta}
Qianru Sun, Yaoyao Liu, Tat-Seng Chua, and Bernt Schiele.
\newblock Meta-transfer learning for few-shot learning.
\newblock In {\em Proceedings of the IEEE/CVF Conference on Computer Vision and
  Pattern Recognition}, pages 403--412, 2019.

\bibitem{breck2017ml}
Eric Breck, Shanqing Cai, Eric Nielsen, Michael Salib, and D~Sculley.
\newblock The ml test score: A rubric for ml production readiness and technical
  debt reduction.
\newblock In {\em 2017 IEEE International Conference on Big Data (Big Data)},
  pages 1123--1132. IEEE, 2017.

\bibitem{jackson2019style}
Philip~TG Jackson, Amir~Atapour Abarghouei, Stephen Bonner, Toby~P Breckon, and
  Boguslaw Obara.
\newblock Style augmentation: data augmentation via style randomization.
\newblock In {\em CVPR Workshops}, pages 83--92, 2019.

\bibitem{Mikolajczyk_AgaMiko_data-augmentation-review_A_comprehensive_2021}
Agnieszka Mikolajczyk.
\newblock {AgaMiko/data-augmentation-review: A comprehensive review of data
  augmentation resources}.
\newblock 9 2021.

\bibitem{zhong2020random}
Zhun Zhong, Liang Zheng, Guoliang Kang, Shaozi Li, and Yi~Yang.
\newblock Random erasing data augmentation.
\newblock In {\em Proceedings of the AAAI Conference on Artificial
  Intelligence}, volume~34, pages 13001--13008, 2020.

\bibitem{tremblay2018training}
Jonathan Tremblay, Aayush Prakash, David Acuna, Mark Brophy, Varun Jampani, Cem
  Anil, Thang To, Eric Cameracci, Shaad Boochoon, and Stan Birchfield.
\newblock Training deep networks with synthetic data: Bridging the reality gap
  by domain randomization.
\newblock In {\em Proceedings of the IEEE conference on computer vision and
  pattern recognition workshops}, pages 969--977, 2018.

\bibitem{jabbar2021survey}
Abdul Jabbar, Xi~Li, and Bourahla Omar.
\newblock A survey on generative adversarial networks: Variants, applications,
  and training.
\newblock {\em ACM Computing Surveys (CSUR)}, 54(8):1--49, 2021.

\bibitem{gatys2015neural}
Leon~A Gatys, Alexander~S Ecker, and Matthias Bethge.
\newblock A neural algorithm of artistic style.
\newblock {\em Journal of Vision}, 16(12), 2015.

\bibitem{hertzmann2001image}
Aaron Hertzmann, Charles~E Jacobs, Nuria Oliver, Brian Curless, and David~H
  Salesin.
\newblock Image analogies.
\newblock In {\em Proceedings of the 28th annual conference on Computer
  graphics and interactive techniques}, pages 327--340, 2001.

\bibitem{shen2016automatic}
Xiaoyong Shen, Aaron Hertzmann, Jiaya Jia, Sylvain Paris, Brian Price, Eli
  Shechtman, and Ian Sachs.
\newblock Automatic portrait segmentation for image stylization.
\newblock In {\em Computer Graphics Forum}, volume~35, pages 93--102. Wiley
  Online Library, 2016.

\bibitem{selim2016painting}
Ahmed Selim, Mohamed Elgharib, and Linda Doyle.
\newblock Painting style transfer for head portraits using convolutional neural
  networks.
\newblock {\em ACM Transactions on Graphics (ToG)}, 35(4):1--18, 2016.

\bibitem{huang2017arbitrary}
Xun Huang and Serge Belongie.
\newblock Arbitrary style transfer in real-time with adaptive instance
  normalization.
\newblock In {\em Proceedings of the IEEE international conference on computer
  vision}, pages 1501--1510, 2017.

\bibitem{chen2018stereoscopic}
Dongdong Chen, Lu~Yuan, Jing Liao, Nenghai Yu, and Gang Hua.
\newblock Stereoscopic neural style transfer.
\newblock In {\em Proceedings of the IEEE Conference on Computer Vision and
  Pattern Recognition}, pages 6654--6663, 2018.

\bibitem{huang2017real}
Haozhi Huang, Hao Wang, Wenhan Luo, Lin Ma, Wenhao Jiang, Xiaolong Zhu, Zhifeng
  Li, and Wei Liu.
\newblock Real-time neural style transfer for videos.
\newblock In {\em Proceedings of the IEEE Conference on Computer Vision and
  Pattern Recognition}, pages 783--791, 2017.

\bibitem{jing2019neural}
Yongcheng Jing, Yezhou Yang, Zunlei Feng, Jingwen Ye, Yizhou Yu, and Mingli
  Song.
\newblock Neural style transfer: A review.
\newblock {\em IEEE transactions on visualization and computer graphics},
  26(11):3365--3385, 2019.

\bibitem{amini2015semi}
Massih-Reza Amini and Nicolas Usunier.
\newblock Semi-supervised learning.
\newblock In {\em Learning with Partially Labeled and Interdependent Data},
  pages 33--61. Springer, 2015.

\bibitem{shabtai2012survey}
Asaf Shabtai, Yuval Elovici, and Lior Rokach.
\newblock {\em A survey of data leakage detection and prevention solutions}.
\newblock Springer Science \& Business Media, 2012.

\bibitem{kukavcka2017regularization}
Jan Kuka{\v{c}}ka, Vladimir Golkov, and Daniel Cremers.
\newblock Regularization for deep learning: A taxonomy.
\newblock {\em arXiv preprint arXiv:1710.10686}, 2017.

\bibitem{tschandl2018ham10000}
Philipp Tschandl, Cliff Rosendahl, and Harald Kittler.
\newblock The ham10000 dataset, a large collection of multi-source
  dermatoscopic images of common pigmented skin lesions.
\newblock {\em Scientific data}, 5(1):1--9, 2018.

\bibitem{codella2018skin}
Noel~CF Codella, David Gutman, M~Emre Celebi, Brian Helba, Michael~A Marchetti,
  Stephen~W Dusza, Aadi Kalloo, Konstantinos Liopyris, Nabin Mishra, Harald
  Kittler, et~al.
\newblock Skin lesion analysis toward melanoma detection: A challenge at the
  2017 international symposium on biomedical imaging (isbi), hosted by the
  international skin imaging collaboration (isic).
\newblock In {\em 2018 IEEE 15th international symposium on biomedical imaging
  (ISBI 2018)}, pages 168--172. IEEE, 2018.

\bibitem{combalia2019bcn20000}
Marc Combalia, Noel~CF Codella, Veronica Rotemberg, Brian Helba, Veronica
  Vilaplana, Ofer Reiter, Cristina Carrera, Alicia Barreiro, Allan~C Halpern,
  Susana Puig, et~al.
\newblock Bcn20000: Dermoscopic lesions in the wild.
\newblock {\em arXiv preprint arXiv:1908.02288}, 2019.

\bibitem{ramella2021hair}
Giuliana Ramella.
\newblock Hair removal combining saliency, shape and color.
\newblock {\em Applied Sciences}, 11(1):447, 2021.

\bibitem{tan2019EfficientNet}
Mingxing Tan and Quoc Le.
\newblock Efficientnet: Rethinking model scaling for convolutional neural
  networks.
\newblock In {\em International Conference on Machine Learning}, pages
  6105--6114. PMLR, 2019.

\bibitem{ha2020identifying}
Qishen Ha, Bo~Liu, and Fuxu Liu.
\newblock Identifying melanoma images using efficientnet ensemble: Winning
  solution to the siim-isic melanoma classification challenge.
\newblock {\em arXiv preprint arXiv:2010.05351}, 2020.

\bibitem{johnson2016perceptual}
Justin Johnson, Alexandre Alahi, and Li~Fei-Fei.
\newblock Perceptual losses for real-time style transfer and super-resolution.
\newblock In {\em European conference on computer vision}, pages 694--711.
  Springer, 2016.

\bibitem{dumoulin2016learned}
Vincent Dumoulin, Jonathon Shlens, and Manjunath Kudlur.
\newblock A learned representation for artistic style.
\newblock {\em ICLR}, 2017.

\bibitem{julesz1983human}
Bela Julesz and James~R Bergen.
\newblock Human factors and behavioral science: Textons, the fundamental
  elements in preattentive vision and perception of textures.
\newblock {\em Bell System Technical Journal}, 62(6):1619--1645, 1983.

\bibitem{chen2017stylebank}
Dongdong Chen, Lu~Yuan, Jing Liao, Nenghai Yu, and Gang Hua.
\newblock Stylebank: An explicit representation for neural image style
  transfer.
\newblock In {\em Proceedings of the IEEE conference on computer vision and
  pattern recognition}, pages 1897--1906, 2017.

\bibitem{chen2016fast}
Tian~Qi Chen and Mark Schmidt.
\newblock Fast patch-based style transfer of arbitrary style.
\newblock {\em arXiv preprint arXiv:1612.04337}, 2016.

\end{thebibliography}

\end{document}